\documentclass{cohere}
\usepackage{bbm}

\usepackage{xcolor}
\definecolor{red}{rgb}{0.84,0.153,0.153}
\definecolor{green}{rgb}{0.172,0.627,0.172}

\title{FAIR-Ensemble: When Fairness Naturally Emerges From Deep Ensembling}
\author{
    name={Wei-Yin Ko*},
    affiliation={Cohere For AI Community}
}
\author{
    name={Daniel D'souza*},
    affiliation={Cohere For AI Community},
}
\author{
    name={Karina Nguyen},
    affiliation={UC Berkeley, Cohere For AI Community},
}
\author{
    name={Randall Balestriero},
    affiliation={Cohere For AI Community},
}
\author{
    name={Sara Hooker},
    affiliation={Cohere For AI},
}

\date{\today}

\abstract{Ensembling multiple Deep Neural Networks (DNNs) is a simple and effective way to improve top-line metrics and to outperform a larger single model. In this work, we go beyond top-line metrics and instead explore the impact of ensembling on subgroup performances. Surprisingly, we observe that even with a simple homogeneous ensemble --all the individual DNNs share the same training set, architecture, and design choices-- the minority group performance disproportionately improves with the number of models compared to the majority group, i.e. fairness naturally emerges from ensembling. Even more surprising, we find that this gain keeps occurring even when a large number of models is considered, e.g. $20$, despite the fact that the average performance of the ensemble plateaus with fewer models. Our work establishes that simple DNN ensembles can be a powerful tool for alleviating disparate impact from DNN classifiers, thus curbing algorithmic harm. We also explore why this is the case. We find that even in homogeneous ensembles, varying the sources of stochasticity through parameter initialization, mini-batch sampling, and data-augmentation realizations, results in different fairness outcomes.

}

\begin{document}

\section{Introduction}
\label{sec:intro}

Deep Neural Networks (DNNs) are powerful function approximators that outperform other alternatives on a variety of tasks \citep{transformer,arulkumaran2017brief,hinton2012deep,resnet}. To further boost performance, a simple and popular recipe is to average the predictions of multiple DNNs, each trained independently from the others to solve the given task, this is known as {\em model ensembling} \citep{breiman2001random,dietterich2000ensemble}.

By averaging independently trained models, one avoids single model symptomatic mistakes by relying on the wisdom of the crowd to improve generalization performance, regardless of the type of model being employed. While existing work has focused on improvements towards aggregate performance \citep{fort2019deep,gupta2022ensembling,Opitz_1999} or gains in efficiency over a single larger model \citep{wang2020,wortsman2022model}, there has been limited consideration of how sensitive ensembling performance is on certain subsets of the data distribution. 

\begin{figure*}[ht]
    \centering
    \begin{minipage}{0.49\linewidth}
    \centering
    \underline{CIFAR100}
    \includegraphics[width=\linewidth]{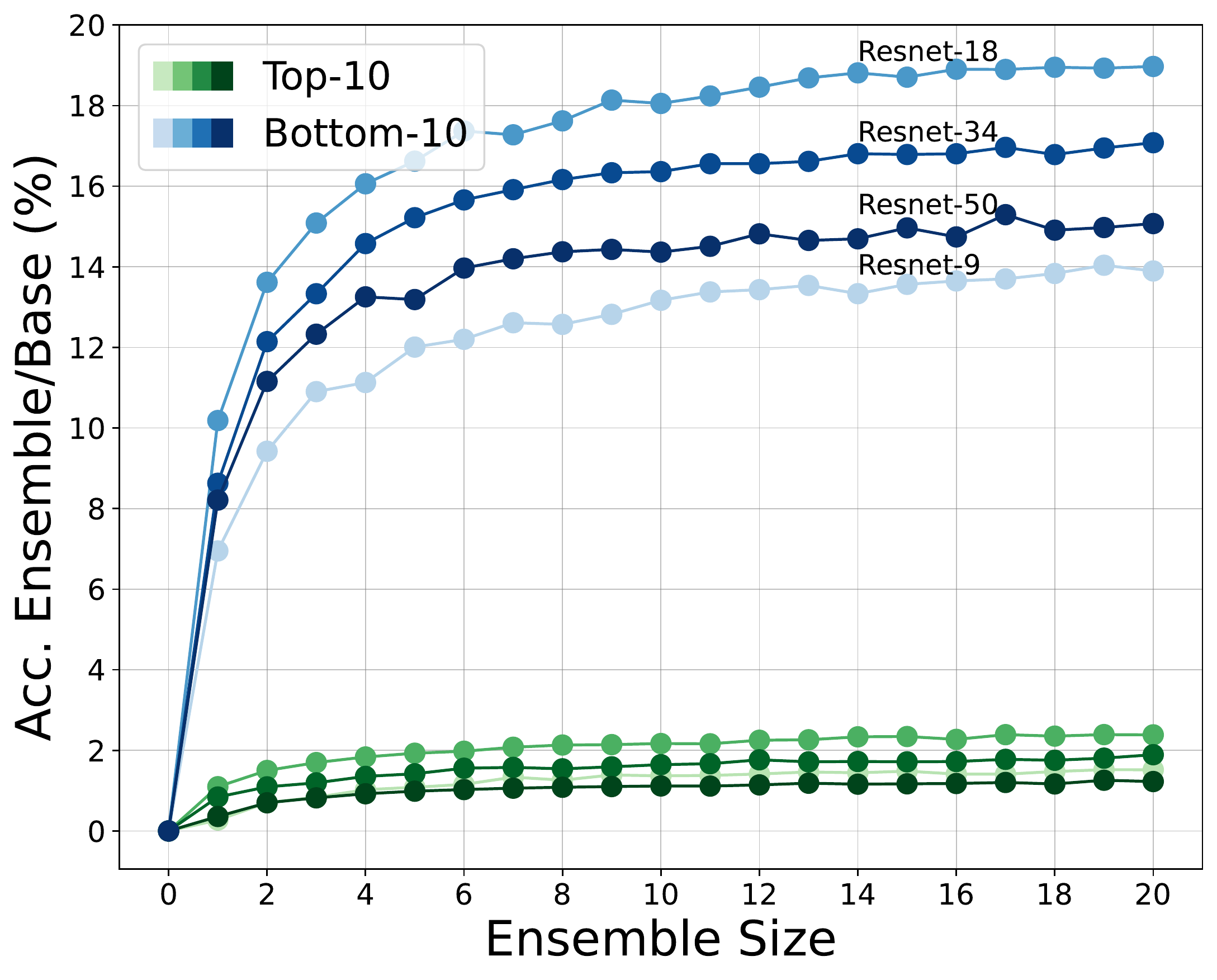}\\
    \vspace{-0.3cm}
    \end{minipage}
    \begin{minipage}{0.49\linewidth}
    \centering
    \underline{TinyImageNet}
    \includegraphics[width=\linewidth]{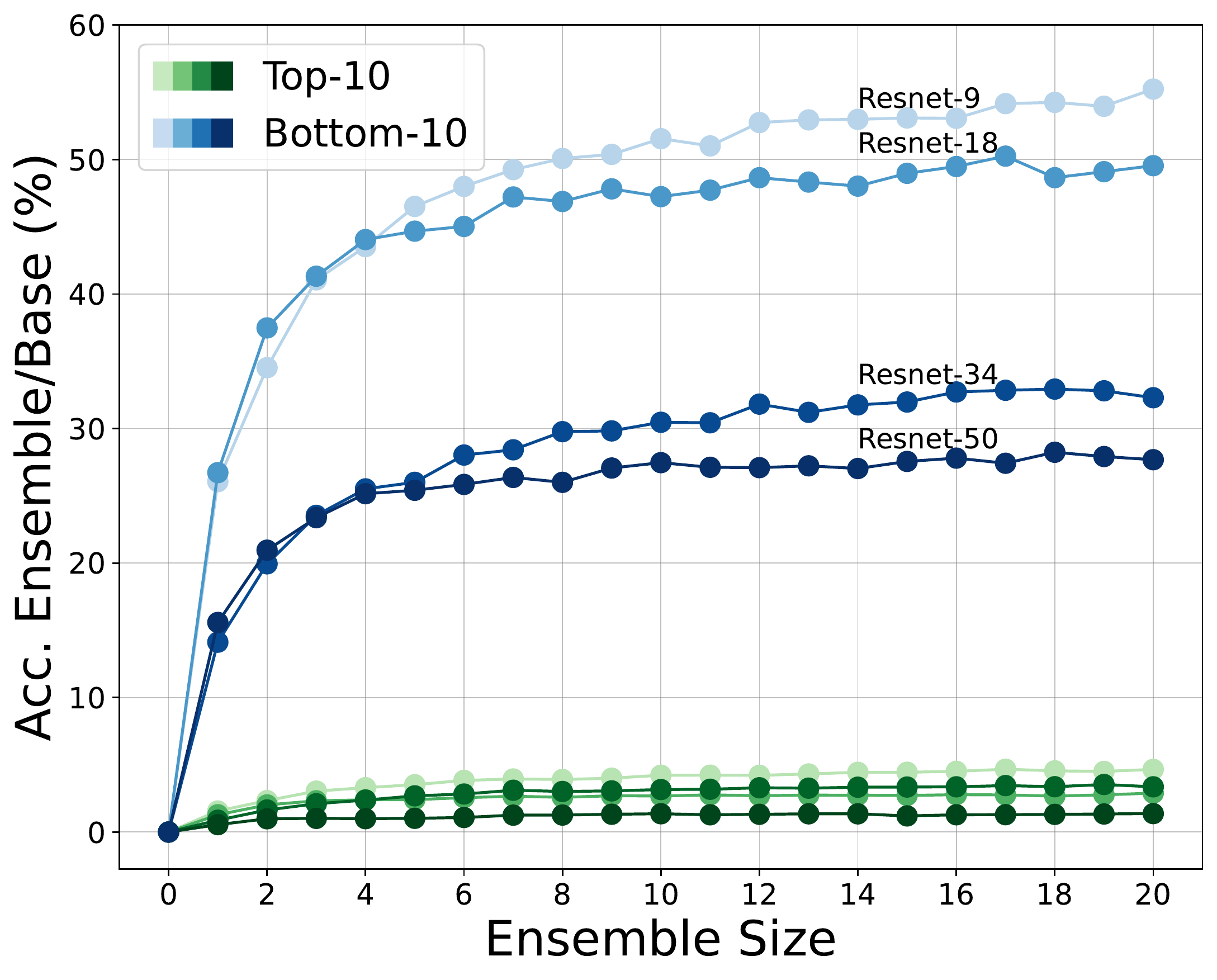}\\
    \vspace{-0.3cm}
    \end{minipage}
    \caption{\small Relative Accuracy for Top-K/Bottom-K. Plot of the ratio of the homogeneous ensemble accuracy over a single base model ({\bf y-axis}) illustrates strong benefits for the  {\color{blue}bottom-k} group of ensembling while the {\color{green}top-k} group only marginally benefits.}
    \label{fig:teaser_percent}
\end{figure*}

Understanding performance on subgroups is a frequent concern from a fairness perspective. A common fairness objective is mitigating disparate impact \citep{kleinberg2016,zafar2015} where a class or subgroup of the dataset presents far higher error rates than other subsets of the distribution. In particular, and as we will thoroughly describe in \cref{sec:background}, many strategies have emerged to improve fairness by designing novel ensembling strategies based on fairness measures obtained from labeled attributes. In this study, we take a step back and focus on studying the fairness benefits of the simplest ensembling strategy: homogeneous ensembles. In this setting, the individual models in the ensemble all have the same architecture and hyperparameters. They are also trained with the same optimizer, data-augmentations, and training set.

Our results are surprising: despite the absence of "diversity" in the models being trained in the homogeneous ensemble, the only sources of randomness are (i) the parameters' initialization, (ii) the realizations of the data-augmentations, and (iii) the ordering of the mini-batches. The final predictions are diverse enough to provide substantial improvements for both the minority groups and the bottom-k classes upon which a single model performs badly. This emergence of fairness is observed consistently across thousands of experiments on popular architectures (ResNet9/18/34/50, VGG16, MLPMixer, ViTs) and datasets (CIFAR10/100/100-C, TinyImagenet, CelebA) (\cref{sec:fair_ensemble}). The first important conclusion unlocked by our thorough empirical validation is that one may effectively improve minority group performance by using the same architecture and hyperparameters for each individual model without the need to observe corresponding labeled attributes. A second crucial finding is that solely controlling for initialization, batch ordering, and data-augmentation realizations is already enough to make training episodes produce models that are complementary with each other. Other factors such as architectures, optimizers, or data-augmentation families may not be the most important variables to produce fair ensemble (\cref{sec:ensemble_bottom_why}). The last interesting observation is that, as a function of the number of models in the homogeneous ensemble, the average performance quickly plateaus after $4$ to $5$ models, but the bottom-k group performance keeps increasing steadily for up to $50$ models. In short, when performing deep ensembling, one should employ as many models as possible--even beyond the point at which the average performance plateaus--in order to produce a final ensemble with as much fairness as possible. Beyond fairness of homogeneous deep ensembles, our empirical study also offers a rich variety of new observations e.g., tying the severity of image corruption to the relative benefits that emerges from homogeneous deep ensembles. 

\textbf{Our contributions can be enumerated as follows}:
\begin{enumerate}
\itemsep0pt
    \item We demonstrate that simple homogeneous deep ensembles trained with the same objective, architecture and optimization settings minimize worst-case error. This holds in both balanced and imbalanced datasets with protected attributes that the model is not trained on. 
    \item We further perform controlled sensitivity experiments where constructed class imbalance and data perturbation is applied (\cref{sec:fair_ensemble}). We observe that homogeneous ensembles continue to improve fairness and, in particular, the bottom-k group benefits more and more with the size of the ensemble compared to the top-k group as the severity of the corruption increases. These observations are held even when the protected attribute is imbalanced and underrepresented, such as in our CelebA experiments.
    \item We further dive into possible causes for this emergence of fairness in homogeneous deep ensembles by measuring model disagreement (\cref{sec:churn}) and by ablating for the different sources of randomness, e.g., weight-initialization (\cref{sec:stochasticity}). We obtain interesting results that suggest certain sources of stochasticity such as mini-batch ordering or data-augmentation realizations are enough to bring diversity into homogeneous ensembles.
\end{enumerate}
The codebase to reproduce our results and figures is available \href{https://github.com/dsouzadaniel/fair_ensemble}{\textbf{here}}

\section{Related Work}
\label{sec:background}
Deep ensembling of Deep Neural Networks (DNNs) is a popular method to improve top-line metrics \citep{2016Lakshminarayanan}. Several works have sought to further improve aggregate performance by amplifying differences between models in the ensemble ranging from varying the data augmentation used for each model \citep{stickland2020}, the architecture \citep{zaidi2021neural}, the hyperparameters \citep{wortsman2022model}, and even the training objectives \citep{jain2020maximizing}. {\em As will become clear, our focus is on the opposite setting where all the models in the ensemble share the same objective, training set, architecture, and optimizer}.

\textbf{Beyond Top-line metrics} Discussions of algorithmic bias often focus on datasets collection and curation \citep{barocas-hardt-narayanan,zhao-etal-2017-men,shankar2017no}, with limited work to-date understanding the role of model design or optimization choices on amplifying or curbing bias \citep{ogueji2022,hooker2019,balestriero2022effects}. Consistent with this, there has been limited work to-date on understanding the implications of ensembling on subgroup error. \citep{Nina2017} points out the theoretical possibility of using an ensemble of randomly selected candidate models to improve fairness, however no empirical validation was presented. \citep{8995403} considers AdaBoost \citep{Freund1995ADG} ensembles and shows that upweighting unfairly predicted examples reaches higher fairness. \citep{kenfack2021impact, chen2022maat} propose explicit schemes to induce fairness by designing heterogeneous ensembles, and \citep{gohar2023towards} provides ensemble design suggestions in heterogeneous ensembles. Recently, \citep{cooper2023arbitrariness} proposed a self-consistency metric for measuring the arbitrariness of single model outputs and provided a modified bagging solution specifically designed to mitigate the arbitrariness from model predictions. {\em  In contrast, our goal is to demonstrate how the simplest homogeneous ensembling strategy where each model is trained independently and with identical settings naturally exhibit fairness benefits without having to measure or have labels for the minority attributes}.

\textbf{Understanding why ensembling benefits subgroup performance.} Several works to date have sought to understand why weight averaging performs well and improves top-line metrics \citep{gupta2022ensembling}. However, few to our knowledge have sought to understand why ensembles disproportionately benefit bottom-k and minority group performance. In particular, \citep{rame2022diverse} explores why weight averaging performs well on out-of-distribution data, relating variance to diversity shift. {\em In this work, we instead explore how individual sources of inherent stochasticity in uniform homogeneous ensembles impact subgroup performance.}

In this work, we consider the impact of ensembling on both \textit{balanced} and \textit{imbalanced} subgroups. Fairness considerations emerge for both groups. Real world data tends to be imbalanced, where infrequent events and minority groups are under-represented in the data collection processes. This leads to representational disparity \citep{hashimoto18a} where the under-represented group consequently experiences higher error rates. Even when training sets are balanced, with an equivalent number of training data points, certain features may be imbalanced leading to a long-tail within a balanced class. Both settings can result in \textit{disparate impact}, where error rates for either a class or a subgroup are far higher \citep{Chatterjee2020Coherent,NEURIPS2020_1e14bfe2}. 
This notion of unfairness is widely documented in machine learning systems: \citep{buolamwini2018gender} find that facial analysis datasets reflect a preponderance of lighter-skinned subjects, with far higher model error rates for dark skinned women. \citep{shankar2017no} show that models trained on datasets with limited geo-diversity show sharp degradation on data drawn from other locales. Word frequency co-occurrences within text datasets frequently reflect social biases relating to gender, race and disability \citep{Garg2017,zhao-etal-2018-gender,bolukbasi2016man,basta-etal-2019-evaluating}. 

In the following \cref{sec:fair_ensemble}, we will study how the randomness stemming from the random initialization, data-augmentation realization, or mini-batch ordering during training may provide enough diversity in homogeneous deep ensembles for fairness to naturally emerge. The why is left for \cref{sec:ensemble_bottom_why}. 

\begin{figure*}[t!]

    \centering
    \begin{minipage}{0.01\linewidth}
    \rotatebox{90}{\small ensemble/base}
    \end{minipage}
    \begin{minipage}{0.98\linewidth}
    \begin{subfigure}{0.16\linewidth}
        \centering
        ResNet9\\
        \includegraphics[width=\linewidth]{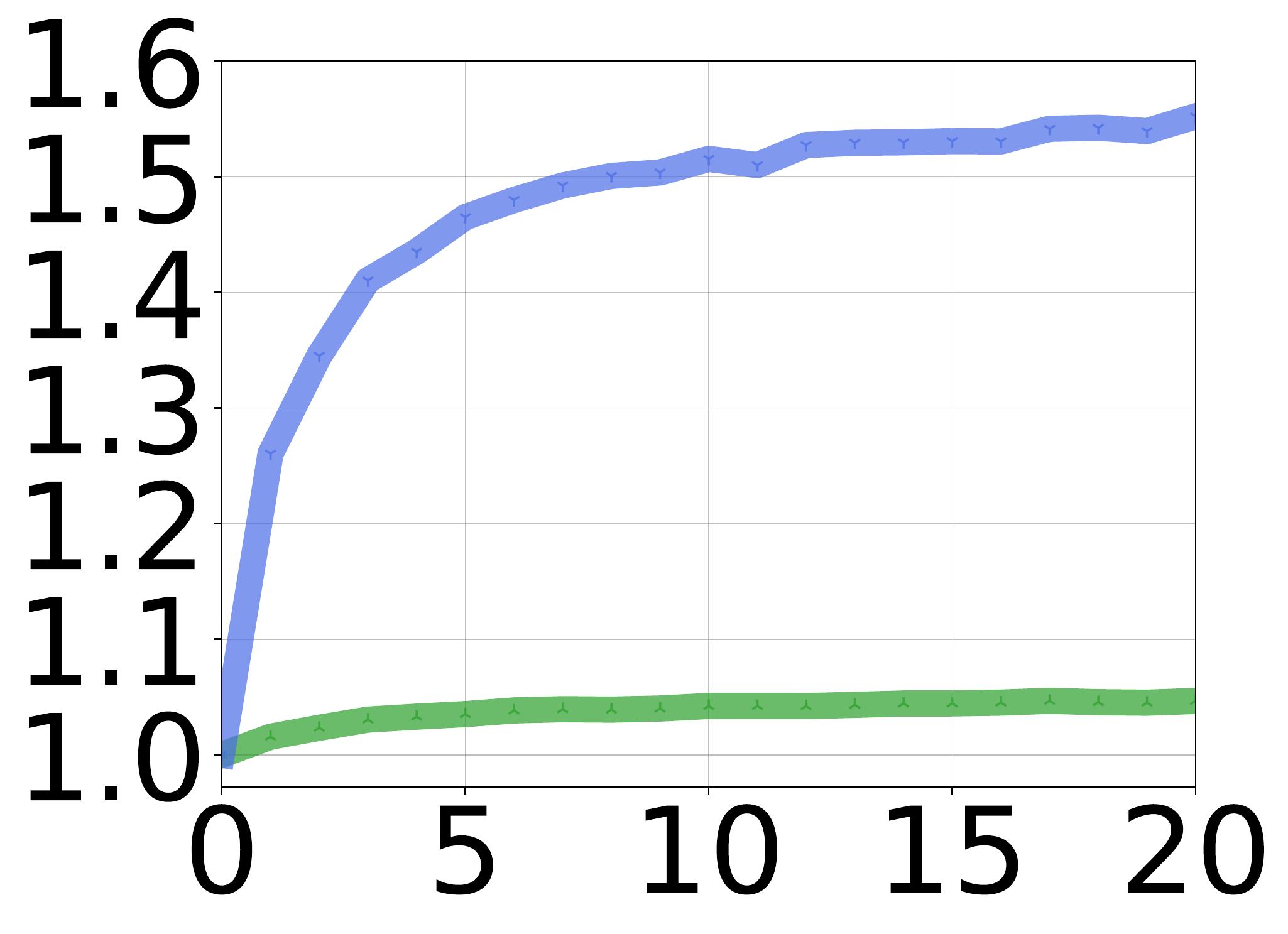}
        \\[-0.7em]
        {\hspace{0.2cm}\scriptsize models in ensemble}
    \end{subfigure}
    \begin{subfigure}{0.16\linewidth}
        \centering
        ResNet18\\
        \includegraphics[width=\linewidth]{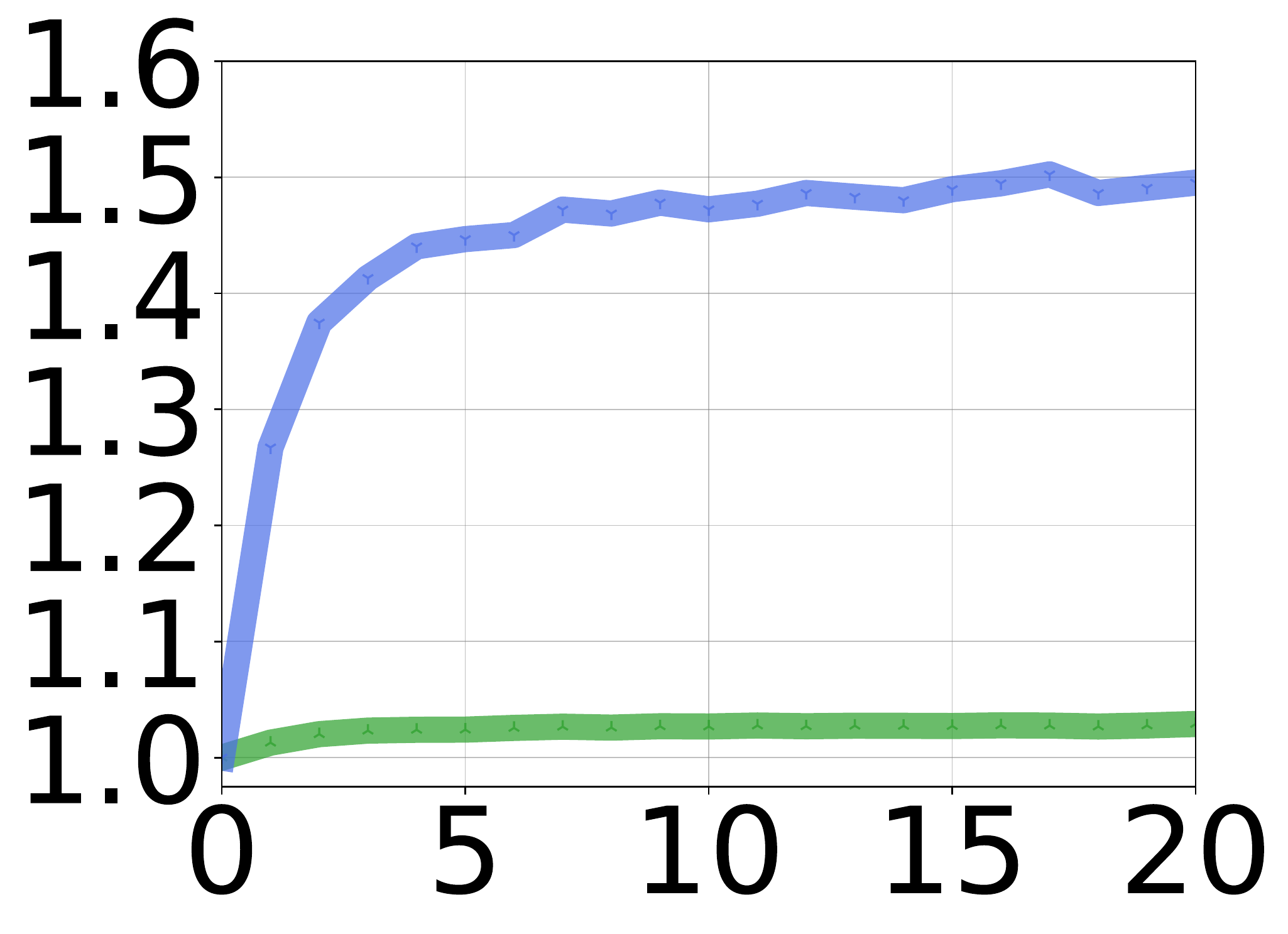}
        \\[-0.7em]
        {\hspace{0.2cm}\scriptsize models in ensemble}
    \end{subfigure}
    \begin{subfigure}{0.16\linewidth}
        \centering
        ResNet34\\
        \includegraphics[width=\linewidth]{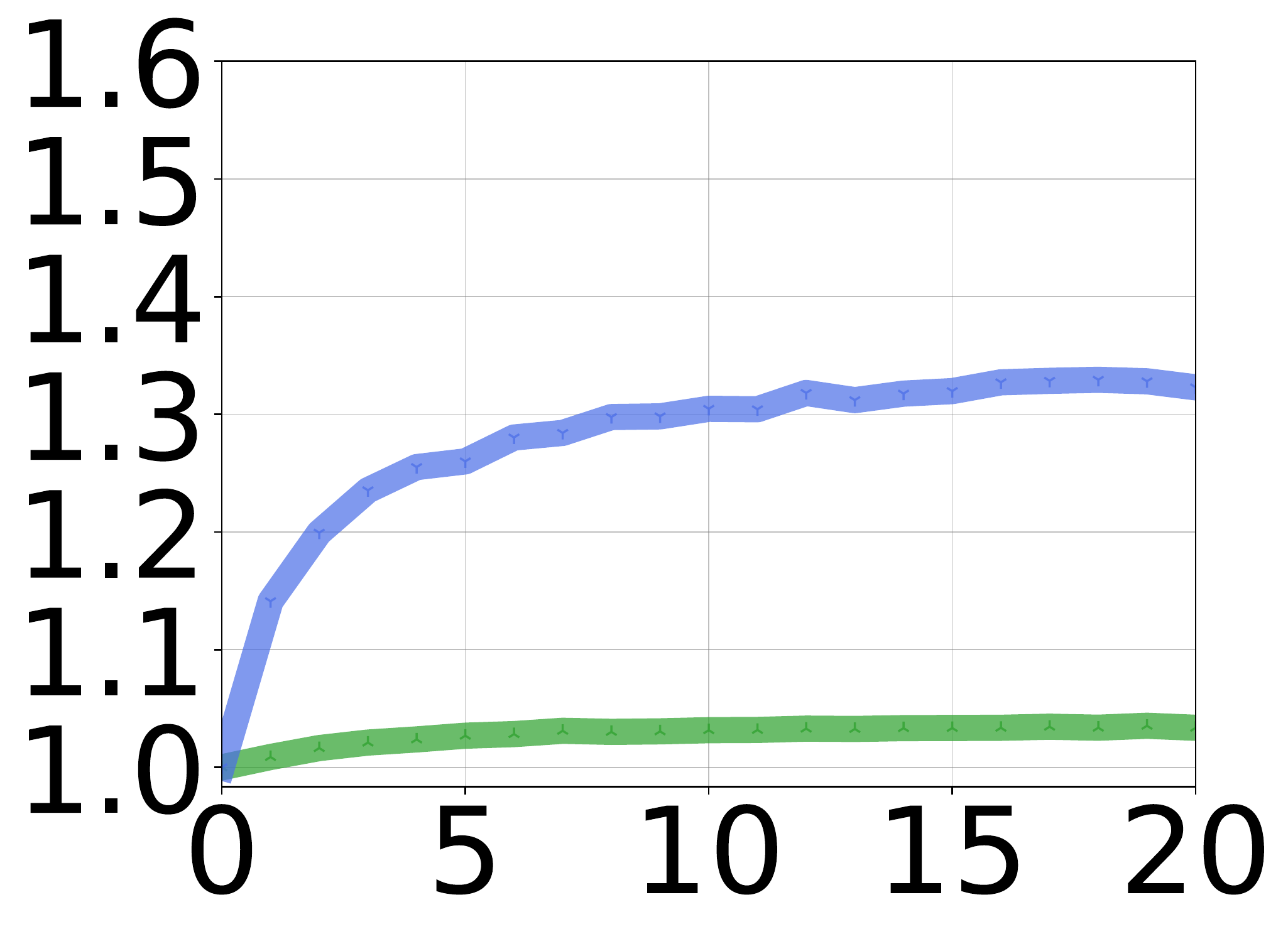}
        \\[-0.7em]
        {\hspace{0.2cm}\scriptsize models in ensemble}
    \end{subfigure}
    \begin{subfigure}{0.16\linewidth}
        \centering
        ResNet50\\
        \includegraphics[width=\linewidth]{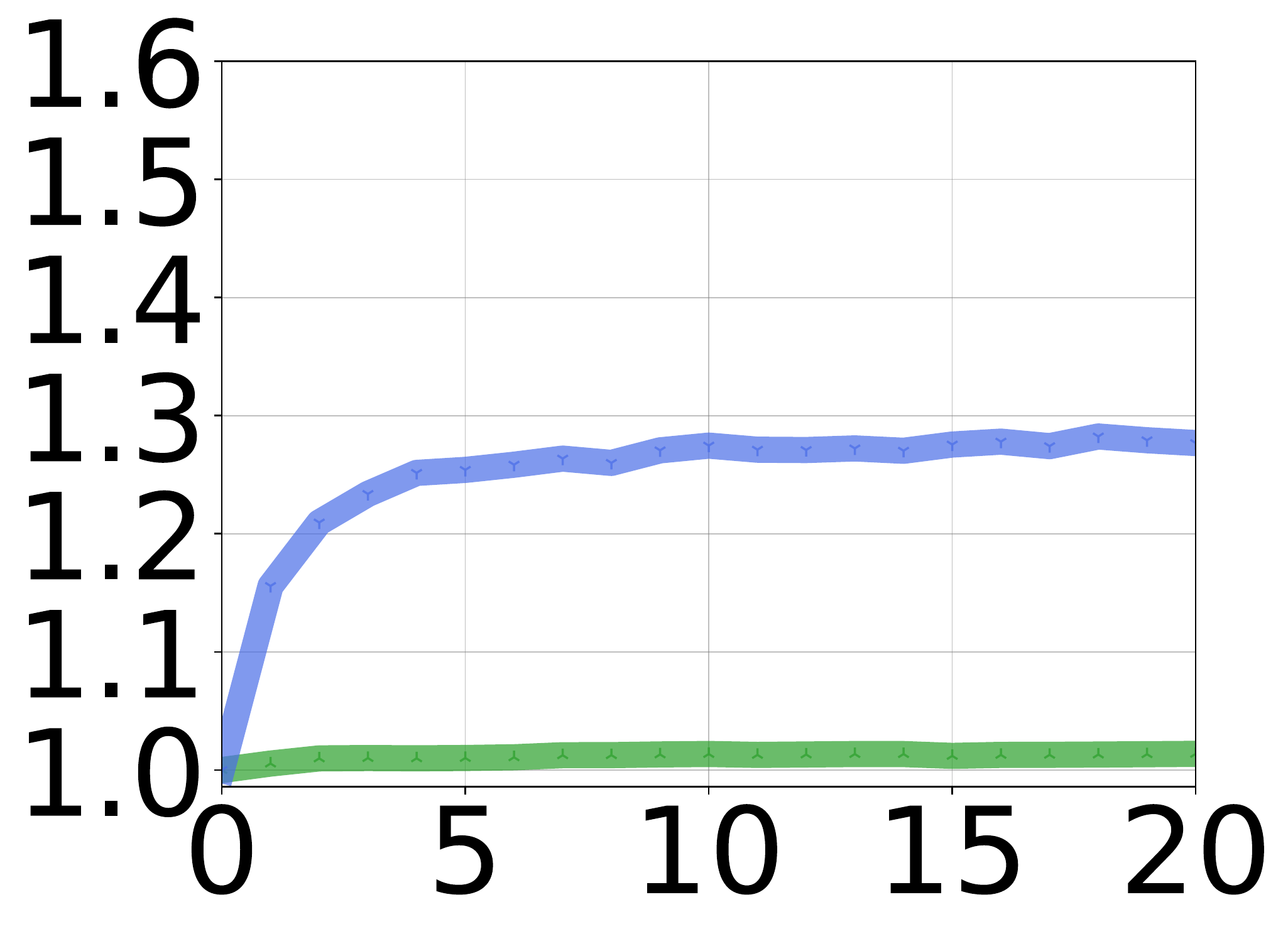}
        \\[-0.7em]
        {\hspace{0.2cm}\scriptsize models in ensemble}
    \end{subfigure}
    \begin{subfigure}{0.16\linewidth}
        \centering
        VGG16\\
        \includegraphics[width=\linewidth]{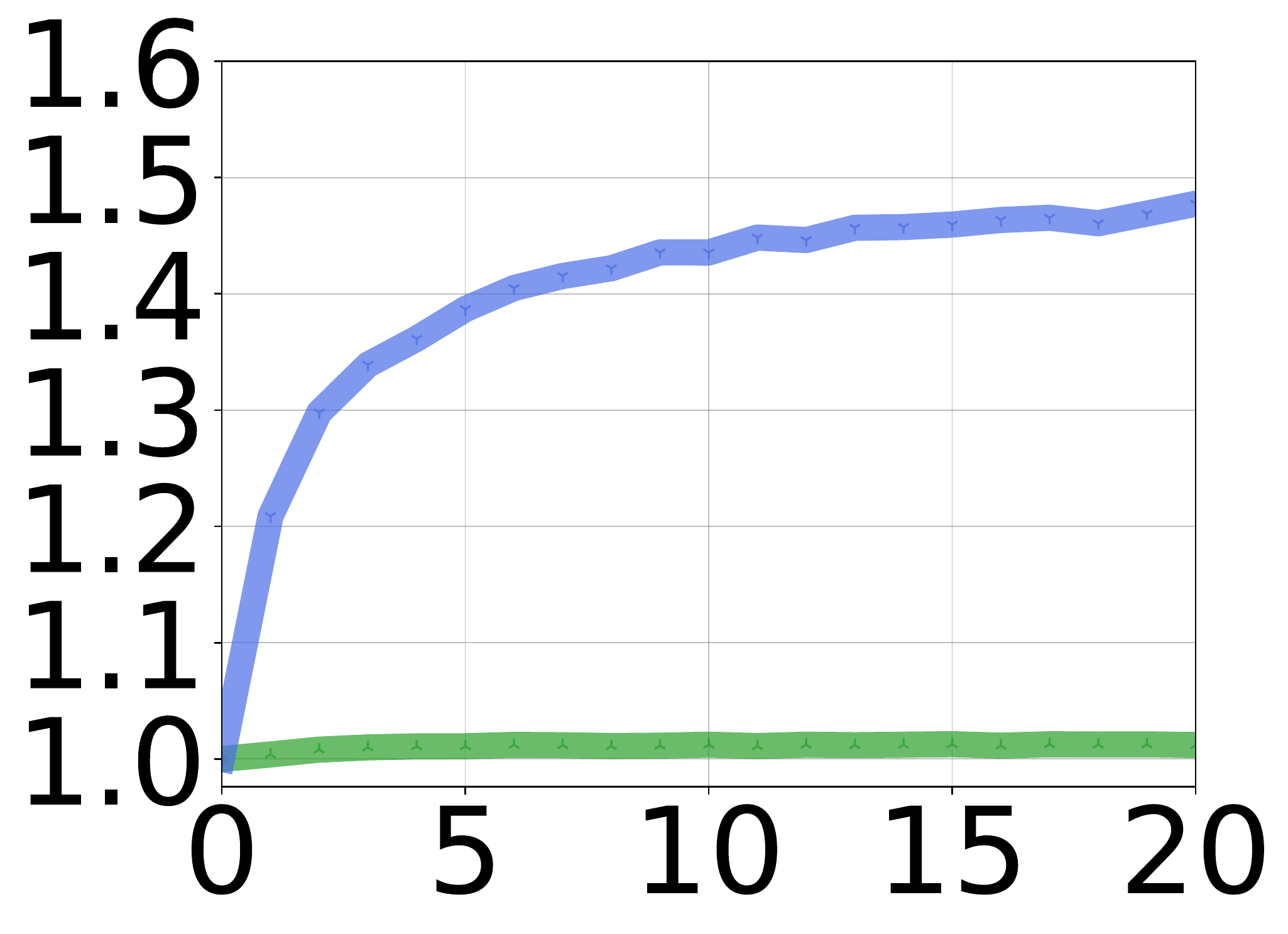}
        \\[-0.7em]
        {\hspace{0.2cm}\scriptsize models in ensemble}
    \end{subfigure}
    \begin{subfigure}{0.16\linewidth}
        \centering
        ViT\\
        \includegraphics[width=\linewidth]{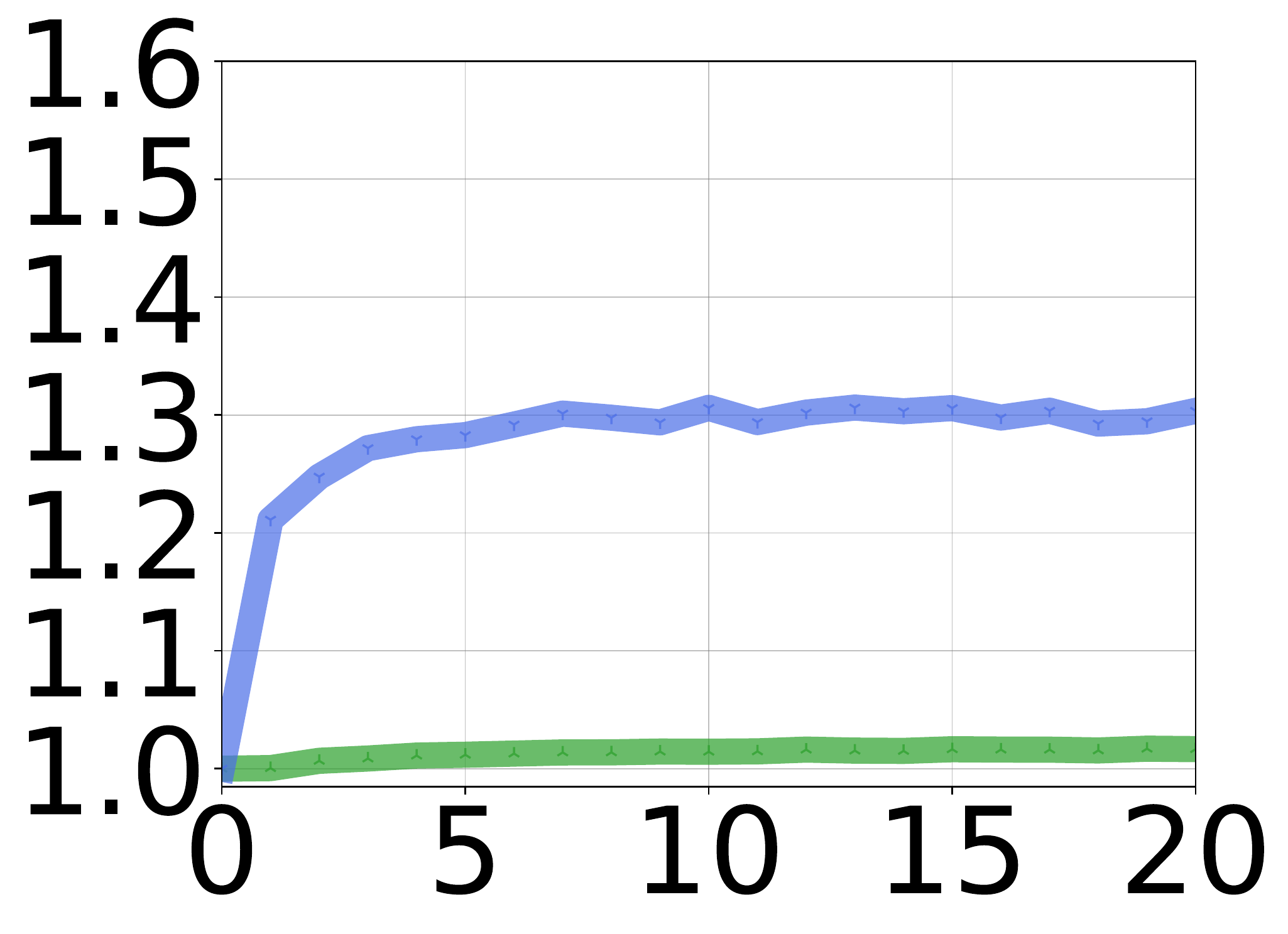}
        \\[-0.7em]
        {\hspace{0.2cm}\scriptsize models in ensemble}
    \end{subfigure}

    \end{minipage}
    \\
\vspace{-0.1cm}
    \caption{\small Test set accuracy gain as a ratio of ensemble accuracy \% over the singular base model ({\bf y-axis}) by group ({\color{green}top-k } and {\color{blue}bottom-k}) for TinyImageNet for different architectures ({\bf columns}) with varying the number of models within the homogeneous ensemble grows ({\bf x-axis}). 
    We clearly observe that as the number of models within the homogeneous ensemble grows, the bottom-k group performance improves. In particular, the bottom-k group's accuracy gain outgrow the top-k group's. This occurs despite the fact that the models within the ensemble are all employing the same hyperparameters, thus inherently share the same functional biases.
    The absolute accuracies are provided in \cref{tab:all_summary} below, and CIFAR100 results are in \cref{fig:fairness_c100}.
    }
    \label{fig:fairness}
\end{figure*}

\section{FAIR-Ensemble: When Homogeneous Ensembles Disproportionately Benefit Minority Groups}
\label{sec:fair_ensemble}

Throughout our study, we will consider a DNN to be a mapping $f_{\theta}: \mathcal{X} \mapsto \mathcal{Y}$ with trainable weights $\theta \in \Theta$. The training dataset $\mathcal{D}$ consists of $N$ data points $\mathcal{D}=\left\{\mathbf{x}_n, y_n\right\}_{n=1}^N$.  Given the training dataset $\mathcal{D}$, the trainable weights are optimized by minimizing an objective function. 
We denote a homogeneous ensemble of $m$ classification models by $\left\{{f_\theta}_1, \ldots, {f_\theta}_m\right\}$, where ${f_\theta}_i$ is the $i^{t h}$ model. Each model is trained independently of the others. {\em We will denote by \textbf{homogeneous ensemble} the setting where the same model architecture, hyperparameters, optimizer, and training set are employed for each model of the ensemble.}

\subsection{Experimental Set-up}
\label{sec:experiment_setup}

\textbf{Experimental set-up:}~we evaluate homogeneous ensembles on CIFAR100 \citep{krizhevsky2009learning} and TinyImageNet \citep{russakovsky2015imagenet} datasets across various architectures: ResNet9/18/34/50 \citep{he2016deep}, VGG16 \citep{simonyan2014very}, MLP-Mixer \citep{tolstikhin2021mlp} and ViT \citep{dosovitskiy2020image}. Training and implementation details are provided in \cref{sec:experiments}. Whenever we report results on the homogeneous ensemble, unless the number of models is explicitly stated, it will comprise of 20 models. Each model is trained independently as in\citep{breiman2001random,lee2015}, i.e. we do not control for any of the remaining sources of randomness as this will be explored exclusively within \cref{sec:stochasticity}.
\\
\textbf{Balanced Dataset Sub-Groups:}~for top-k and bottom-k, we calculate the class accuracy of the base model and find the best and worst $K$ ($K=10$) performing classes and track the associated classes as bottom-k and top-k groups. We then proceed to measure how performance on these groups changes as a function of the homogeneous ensemble size. We highlight that although we leverage $K=10$ in many experiments, the precise choice of $K$ does not impact our findings, as demonstrated in \cref{fig:Tinyimagenet_K} and \cref{fig:C100_K}.
\\
\textbf{Imbalanced Dataset Sub-Groups:}~we consider a setting where the protected attribute is an underlying variable different from the classification target. Similar to the setup in \citep{hooker2019compressed,veldanda2022fairness}, we treat CelebA\citep{liu2015faceattributes} as a binary classification problem where the task is predicting hair color $\mathcal{Y}$ =\{blonde, dark haired\} and the sensitive attribute is gender. In this dataset, blonde individuals constitute only 15\% of which a mere 6\% are males. Hence, blonde male is an underrepresented attribute. We then proceed to measure how performance on the protected \textbf{gender:male} attribute varies as a function of ensemble size.

Given the above experimental details, we can now proceed to present our core observations that tie the homogeneous ensemble size with its fairness benefits.

\subsection{Observing Disproportionate Benefits For Bottom-K Groups}
\label{sec:ensemble_bottom}

\begin{table*}[t!]
\setlength\tabcolsep{0.32em}
\caption{\small Depiction of the average and per-group ({\color{green}top-k } and {\color{blue}bottom-k}) absolute test set accuracies corresponding to the models and datasets depicted in \cref{fig:fairness} above and \cref{fig:fairness_c100} in the Appendix, again the homogeneous ensemble consists of $20$ models. We clearly observe that fairness naturally emerges through ensembling i.e. the bottom-k group substantially benefits from homogeneous ensembling compared to the top-k group.}
\vspace{-0.2cm}
\resizebox{\textwidth}{!}{%
\begin{tabular}{l ccc ccc ccc ccc}
\multicolumn{1}{c}{}&\multicolumn{6}{c}{CIFAR100} &\multicolumn{6}{c}{TinyImageNet}\\
\cmidrule[\heavyrulewidth]{2-13}
\multicolumn{1}{c}{} & \multicolumn{3}{c}{Ensemble}& \multicolumn{3}{c}{Single}& \multicolumn{3}{c}{Ensemble}& \multicolumn{3}{c}{Single}\\\cmidrule[\heavyrulewidth]{1-13}
Arch. & mean&	\color{green}top-k&	\color{blue}bottom-k & mean & \color{green}top-k & \color{blue}bottom-k& mean&	\color{green}top-k&	\color{blue}bottom-k & mean & \color{green}top-k & \color{blue}bottom-k\\\cmidrule[\heavyrulewidth]{1-13}
ResNet9  &{ 77.01} & { 92.18} & { 58.43} & { 72.21} & { 90.80} & { 51.30} & { 58.29} & { 86.66} & { 23.60} & { 50.71} & { 82.80} & { 15.20} \\
ResNet18 &{ 78.15} & { 94.19} & { 59.13} & { 73.57} & { 92.00} & { 49.70} & { 56.50} & { 86.64} & { 24.82} & { 49.29} & { 84.20} & { 16.60} \\
ResNet34 &{ 78.68} & { 93.84} & { 58.89} & { 74.26} & { 92.10} & { 50.30} & { 58.89} & { 87.44} & { 27.25} & { 52.18} & { 84.60} & { 20.60} \\
ResNet50 &{ 77.94} & { 93.53} & { 58.34} & { 74.88} & { 92.40} & { 50.70} & { 60.35} & { 87.38} & { 28.09} & { 55.00} & { 86.20} & { 22.00} \\
VGG16  &{ 76.95} & { 92.88} & { 57.32} & { 71.24} & { 91.50} & { 44.40} & { 67.04} & { 90.27} & { 38.71} & { 60.36} & { 89.20} & { 26.20} \\
MLPMixer/ViT &{ 66.69} & { 87.95} & { 40.93} & { 60.25} & { 84.50} & { 33.00} & { 56.97} & { 85.60} & { 22.42} & { 51.23} & { 84.20} & { 17.20}
\\ \bottomrule
\end{tabular}%
}
\vspace{-0.3cm}
\label{tab:all_summary}
\end{table*}

\textbf{Impact on bottom-k classes:}~in \cref{fig:teaser_percent} and \cref{fig:fairness}, we plot the relative gain in accuracy, i.e., the ratio between the homogeneous ensemble and base model performance on top-k/bottom-k groups, for each model architecture and dataset. Therefore answering the question: \textit{what is the relative improvement in performance of using a homogeneous ensemble over a single model?} Across models and datasets, there is a disproportionate benefit for the bottom-k performance. For CIFAR100, this benefit ranges from 14\%-29\% for bottom-k across different architectures compared to 1\%-4\% for top-k. For TinyImageNet the benefits are even more pronounced with a maximum gain of 55\% for bottom-k compared to 5\% for top-k across different architectures. We also provide in \cref{tab:all_summary} the absolute per-group accuracy and average performances for the corresponding models and datasets. For example, we observe a gain of more than 10\% in absolute accuracy for the bottom-k classes against a gain of around 4\% for the top-k group across settings.
As a result, we obtain that {\em even when ensembling models that share all their hyperparameters, data, and training settings, fairness naturally emerges}. Given these observations, one may wonder how does the number of models in the homogeneous ensemble impact fairness benefits. In \cref{fig:fairness} and \cref{fig:fairness_c100}, we plot fairness impact as a function of $m$, the number of models being used. A key observation we obtain is that {\em while the top-k group's performance plateaus rapidly for small $m$, the bottom-k group still exhibits improvements when reaching $m=20$}. We further explore increases of $m$ in the Appendix, where we consider up to 50 model ensembles (see \cref{fig:ensemble_size_comp}). In both TinyImageNet and CIFAR100 datasets, the absolute accuracy improvements of architectures such as ResNet9, ResNet50, and VGG16 all slowly plateaued as $m \rightarrow 50$; we also present the relative test set accuracies in \cref{fig:CIFAR100_ratio,fig:TinyImageNet_ratio}. 
For the test set accuracy performance between the top-k and the bottom-k groups over ensemble size and associated absolute errors, please refer to \cref{fig:CIFAR100_dualaxis} and \cref{fig:TinyImageNet_dualaxis}.

\begin{figure*}[t!]
    \centering
    \begin{minipage}{0.49\linewidth}
    \includegraphics[width=\linewidth]{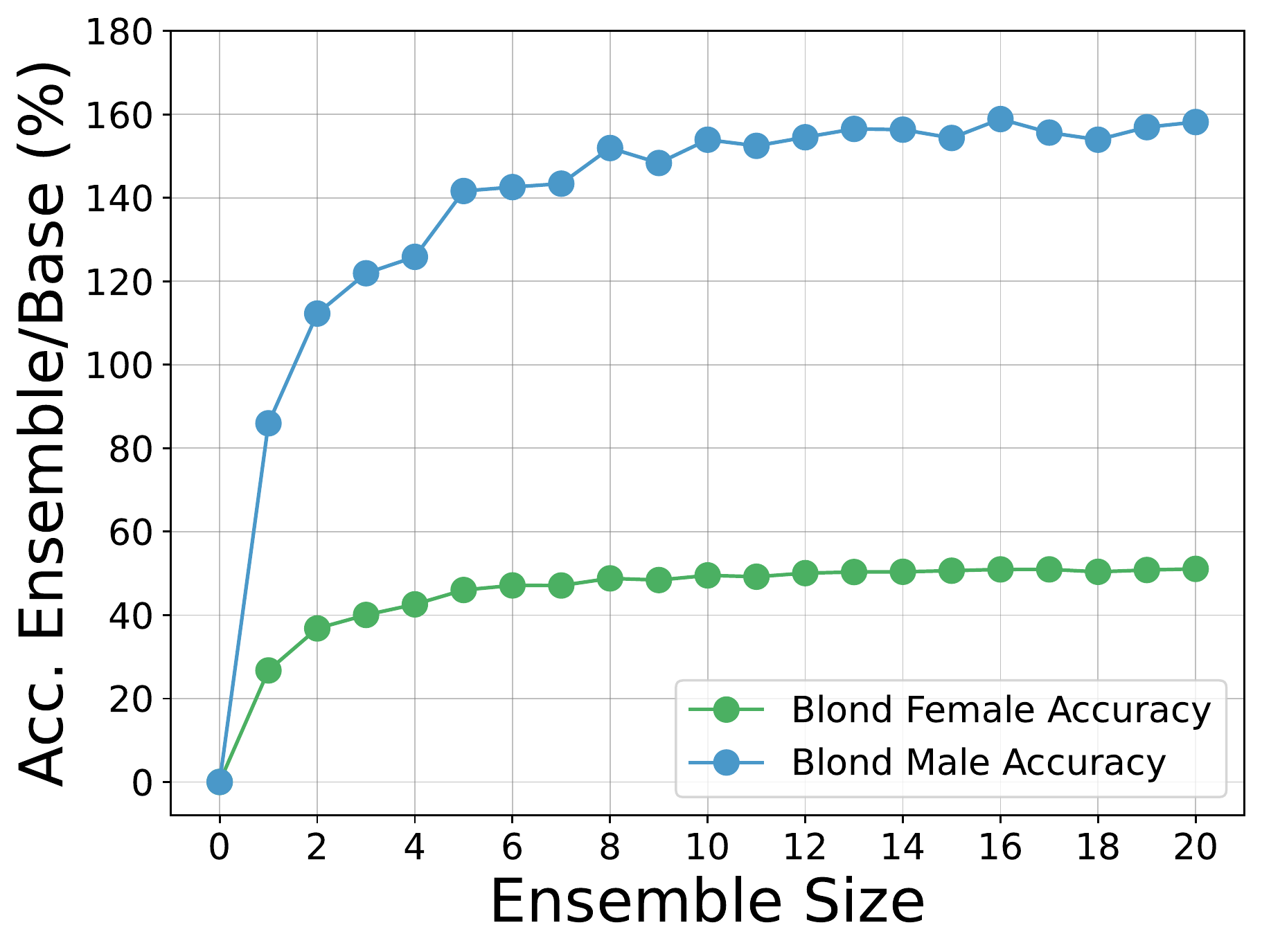}\\
    \vspace{-0.3cm}
    \end{minipage}
    \begin{minipage}{0.49\linewidth}
    \includegraphics[width=\linewidth]{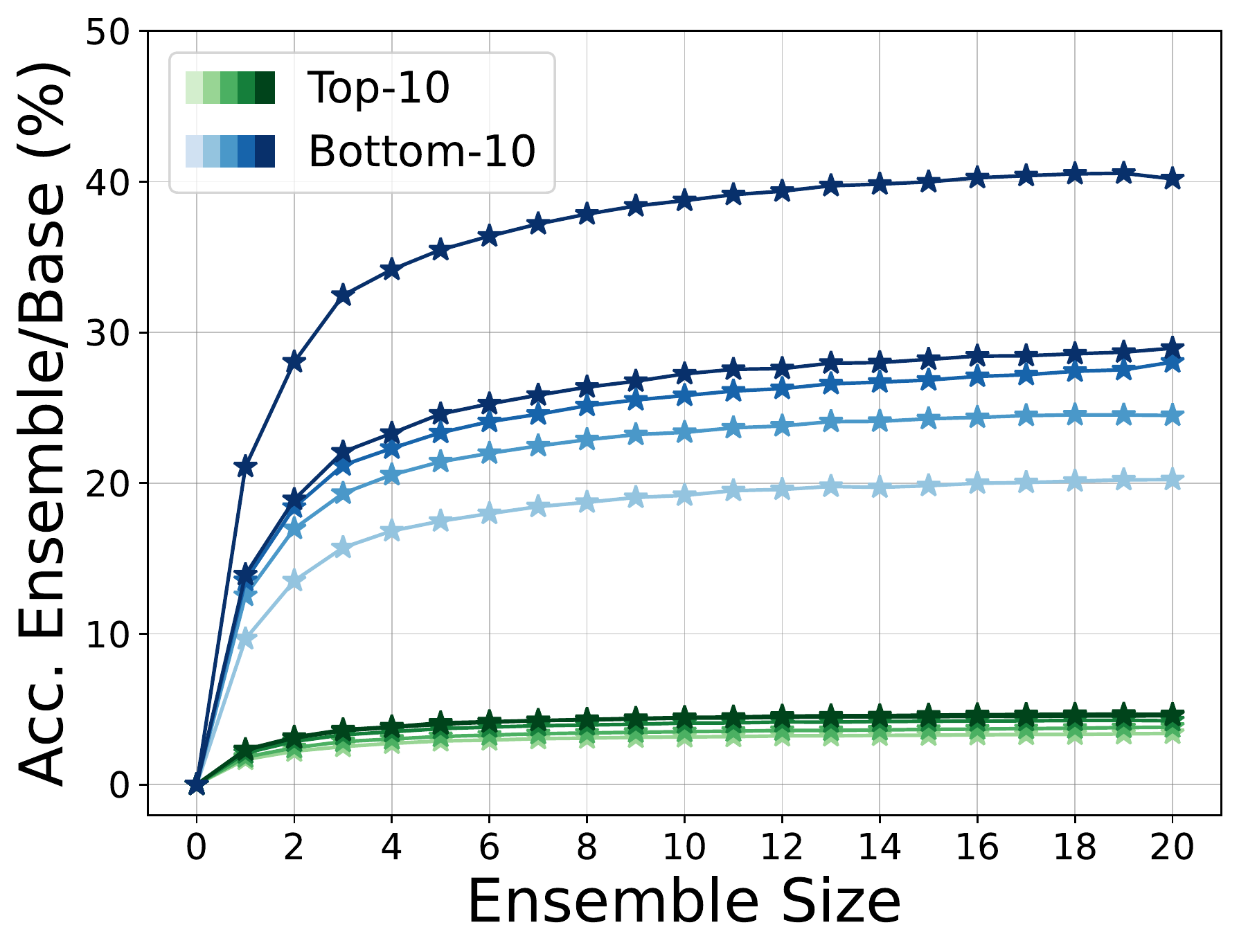}\\
    \vspace{-0.3cm}
    \end{minipage}
    
    \begin{minipage}{\linewidth}
            \caption{\small {\bf Left:} CelebA test set accuracy gain as a ratio of ensemble accuracy \% over the singular base model ({\bf y-axis}) by group ({\color{green}majority } and {\color{blue}minority}). Male is the protected attribute and Blond Males are extremely underrepresented in the training data. Nevertheless, we clearly observe that as the number of models within the homogeneous ensemble grows ({\bf x-axis)}, the protected attribute group's classification accuracy outgrows the majority group's. 
            {\bf Right:} CIFAR100-C test set accuracy gain ratio (same as {\bf left} per-group {\color{green}top-k } and {\color{blue}bottom-k}) of the homogeneous ensemble as the number of models being aggregated increases ({\bf x-axis}) for varying severity of corruption levels in CIFAR100-C (recall \cref{fig:cifar100c_samples}) from light to dark color shading. A striking observation is that not only does homogeneous DNN ensembling improves fairness by increasing the performance on the bottom group more drastically than on the top group, this effect is even more prominent at higher corruption severity levels.}
            \label{fig:cifar100c_corruption_severity}\label{fig:celeba_summary}
    \end{minipage}
    \vspace{-0.7cm}
\end{figure*}

\textbf{Controlled Experiment: CelebA}~Beyond looking at the top-k and bottom-k classes, we leverage the CelebA dataset which contains fine-grained attributes to study the fairness impact of homogeneous ensembles. Using the ResNet18 architecture, we train $20$ models and measure their performances on the protected \textbf{gender:male} attribute. Employing homogeneous ensembles, we observe the average performance for the Blonde classification task to increase from 92.02\% to 94.04\%. Furthermore, for the protected gender attribute, we see the average performance increase from 9.44\% to 21.80\%, a considerable benefit that alleviates the disparate impact on an under-represented attribute. As we previously observed, homogeneous ensembles provide a disproportionate accuracy gain in the minority subgroup as further depicted in \cref{fig:celeba_summary,fig:celeba}. 

\textbf{Controlled Experiment: CIFAR100-C}~\citep{hendrycks2018benchmarking} is an artificially constructed dataset of 19 individual corruptions on the CIFAR100 Test Dataset as depicted in \cref{fig:cifar100c_samples}, each with a severity level ranging from 1 to 5. Our goal is to understand the relation between fairness benefits for the bottom-k group and severity of the input corruption. We thus propose to benchmark our homogeneous ensembles on all severity levels, and for completeness, we benchmark and average performance across all corruptions for each severity level. In \cref{fig:cifar100c_corruption_severity}, we depict the gain in test-set accuracy achieved by the top-k and bottom-k (K=10) classes as the ensemble size ($m$) increases \textit{relative} to a single model. We see that, consistent with earlier results, gains on top-k plateau earlier as the size of the ensemble increases. However, {\em the benefits of homogeneous ensembles are even more pronounced when the data is increasingly corrupted}. We observe in \cref{fig:Cifar100C_AllArch} that the largest fairness benefits occur with the maximum severity, with a maximum relative gain of 40.17\% for severity 5 vs 20.18\% for severity 1.

\begin{figure*}[t!]
    \centering
    \begin{minipage}{0.49\linewidth}
    \centering
    \underline{CIFAR100}
    \includegraphics[width=\linewidth]{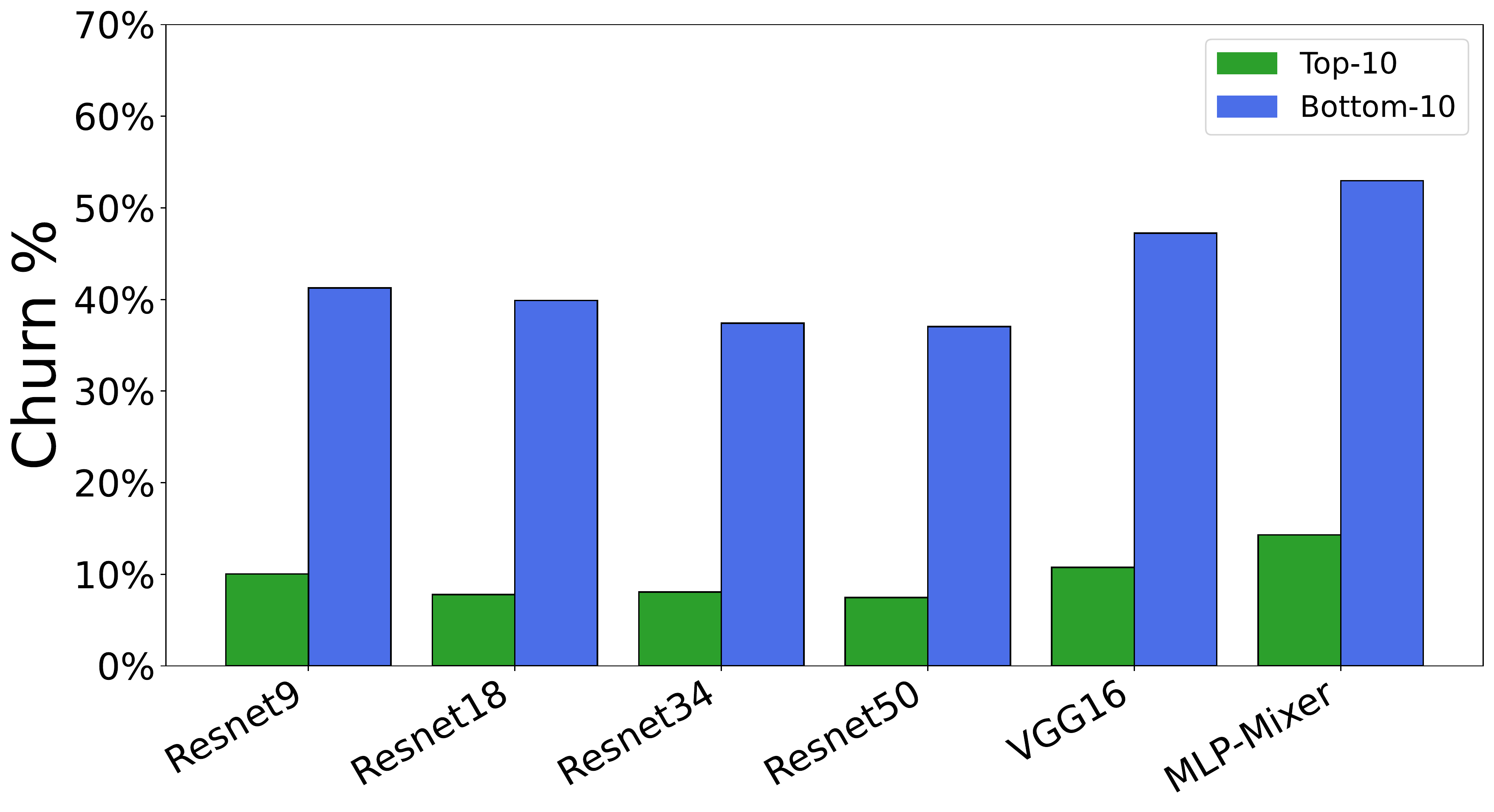}
    \end{minipage}
    \begin{minipage}{0.49\linewidth}
    \centering
    \underline{TinyImageNet}
    \includegraphics[width=\linewidth]{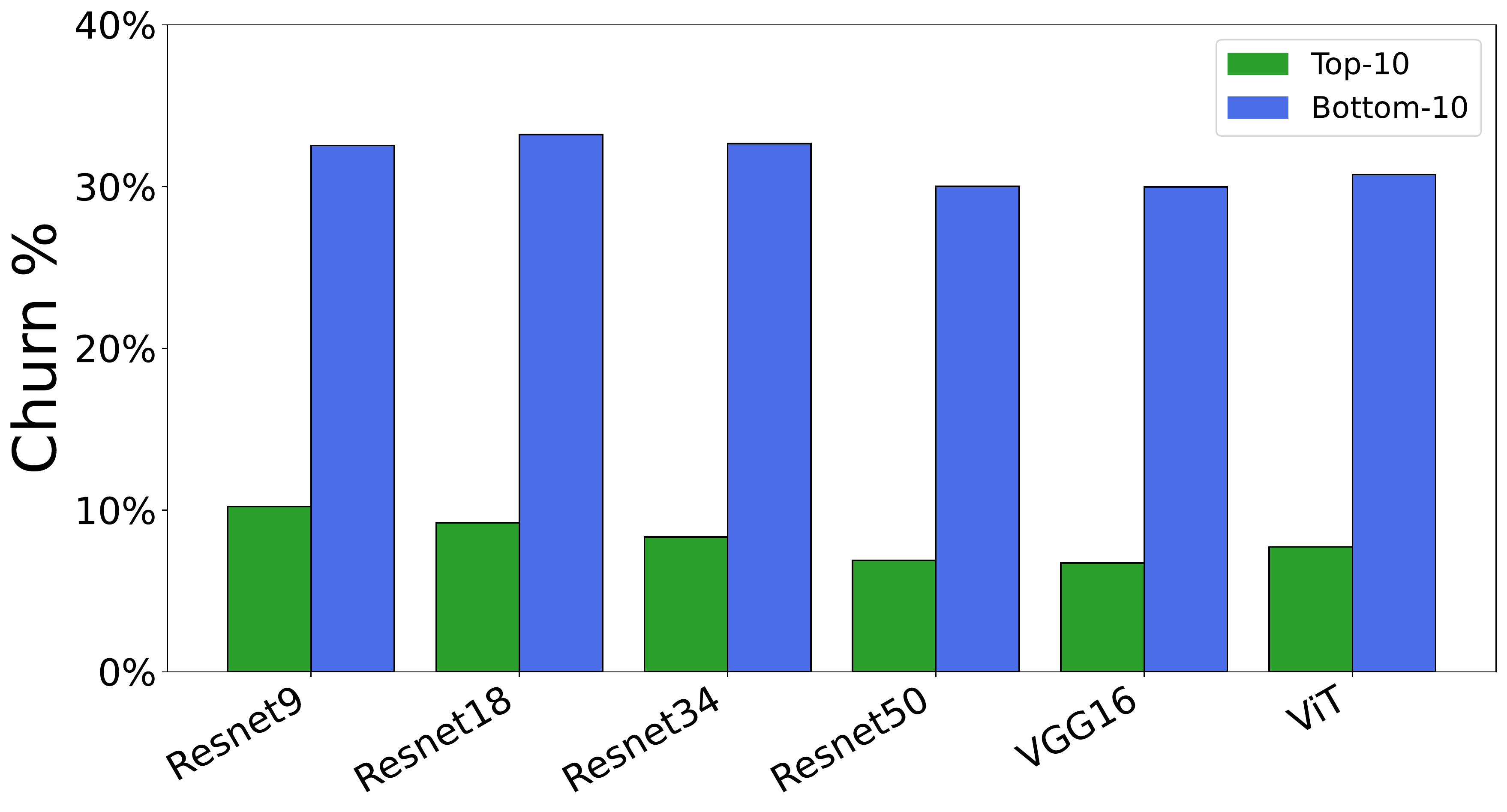}
    \end{minipage}\\
	\vspace{-0.3cm}
    \caption{\small Depiction of churn results across models and datasets. The results demonstrate that churn is significantly higher for the {\color{blue}bottom-k} group compared to the {\color{green}top-k} group, indicating that ensembling these models disproportionately impacts the {\color{blue}bottom-k} group (as defined in \cref{eq:churn}). The difference in churn between {\color{blue}bottom-k} and {\color{green}top-k} groups varies based on model architecture, suggesting that some homogeneous ensembles achieve more fairness than others.
    }
    \label{fig:churn}
    \vspace{-0.6cm}
\end{figure*}

\vspace{-0.2cm}
\section{Why Homogeneous Ensembles Improve Fairness}
\label{sec:ensemble_bottom_why}
\vspace{-0.2cm}

We established in the previous \cref{sec:fair_ensemble} that homogeneous ensembles overly benefit minority sub-group performance. However, it is still unclear why. In this section, we take a step towards understanding that effect through the scope of model disagreement, and in particular how the only three sources of stochasticity in homogeneous ensemble may impact those results.

\vspace{-0.2cm}
\subsection{Difference in Churn Between Models Explains Ensemble Fairness} \label{sec:churn}
\vspace{-0.2cm}

It might not be clear a priori how to explain the disparate impact of homogeneous deep ensembling in bottom-k groups compared to top-k groups, as we observed in the previous \cref{sec:fair_ensemble}, however we do know that such benefit only appears if the individual models do not all predict the same class, i.e., there is disagreement between models. One popular metric of model disagreement known as the {\em churn} will provide us with an obvious yet quantifiable answer.

{\bf Experiment set-up.}~To understand the benefit of model ensembling one has to recall that if all the models within the ensemble agree, then there will not be any benefit to aggregating the individual predictions. Hence, model disagreement is a key metric that will explain the stark change in performance that our homogeneous DNN ensembles have shown on the bottom-k group. We consider differences in churn between top-k and bottom-k. We also recall that the predictive churn is a measure of predictive divergence between two models. There are several different proposed definitions of predictive churn \citep{chen2020point,Shamir2020,Snapp2021}; we will employ the one that is defined on two models $f_1$ and $f_2$ as done by \citep{NIPS2016_dc5c768b} as the fraction of test examples for which the two models disagree:
\begin{align}
C(f_1, f_2) &= \mathbb{E}_{\mathcal{X}}\big[\mathbbm{1}_{\{\hat{\mathcal{Y}}_{x; f_1} \neq \hat{\mathcal{Y}}_{x; f_2}\}}\big],\label{eq:churn}
\end{align}
where $\mathbbm{1}$ is the indicator. For an ensemble with more than two models, we will report the average churn computed across $100$ randomly sampled (without replacement) pairs of models. As a further motivation to employ \cref{eq:churn}, we provide in \cref{fig:churn_correlation} the strong correlation between Churn(\%) and Test accuracy improvement(\%) for various architectures on both CIFAR100 and TinyImageNet. In fact, the Pearson correlation coefficient (a maximum score of 1 indicates perfect positive correlation) between churn and test set accuracy are $0.975$ for CIFAR100 and $0.93$ for TinyImageNet i.e., a greater value for \cref{eq:churn} is an informative proxy on the impact toward test set accuracy.

{\bf Observations.}~ In \cref{fig:churn}, we report churn for various architectures on CIFAR100 and TinyImagenet. We observe that architectures differ in the overall level of churn, but a consistent observation across architectures emerges: there are large gaps in the level of churn between top-k and bottom-k. For example, on ResNet18 for TinyImageNet the difference is churn of 9.22\% and 33.21\% for top-k and bottom-k respectively, while it is 7.78\% and 39.89\% for top-k and bottom-k for CIFAR100. In short, the models disagree much more when looking at samples belonging to the bottom-k groups than when looking at samples belonging to the top-k groups. In fact, when looking at the samples of the bottom-k classes, the models vary in which samples are incorrectly classified (by definition of churn, please see \cref{eq:churn}). As a result, that group benefits much more from homogeneous ensembling. From these observations, it becomes clear that poor performance from individual models on the bottom-k subgroups does not stem from a systemic failure and can thus be overcome through homogeneous ensembling.

\begin{figure*}[ht]
    \centering
    \begin{minipage}{0.01\linewidth}
    \rotatebox{90}{\small test accuracy \%}
    \end{minipage}
    \begin{minipage}{0.24\linewidth}
    \centering \small
     \strut    \subcaption{Batch Order} 

        \includegraphics[width=\linewidth]{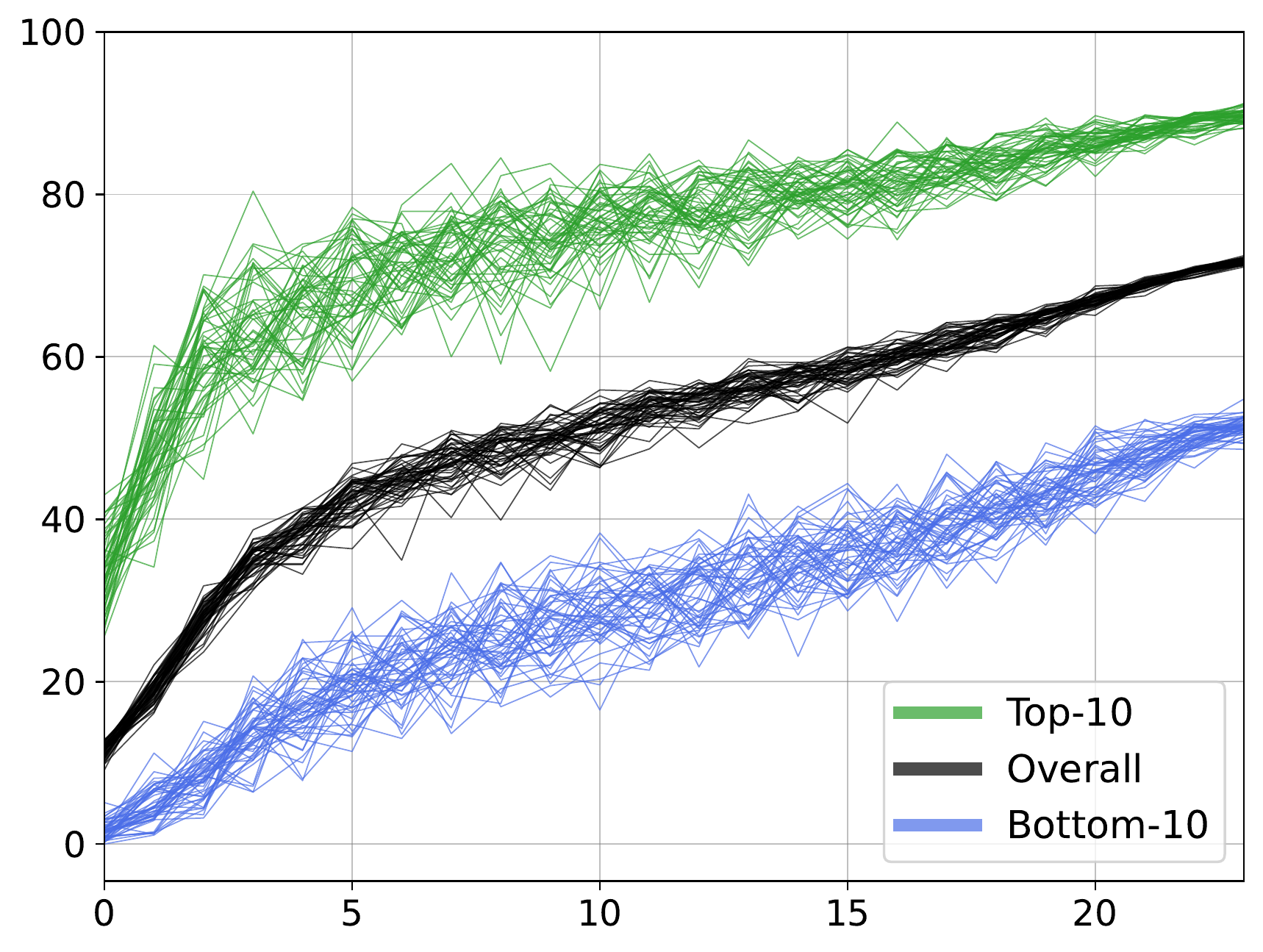}\\[-0.7em]
        {\small epochs}
    \end{minipage}
    \begin{minipage}{0.24\linewidth}
    \centering \small
     \strut    \subcaption{Data-Augmentation} 

        \includegraphics[width=\linewidth]{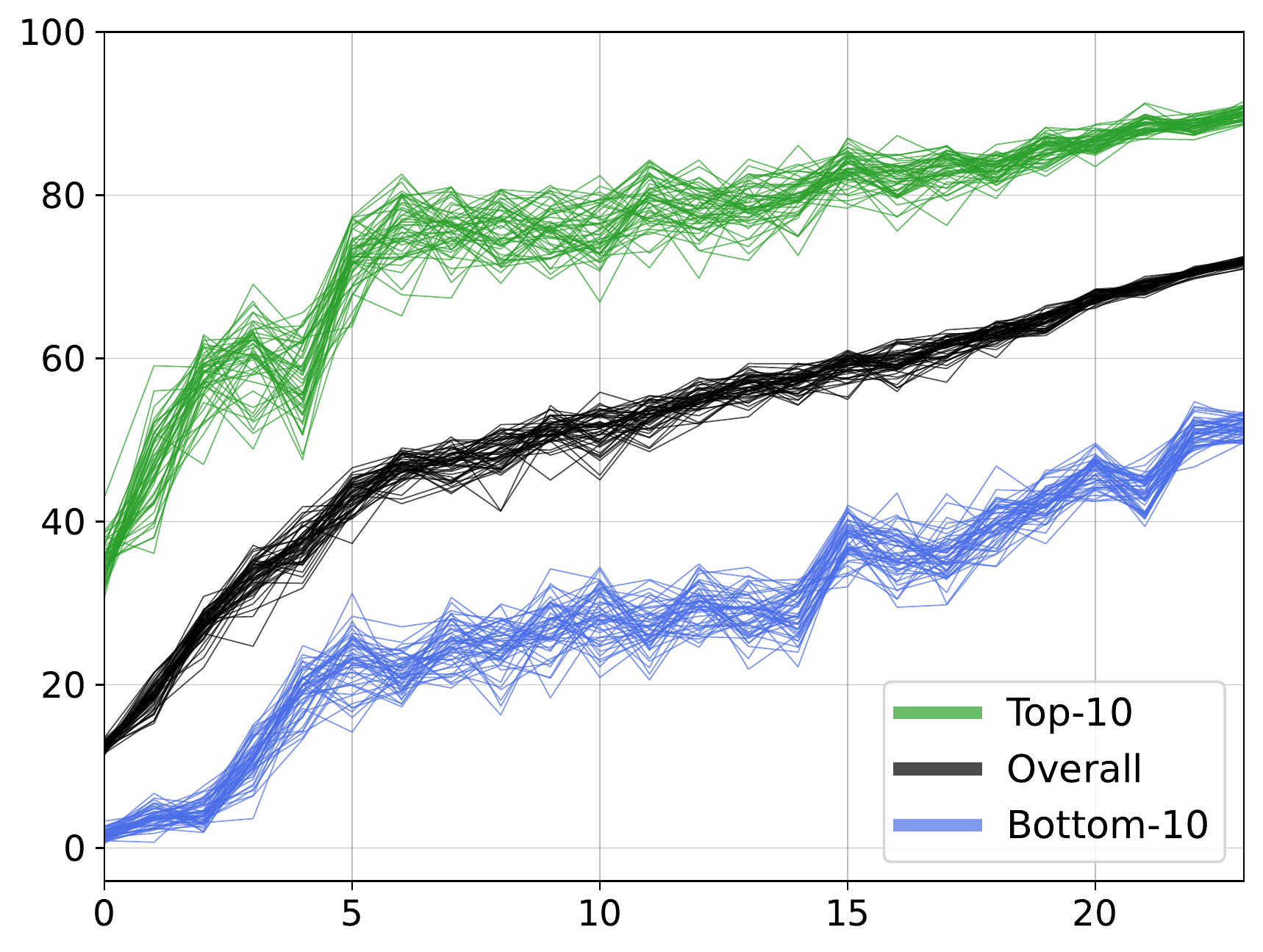}\\[-0.7em]
        {\small epochs}
    \end{minipage}
    \begin{minipage}{0.24\linewidth}
    \centering \small
        \strut      \subcaption{Batch Order}
        \includegraphics[width=\linewidth]{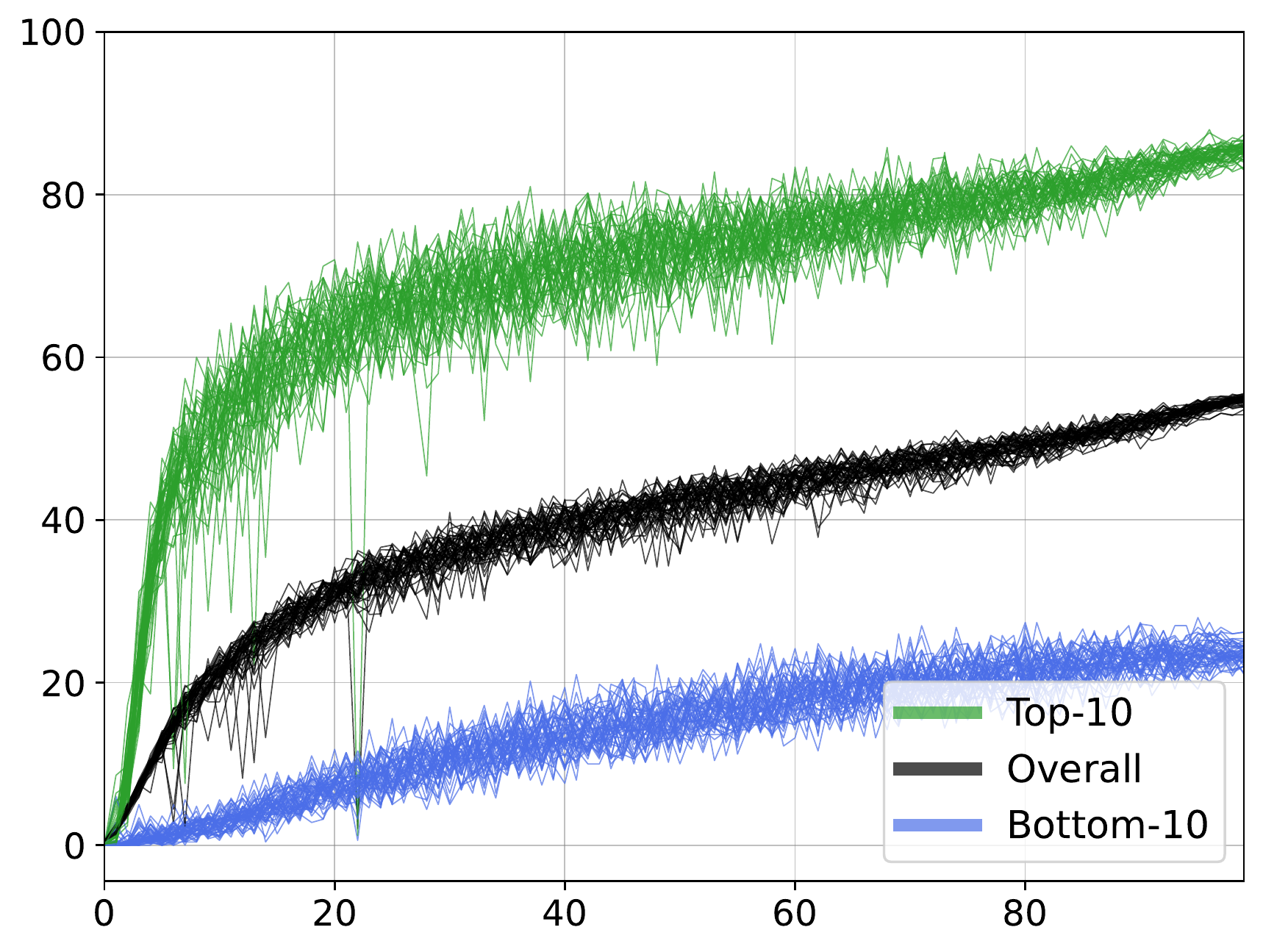}\\[-0.7em]
        {\small epochs}
    \end{minipage}
    \begin{minipage}{0.24\linewidth}
    \centering \small
     \strut    \subcaption{Model Initialization} 
        \includegraphics[width=\linewidth]{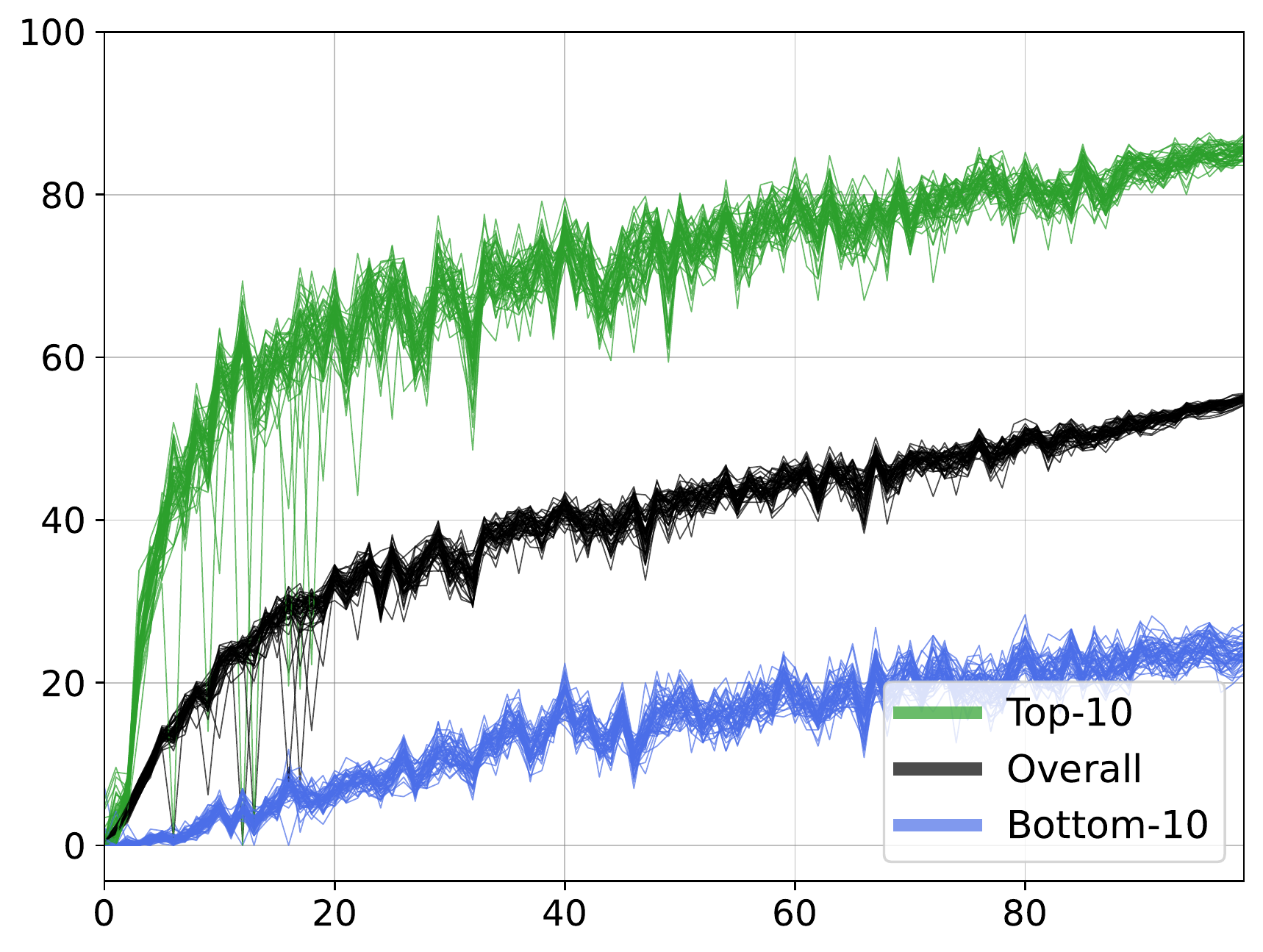}\\[-0.7em]
        {\small epochs}
    \end{minipage}\\[0em]
    \begin{subfigure}{0.24\linewidth}
		\centering
    	\includegraphics[width=1.04\linewidth]{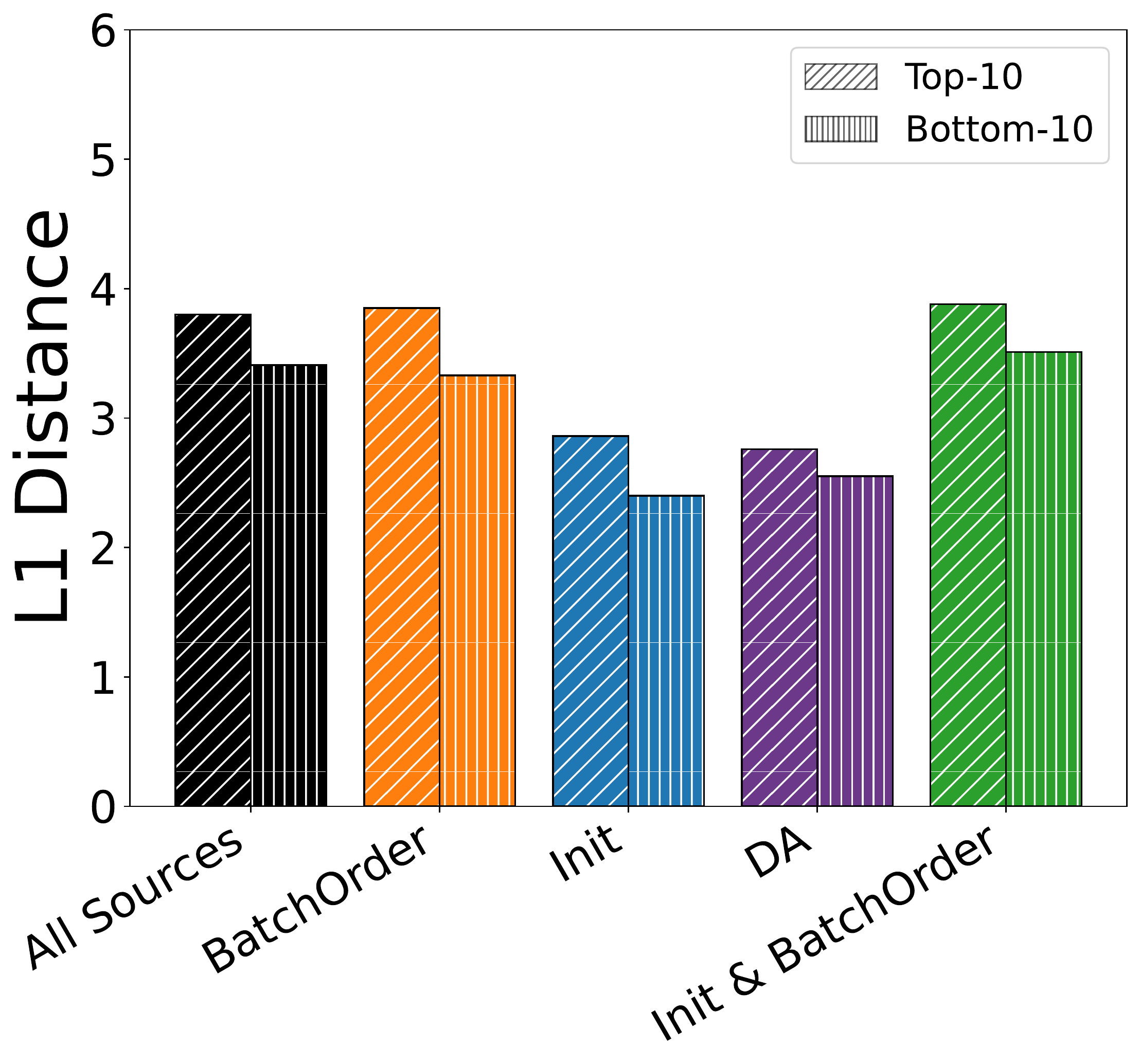}
        \label{fig:Cifar100_resnet9_stochastic_metrics_L2Distance}
	\end{subfigure}
 	\begin{subfigure}{0.24\linewidth}
		\centering
    	\includegraphics[width=1.04\linewidth]{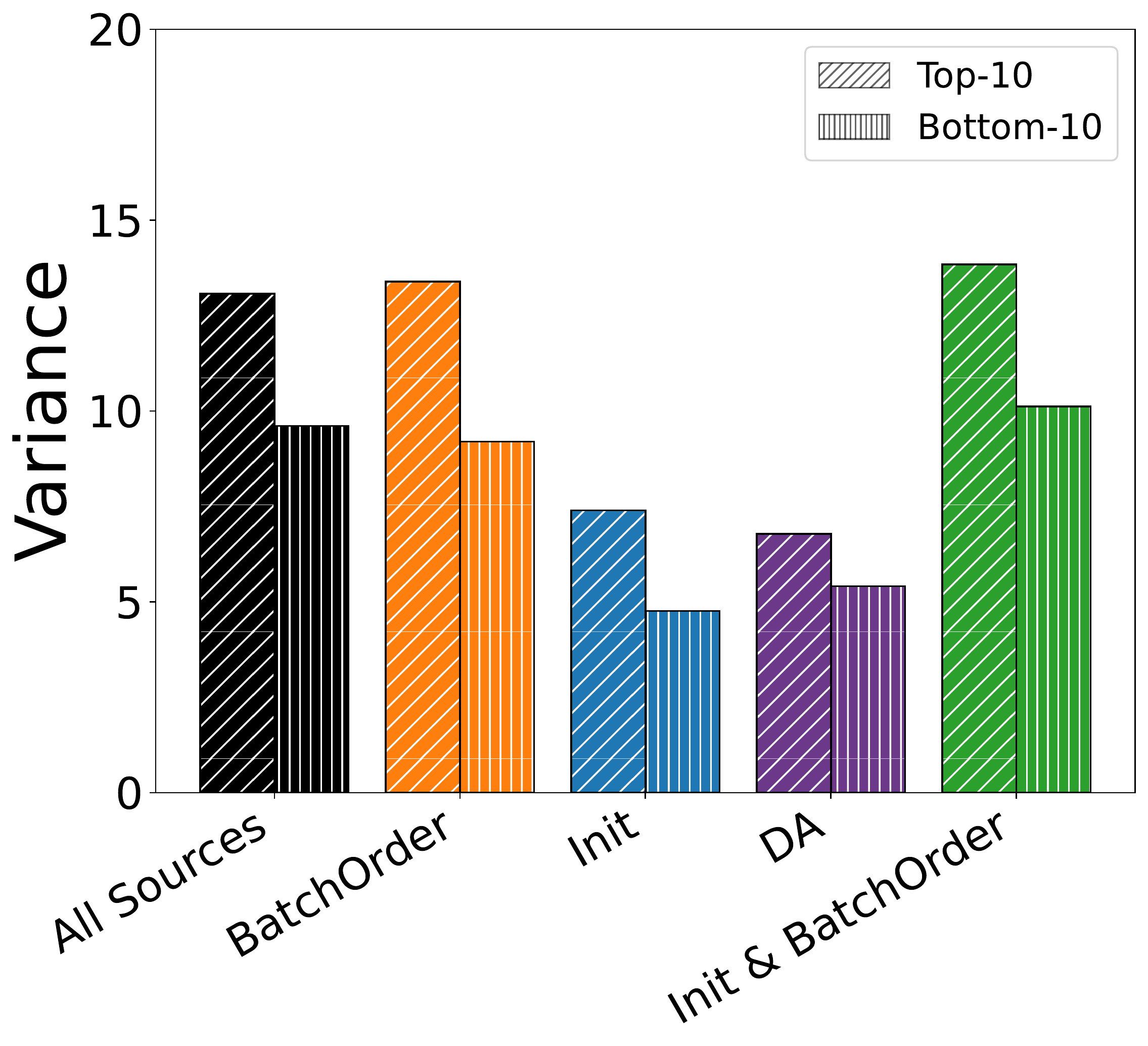}
        \label{fig:Cifar100_resnet9_stochastic_metrics_VARIANCE}
	\end{subfigure}
 	\begin{subfigure}{0.24\linewidth}
		\centering
    	\includegraphics[width=1.04\linewidth]{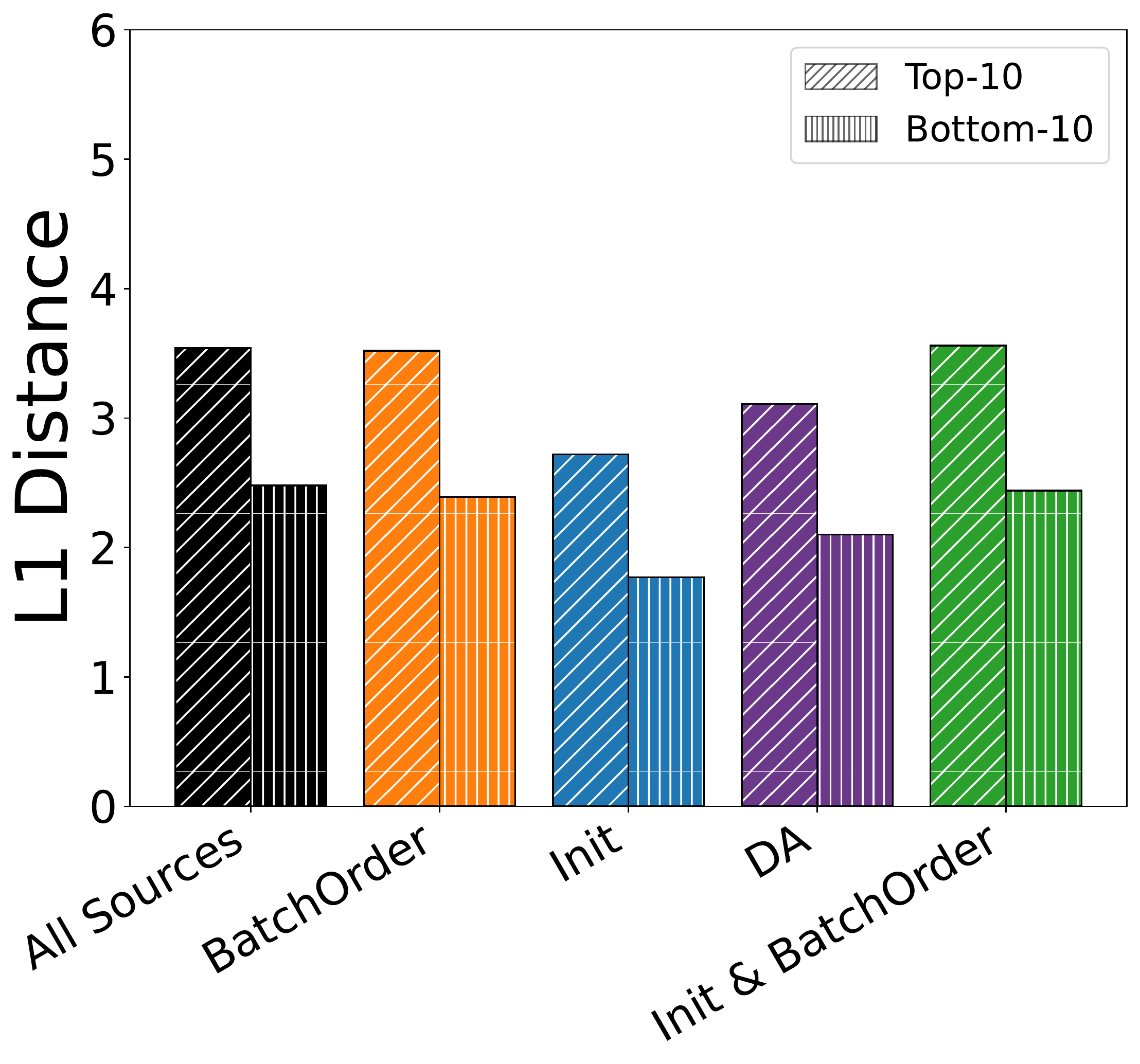}
        \label{fig:Tinyimagenet_resnet50_stochastic_metrics_L2Distance}
	\end{subfigure}
 	\begin{subfigure}{0.24\linewidth}
		\centering
    	\includegraphics[width=1.04\linewidth]{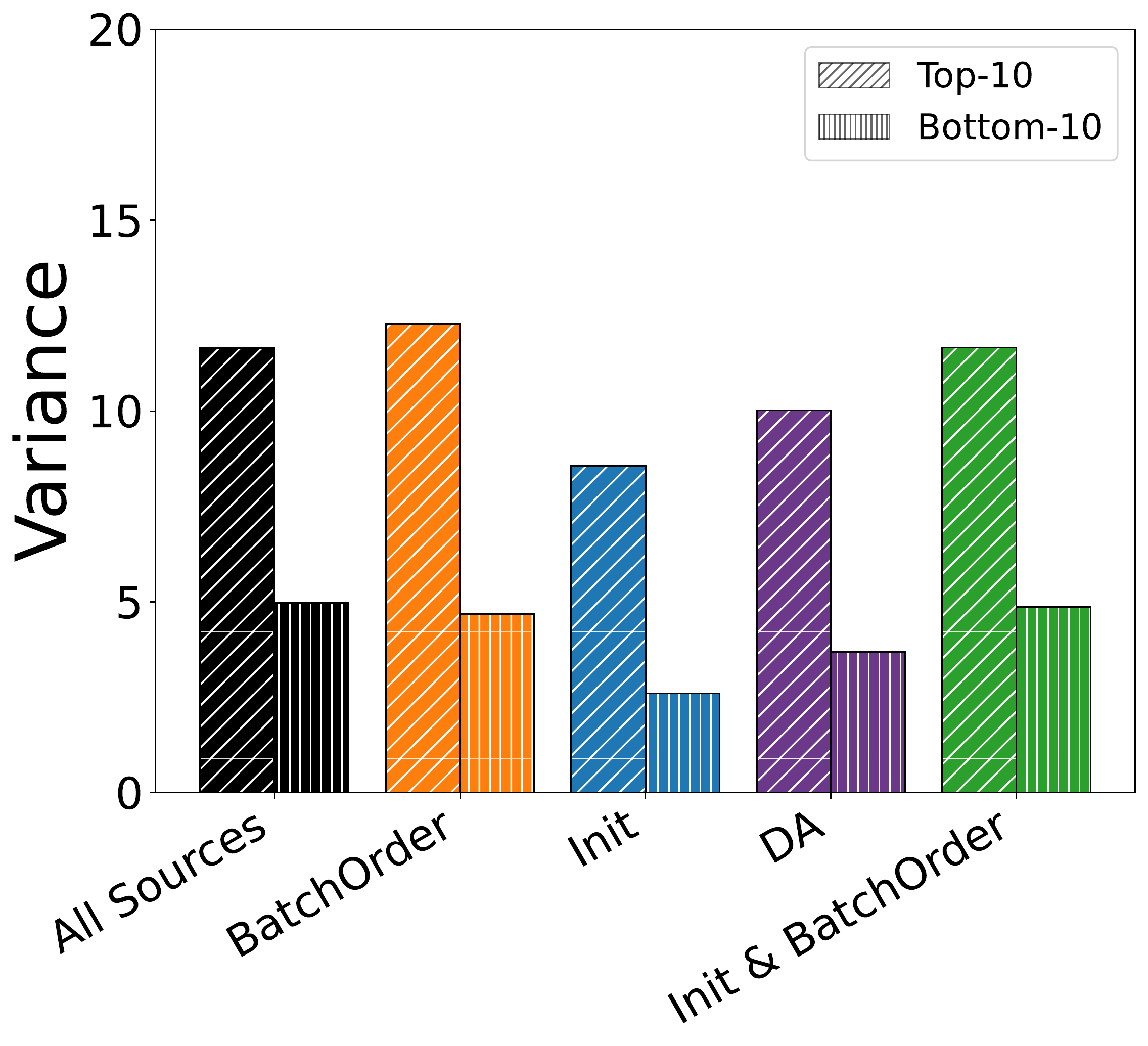}
        \label{fig:Tinyimagenet_resnet50_stochastic_metrics_VARIANCE}
	\end{subfigure}
	\vspace{-0.7cm}
    \caption{\small Depiction of multiple individual training episodes of a ResNet9 model on CIFAR100 ({\bf two left columns}) and ResNet50 model on TinyImageNet ({\bf two right columns}).  We clearly observe that varying one factor of stochasticity at a time highlights which ones provide the most randomness between training episodes. In this setting, we see that batch ordering is the main source. On the other hand, model-init and data-augmentation have little effect and we even observe very similar trends at different epochs between the individual runs.}
    \label{fig:training_dynamics}
\end{figure*}

\vspace{-0.2cm}
\subsection{Characterizing Stochasticity In Deep Neural Networks Training}
\label{sec:stochasticity}
\vspace{-0.2cm}

While \cref{sec:fair_ensemble} demonstrated the fairness benefits of homogeneous ensembles, and \cref{sec:churn} linked those improvements to increased disagreement between the individual models for the minority group and bottom-k classes, one question remains unanswered: what drives models trained with the same hyperparameters, optimizers, architectures, and training data to end-up disagreeing? This is what we propose to answer in this section by controlling each of the possible sources of randomness that impact training of the individual models.

\begin{figure*}[t!]
	\centering
    \begin{minipage}{0.01\linewidth}
        \rotatebox{90}{\hspace{1cm} \% accuracy diff.}
    \end{minipage}
    \hspace{0.1cm}
    \begin{minipage}{0.97\linewidth}
	\begin{subfigure}{0.32\linewidth}
		\centering
    	\includegraphics[width=1.0\linewidth]{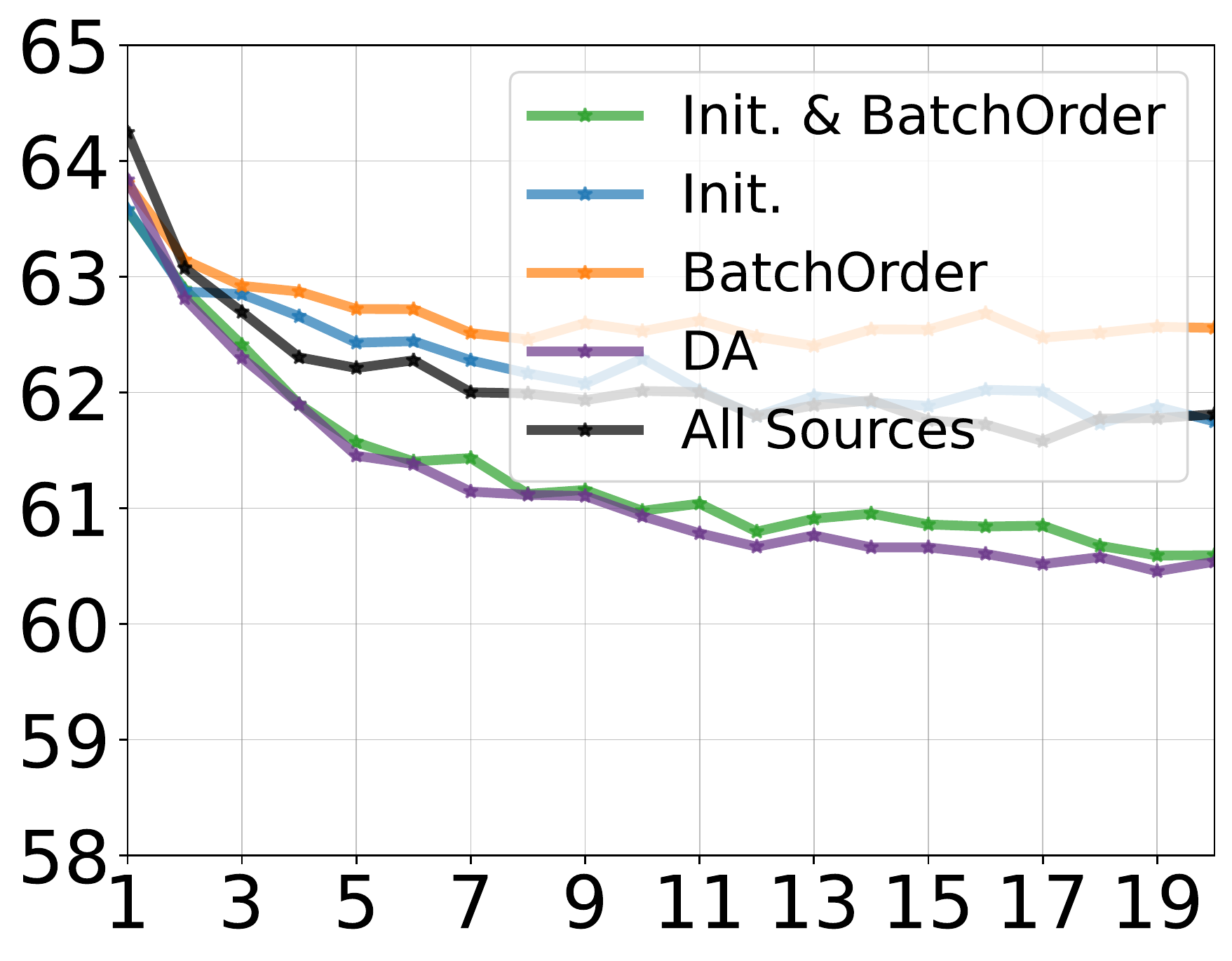}\\[-0.7em]
    	models in ensemble
        \label{fig:Tinyimagenet_res1_20_diff}
	\end{subfigure}
 	\begin{subfigure}{0.32\linewidth}
		\centering
    	\includegraphics[width=1.0\linewidth]{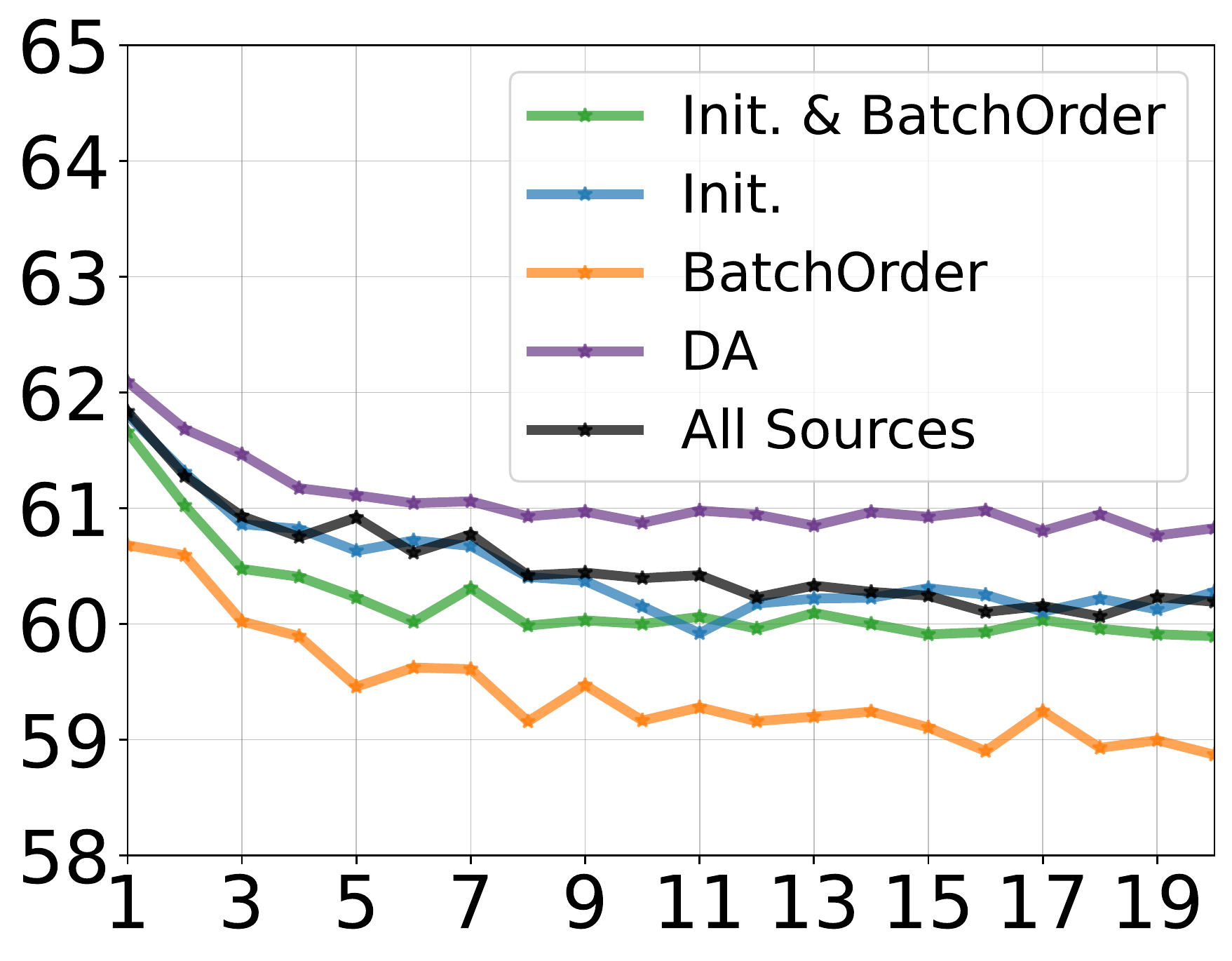}\\[-0.7em]
    	models in ensemble
        \label{fig:Tinyimagenet_res34_20_diff}
	\end{subfigure}
 	\begin{subfigure}{0.32\linewidth}
		\centering
    	\includegraphics[width=1.0\linewidth]{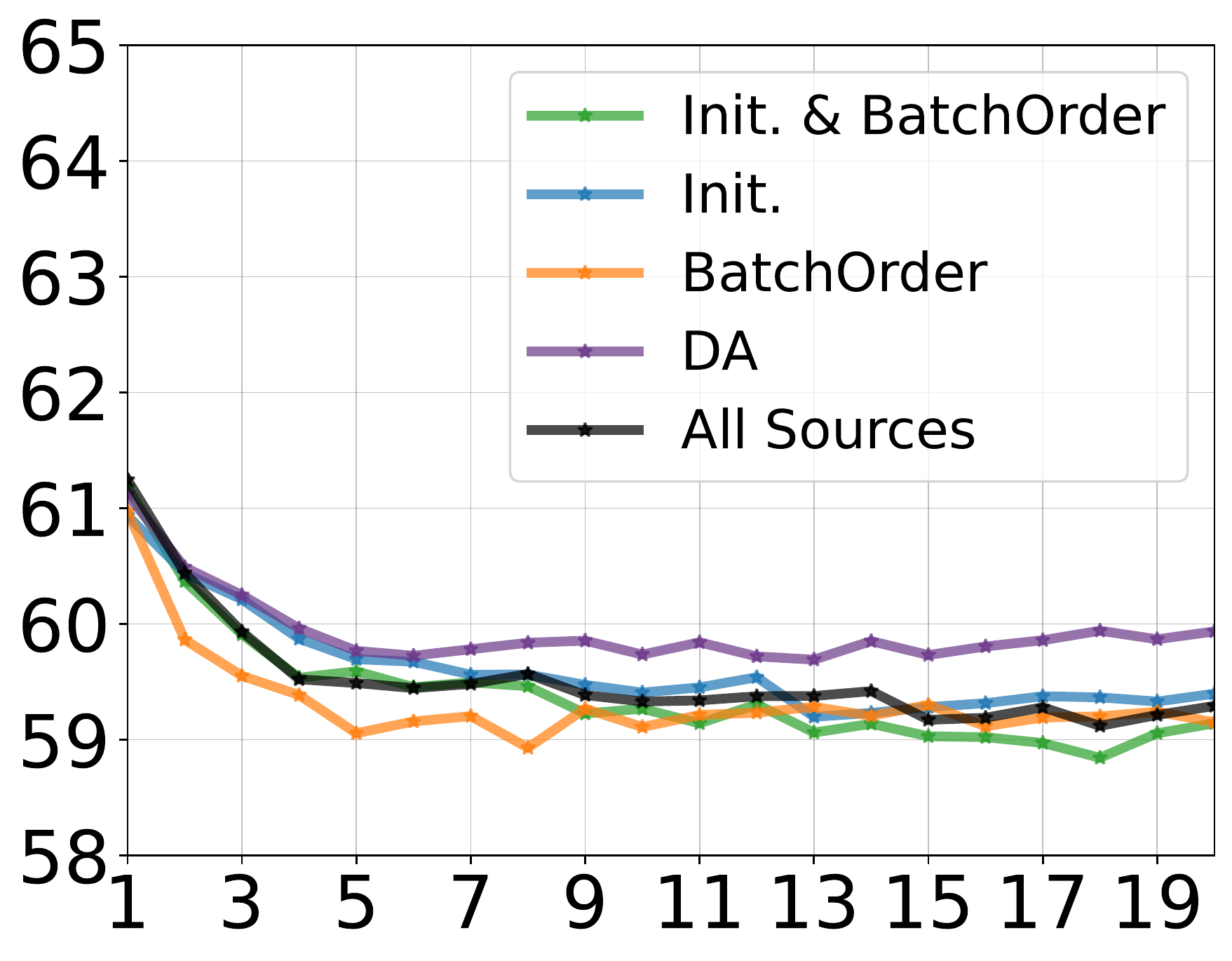}\\[-0.7em]
    	models in ensemble
        \label{fig:Tinyimagenet_res50_20_diff}
	\end{subfigure}
 \end{minipage}
	\vspace{-0.2cm}
	\caption{\small Accuracy \% difference between top and bottom 10 classes, for ResNet18 ({\bf left}), 34 ({\bf middle}), and 50 ({\bf right}) for TinyImageNet. We clearly observe that once we control for the different sources of stochasticity, it is possible to skew the ensemble to favor the bottom group, in which case fairness is further amplified compared to the baseline ensemble. Although the trends seem mostly consistent across architectures of the same family (ResNets) and datasets, it is not necessarily the case between architecture families: see \cref{fig:resnet_family_diff_c100} for ResNet-CIFAR100, \cref{fig:training_dynamics_extra} for MLPMixer/VGG16-CIFAR100 andViT/VGG16-TinyImagenet.
	}
	\label{fig:resnet_family_diff}
\end{figure*}

 To understand more what introduces the most significant levels of stochasticity, we first explore how different sources of randomness impact the training trajectories of DNNs. In particular, for homogeneous ensembles there are only three source of randomness: (i) \textit{Random Initialization} \citep{pmlr-v9-glorot10a,resnet}, (ii) \textit{Data augmentation} realizations \citep{kukacka2017regularization,Hern_ndez_Garc_a_2018}, and (iii) \textit{Data shuffling and ordering} \citep{smith2018dont,shumailov2021manipulating}. Clearly, if a source introduces low randomness, different training episodes will produce models with low disagreement and thus low fairness benefits. 

{\bf Experiment set-up.}~To isolate the impact of the different sources of stochasticity, we propose an thorough ablation study of the following sources:
{\em Change Model Initialization} (\texttt{Init}): for this ablation, we change the model initialization weights by changing the torch seed for each model before the model is instantiated.
{\em Change Batch Ordering} (\texttt{BatchOrder}): for this ablation, we change the ordering of image data in each mini-batch by changing the seed for the dataloader for each model training.
{\em Change Model Initialization and Batch Ordering} (\texttt{Init \& BatchOrder}): for this ablation, both the model initialization and batch ordering are changed for each model training.
{\em Change Data Augmentation} (\texttt{DA}): for this ablation, only the randomness in the data augmentation (e.g. probability of random flips, probability of CutMix\citep{yun2019cutmix}, etc.) is changed. The relevant torch and numpy seeds are changed right before instantiating the data augmentation pipeline. Custom fixed-seed data augmentations is also used.
{\em Change Model Initialization, Batch Ordering and Data Augmentation} (\texttt{All Sources}): for this ablation, the model initialization, batch ordering and data augmentation seeds are changed for each model training--this ablation represents the standard homogeneous ensemble of \cref{sec:fair_ensemble}.
A last source of randomness can emerge from hardware or software choices and round-off errors \citep{MLSYS2022_757b505c,shallue2019measuring} which we found to be negligible compared to the others. In addition to providing training curves evolution for each ablation, we also use two quantitative metrics. First, we will leverage the \texttt{L1-Distance} of the accuracy trajectories during training, which is calculated for every epoch by averaging the absolute distance in accuracy among the ensemble members and averaging these values across the training epochs. Second, we will leverage the \texttt{Variance} of the different training episodes' accuracy at each epoch and then average over all the epochs.

{\bf Observations.}~In \cref{fig:training_dynamics}, we plot these measures of stochasticity for both CIFAR100 and TinyImageNet on different DNNs. We observe that the single sources of noise dominate, such that the ablations themselves equate to the level of noise in the DNN with all sources of noise present. In particular, we observe one striking phenomenon: the variation of the data ordering within each epoch between training trajectories \texttt{BatchOrder} is the main source of randomness. It is equivalent to the level of noise we observe for the DNN with all sources of noise \texttt{All Sources}, and the DNN with the ablation \texttt{Init \& BatchOrder}. As seen in \cref{fig:training_dynamics} when the batch ordering is kept the same across training episodes, varying the data-augmentation and/or the model initialization has very little impact.

\begin{figure*}[ht!]
    \centering
    \begin{minipage}{0.65\linewidth}
    \begin{minipage}{0.49\linewidth}
    \centering
    \underline{\small CelebA : Blonde}
    \includegraphics[width=\linewidth]{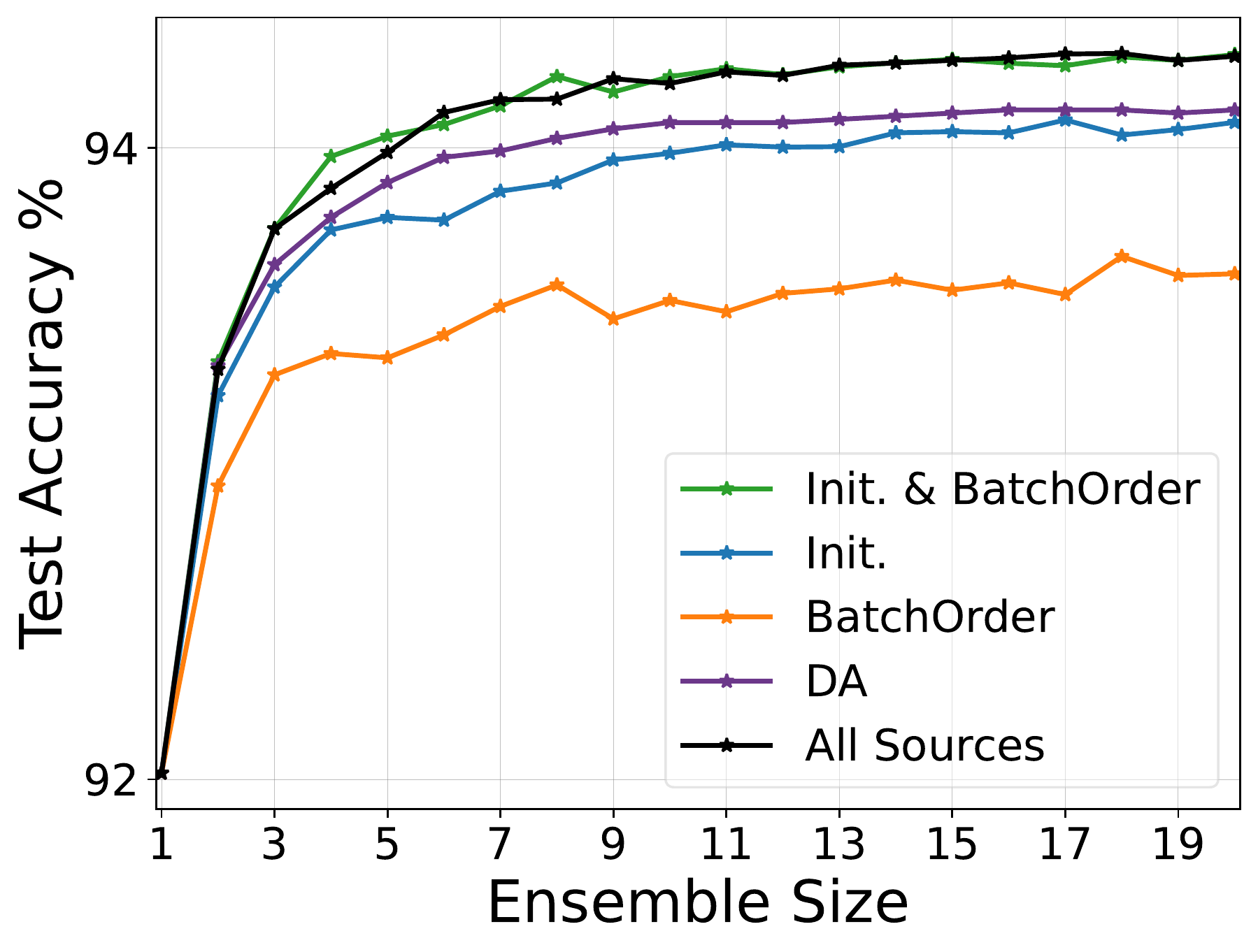}
    \end{minipage}
    \begin{minipage}{0.49\linewidth}
    \centering
    \underline{\small CelebA : Blonde Male}
    \includegraphics[width=\linewidth]{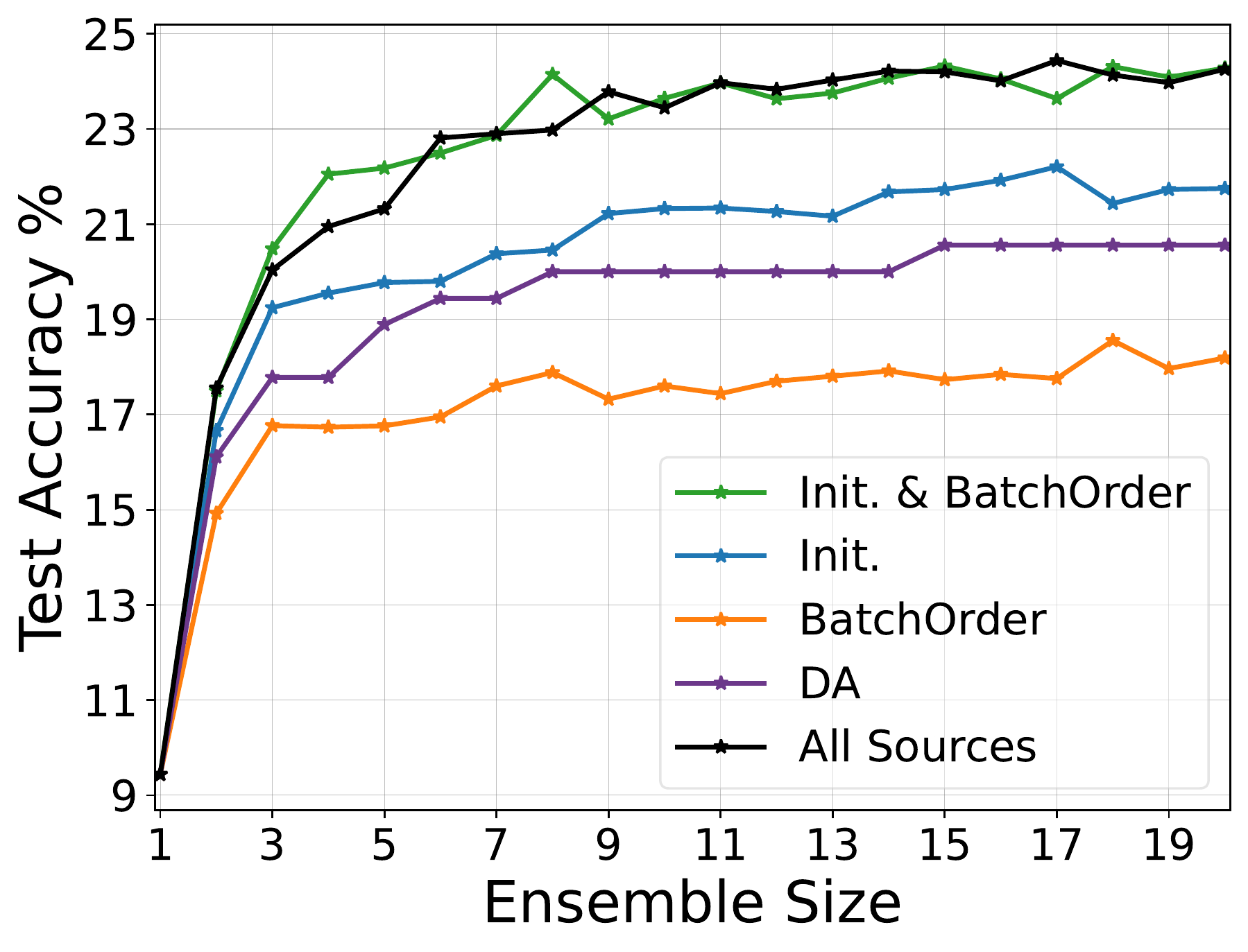}
    \end{minipage}
    \end{minipage}
    \begin{minipage}{0.34\linewidth}
    \caption{\small 20 Model Ensemble Performance (ResNet18) on Blond ({\bf left}) and Blond Male ({\bf right}) Classification in CelebA. In the overall blond classification, the max accuracy increased up to 2.27\% using homogeneous ensembling, whereas for the Blond Male minority group the max accuracy increased dramatically--up to 14.99\%--in the same ensembles.}\label{fig:celeba}
    \end{minipage}
    \vspace{-0.7cm}
\end{figure*}

\vspace{-0.2cm}
\subsection{Can Different Sources of Stochasticity Improve Homogeneous Deep Ensemble Fairness?} \label{sec:further}
\vspace{-0.2cm}

The last important point that needs to be addressed is to relate the amount of randomness that each of the three sources introduce (recall \cref{sec:stochasticity}) with the actual fairness benefits of the homogeneous ensemble. In fact, \cref{fig:training_dynamics} did emphasize how each source of randomness provides different training dynamics and levels of disagreements, which are the cause of the final fairness outcomes.

In \cref{fig:resnet_family_diff}, we depict the \textbf{accuracy difference} between average top-k and bottom-k. A value of $0$ indicates that the model performs equally on both top-k and bottom-k classes. We observe that for the majority of dataset/architecture combinations, batch ordering minimizes the gap between top and bottom-k class accuracy. Surprisingly, the resulting fairness level is even greater than when employing all the source of stochasticity, i.e., {\em it is possible to further improve the emergence of fairness in homogeneous ensembles solely by varying the batch ordering between the individual models}. In \cref{fig:celeba}, we observe that although gains quickly plateau for the Blonde category in all sources, the stochasticity introduced by initialization and batch ordering \texttt{Init \& BatchOrder} matches, and sometimes outperforms the noise ablation on the minority group performance. There is one exception to this, as we see that data-augmentation variation for ResNet18 on TinyImageNet creates the largest decrease. This observation is aligned with prior studies which compared the variability of a learned representation as a function of the different sources of stochasticity present during training. 

In Appendix B. of \citep{fort2019deep}, the authors note that at higher learning rates, mini-batch shuffling adds more randomness than model initialization due to gradient noise. Since our experiments for CIFAR-100 and TinyImageNet use higher learning rates, this is in line with the observations from \citep{fort2019deep}. Additionally, we also perform an ablation on learning rates \cref{fig:TinyImagenet_lr_wd} in Appendix H where one can clearly see the impact of different hyperparameters onto the final conclusions. There are also several works to-date that have considered how stochasticity can impact top-line metrics \citep{impactofnondeterminismrl}. Most relevant to our work is \citep{qian2021my,MLSYS2022_757b505c,madhyastha2019model,summers2021nondeterminism} that evaluates how stochasticity in training impacts fairness in DNN Systems. However, all the existing works have restricted their treatment to a single model setting, and do not evaluate the impact of ensembling.

\vspace{-0.3cm}
\section{Conclusion and Future Work} 
\vspace{-0.2cm}

In this work, we establish that while ensembling DNNs is often seen as a method of improving average performance, it can also provide significant fairness gains--even when the apparent diversity of the individual models is limited, e.g., only varying through the batch ordering or parameter initialization. Our method does not need fine-grained label information about the root cause of unfair predictions. Regardless of the actual attribute that may be the source or recipient of unfair performances, FAIR-ensembles will improve the bottom group performance. This suggests that homogeneous ensembles are a powerful tool to improve fairness outcomes in sensitive domains where human welfare is at risk, as long as the number of employed models is pushed further even after the average performance plateaus (recall \cref{sec:fair_ensemble}). Our observations led us to precisely understand the cause for the fairness emergence. In short, by controlling the different sources of randomness, we were not only able to measure the impact of each source onto the final ensemble diversity, but we were also able to pinpoint initialization and batch ordering as the main source of diversity. We hope that our observations will open the door to address fairness in homogeneous ensemble through a principled and carefully designed control of the sources of stochasticity in DNN training.

 {\bf Limitations: Validity on non-DNN models/non-image datasets.}~While our study focuses on image datasets and DNNs, we found that the fairness benefits of homogeneous ensembles extend beyond such settings. For example, we have conducted additional experiments on the Adult Census Income dataset\citep{misc_adult_2} using both a 3-layer multi-layer perceptron (MLP) model and a Decision Trees model. In the MLP setting, we used both race and sex as sensitive attributes. We trained on the remaining 12 features to predict income level $>$\$50k. Using the same ablations as our DNN experiments we report in \cref{tab:adult_census_mlp}: homogeneous ensembling improves the $>$\$50k Amer-Indian-Eskimo subgroup prediction performance by $2.53$\%. 
 As for the Decision Trees, we limited the max depth to be 10 and used the random state, which affects the random feature permutation, as the source of stochasticity to control. The results, shown in \cref{tab:adult_census_decision_trees}, depict improved fairness for Black and Amer-Indian-Eskimo. The $>$\$50k Other races subgroup had an outsized improvement with a $3.84$\% increase in accuracy over the base model. This motivates the need to develop novel theories explaining the fairness benefits of homogeneous ensembles, as those benefits are not limited to DNNs or image datasets.

\newpage

\bibliography{egbib}

\begin{thebibliography}{74}
\providecommand{\natexlab}[1]{#1}
\providecommand{\url}[1]{\texttt{#1}}
\expandafter\ifx\csname urlstyle\endcsname\relax
  \providecommand{\doi}[1]{doi: #1}\else
  \providecommand{\doi}{doi: \begingroup \urlstyle{rm}\Url}\fi

\bibitem[Arulkumaran et~al.(2017)Arulkumaran, Deisenroth, Brundage, and
  Bharath]{arulkumaran2017brief}
Kai Arulkumaran, Marc~Peter Deisenroth, Miles Brundage, and Anil~Anthony
  Bharath.
\newblock A brief survey of deep reinforcement learning.
\newblock \emph{arXiv preprint arXiv:1708.05866}, 2017.

\bibitem[Balestriero et~al.(2022)Balestriero, Bottou, and
  LeCun]{balestriero2022effects}
Randall Balestriero, Leon Bottou, and Yann LeCun.
\newblock The effects of regularization and data augmentation are class
  dependent.
\newblock \emph{Advances in Neural Information Processing Systems},
  35:\penalty0 37878--37891, 2022.

\bibitem[Barocas et~al.(2019)Barocas, Hardt, and
  Narayanan]{barocas-hardt-narayanan}
Solon Barocas, Moritz Hardt, and Arvind Narayanan.
\newblock \emph{Fairness and Machine Learning: Limitations and Opportunities}.
\newblock fairmlbook.org, 2019.
\newblock \url{http://www.fairmlbook.org}.

\bibitem[Basta et~al.(2019)Basta, Costa-juss{\`a}, and
  Casas]{basta-etal-2019-evaluating}
Christine Basta, Marta~R. Costa-juss{\`a}, and Noe Casas.
\newblock Evaluating the underlying gender bias in contextualized word
  embeddings.
\newblock In \emph{Proceedings of the First Workshop on Gender Bias in Natural
  Language Processing}, pp.\  33--39, Florence, Italy, August 2019. Association
  for Computational Linguistics.
\newblock \doi{10.18653/v1/W19-3805}.
\newblock URL \url{https://aclanthology.org/W19-3805}.

\bibitem[Becker \& Kohavi(1996)Becker and Kohavi]{misc_adult_2}
Barry Becker and Ronny Kohavi.
\newblock {Adult}.
\newblock UCI Machine Learning Repository, 1996.
\newblock {DOI}: https://doi.org/10.24432/C5XW20.

\bibitem[Bhaskaruni et~al.(2019)Bhaskaruni, Hu, and Lan]{8995403}
Dheeraj Bhaskaruni, Hui Hu, and Chao Lan.
\newblock Improving prediction fairness via model ensemble.
\newblock In \emph{2019 IEEE 31st International Conference on Tools with
  Artificial Intelligence (ICTAI)}, pp.\  1810--1814, 2019.
\newblock \doi{10.1109/ICTAI.2019.00273}.

\bibitem[Bolukbasi et~al.(2016)Bolukbasi, Chang, Zou, Saligrama, and
  Kalai]{bolukbasi2016man}
Tolga Bolukbasi, Kai-Wei Chang, James~Y Zou, Venkatesh Saligrama, and Adam~T
  Kalai.
\newblock Man is to computer programmer as woman is to homemaker? debiasing
  word embeddings.
\newblock \emph{Advances in neural information processing systems}, 29, 2016.

\bibitem[Breiman(2001)]{breiman2001random}
Leo Breiman.
\newblock Random forests.
\newblock \emph{Machine learning}, 45\penalty0 (1):\penalty0 5--32, 2001.

\bibitem[Buolamwini \& Gebru(2018)Buolamwini and Gebru]{buolamwini2018gender}
Joy Buolamwini and Timnit Gebru.
\newblock Gender shades: Intersectional accuracy disparities in commercial
  gender classification.
\newblock In \emph{Conference on fairness, accountability and transparency},
  pp.\  77--91, 2018.

\bibitem[Chatterjee(2020)]{Chatterjee2020Coherent}
Satrajit Chatterjee.
\newblock Coherent gradients: An approach to understanding generalization in
  gradient descent-based optimization.
\newblock \emph{arXiv preprint arXiv:2002.10657}, 2020.

\bibitem[Chen et~al.(2020)Chen, Wang, Lin, Cheng, Hong, Chi, and
  Cui]{chen2020point}
Zhe Chen, Yuyan Wang, Dong Lin, Derek~Zhiyuan Cheng, Lichan Hong, Ed~H. Chi,
  and Claire Cui.
\newblock Beyond point estimate: Inferring ensemble prediction variation from
  neuron activation strength in recommender systems, 2020.

\bibitem[Chen et~al.(2022)Chen, Zhang, Sarro, and Harman]{chen2022maat}
Zhenpeng Chen, Jie~M Zhang, Federica Sarro, and Mark Harman.
\newblock Maat: a novel ensemble approach to addressing fairness and
  performance bugs for machine learning software.
\newblock In \emph{Proceedings of the 30th ACM Joint European Software
  Engineering Conference and Symposium on the Foundations of Software
  Engineering}, pp.\  1122--1134, 2022.

\bibitem[Cooper et~al.(2023)Cooper, Lee, Choksi, Barocas, Sa, Grimmelmann,
  Kleinberg, Sen, and Zhang]{cooper2023arbitrariness}
A.~Feder Cooper, Katherine Lee, Madiha Choksi, Solon Barocas, Christopher~De
  Sa, James Grimmelmann, Jon Kleinberg, Siddhartha Sen, and Baobao Zhang.
\newblock Arbitrariness and prediction: The confounding role of variance in
  fair classification, 2023.

\bibitem[Cubuk et~al.(2018)Cubuk, Zoph, Mane, Vasudevan, and
  Le]{cubuk2018autoaugment}
Ekin~D Cubuk, Barret Zoph, Dandelion Mane, Vijay Vasudevan, and Quoc~V Le.
\newblock Autoaugment: Learning augmentation policies from data.
\newblock \emph{arXiv preprint arXiv:1805.09501}, 2018.

\bibitem[Dietterich(2000)]{dietterich2000ensemble}
Thomas~G Dietterich.
\newblock Ensemble methods in machine learning.
\newblock In \emph{International workshop on multiple classifier systems}, pp.\
   1--15. Springer, 2000.

\bibitem[Dosovitskiy et~al.(2020)Dosovitskiy, Beyer, Kolesnikov, Weissenborn,
  Zhai, Unterthiner, Dehghani, Minderer, Heigold, Gelly,
  et~al.]{dosovitskiy2020image}
Alexey Dosovitskiy, Lucas Beyer, Alexander Kolesnikov, Dirk Weissenborn,
  Xiaohua Zhai, Thomas Unterthiner, Mostafa Dehghani, Matthias Minderer, Georg
  Heigold, Sylvain Gelly, et~al.
\newblock An image is worth 16x16 words: Transformers for image recognition at
  scale.
\newblock \emph{arXiv preprint arXiv:2010.11929}, 2020.

\bibitem[Feldman \& Zhang(2020)Feldman and Zhang]{NEURIPS2020_1e14bfe2}
Vitaly Feldman and Chiyuan Zhang.
\newblock What neural networks memorize and why: Discovering the long tail via
  influence estimation.
\newblock In H.~Larochelle, M.~Ranzato, R.~Hadsell, M.~F. Balcan, and H.~Lin
  (eds.), \emph{Advances in Neural Information Processing Systems}, volume~33,
  pp.\  2881--2891. Curran Associates, Inc., 2020.
\newblock URL
  \url{https://proceedings.neurips.cc/paper/2020/file/1e14bfe2714193e7af5abc64ecbd6b46-Paper.pdf}.

\bibitem[Fort et~al.(2019)Fort, Hu, and Lakshminarayanan]{fort2019deep}
Stanislav Fort, Huiyi Hu, and Balaji Lakshminarayanan.
\newblock Deep ensembles: A loss landscape perspective.
\newblock \emph{arXiv preprint arXiv:1912.02757}, 2019.

\bibitem[Freund \& Schapire(1995)Freund and Schapire]{Freund1995ADG}
Yoav Freund and Robert~E. Schapire.
\newblock A decision-theoretic generalization of on-line learning and an
  application to boosting.
\newblock In \emph{EuroCOLT}, 1995.

\bibitem[Garg et~al.(2017)Garg, Schiebinger, Jurafsky, and Zou]{Garg2017}
Nikhil Garg, Londa Schiebinger, Dan Jurafsky, and James Zou.
\newblock Word embeddings quantify 100 years of gender and ethnic stereotypes.
\newblock \emph{Proceedings of the National Academy of Sciences}, 115, 11 2017.
\newblock \doi{10.1073/pnas.1720347115}.

\bibitem[Glorot \& Bengio(2010)Glorot and Bengio]{pmlr-v9-glorot10a}
Xavier Glorot and Yoshua Bengio.
\newblock Understanding the difficulty of training deep feedforward neural
  networks.
\newblock In Yee~Whye Teh and Mike Titterington (eds.), \emph{Proceedings of
  the Thirteenth International Conference on Artificial Intelligence and
  Statistics}, volume~9 of \emph{Proceedings of Machine Learning Research},
  pp.\  249--256, Chia Laguna Resort, Sardinia, Italy, 13--15 May 2010. PMLR.
\newblock URL \url{http://proceedings.mlr.press/v9/glorot10a.html}.

\bibitem[Gohar et~al.(2023)Gohar, Biswas, and Rajan]{gohar2023towards}
Usman Gohar, Sumon Biswas, and Hridesh Rajan.
\newblock Towards understanding fairness and its composition in ensemble
  machine learning.
\newblock In \emph{2023 IEEE/ACM 45th International Conference on Software
  Engineering (ICSE)}, pp.\  1533--1545. IEEE, 2023.

\bibitem[Grgić-Hlača et~al.(2017)Grgić-Hlača, Zafar, Gummadi, and
  Weller]{Nina2017}
Nina Grgić-Hlača, Muhammad~Bilal Zafar, Krishna~P. Gummadi, and Adrian
  Weller.
\newblock On fairness, diversity and randomness in algorithmic decision making,
  2017.
\newblock URL \url{https://arxiv.org/abs/1706.10208}.

\bibitem[Gupta et~al.(2022)Gupta, Smith, Adlam, and
  Mariet]{gupta2022ensembling}
Neha Gupta, Jamie Smith, Ben Adlam, and Zelda Mariet.
\newblock Ensembling over classifiers: a bias-variance perspective.
\newblock \emph{arXiv preprint arXiv:2206.10566}, 2022.

\bibitem[Hashimoto et~al.(2018)Hashimoto, Srivastava, Namkoong, and
  Liang]{hashimoto18a}
Tatsunori Hashimoto, Megha Srivastava, Hongseok Namkoong, and Percy Liang.
\newblock Fairness without demographics in repeated loss minimization.
\newblock In Jennifer Dy and Andreas Krause (eds.), \emph{Proceedings of the
  35th International Conference on Machine Learning}, volume~80 of
  \emph{Proceedings of Machine Learning Research}, pp.\  1929--1938,
  Stockholmsmässan, Stockholm Sweden, 10--15 Jul 2018. PMLR.
\newblock URL \url{http://proceedings.mlr.press/v80/hashimoto18a.html}.

\bibitem[He et~al.(2016{\natexlab{a}})He, Zhang, Ren, and Sun]{he2016deep}
Kaiming He, Xiangyu Zhang, Shaoqing Ren, and Jian Sun.
\newblock Deep residual learning for image recognition.
\newblock In \emph{CVPR}, 2016{\natexlab{a}}.

\bibitem[He et~al.(2016{\natexlab{b}})He, Zhang, Ren, and Sun]{resnet}
Kaiming He, Xiangyu Zhang, Shaoqing Ren, and Jian Sun.
\newblock Deep residual learning for image recognition.
\newblock In \emph{2016 {IEEE} Conference on Computer Vision and Pattern
  Recognition, {CVPR}}, 2016{\natexlab{b}}.

\bibitem[Hendrycks \& Dietterich(2018)Hendrycks and
  Dietterich]{hendrycks2018benchmarking}
Dan Hendrycks and Thomas~G Dietterich.
\newblock Benchmarking neural network robustness to common corruptions and
  surface variations.
\newblock \emph{arXiv preprint arXiv:1807.01697}, 2018.

\bibitem[Hernández-García \& König(2018)Hernández-García and
  König]{Hern_ndez_Garc_a_2018}
Alex Hernández-García and Peter König.
\newblock Further advantages of data augmentation on convolutional neural
  networks.
\newblock \emph{Lecture Notes in Computer Science}, pp.\  95–103, 2018.
\newblock ISSN 1611-3349.
\newblock \doi{10.1007/978-3-030-01418-6_10}.
\newblock URL \url{http://dx.doi.org/10.1007/978-3-030-01418-6_10}.

\bibitem[Hinton et~al.(2012)Hinton, Deng, Yu, Dahl, Mohamed, Jaitly, Senior,
  Vanhoucke, Nguyen, Sainath, et~al.]{hinton2012deep}
Geoffrey Hinton, Li~Deng, Dong Yu, George~E Dahl, Abdel-rahman Mohamed, Navdeep
  Jaitly, Andrew Senior, Vincent Vanhoucke, Patrick Nguyen, Tara~N Sainath,
  et~al.
\newblock Deep neural networks for acoustic modeling in speech recognition: The
  shared views of four research groups.
\newblock \emph{IEEE Signal processing magazine}, 29\penalty0 (6):\penalty0
  82--97, 2012.

\bibitem[{Hooker} et~al.(2019){Hooker}, {Courville}, {Clark}, {Dauphin}, and
  {Frome}]{hooker2019}
Sara {Hooker}, Aaron {Courville}, Gregory {Clark}, Yann {Dauphin}, and Andrea
  {Frome}.
\newblock {What Do Compressed Deep Neural Networks Forget?}
\newblock \emph{arXiv e-prints}, art. arXiv:1911.05248, November 2019.

\bibitem[Hooker et~al.(2019)Hooker, Courville, Clark, Dauphin, and
  Frome]{hooker2019compressed}
Sara Hooker, Aaron Courville, Gregory Clark, Yann Dauphin, and Andrea Frome.
\newblock What do compressed deep neural networks forget?
\newblock \emph{arXiv}, 2019.

\bibitem[Howard \& Ruder(2018)Howard and Ruder]{howard2018universal}
Jeremy Howard and Sebastian Ruder.
\newblock Universal language model fine-tuning for text classification.
\newblock \emph{arXiv preprint arXiv:1801.06146}, 2018.

\bibitem[Jain et~al.(2020)Jain, Liu, Mueller, and Gifford]{jain2020maximizing}
Siddhartha Jain, Ge~Liu, Jonas Mueller, and David Gifford.
\newblock Maximizing overall diversity for improved uncertainty estimates in
  deep ensembles.
\newblock In \emph{Proceedings of the AAAI Conference on Artificial
  Intelligence}, volume~34, pp.\  4264--4271, 2020.

\bibitem[Kenfack et~al.(2021)Kenfack, Khan, Kazmi, Hussain, Oracevic, and
  Khattak]{kenfack2021impact}
Patrik~Joslin Kenfack, Adil~Mehmood Khan, SM~Ahsan Kazmi, Rasheed Hussain, Alma
  Oracevic, and Asad~Masood Khattak.
\newblock Impact of model ensemble on the fairness of classifiers in machine
  learning.
\newblock In \emph{2021 International conference on applied artificial
  intelligence (ICAPAI)}, pp.\  1--6. IEEE, 2021.

\bibitem[Kingma \& Ba(2014)Kingma and Ba]{kingma2014adam}
Diederik~P Kingma and Jimmy Ba.
\newblock Adam: A method for stochastic optimization.
\newblock \emph{arXiv preprint arXiv:1412.6980}, 2014.

\bibitem[Kleinberg et~al.(2016)Kleinberg, Mullainathan, and
  Raghavan]{kleinberg2016}
Jon~M. Kleinberg, Sendhil Mullainathan, and Manish Raghavan.
\newblock Inherent trade-offs in the fair determination of risk scores.
\newblock \emph{CoRR}, abs/1609.05807, 2016.
\newblock URL \url{http://arxiv.org/abs/1609.05807}.

\bibitem[Krizhevsky et~al.(2009)Krizhevsky, Hinton,
  et~al.]{krizhevsky2009learning}
Alex Krizhevsky, Geoffrey Hinton, et~al.
\newblock Learning multiple layers of features from tiny images.
\newblock 2009.

\bibitem[Kukačka et~al.(2017)Kukačka, Golkov, and
  Cremers]{kukacka2017regularization}
Jan Kukačka, Vladimir Golkov, and Daniel Cremers.
\newblock Regularization for deep learning: A taxonomy, 2017.

\bibitem[{Lakshminarayanan} et~al.(2016){Lakshminarayanan}, {Pritzel}, and
  {Blundell}]{2016Lakshminarayanan}
Balaji {Lakshminarayanan}, Alexander {Pritzel}, and Charles {Blundell}.
\newblock {Simple and Scalable Predictive Uncertainty Estimation using Deep
  Ensembles}.
\newblock \emph{arXiv e-prints}, art. arXiv:1612.01474, Dec 2016.

\bibitem[Lee et~al.(2015)Lee, Purushwalkam, Cogswell, Crandall, and
  Batra]{lee2015}
Stefan Lee, Senthil Purushwalkam, Michael Cogswell, David~J. Crandall, and
  Dhruv Batra.
\newblock Why {M} heads are better than one: Training a diverse ensemble of
  deep networks.
\newblock \emph{CoRR}, abs/1511.06314, 2015.
\newblock URL \url{http://arxiv.org/abs/1511.06314}.

\bibitem[Liu et~al.(2015)Liu, Luo, Wang, and Tang]{liu2015faceattributes}
Ziwei Liu, Ping Luo, Xiaogang Wang, and Xiaoou Tang.
\newblock Deep learning face attributes in the wild.
\newblock In \emph{Proceedings of International Conference on Computer Vision
  (ICCV)}, December 2015.

\bibitem[Loshchilov \& Hutter(2016)Loshchilov and Hutter]{loshchilov2016sgdr}
Ilya Loshchilov and Frank Hutter.
\newblock Sgdr: Stochastic gradient descent with warm restarts.
\newblock \emph{arXiv preprint arXiv:1608.03983}, 2016.

\bibitem[Loshchilov \& Hutter(2017)Loshchilov and
  Hutter]{loshchilov2017decoupled}
Ilya Loshchilov and Frank Hutter.
\newblock Decoupled weight decay regularization.
\newblock \emph{arXiv preprint arXiv:1711.05101}, 2017.

\bibitem[Madhyastha \& Jain(2019)Madhyastha and Jain]{madhyastha2019model}
Pranava Madhyastha and Rishabh Jain.
\newblock On model stability as a function of random seed.
\newblock \emph{arXiv preprint arXiv:1909.10447}, 2019.

\bibitem[Milani~Fard et~al.(2016)Milani~Fard, Cormier, Canini, and
  Gupta]{NIPS2016_dc5c768b}
Mahdi Milani~Fard, Quentin Cormier, Kevin Canini, and Maya Gupta.
\newblock Launch and iterate: Reducing prediction churn.
\newblock In D.~Lee, M.~Sugiyama, U.~Luxburg, I.~Guyon, and R.~Garnett (eds.),
  \emph{Advances in Neural Information Processing Systems}, volume~29. Curran
  Associates, Inc., 2016.
\newblock URL
  \url{https://proceedings.neurips.cc/paper/2016/file/dc5c768b5dc76a084531934b34601977-Paper.pdf}.

\bibitem[Nagarajan et~al.(2018)Nagarajan, Warnell, and
  Stone]{impactofnondeterminismrl}
Prabhat Nagarajan, Garrett Warnell, and Peter Stone.
\newblock The impact of nondeterminism on reproducibility in deep reinforcement
  learning.
\newblock In \emph{Reproducibility in ML Workshop at the 35th International
  Conference on Machine Learning, {ICML}}, 2018.

\bibitem[Ogueji et~al.(2022)Ogueji, Ahia, Onilude, Gehrmann, Hooker, and
  Kreutzer]{ogueji2022}
Kelechi Ogueji, Orevaoghene Ahia, Gbemileke Onilude, Sebastian Gehrmann, Sara
  Hooker, and Julia Kreutzer.
\newblock Intriguing properties of compression on multilingual models, 2022.
\newblock URL \url{https://arxiv.org/abs/2211.02738}.

\bibitem[Opitz \& Maclin(1999)Opitz and Maclin]{Opitz_1999}
D.~Opitz and R.~Maclin.
\newblock Popular ensemble methods: An empirical study.
\newblock \emph{Journal of Artificial Intelligence Research}, 11:\penalty0
  169--198, aug 1999.
\newblock \doi{10.1613/jair.614}.
\newblock URL \url{https://doi.org/10.1613%2Fjair.614}.

\bibitem[Qian et~al.(2021)Qian, Pham, Lutellier, Hu, Kim, Tan, Yu, Chen, and
  Shah]{qian2021my}
Shangshu Qian, Viet~Hung Pham, Thibaud Lutellier, Zeou Hu, Jungwon Kim, Lin
  Tan, Yaoliang Yu, Jiahao Chen, and Sameena Shah.
\newblock Are my deep learning systems fair? an empirical study of fixed-seed
  training.
\newblock \emph{Advances in Neural Information Processing Systems},
  34:\penalty0 30211--30227, 2021.

\bibitem[Rame et~al.(2022)Rame, Kirchmeyer, Rahier, Rakotomamonjy, Gallinari,
  and Cord]{rame2022diverse}
Alexandre Rame, Matthieu Kirchmeyer, Thibaud Rahier, Alain Rakotomamonjy,
  Patrick Gallinari, and Matthieu Cord.
\newblock Diverse weight averaging for out-of-distribution generalization.
\newblock \emph{arXiv preprint arXiv:2205.09739}, 2022.

\bibitem[Russakovsky et~al.(2015)Russakovsky, Deng, Su, Krause, Satheesh, Ma,
  Huang, Karpathy, Khosla, Bernstein, et~al.]{russakovsky2015imagenet}
Olga Russakovsky, Jia Deng, Hao Su, Jonathan Krause, Sanjeev Satheesh, Sean Ma,
  Zhiheng Huang, Andrej Karpathy, Aditya Khosla, Michael Bernstein, et~al.
\newblock Imagenet large scale visual recognition challenge.
\newblock In \emph{IJCV}, 2015.

\bibitem[Shallue et~al.(2019)Shallue, Lee, Antognini, Sohl-Dickstein, Frostig,
  and Dahl]{shallue2019measuring}
Christopher~J. Shallue, Jaehoon Lee, Joseph Antognini, Jascha Sohl-Dickstein,
  Roy Frostig, and George~E. Dahl.
\newblock Measuring the effects of data parallelism on neural network training,
  2019.

\bibitem[Shamir \& Coviello(2020)Shamir and Coviello]{Shamir2020}
Gil~I. Shamir and Lorenzo Coviello.
\newblock Anti-distillation: Improving reproducibility of deep networks.
\newblock \emph{CoRR}, abs/2010.09923, 2020.
\newblock URL \url{https://arxiv.org/abs/2010.09923}.

\bibitem[Shankar et~al.(2017)Shankar, Halpern, Breck, Atwood, Wilson, and
  Sculley]{shankar2017no}
Shreya Shankar, Yoni Halpern, Eric Breck, James Atwood, Jimbo Wilson, and
  D~Sculley.
\newblock No classification without representation: Assessing geodiversity
  issues in open data sets for the developing world.
\newblock \emph{arXiv preprint arXiv:1711.08536}, 2017.

\bibitem[Shumailov et~al.(2021)Shumailov, Shumaylov, Kazhdan, Zhao, Papernot,
  Erdogdu, and Anderson]{shumailov2021manipulating}
Ilia Shumailov, Zakhar Shumaylov, Dmitry Kazhdan, Yiren Zhao, Nicolas Papernot,
  Murat~A. Erdogdu, and Ross Anderson.
\newblock Manipulating sgd with data ordering attacks, 2021.

\bibitem[Simonyan \& Zisserman(2014)Simonyan and Zisserman]{simonyan2014very}
Karen Simonyan and Andrew Zisserman.
\newblock Very deep convolutional networks for large-scale image recognition.
\newblock \emph{arXiv preprint arXiv:1409.1556}, 2014.

\bibitem[Smith et~al.(2018)Smith, Kindermans, Ying, and Le]{smith2018dont}
Samuel~L. Smith, Pieter-Jan Kindermans, Chris Ying, and Quoc~V. Le.
\newblock Don't decay the learning rate, increase the batch size, 2018.

\bibitem[Snapp \& Shamir(2021)Snapp and Shamir]{Snapp2021}
Robert~R. Snapp and Gil~I. Shamir.
\newblock Synthesizing irreproducibility in deep networks.
\newblock \emph{CoRR}, abs/2102.10696, 2021.
\newblock URL \url{https://arxiv.org/abs/2102.10696}.

\bibitem[Stickland \& Murray(2020)Stickland and Murray]{stickland2020}
Asa~Cooper Stickland and Iain Murray.
\newblock Diverse ensembles improve calibration.
\newblock \emph{CoRR}, abs/2007.04206, 2020.
\newblock URL \url{https://arxiv.org/abs/2007.04206}.

\bibitem[Summers \& Dinneen(2021)Summers and
  Dinneen]{summers2021nondeterminism}
Cecilia Summers and Michael~J Dinneen.
\newblock Nondeterminism and instability in neural network optimization.
\newblock In \emph{International Conference on Machine Learning}, pp.\
  9913--9922. PMLR, 2021.

\bibitem[Tolstikhin et~al.(2021)Tolstikhin, Houlsby, Kolesnikov, Beyer, Zhai,
  Unterthiner, Yung, Steiner, Keysers, Uszkoreit, et~al.]{tolstikhin2021mlp}
Ilya~O Tolstikhin, Neil Houlsby, Alexander Kolesnikov, Lucas Beyer, Xiaohua
  Zhai, Thomas Unterthiner, Jessica Yung, Andreas Steiner, Daniel Keysers,
  Jakob Uszkoreit, et~al.
\newblock Mlp-mixer: An all-mlp architecture for vision.
\newblock \emph{Advances in Neural Information Processing Systems},
  34:\penalty0 24261--24272, 2021.

\bibitem[Vaswani et~al.(2017)Vaswani, Shazeer, Parmar, Uszkoreit, Jones, Gomez,
  Kaiser, and Polosukhin]{transformer}
Ashish Vaswani, Noam Shazeer, Niki Parmar, Jakob Uszkoreit, Llion Jones,
  Aidan~N. Gomez, Lukasz Kaiser, and Illia Polosukhin.
\newblock {A}ttention is {A}ll you {N}eed.
\newblock In \emph{Advances in Neural Information Processing Systems 30: Annual
  Conference on Neural Information Processing Systems 2017, 4-9 December 2017,
  Long Beach, CA, {USA}}, pp.\  6000--6010, 2017.

\bibitem[Veldanda et~al.(2022)Veldanda, Brugere, Chen, Dutta, Mishler, and
  Garg]{veldanda2022fairness}
Akshaj~Kumar Veldanda, Ivan Brugere, Jiahao Chen, Sanghamitra Dutta, Alan
  Mishler, and Siddharth Garg.
\newblock Fairness via in-processing in the over-parameterized regime: A
  cautionary tale.
\newblock \emph{arXiv preprint arXiv:2206.14853}, 2022.

\bibitem[Wang et~al.(2020)Wang, Kondratyuk, Kitani, Movshovitz{-}Attias, and
  Eban]{wang2020}
Xiaofang Wang, Dan Kondratyuk, Kris~M. Kitani, Yair Movshovitz{-}Attias, and
  Elad Eban.
\newblock Multiple networks are more efficient than one: Fast and accurate
  models via ensembles and cascades.
\newblock \emph{CoRR}, abs/2012.01988, 2020.
\newblock URL \url{https://arxiv.org/abs/2012.01988}.

\bibitem[Wortsman et~al.(2022)Wortsman, Ilharco, Gadre, Roelofs, Gontijo-Lopes,
  Morcos, Namkoong, Farhadi, Carmon, Kornblith, et~al.]{wortsman2022model}
Mitchell Wortsman, Gabriel Ilharco, Samir~Ya Gadre, Rebecca Roelofs, Raphael
  Gontijo-Lopes, Ari~S Morcos, Hongseok Namkoong, Ali Farhadi, Yair Carmon,
  Simon Kornblith, et~al.
\newblock Model soups: averaging weights of multiple fine-tuned models improves
  accuracy without increasing inference time.
\newblock In \emph{International Conference on Machine Learning}, pp.\
  23965--23998. PMLR, 2022.

\bibitem[Yun et~al.(2019)Yun, Han, Oh, Chun, Choe, and Yoo]{yun2019cutmix}
Sangdoo Yun, Dongyoon Han, Seong~Joon Oh, Sanghyuk Chun, Junsuk Choe, and
  Youngjoon Yoo.
\newblock Cutmix: Regularization strategy to train strong classifiers with
  localizable features.
\newblock In \emph{Proceedings of the IEEE/CVF international conference on
  computer vision}, pp.\  6023--6032, 2019.

\bibitem[Zafar et~al.(2015)Zafar, Valera, Rodriguez, and Gummadi]{zafar2015}
Muhammad~Bilal Zafar, Isabel Valera, Manuel~Gomez Rodriguez, and Krishna~P.
  Gummadi.
\newblock Fairness constraints: Mechanisms for fair classification, 2015.
\newblock URL \url{https://arxiv.org/abs/1507.05259}.

\bibitem[Zaidi et~al.(2021)Zaidi, Zela, Elsken, Holmes, Hutter, and
  Teh]{zaidi2021neural}
Sheheryar Zaidi, Arber Zela, Thomas Elsken, Chris~C Holmes, Frank Hutter, and
  Yee Teh.
\newblock Neural ensemble search for uncertainty estimation and dataset shift.
\newblock \emph{Advances in Neural Information Processing Systems},
  34:\penalty0 7898--7911, 2021.

\bibitem[Zhang et~al.(2017)Zhang, Cisse, Dauphin, and
  Lopez-Paz]{zhang2017mixup}
Hongyi Zhang, Moustapha Cisse, Yann~N Dauphin, and David Lopez-Paz.
\newblock mixup: Beyond empirical risk minimization.
\newblock \emph{arXiv preprint arXiv:1710.09412}, 2017.

\bibitem[Zhao et~al.(2017)Zhao, Wang, Yatskar, Ordonez, and
  Chang]{zhao-etal-2017-men}
Jieyu Zhao, Tianlu Wang, Mark Yatskar, Vicente Ordonez, and Kai-Wei Chang.
\newblock Men also like shopping: Reducing gender bias amplification using
  corpus-level constraints.
\newblock In \emph{Proceedings of the 2017 Conference on Empirical Methods in
  Natural Language Processing}, September 2017.

\bibitem[Zhao et~al.(2018)Zhao, Wang, Yatskar, Ordonez, and
  Chang]{zhao-etal-2018-gender}
Jieyu Zhao, Tianlu Wang, Mark Yatskar, Vicente Ordonez, and Kai-Wei Chang.
\newblock Gender bias in coreference resolution: Evaluation and debiasing
  methods.
\newblock In \emph{Proceedings of the 2018 Conference of the North {A}merican
  Chapter of the Association for Computational Linguistics: Human Language
  Technologies, Volume 2 (Short Papers)}, pp.\  15--20, New Orleans, Louisiana,
  June 2018. Association for Computational Linguistics.
\newblock \doi{10.18653/v1/N18-2003}.
\newblock URL \url{https://aclanthology.org/N18-2003}.

\bibitem[Zhong et~al.(2020)Zhong, Zheng, Kang, Li, and Yang]{zhong2020random}
Zhun Zhong, Liang Zheng, Guoliang Kang, Shaozi Li, and Yi~Yang.
\newblock Random erasing data augmentation.
\newblock In \emph{Proceedings of the AAAI conference on artificial
  intelligence}, volume~34, pp.\  13001--13008, 2020.

\bibitem[Zhuang et~al.(2022)Zhuang, Zhang, Song, and
  Hooker]{MLSYS2022_757b505c}
Donglin Zhuang, Xingyao Zhang, Shuaiwen Song, and Sara Hooker.
\newblock Randomness in neural network training: Characterizing the impact of
  tooling.
\newblock In D.~Marculescu, Y.~Chi, and C.~Wu (eds.), \emph{Proceedings of
  Machine Learning and Systems}, volume~4, pp.\  316--336, 2022.
\newblock URL
  \url{https://proceedings.mlsys.org/paper/2022/file/757b505cfd34c64c85ca5b5690ee5293-Paper.pdf}.

\end{thebibliography}

\clearpage

\newpage
\appendix
\onecolumn

\section{Experimental Setup}
\label{sec:experiments}

\subsection{Sampling}
Given a pool of \textit{M} models, for each ensemble size \textit{S} we sample 100 times with replacement. We then average the accuracy across the 100 samples plus one base model that is shared across all variants. The result at each \textit{S} is reported until an ensemble of \textit{M} models is reached.

\subsection{CIFAR-100 Training}
We use the following architectures: ResNet9 \citep{he2016deep}, VGG16 \citep{simonyan2014very} and MLP-Mixer \citep{tolstikhin2021mlp}. We train them as follows: 

\textbf{ResNet-9} We train the model for 24 steps using Stochastic Gradient Descent (SGD). We implemented standard data augmentation by applying Random Horizontal Flip, Random Translate, and Cutout. We use a Slanted Triangular Learning Rate (SLTR) \citep{howard2018universal}. The top-1 test set accuracy is 72.24\%

\textbf{ResNet18/34/50} For these 3 ResNet architectures, we train the model for 50 epochs using Stochastic Gradient Descent (SGD), batch size of 512, momentum=0.9, and weight decay=0.0005. We implemented standard data augmentation by applying Random Horizontal Flip, Random Crop, Random Affine, and Cutout. We use a combination of warmup for the first 5 epoch and cosine annealing for scheduler. The top-1 test set accuracy for ResNet-18 is 73.56\%, ResNet-34 is 74.24\%, and ResNet-50 is 74.89\%

\textbf{VGG16} We train the model for 130 epochs using Stochastic Gradient Descent (SGD). We implemented standard data augmentation by applying Random Horizontal Flip, Random Crop, and Random Rotation. We use a combination of warmup for 1 epoch and a multi-step scheduler with milestones at steps 60 and 120. The top-1 test set accuracy is 71.23\%

\textbf{MLP-Mixer} We train the model for 300 steps using Adaptive Moment Estimation (Adam) \citep{kingma2014adam}. We implemented standard data augmentation by applying Random Crop, AutoAugment (CIFAR10 Policy) \citep{cubuk2018autoaugment}, and CutMix \citep{yun2019cutmix}. We use a combination of warmup for the first 5 epoch and cosine annealing for scheduler. The top-1 test set accuracy is 60.28\%

\subsection{TinyImageNet Training}
We use the following architectures: ResNets \citep{he2016deep}, VGG-16 \citep{simonyan2014very} and ViT \citep{dosovitskiy2020image}. We train them as follows: 

\textbf{ResNets} We train 3 different architectures from the ResNet family (ResNet18, 34, 50) for 100 steps using Stochastic Gradient Descent (SGD). We implemented standard data augmentation by applying Random Resized Crop and Random Horizontal Flip. We use a Slanted Triangular Learning Rate (SLTR) \citep{howard2018universal}. The top-1 test set accuracy for ResNet-18 is 49.27\%, ResNet-34 is 52.18\%, and ResNet-50 is 54.99\%

\textbf{VGG16} We train the model for 100 steps using Stochastic Gradient Descent (SGD). We implemented standard data augmentation by applying Random Resized Crop and Random Horizontal Flip. We use a Slanted Triangular Learning Rate (SLTR) \citep{howard2018universal}. The top-1 test set accuracy is 60.37\%

\textbf{ViT} We train the model for 100 steps using Adaptive Moment Estimation with decoupled weight decay  (AdamW) \citep{loshchilov2017decoupled}. We implemented standard data augmentation by applying Random Horizontal Flip, Random Resized Crop, AutoAugment \citep{cubuk2018autoaugment}, Random Erasing \citep{zhong2020random}, Cutmix \citep{yun2019cutmix}, and Mixup\citep{zhang2017mixup}. We use a combination of warmup for the first 10 epoch and cosine annealing \citep{loshchilov2016sgdr} for scheduler. The top-1 test set accuracy is 51.21\%

\newpage

\begin{figure*}[ht]
    \section{FAIR-Ensemble: When Homogeneous Ensemble Disproportionately Benefit Minority Groups}
    \subsection{Experimental Setup}
    \centering
    \underline{CIFAR100}\\
    \begin{minipage}{0.01\linewidth}
    \rotatebox{90}{\small ensemble/base}
    \end{minipage}
    \begin{minipage}{0.98\linewidth}
    \begin{subfigure}{0.16\linewidth}
        \centering
        ResNet9\\
        \includegraphics[width=\linewidth]{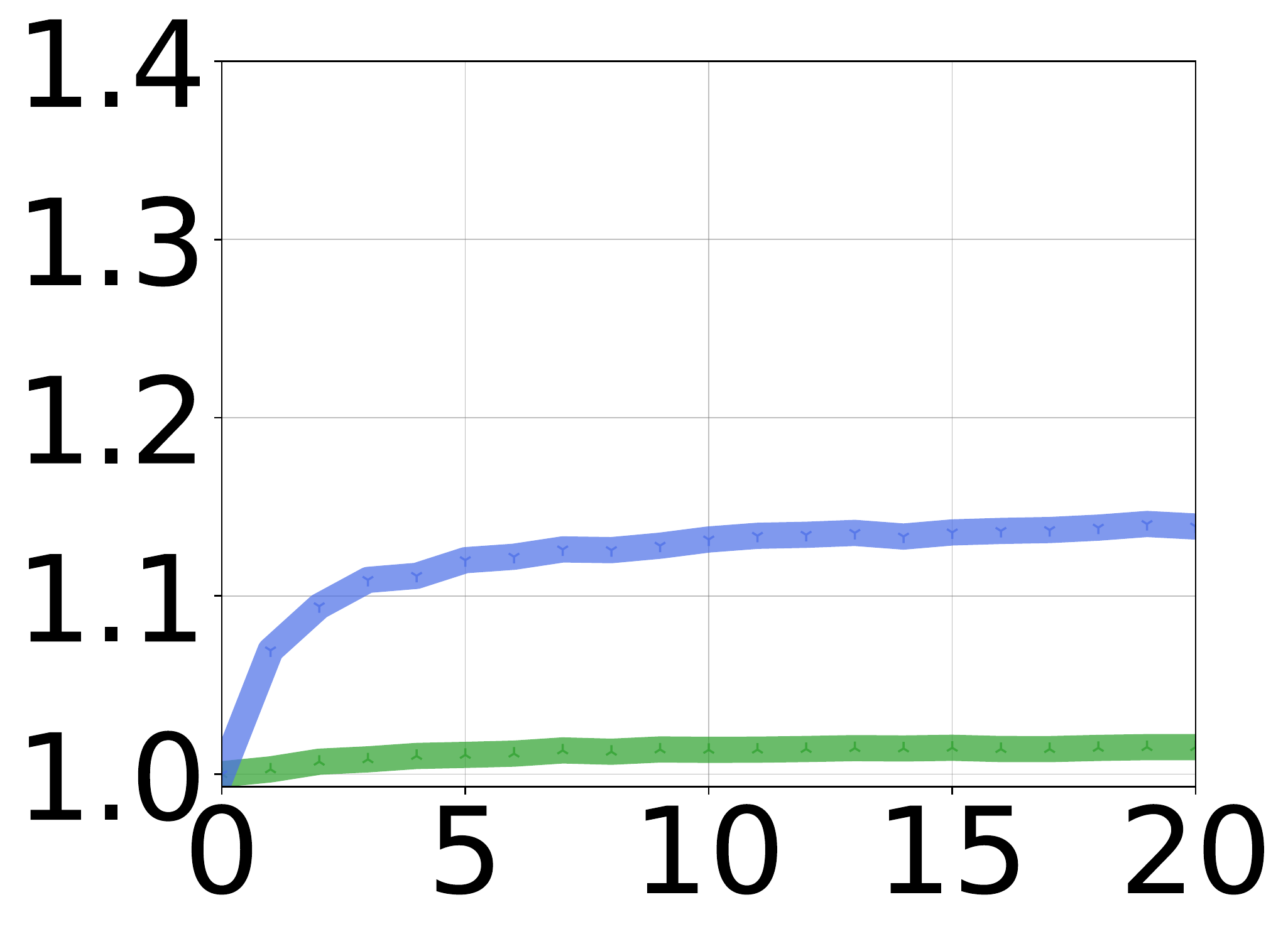}
    \end{subfigure}
    \begin{subfigure}{0.16\linewidth}
        \centering
        ResNet18\\
        \includegraphics[width=\linewidth]{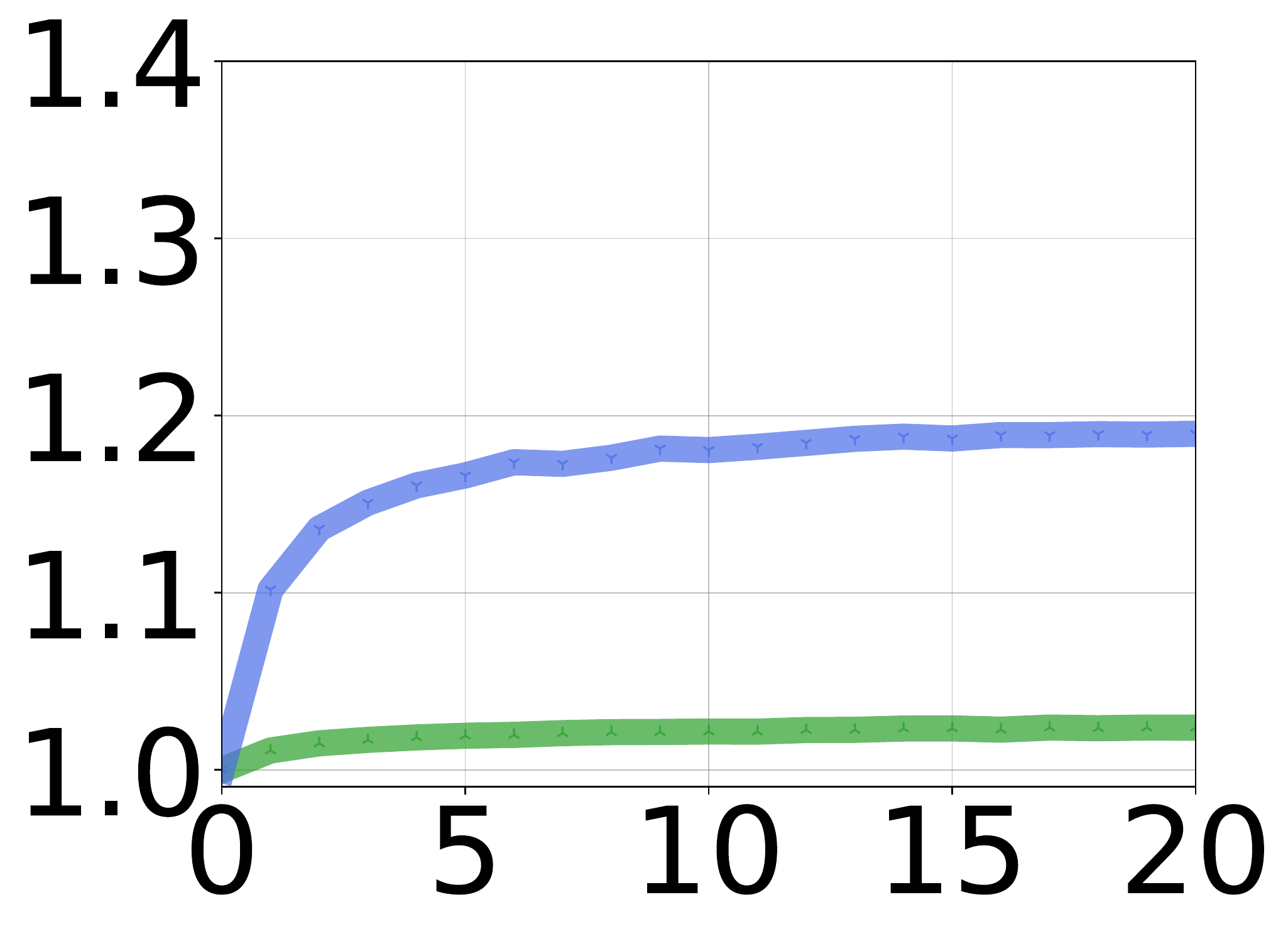}
    \end{subfigure}
    \begin{subfigure}{0.16\linewidth}
        \centering
        ResNet34\\
        \includegraphics[width=\linewidth]{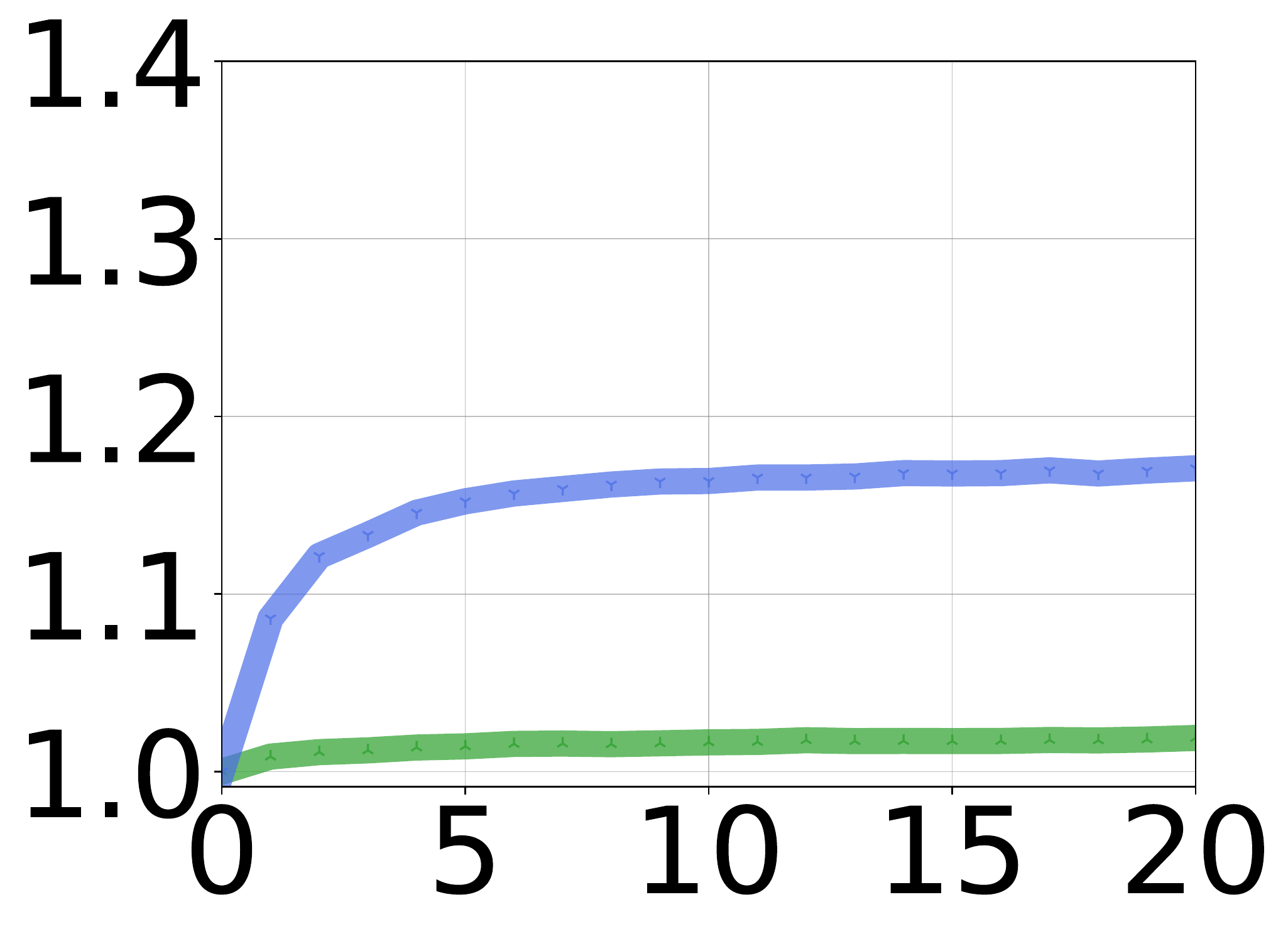}
    \end{subfigure}
    \begin{subfigure}{0.16\linewidth}
        \centering
        ResNet50\\
        \includegraphics[width=\linewidth]{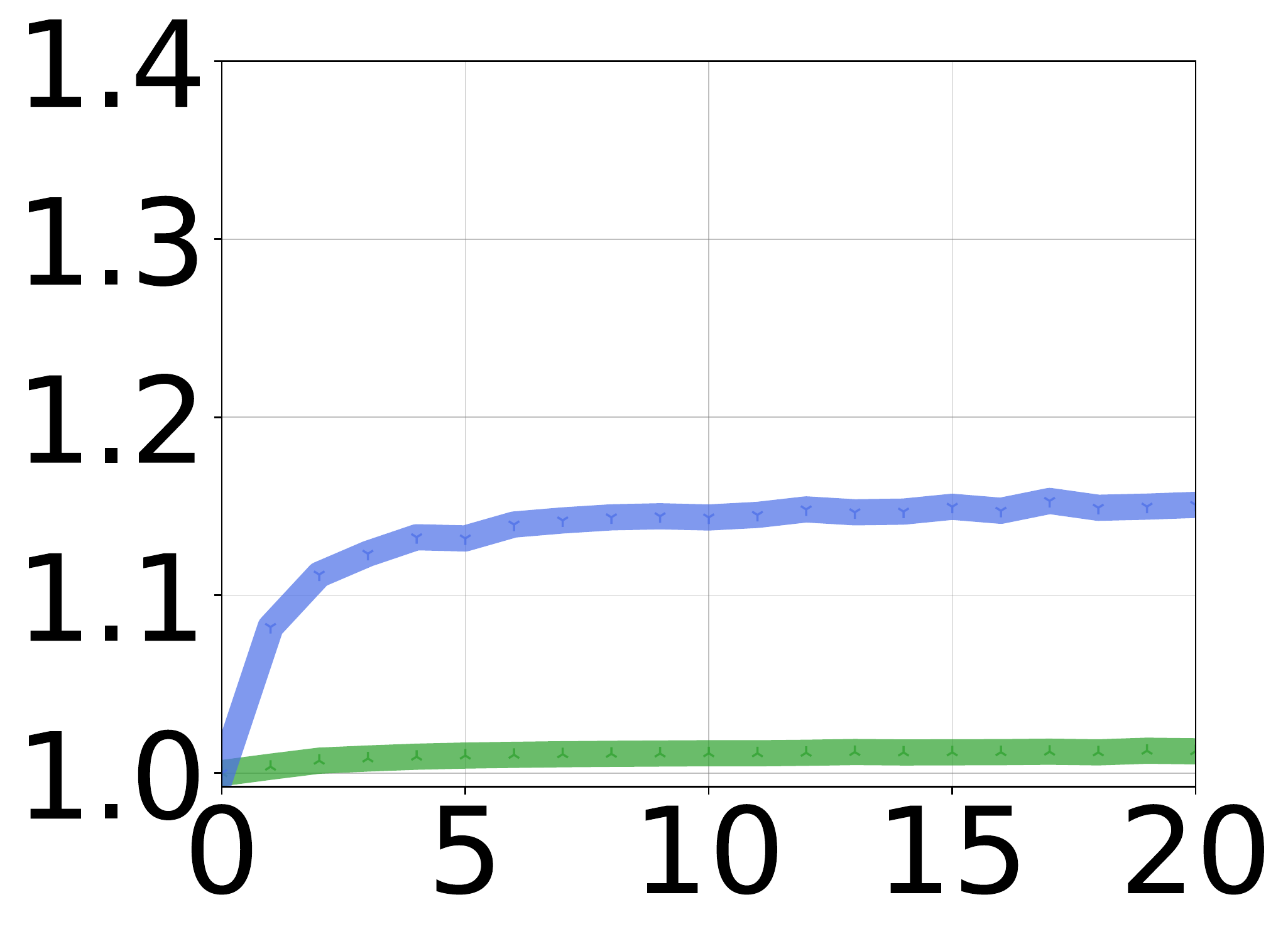}
    \end{subfigure}
    \begin{subfigure}{0.16\linewidth}
        \centering
        VGG16\\
        \includegraphics[width=\linewidth]{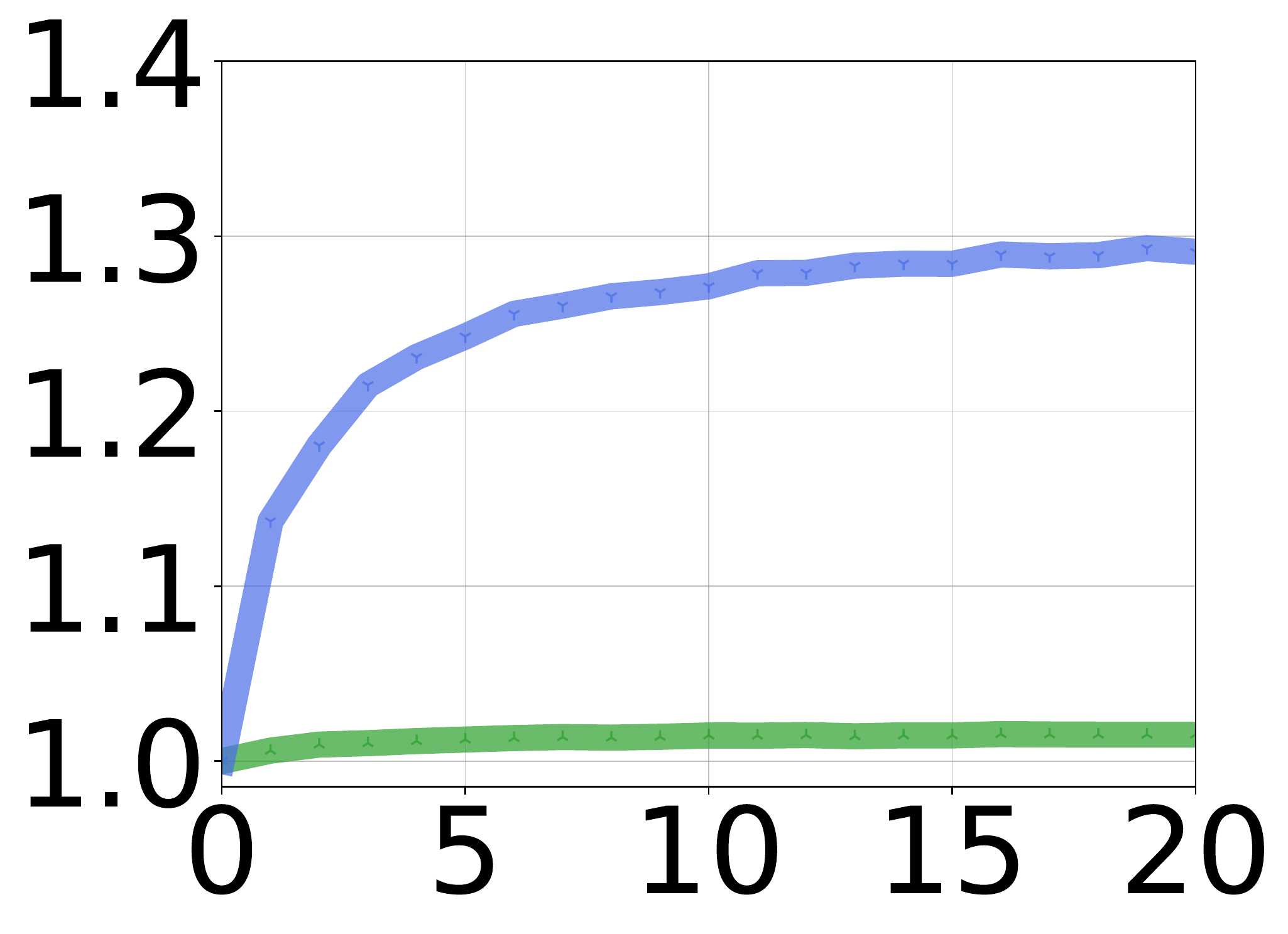}
    \end{subfigure}
    \begin{subfigure}{0.16\linewidth}
        \centering
        MLP-Mixer\\
        \includegraphics[width=\linewidth]{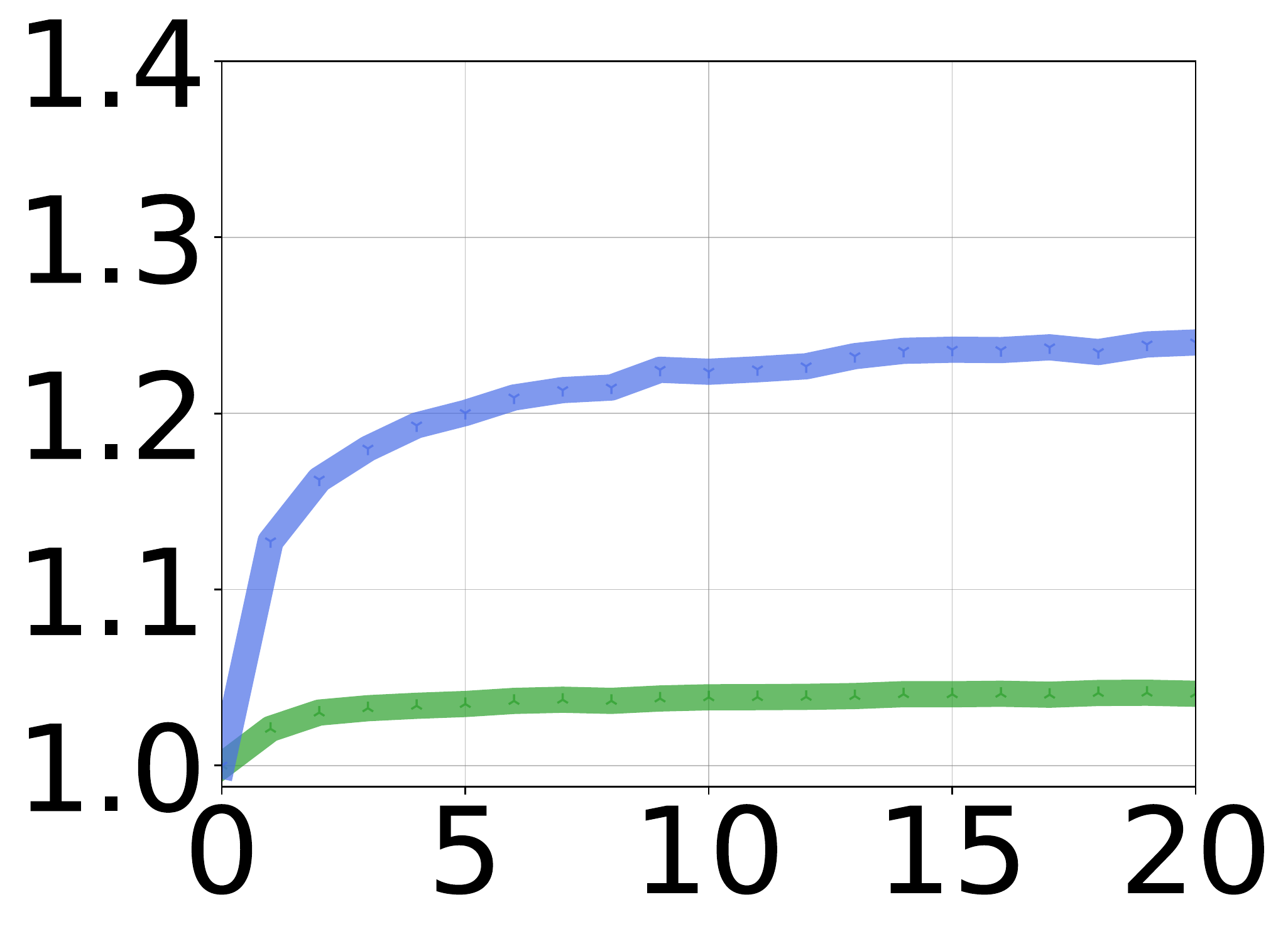}
    \end{subfigure}
    \end{minipage}
    \caption{\small Depiction of the accuracy gain as a ratio of ensemble accuracy \% over the singular base model ({\bf y-axis}) by per-group ({\color{green}top-k } and {\color{blue}bottom-k}) test set accuracies for CIFAR100.}
    \label{fig:fairness_c100}
\end{figure*}
\begin{figure*}[!ht]

\subsection{Balanced Dataset Sub-Groups}
\vspace{0.2cm}
	\centering
     \begin{minipage}{0.01\linewidth}
        \rotatebox{90}{\hspace{0.7cm} test accuracy \%}
    \end{minipage}
    \hspace{0.07cm}
    \begin{minipage}{0.47\linewidth}
	\begin{subfigure}{=\linewidth}
		\centering
    	\includegraphics[width=1.0\linewidth]{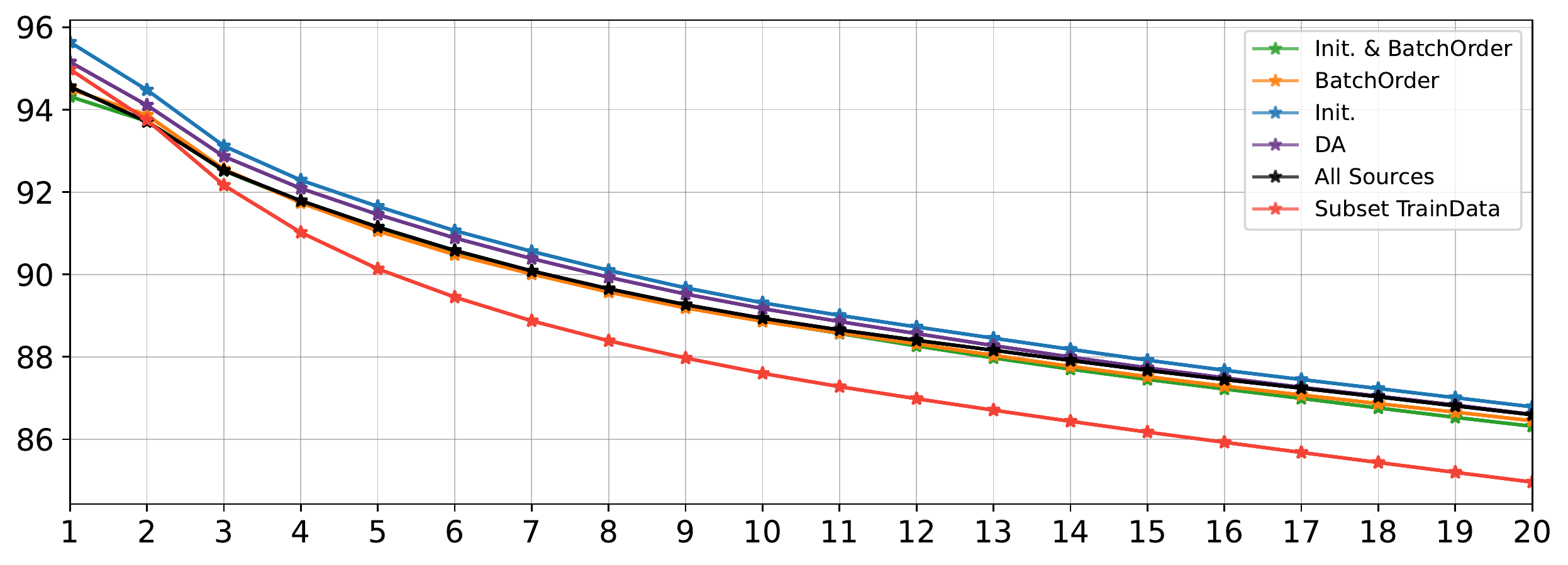}
        \\[-0.5em]
        K size
    	\caption{ResNet-50 Top-K classes}
        \label{fig:Res50_TopK_classes}
	\end{subfigure}
    \end{minipage}
    \begin{minipage}{0.01\linewidth}
        \rotatebox{90}{\hspace{0.7cm} test accuracy \%}
    \end{minipage}
    \hspace{0.07cm}
    \begin{minipage}{0.47\linewidth}
	\begin{subfigure}{=\linewidth}
		\centering
    	\includegraphics[width=1.0\linewidth]{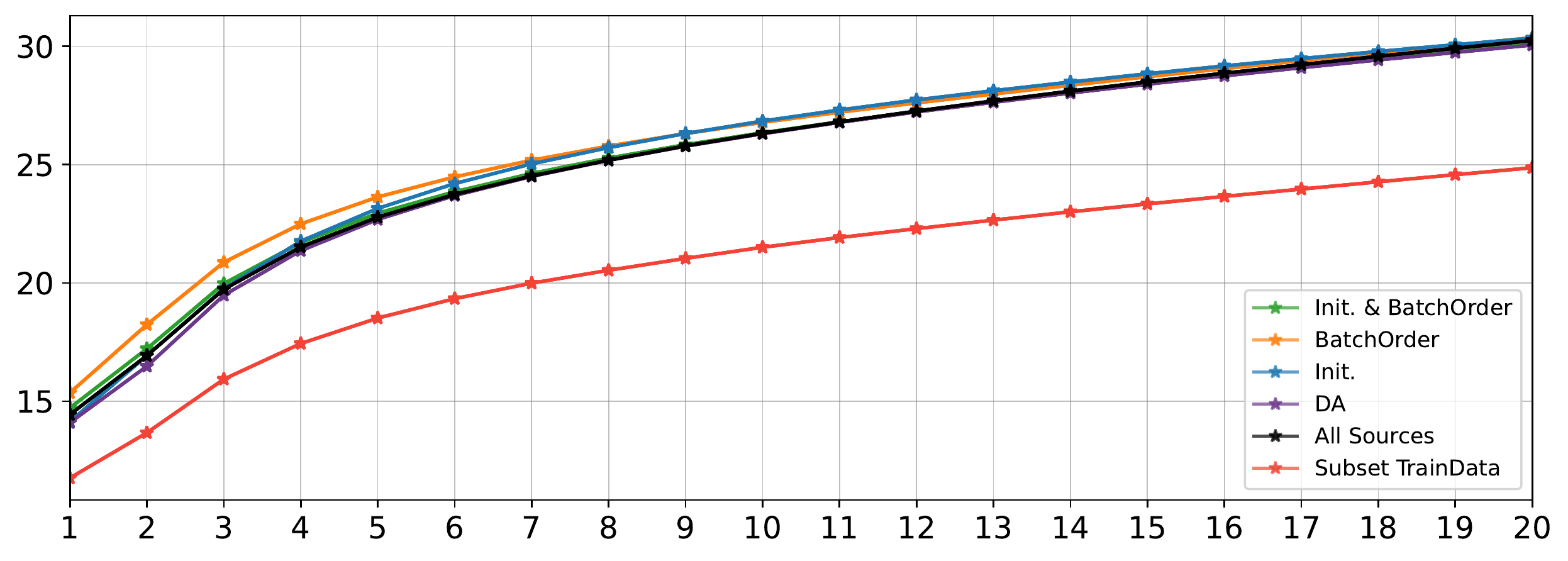}
        \\[-0.5em]
        K size
    	\caption{ResNet-50 Bottom-K classes}
        \label{fig:Res50_BottomK_classes}
	\end{subfigure}
    \end{minipage}
	\caption{
	    Average Top and Bottom-K accuracy as number of K-classes increase. We observe that \textbf{DA} and \textbf{Init.} outperforms \textbf{All Sources} baseline performance in Top-K classes, whereas \textbf{BatchOrder} and \textbf{Init.} outperfroms \textbf{All Sources} on Bottom-K classes. In both top and bottom groups, only the \textbf{TrainSubsetData} variant underperforms \textbf{All Sources}.
	}
	\label{fig:Tinyimagenet_K}
\end{figure*}

\begin{figure*}[ht!]
    \subsection{CIFAR-100}
    \centering
    \underline{ResNet9}\\
    \vspace{0.1cm}

    \begin{minipage}{0.01\linewidth}
        \rotatebox{90}{\hspace{0.2cm} ensemble/base }
    \end{minipage}
    \begin{minipage}{0.98\linewidth}
	\begin{subfigure}{0.24\linewidth}
		\centering
        Init\\
    	\includegraphics[width=1.0\linewidth]{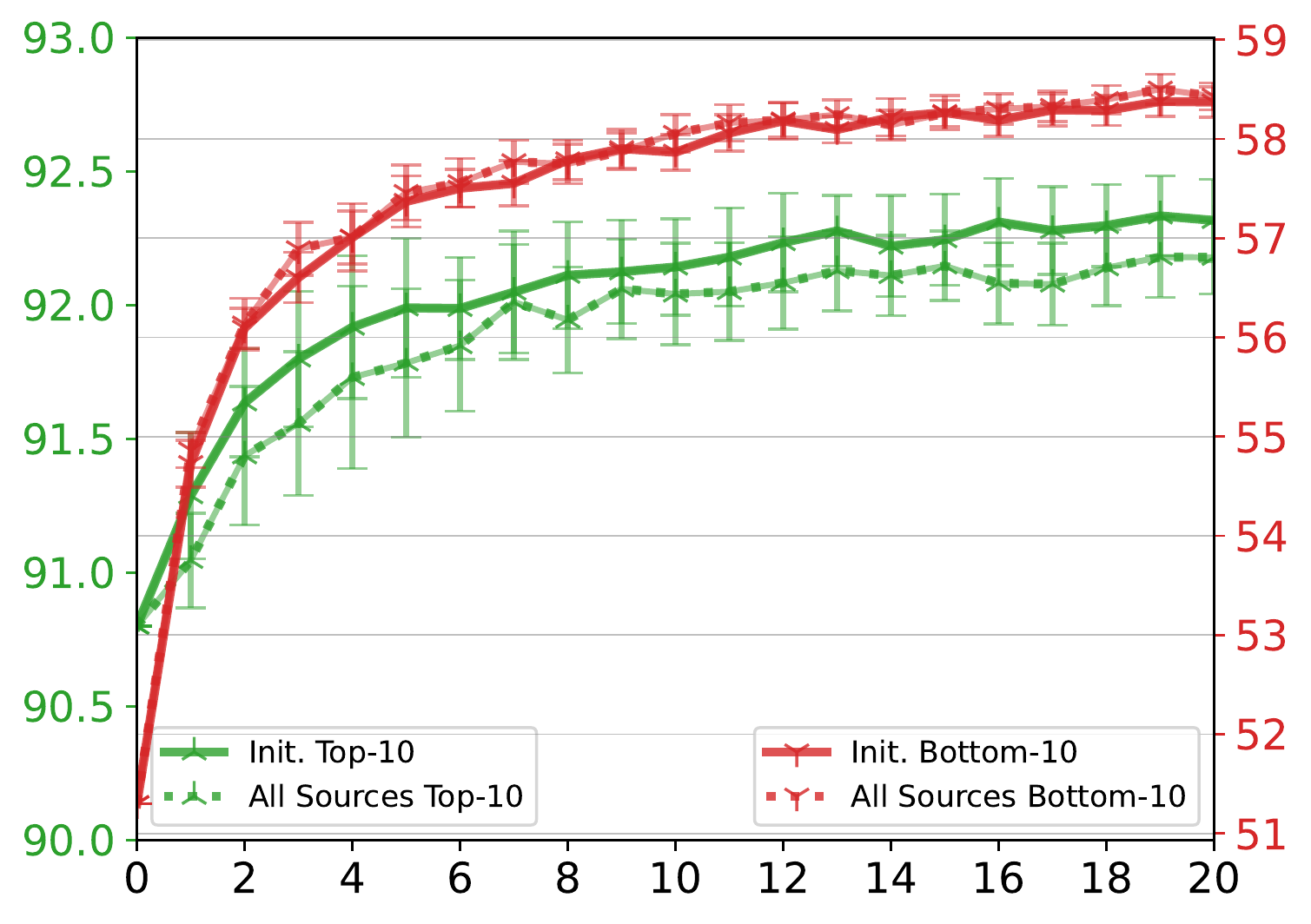}
	\end{subfigure}
 	\begin{subfigure}{0.24\linewidth}
		\centering
        BatchOrder\\
    	\includegraphics[width=1.0\linewidth]{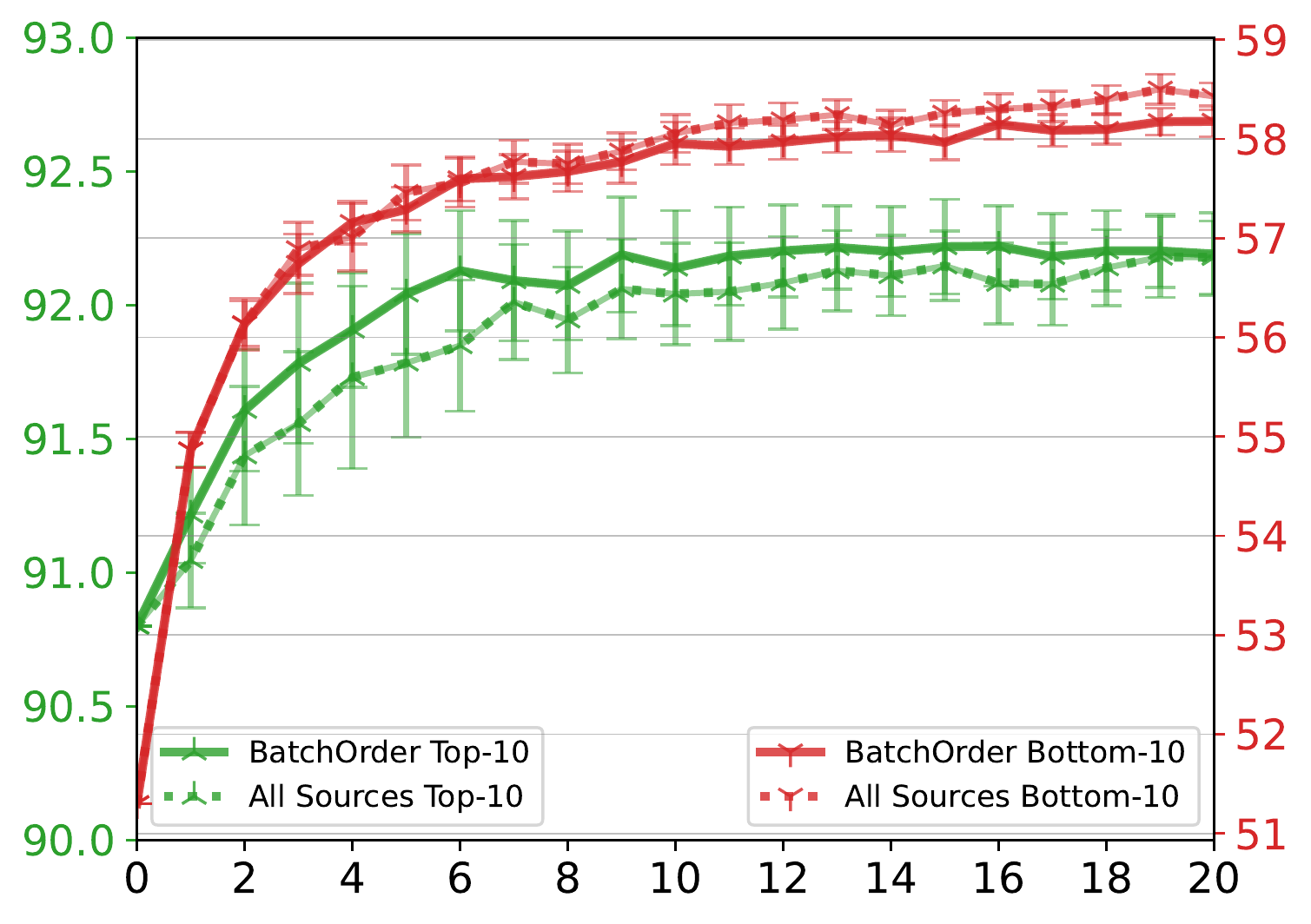}
	\end{subfigure}
	\begin{subfigure}{0.24\linewidth}
		\centering
        DA\\
    	\includegraphics[width=1.0\linewidth]{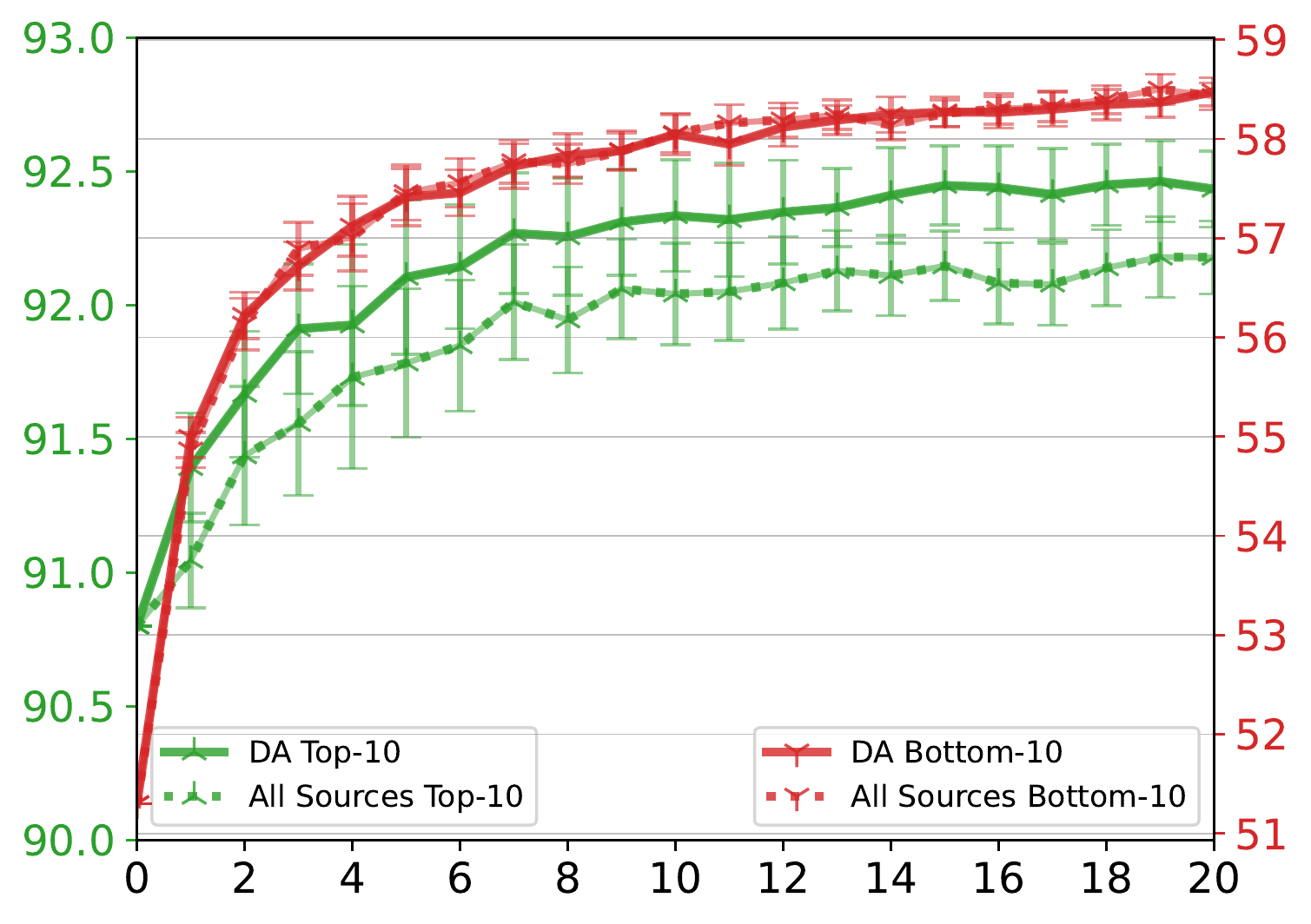}
	\end{subfigure}
 	\begin{subfigure}{0.24\linewidth}
		\centering
        Init \& BatchOrder\\
    	\includegraphics[width=1.0\linewidth]{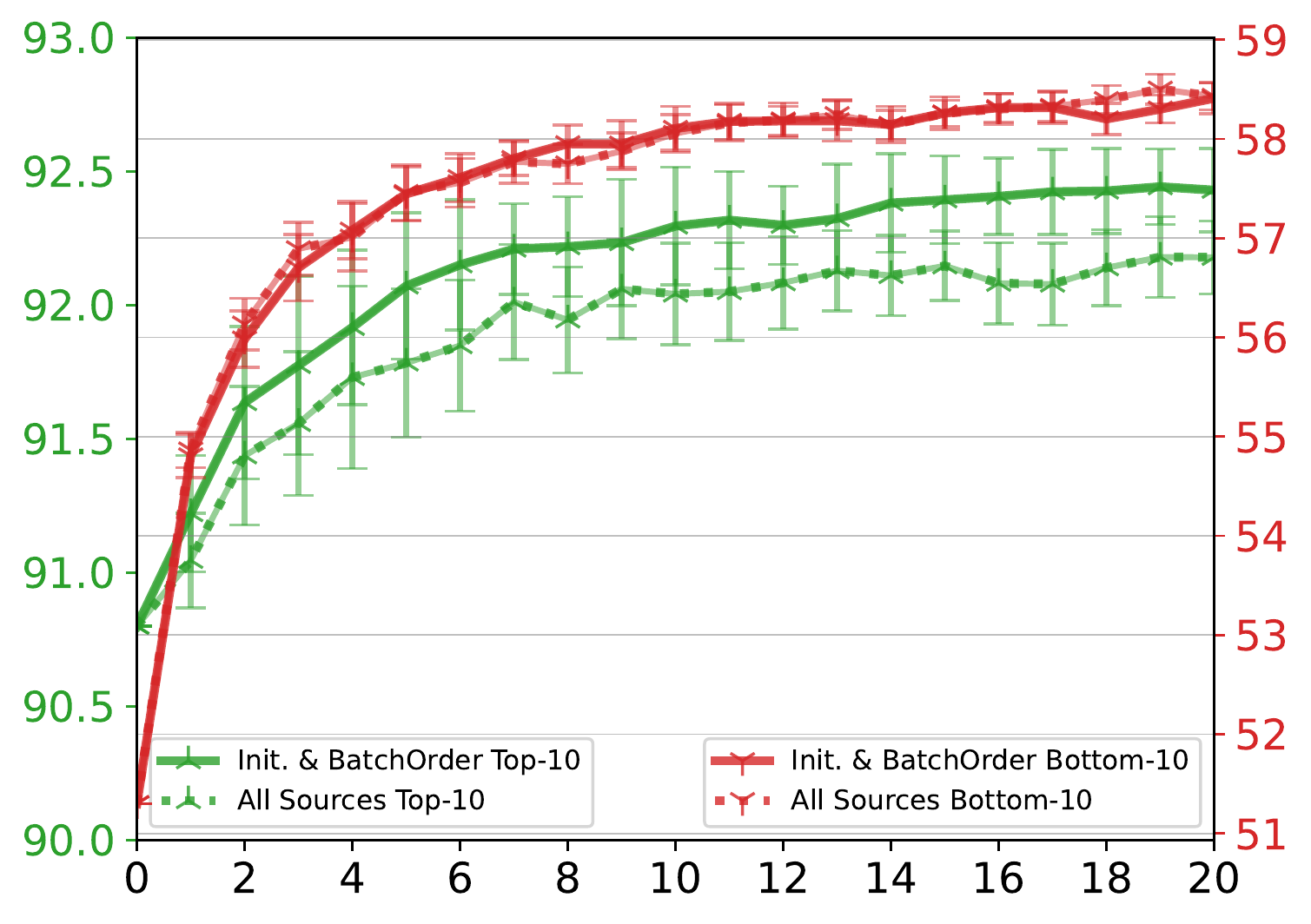}
	\end{subfigure}
    \end{minipage}

    \underline{ResNet18}\\
    \vspace{0.1cm}

    \begin{minipage}{0.01\linewidth}
        \rotatebox{90}{\hspace{0.2cm} ensemble/base }
    \end{minipage}
    \begin{minipage}{0.98\linewidth}
	\begin{subfigure}{0.24\linewidth}
		\centering
        Init\\
    	\includegraphics[width=1.0\linewidth]{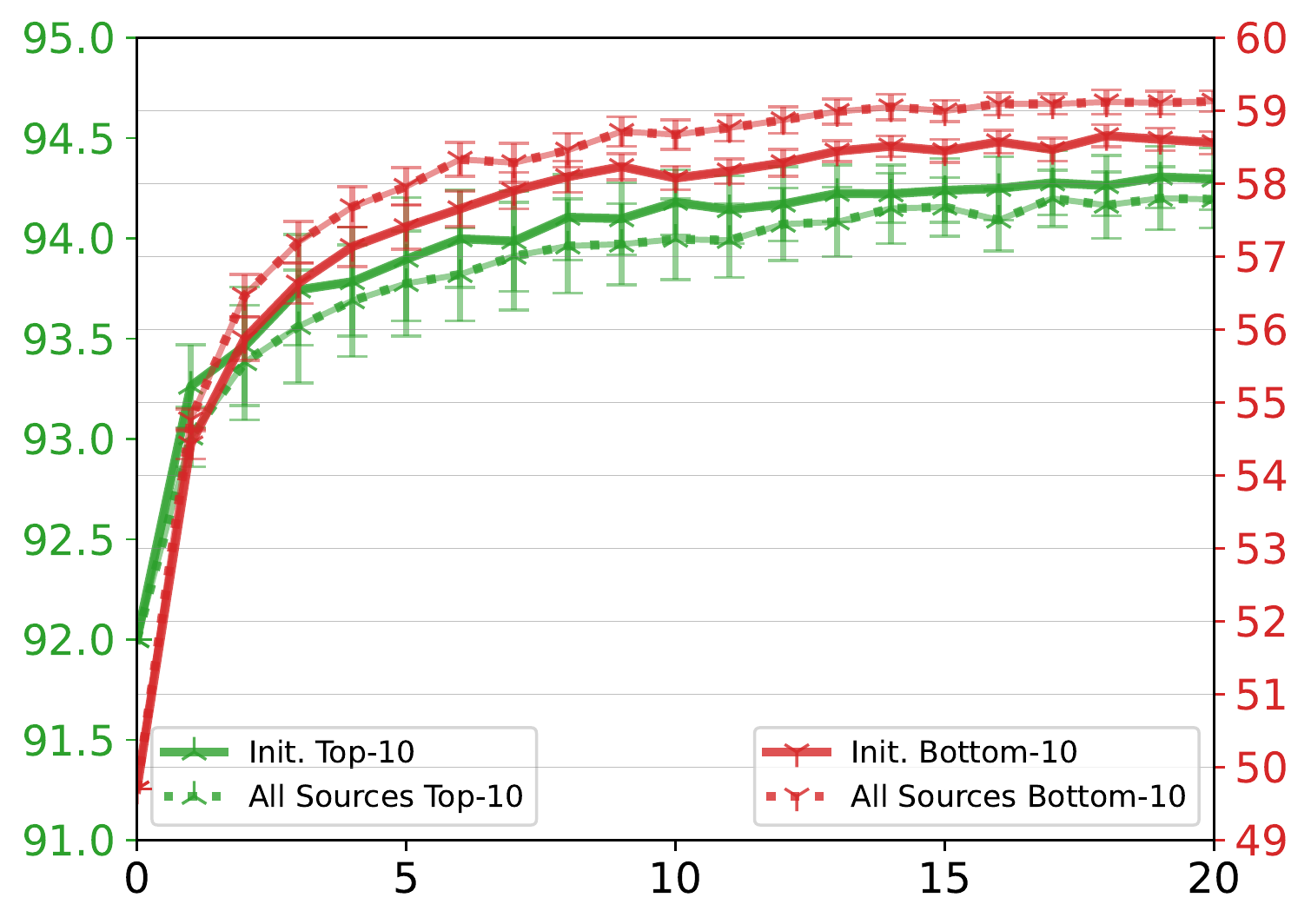}
	\end{subfigure}
 	\begin{subfigure}{0.24\linewidth}
		\centering
        BatchOrder\\
    	\includegraphics[width=1.0\linewidth]{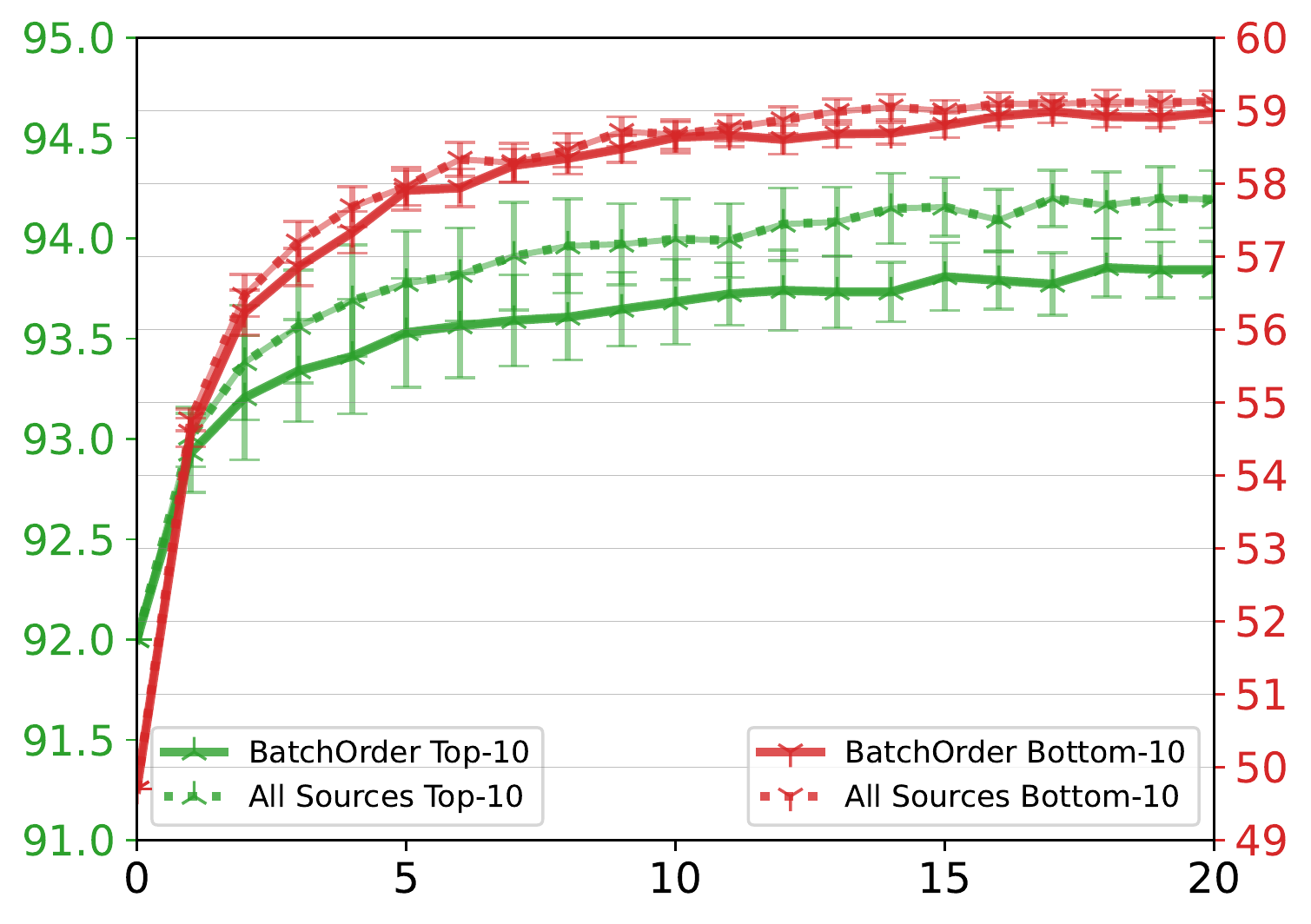}
	\end{subfigure}
	\begin{subfigure}{0.24\linewidth}
		\centering
        DA\\
    	\includegraphics[width=1.0\linewidth]{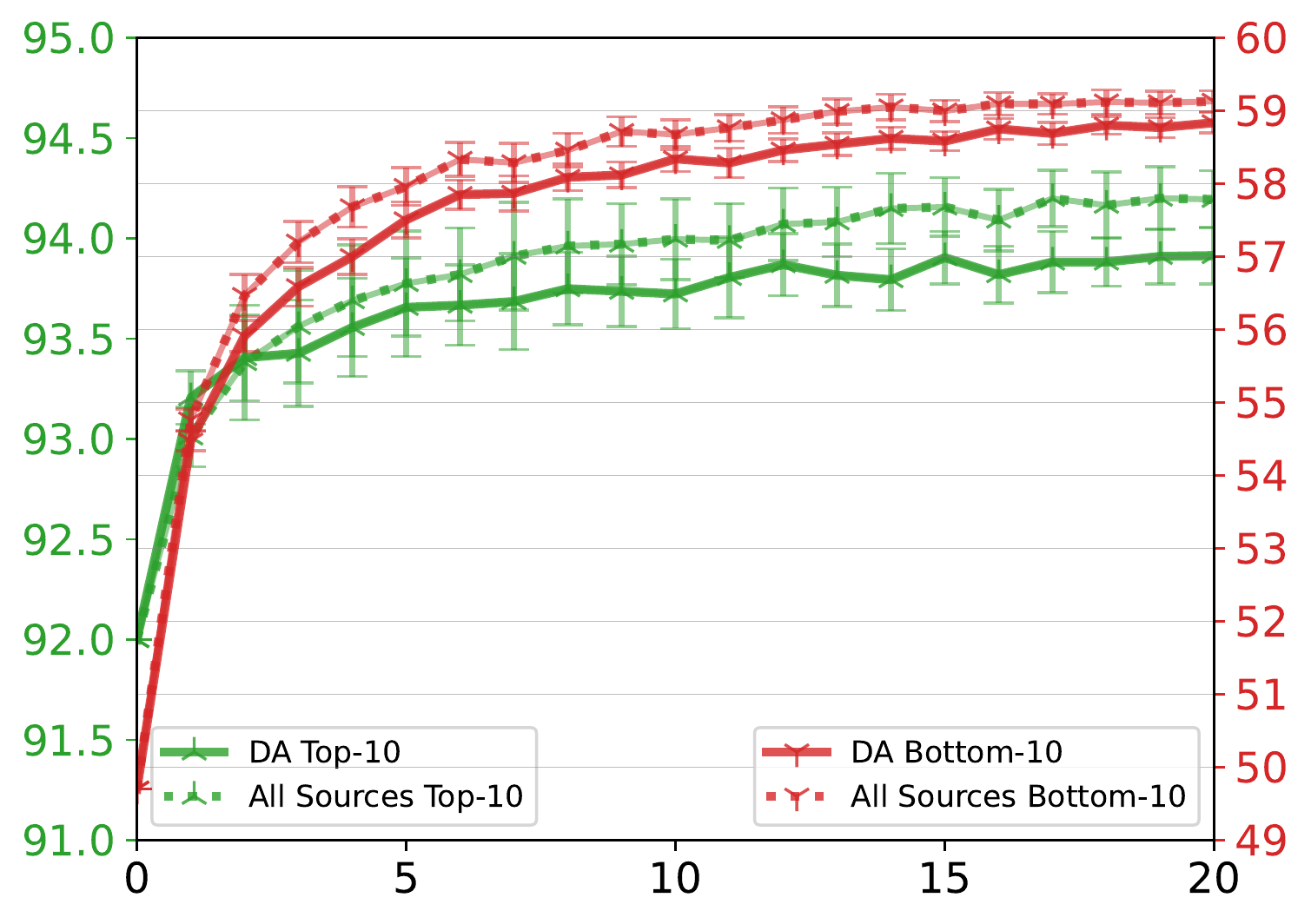}
	\end{subfigure}
 	\begin{subfigure}{0.24\linewidth}
		\centering
        Init \& BatchOrder\\
    	\includegraphics[width=1.0\linewidth]{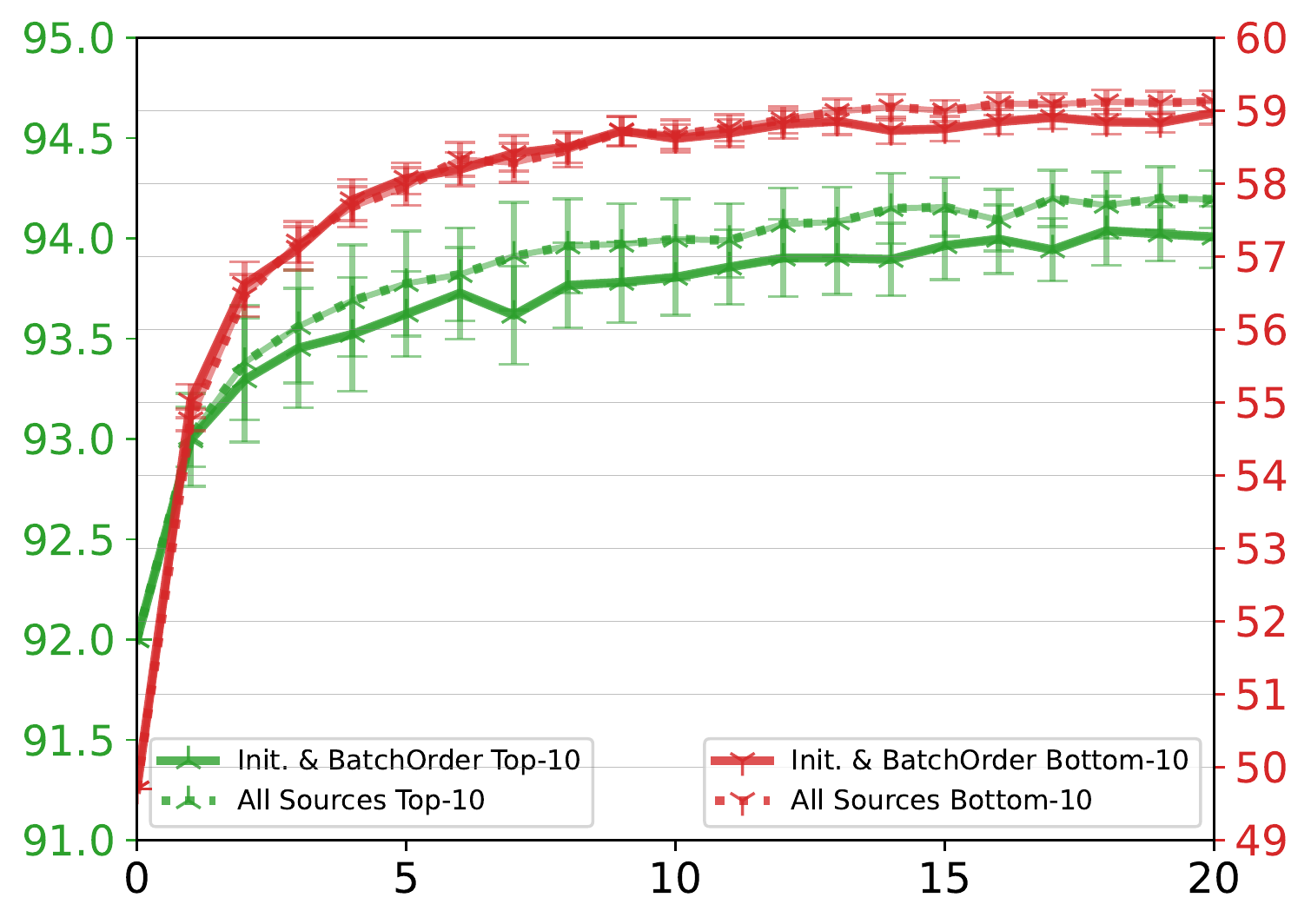}
	\end{subfigure}
    \end{minipage}

    \underline{ResNet34}\\
    \vspace{0.1cm}

    \begin{minipage}{0.01\linewidth}
        \rotatebox{90}{\hspace{0.2cm} ensemble/base }
    \end{minipage}
    \begin{minipage}{0.98\linewidth}
	\begin{subfigure}{0.24\linewidth}
		\centering
        Init\\
    	\includegraphics[width=1.0\linewidth]{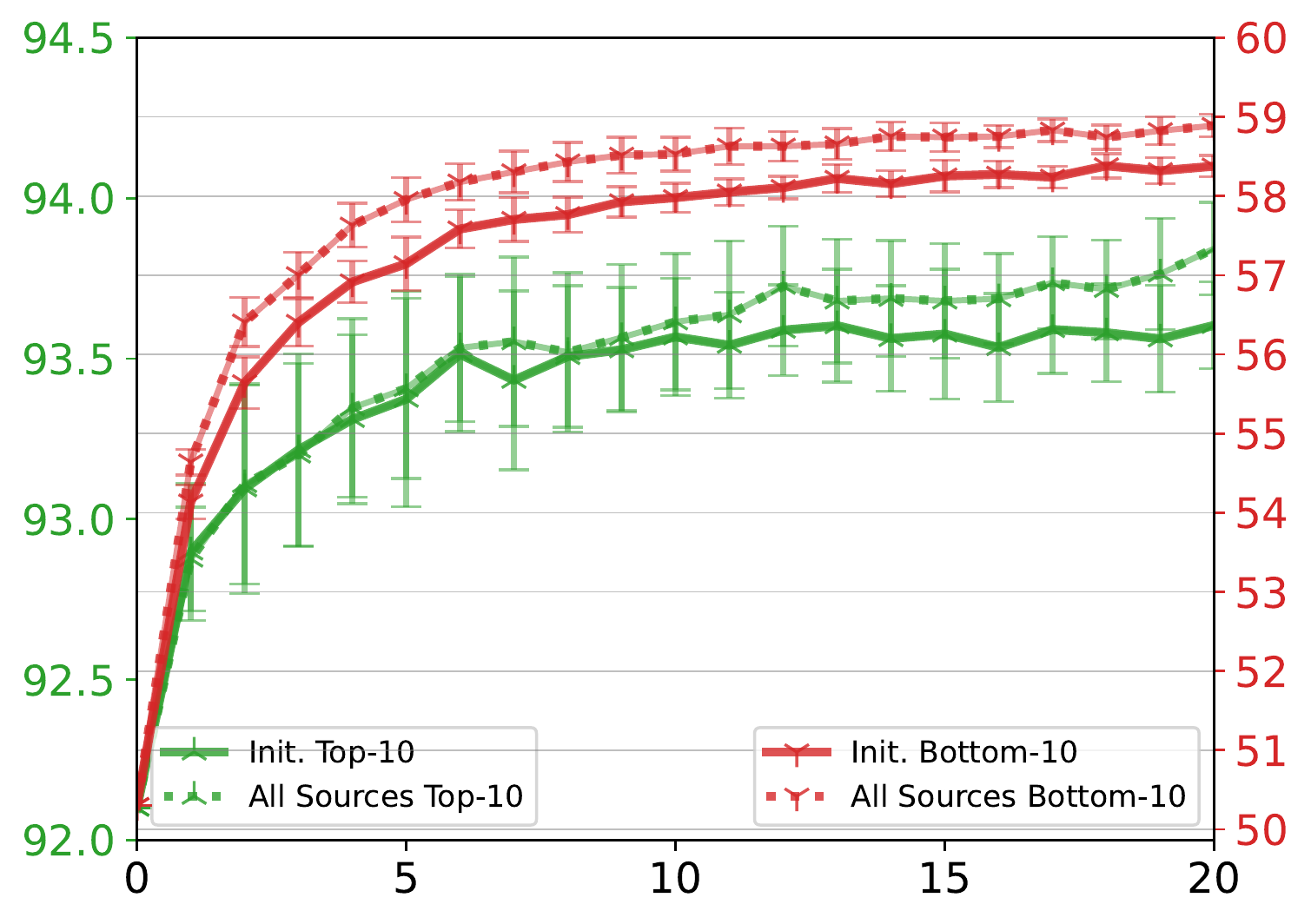}
	\end{subfigure}
 	\begin{subfigure}{0.24\linewidth}
		\centering
        BatchOrder\\
    	\includegraphics[width=1.0\linewidth]{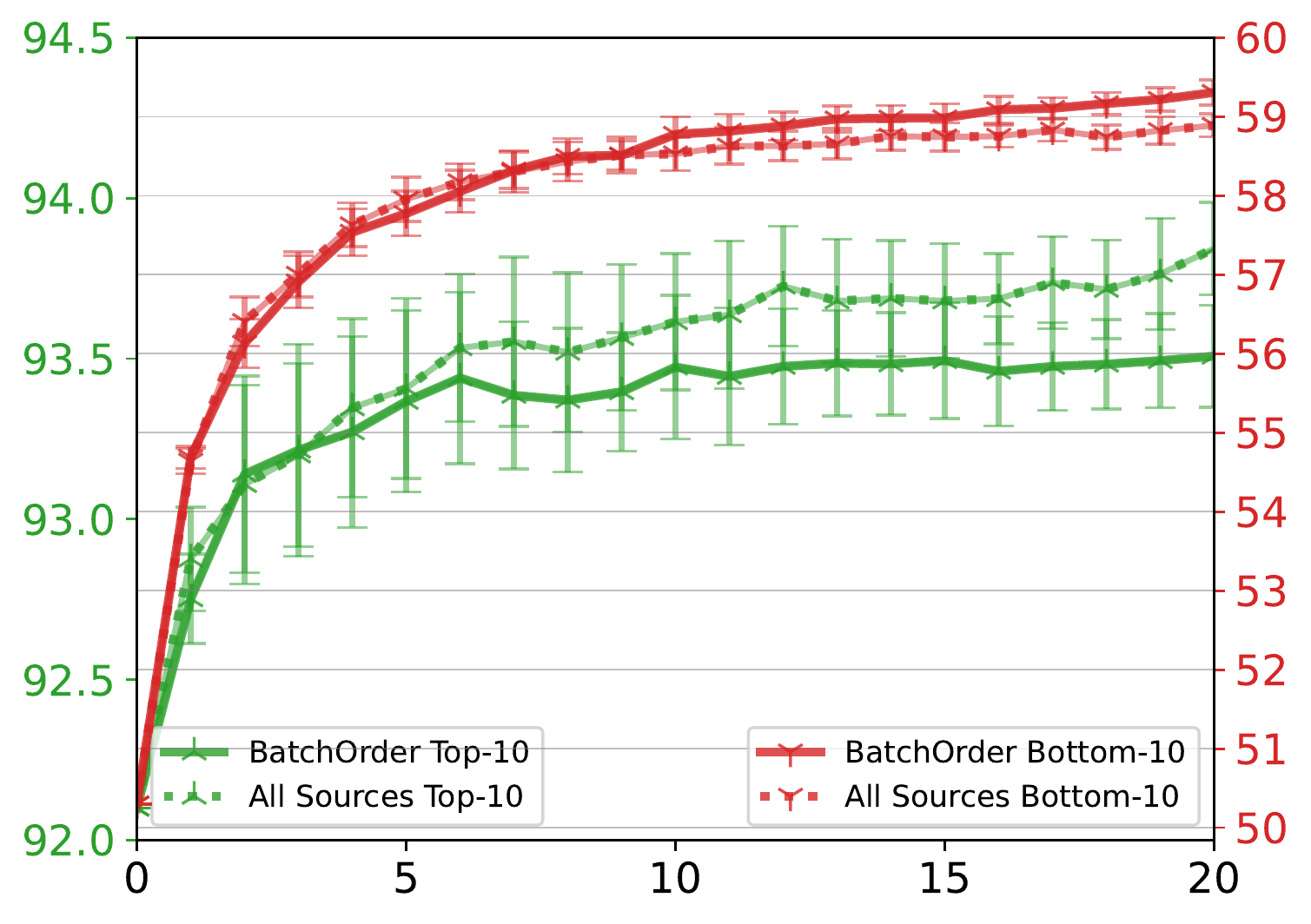}
	\end{subfigure}
	\begin{subfigure}{0.24\linewidth}
		\centering
        DA\\
    	\includegraphics[width=1.0\linewidth]{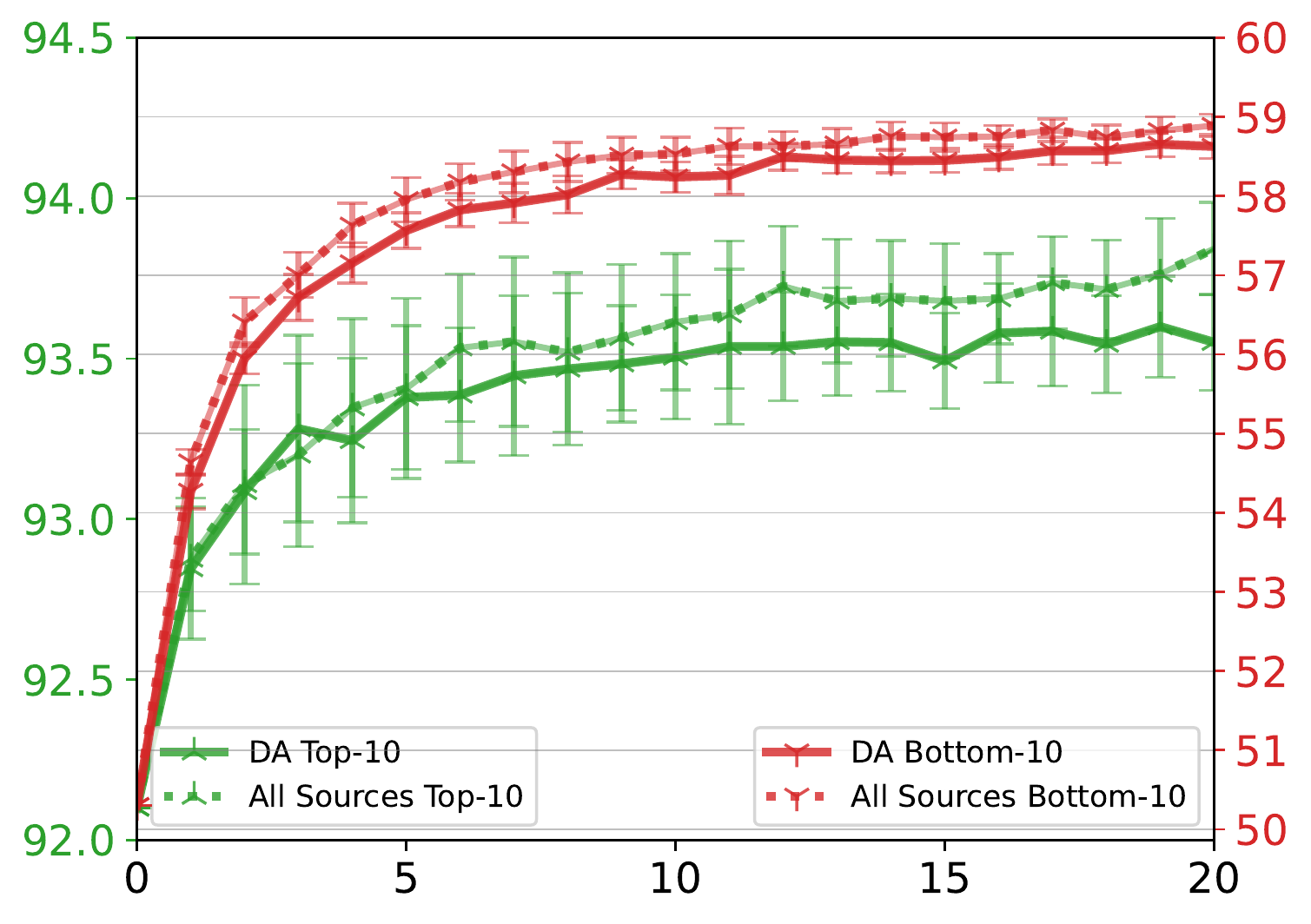}
	\end{subfigure}
 	\begin{subfigure}{0.24\linewidth}
		\centering
        Init \& BatchOrder\\
    	\includegraphics[width=1.0\linewidth]{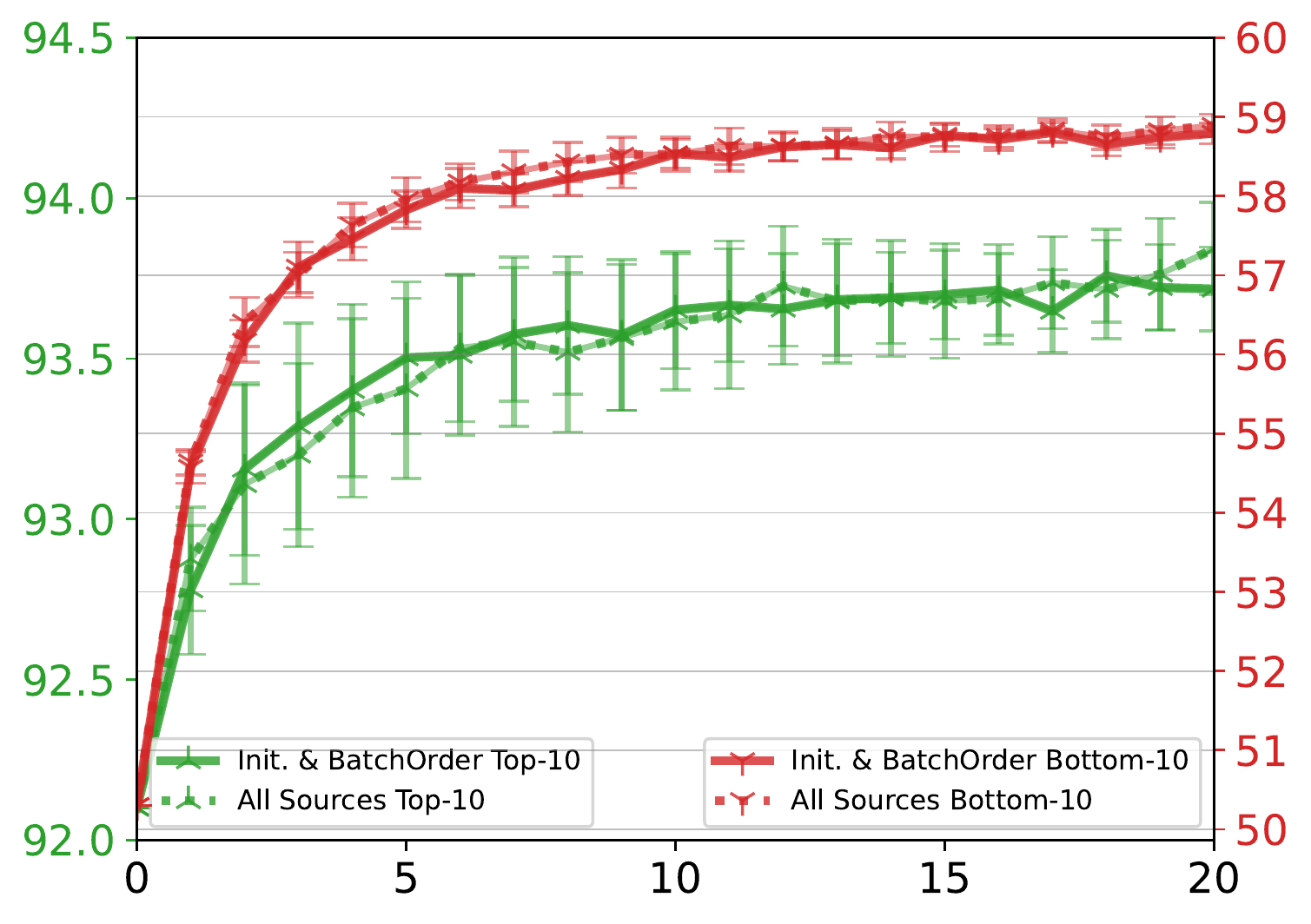}
	\end{subfigure}
    \end{minipage}

    \underline{ResNet50}\\
    \vspace{0.1cm}

    \begin{minipage}{0.01\linewidth}
        \rotatebox{90}{\hspace{0.2cm} ensemble/base }
    \end{minipage}
    \begin{minipage}{0.98\linewidth}
	\begin{subfigure}{0.24\linewidth}
		\centering
        Init\\
    	\includegraphics[width=1.0\linewidth]{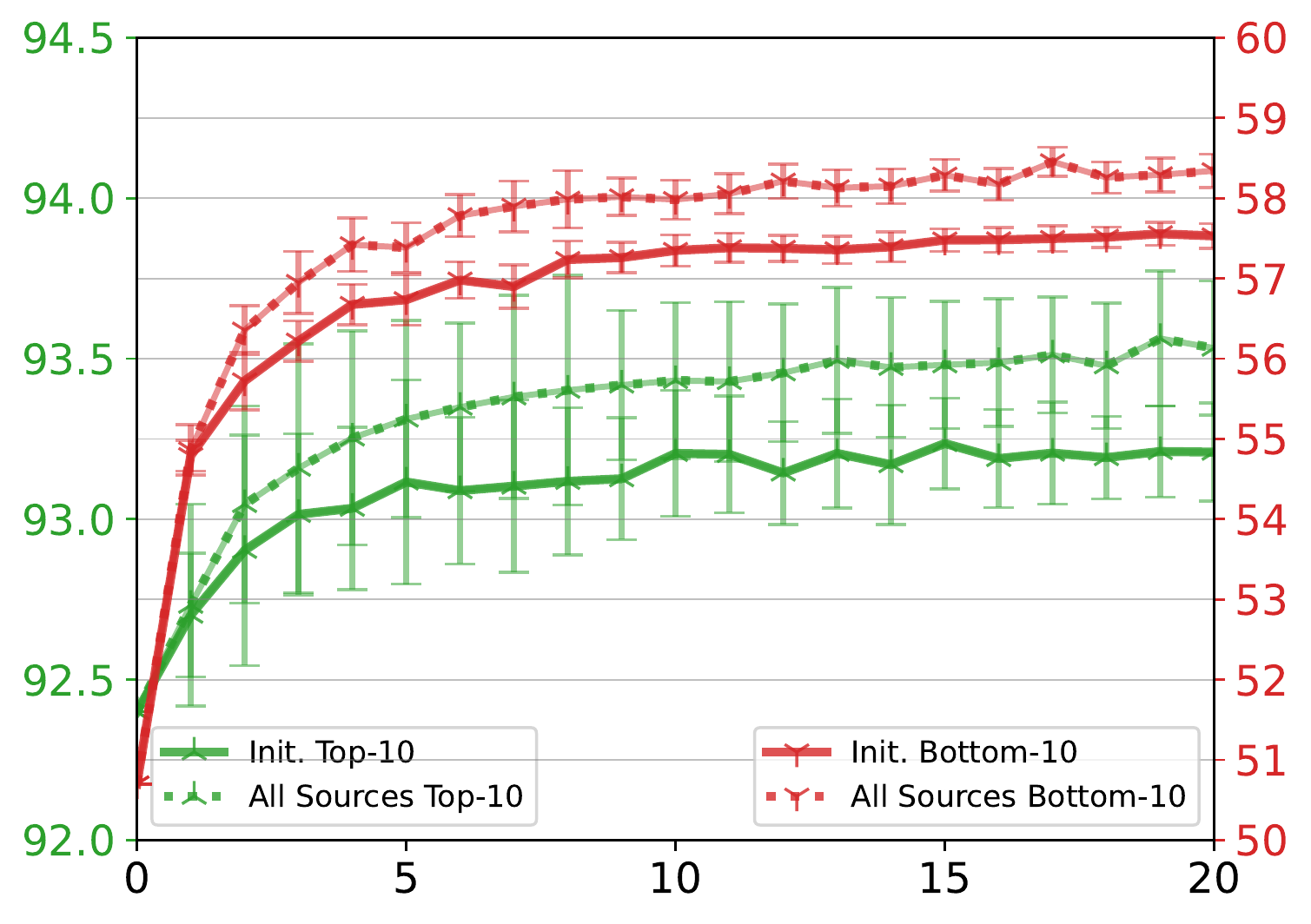}
	\end{subfigure}
 	\begin{subfigure}{0.24\linewidth}
		\centering
        BatchOrder\\
    	\includegraphics[width=1.0\linewidth]{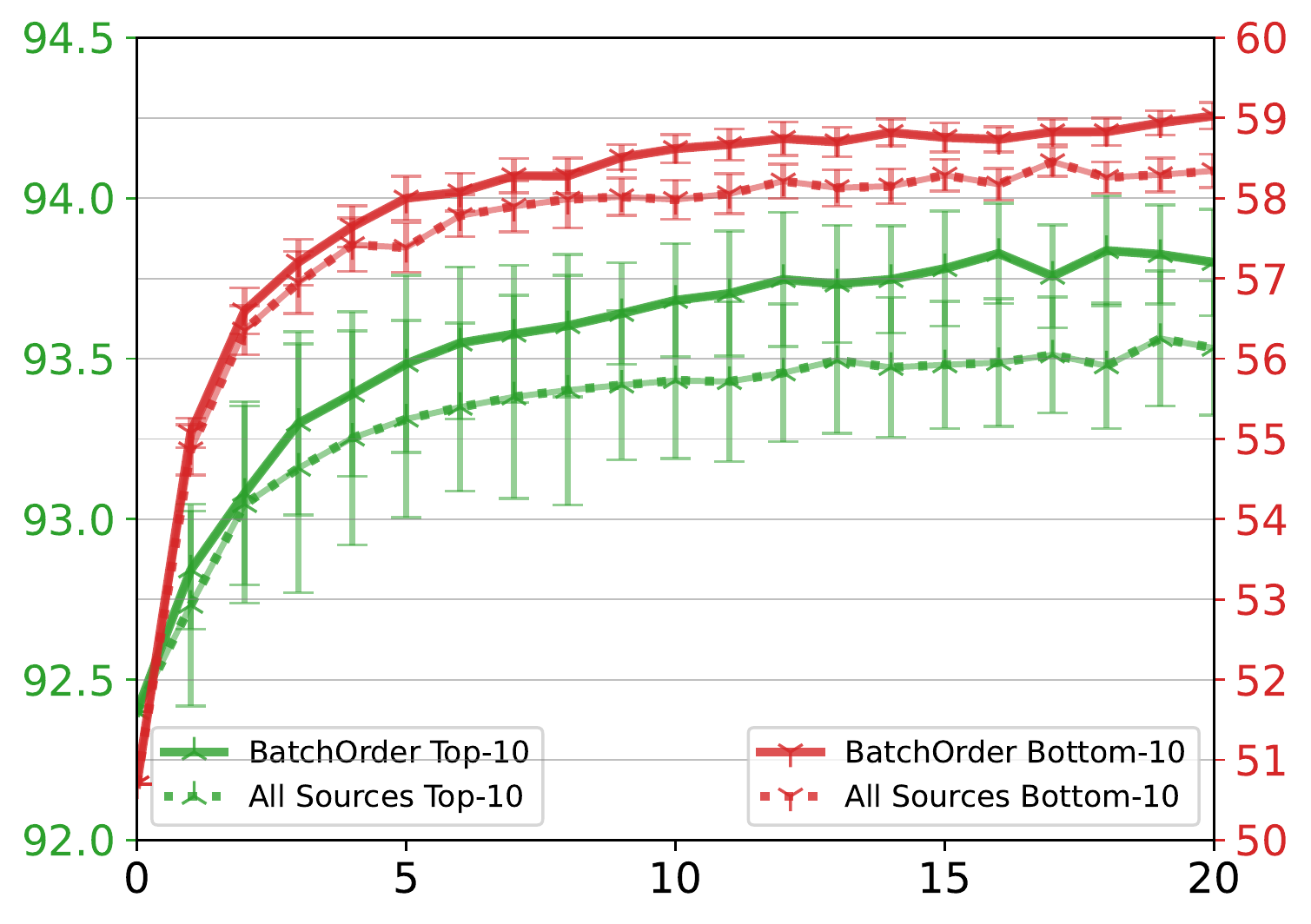}
	\end{subfigure}
	\begin{subfigure}{0.24\linewidth}
		\centering
        DA\\
    	\includegraphics[width=1.0\linewidth]{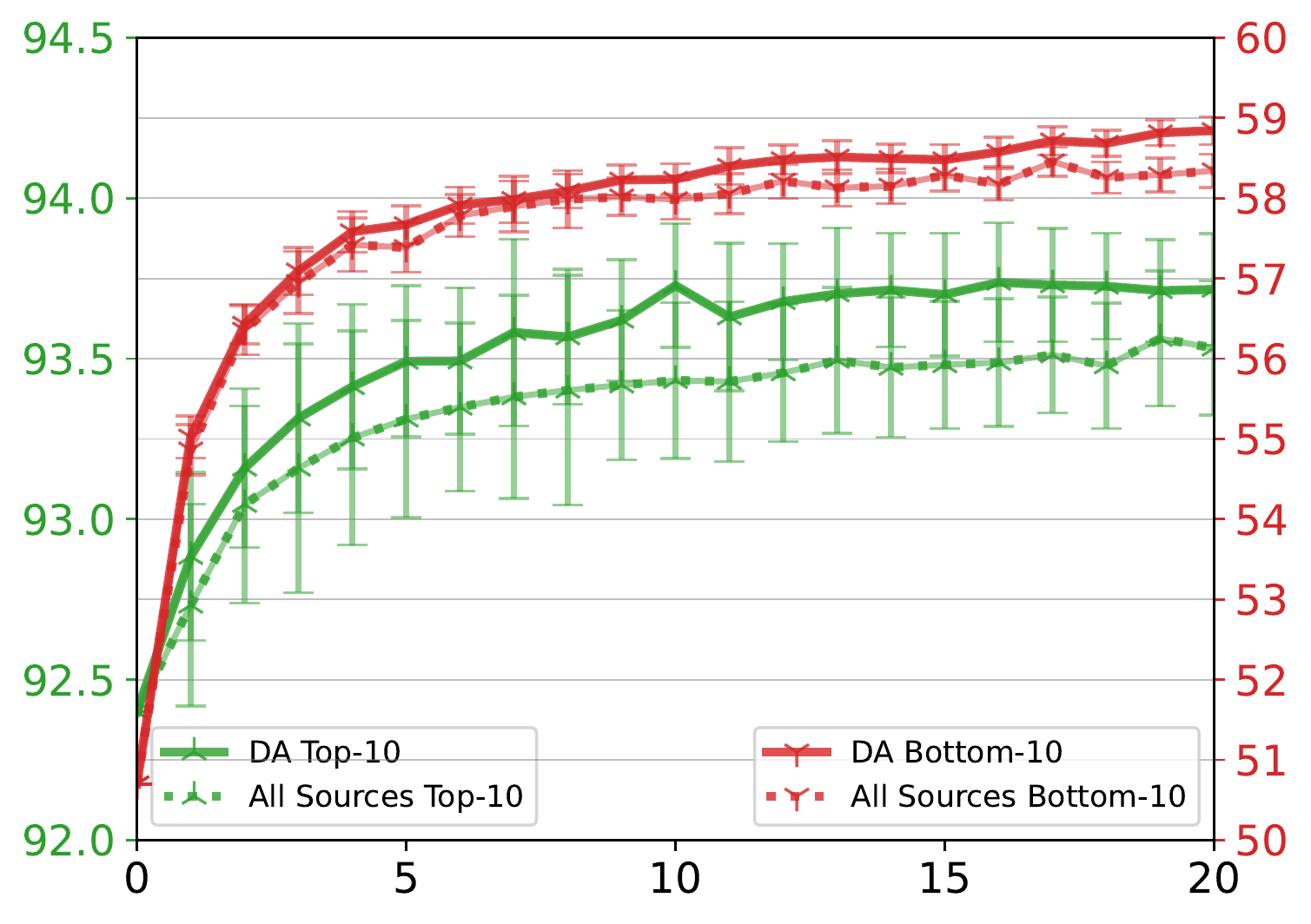}
	\end{subfigure}
 	\begin{subfigure}{0.24\linewidth}
		\centering
        Init \& BatchOrder\\
    	\includegraphics[width=1.0\linewidth]{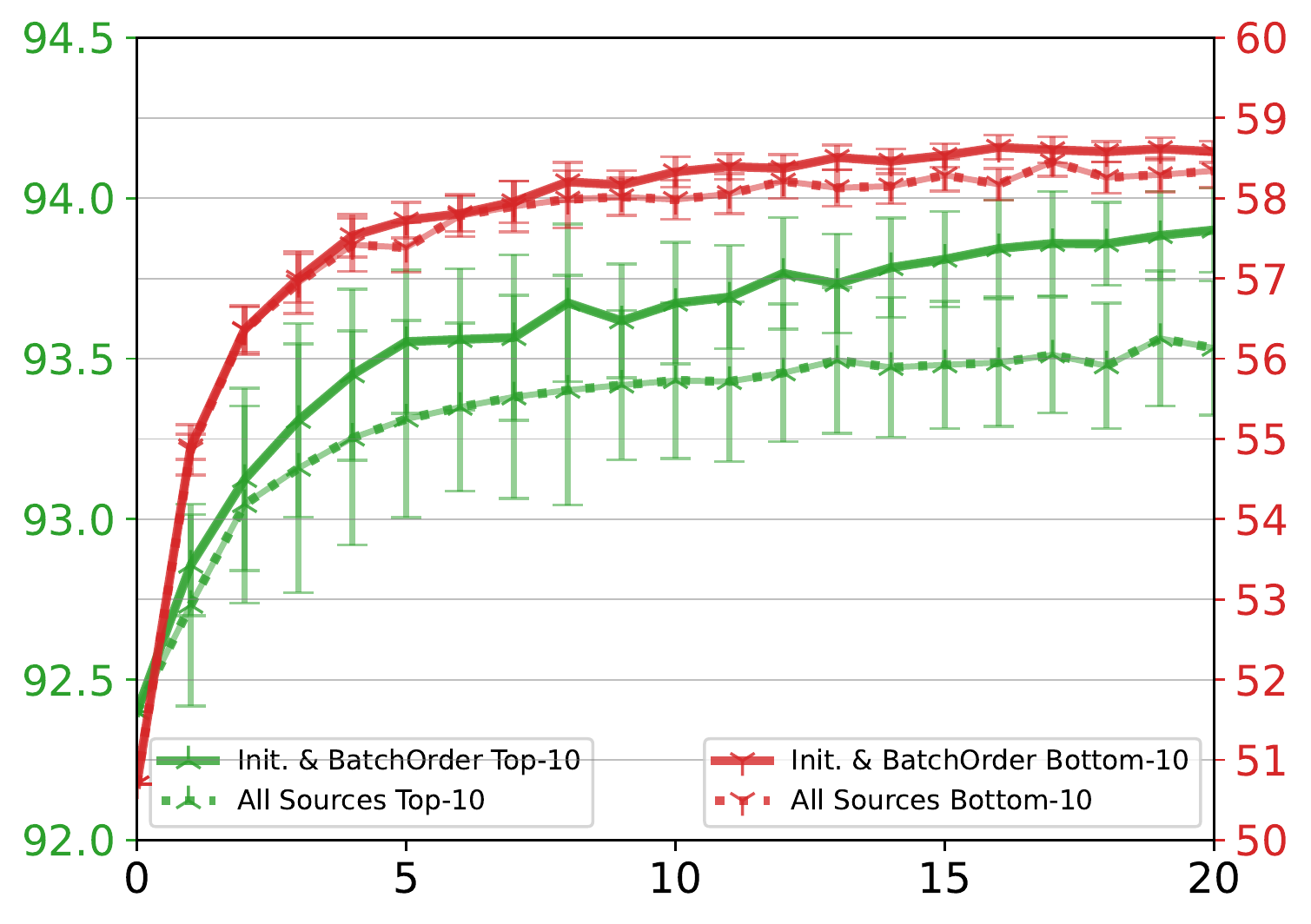}
	\end{subfigure}
    \end{minipage}
\end{figure*}
\begin{figure*}[ht!] \ContinuedFloat
    \centering
    \underline{VGG16}\\
    \vspace{0.1cm}

    \begin{minipage}{0.01\linewidth}
        \rotatebox{90}{\hspace{0.2cm} ensemble/base}
    \end{minipage}
    \begin{minipage}{0.98\linewidth}
	\begin{subfigure}{0.24\linewidth}
		\centering
        Init\\
    	\includegraphics[width=1.0\linewidth]{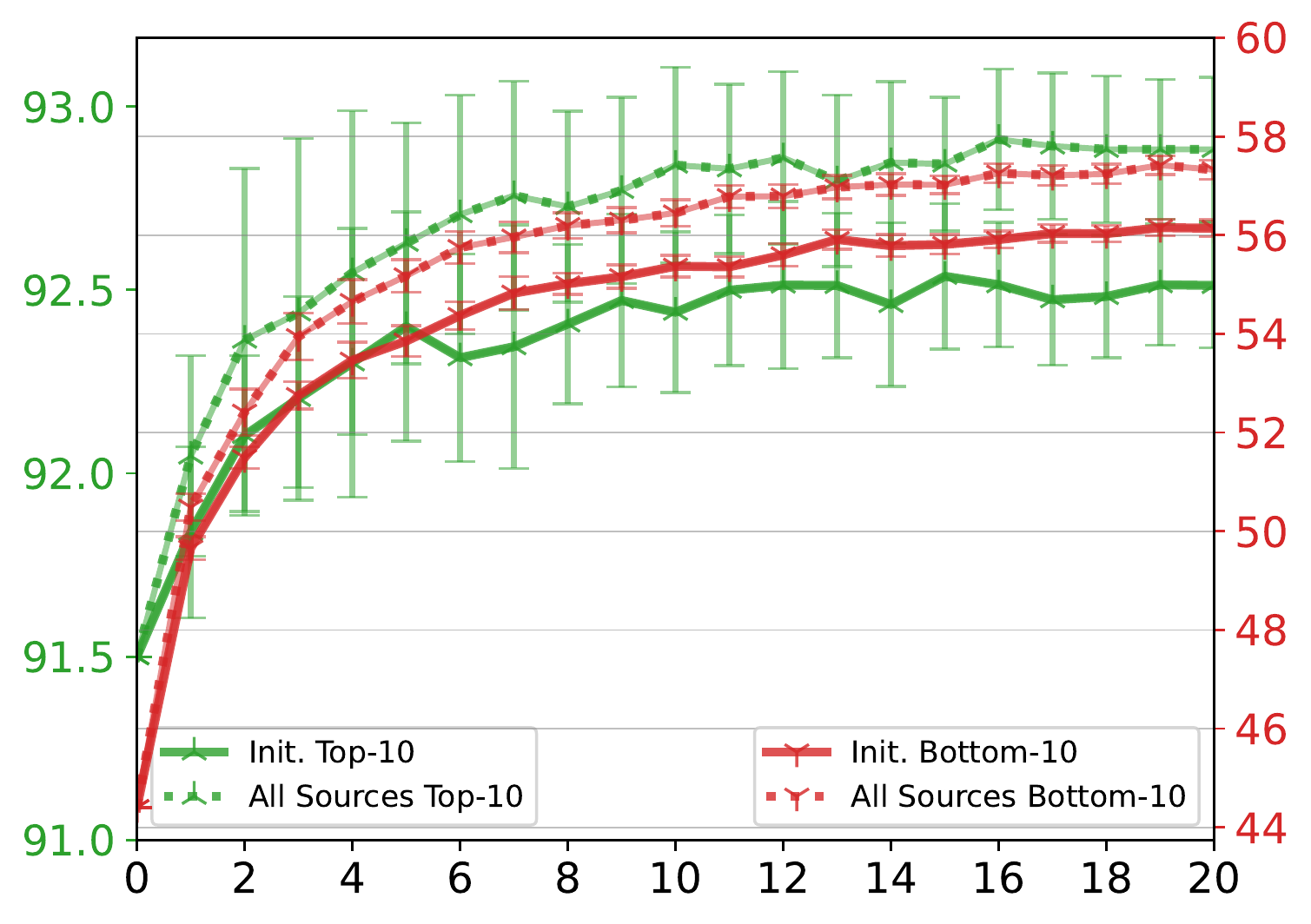}
	\end{subfigure}
 	\begin{subfigure}{0.24\linewidth}
		\centering
        BatchOrder\\
    	\includegraphics[width=1.0\linewidth]{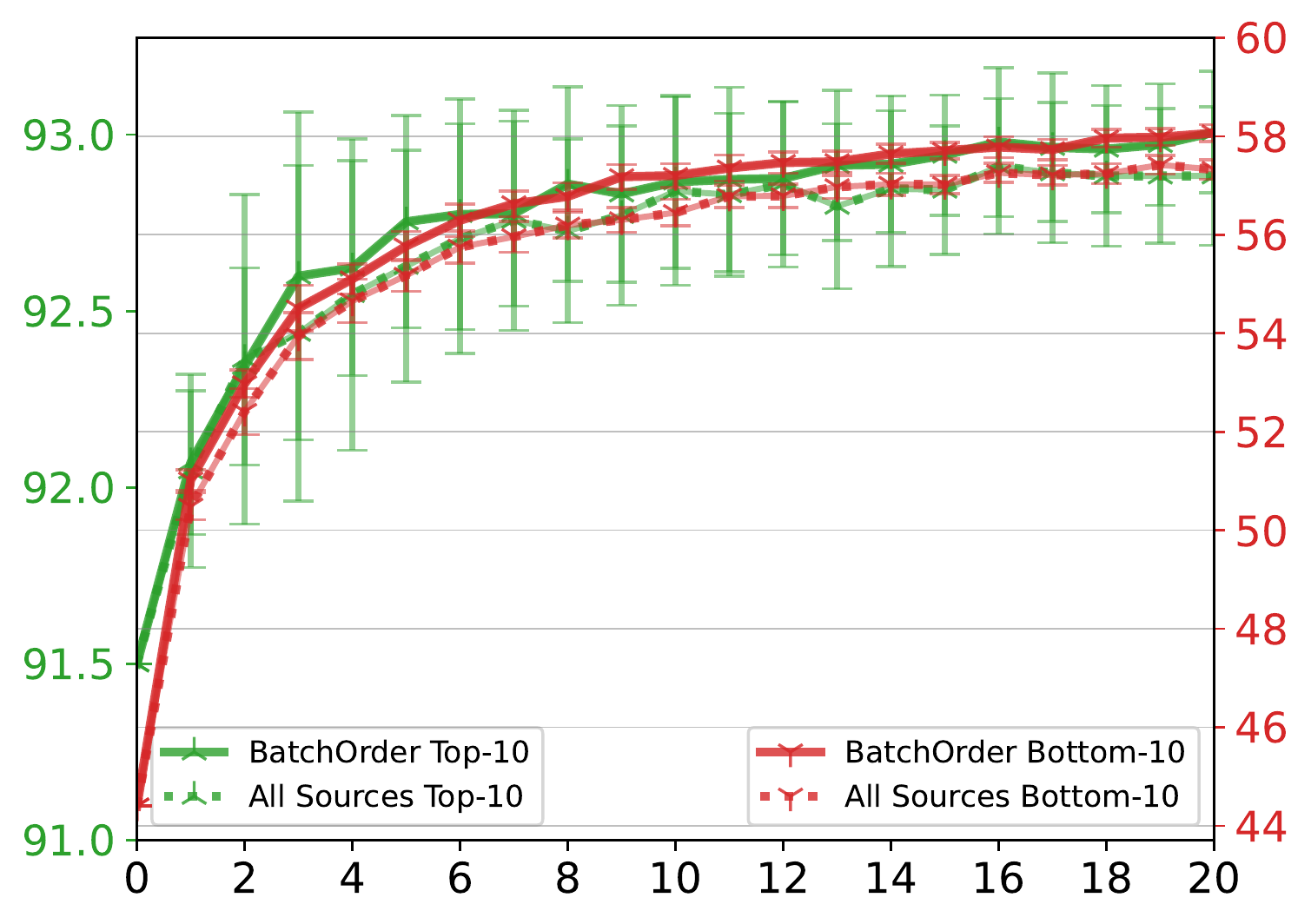}
	\end{subfigure}
	\begin{subfigure}{0.24\linewidth}
		\centering
        DA\\
    	\includegraphics[width=1.0\linewidth]{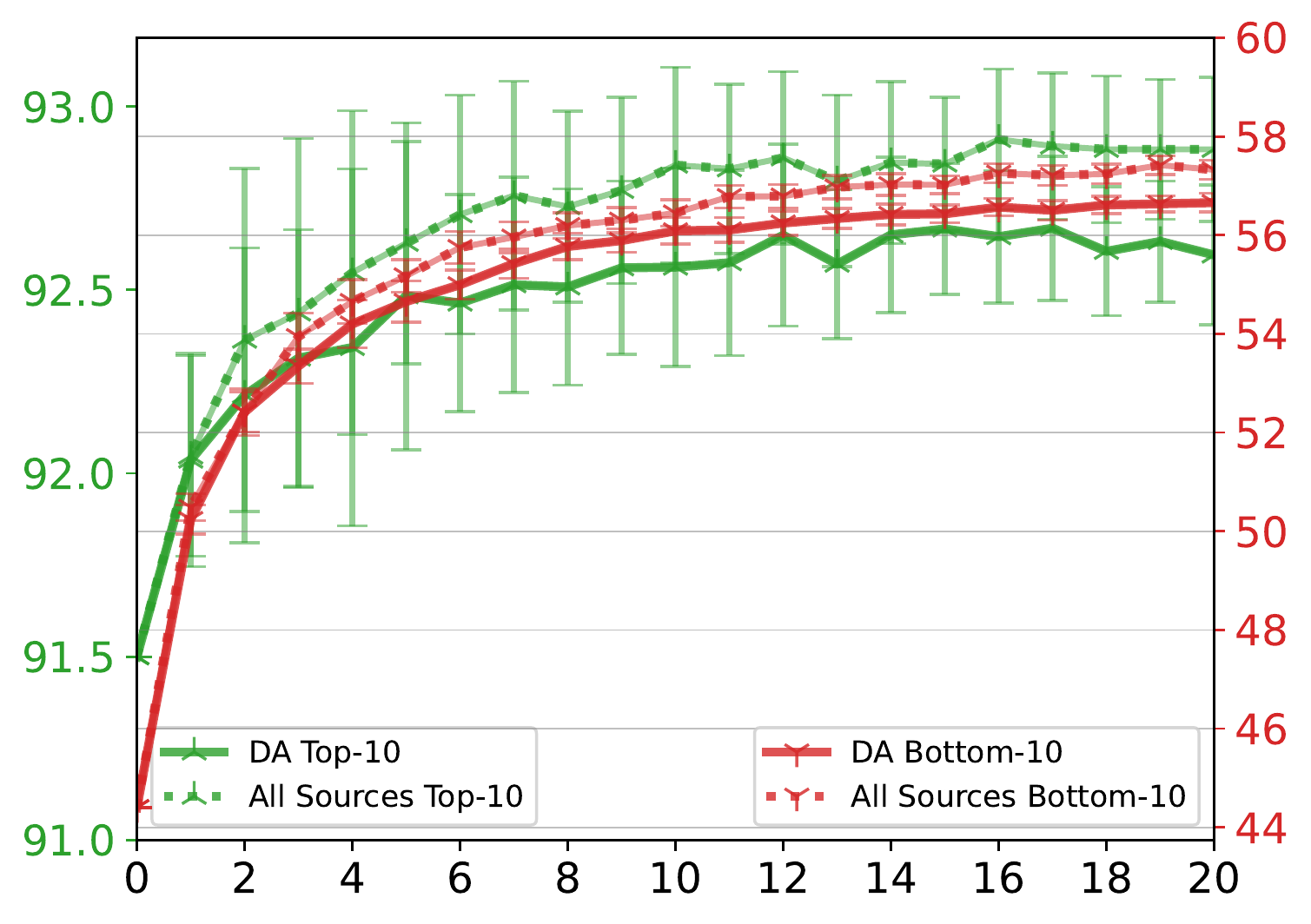}
	\end{subfigure}
 	\begin{subfigure}{0.24\linewidth}
		\centering
        Init \& BatchOrder
    	\includegraphics[width=1.0\linewidth]{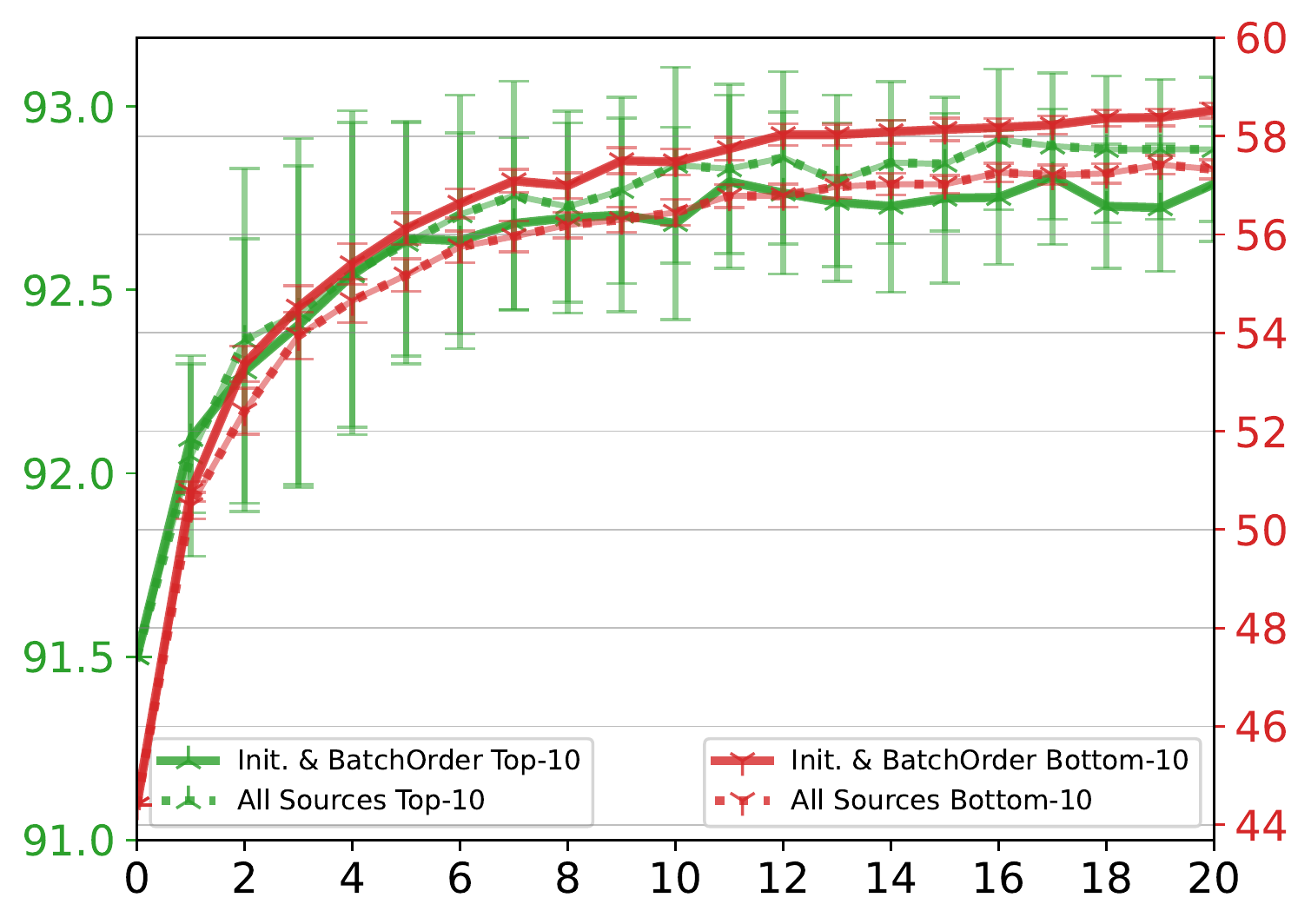}
	\end{subfigure}
    \end{minipage}

    \centering
    \underline{MLP-Mixer}\\
    \vspace{0.1cm}
    \begin{minipage}{0.01\linewidth}
        \rotatebox{90}{ensemble/base}
    \end{minipage}
    \begin{minipage}{0.98\linewidth}
	\begin{subfigure}{0.24\linewidth}
		\centering
        Init\\
    	\includegraphics[width=1.0\linewidth]{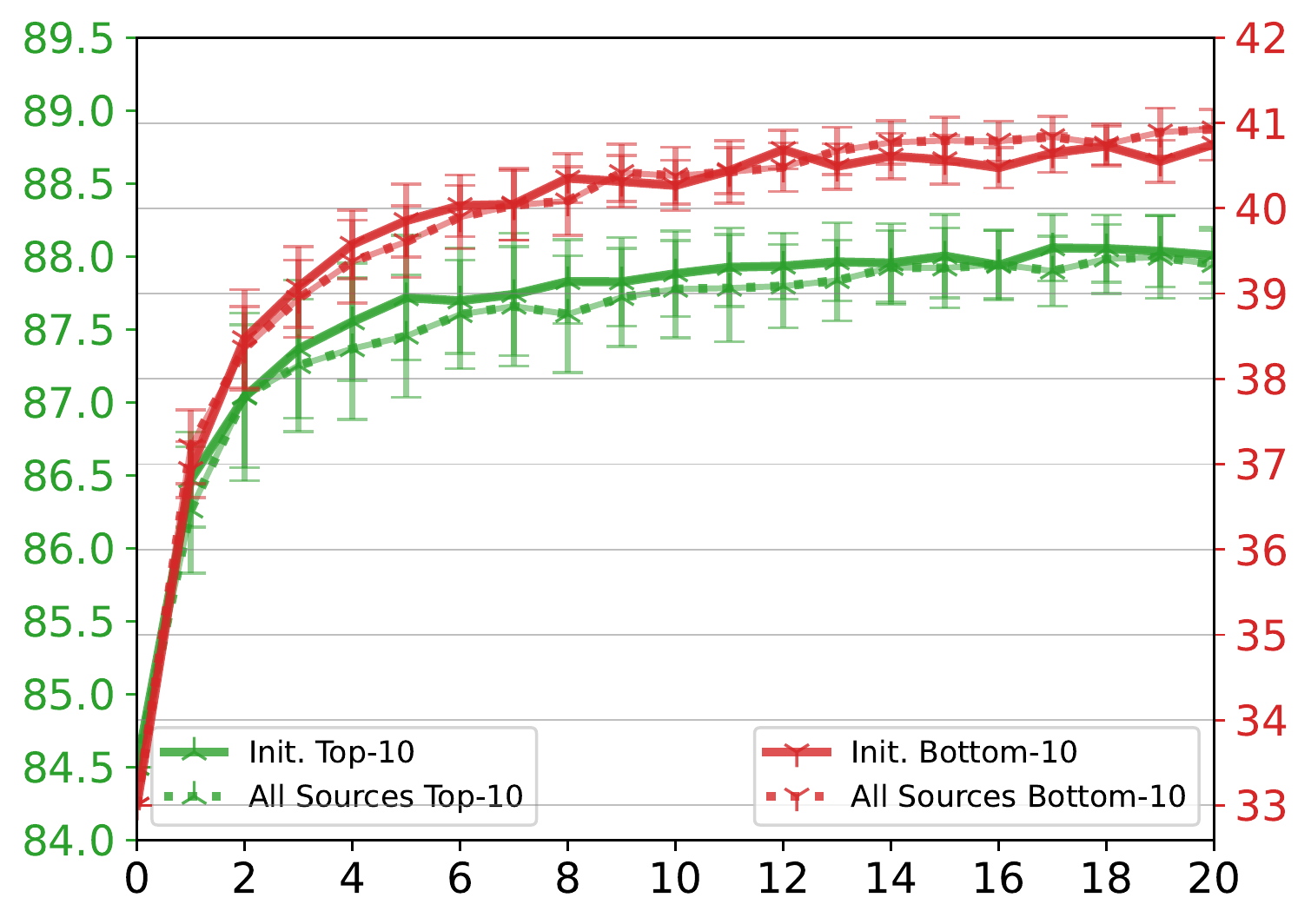}
        \\[-0.7em]
    	{\hspace{0.2cm} \small models in ensemble}
        \label{fig:c100_mlp_mixer_combined_Change_ModelInit}
	\end{subfigure}
 	\begin{subfigure}{0.24\linewidth}
		\centering
        BatchOrder\\
    	\includegraphics[width=1.0\linewidth]{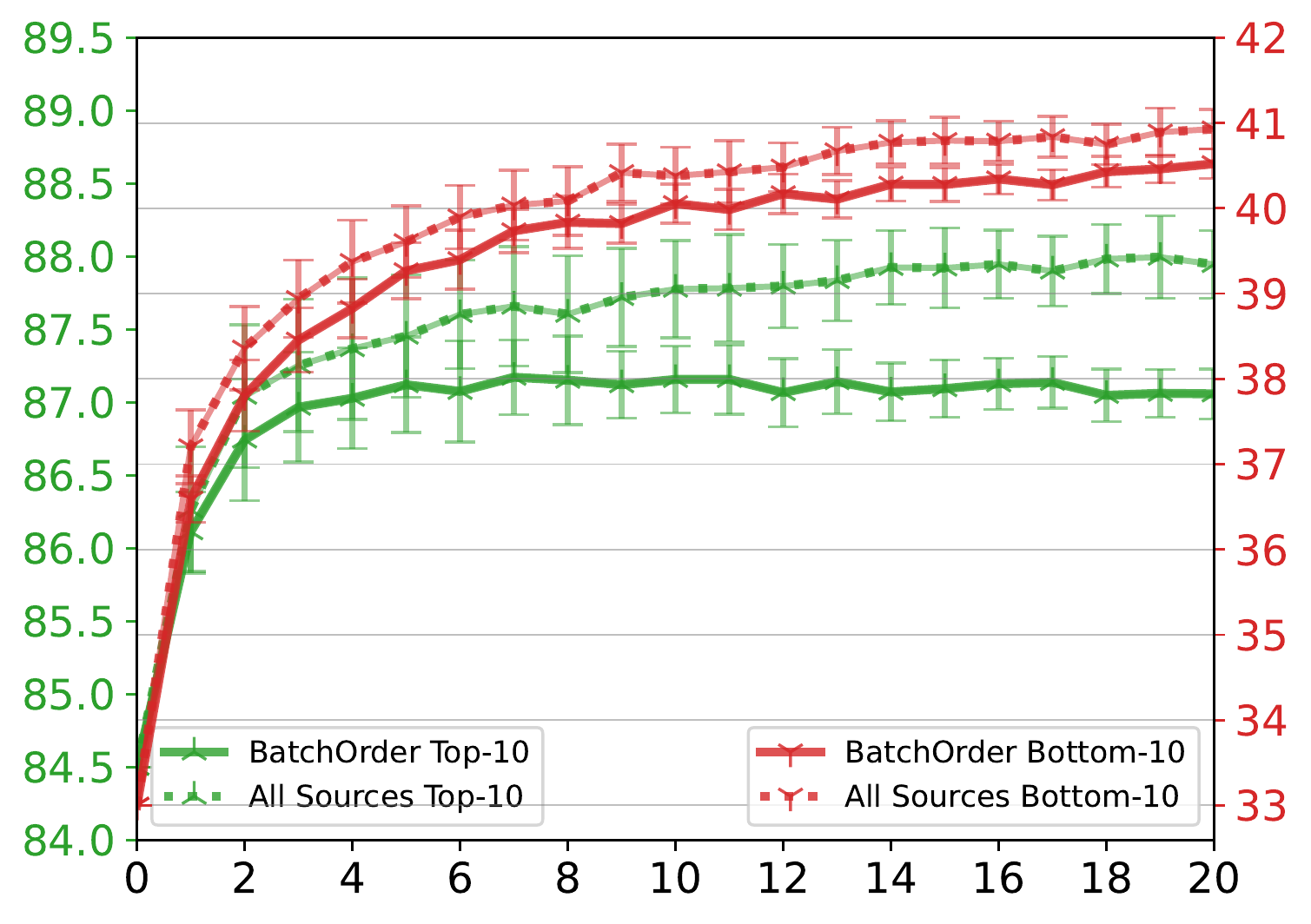}
        \\[-0.7em]
    	{\hspace{0.2cm} \small models in ensemble}
        \label{fig:c100_mlp_mixer_combined_Change_BatchOrder}
	\end{subfigure}
	\begin{subfigure}{0.24\linewidth}
		\centering
        DA\\
    	\includegraphics[width=1.0\linewidth]{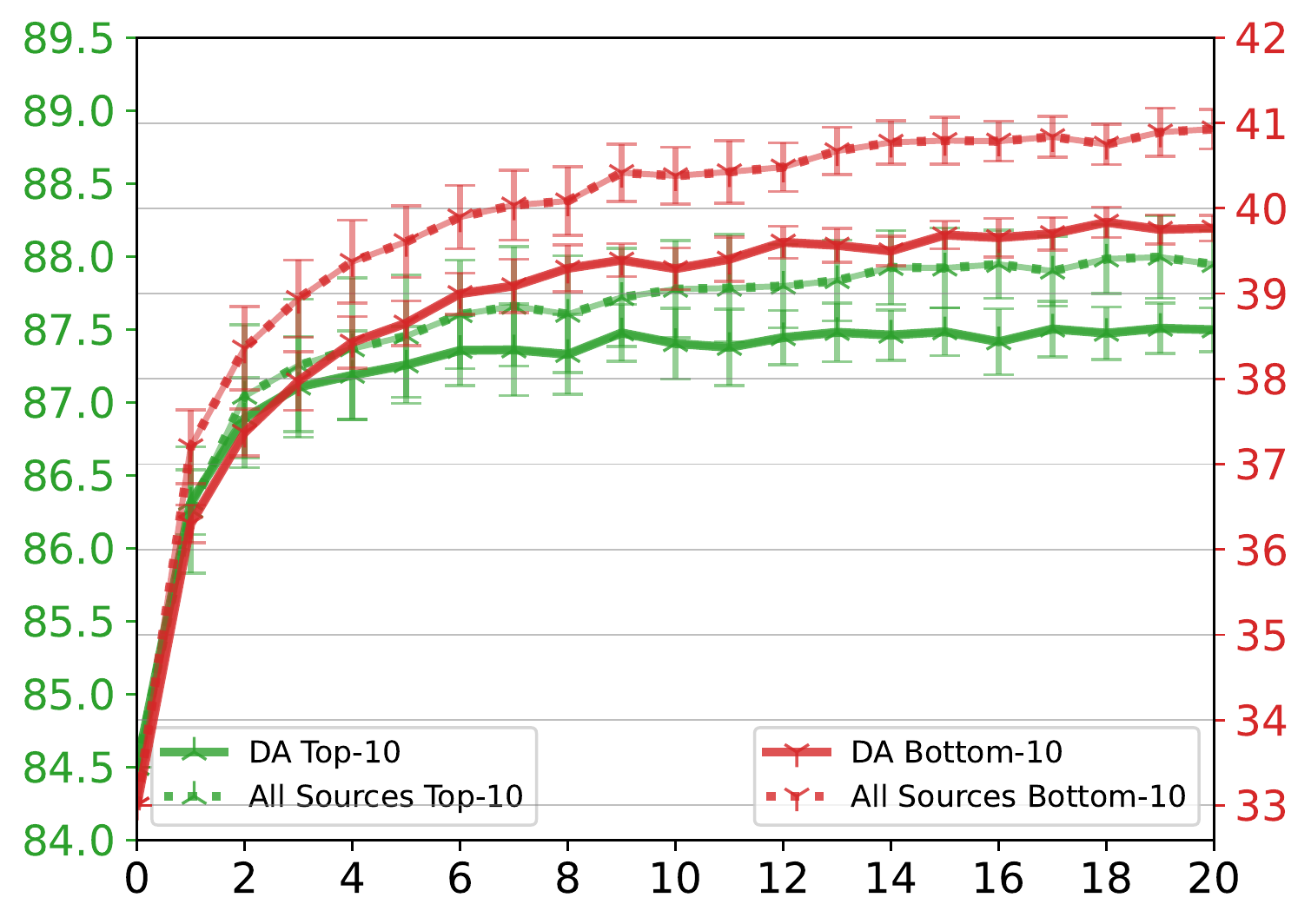}
        \\[-0.7em]
    	{\hspace{0.2cm} \small models in ensemble}
        \label{fig:c100_mlp_mixer_combined_Change_DA}
	\end{subfigure}
 	\begin{subfigure}{0.24\linewidth}
		\centering
        Init \& BatchOrder\\
    	\includegraphics[width=1.0\linewidth]{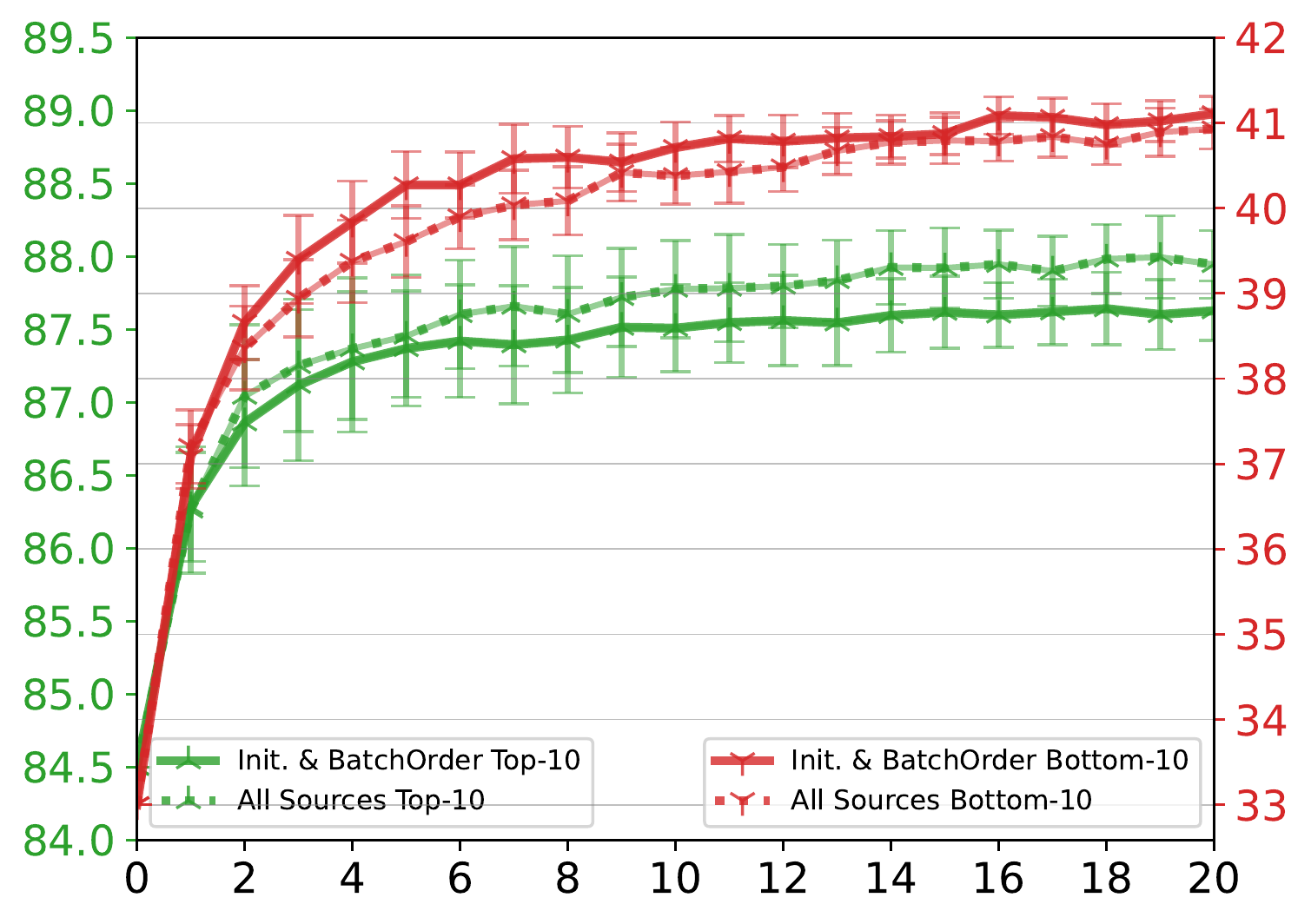}
        \\[-0.7em]
    	{\hspace{0.2cm} \small models in ensemble}
        \label{fig:c100_mlp_mixer_combined_Change_ModelInit_BatchOrder}
	\end{subfigure}
    \end{minipage}

	\caption{ Accuracy for Top-K and Bottom-K across models added to ensemble on CIFAR100}
    \label{fig:CIFAR100_dualaxis}
\end{figure*}

\begin{figure*}[ht!]
    \subsection{TinyImageNet}

    \centering
    \underline{ResNet9}\\
    \vspace{0.1cm}
    
    \begin{minipage}{0.01\linewidth}
        \rotatebox{90}{\hspace{0.2cm} ensemble/base}
    \end{minipage}
    \begin{minipage}{0.98\linewidth}
	\begin{subfigure}{0.24\linewidth}
		\centering
        Init\\
    	\includegraphics[width=1.0\linewidth]{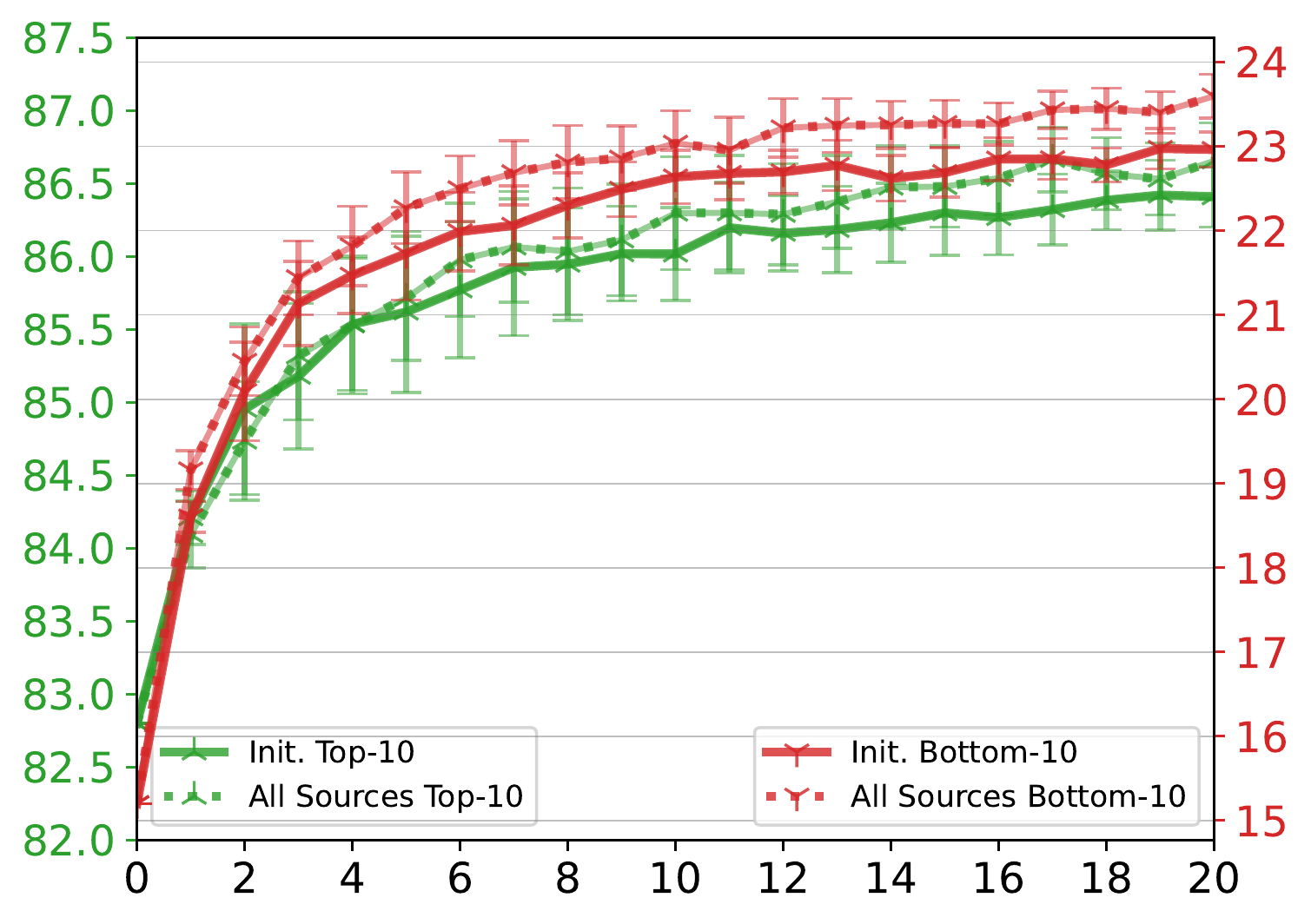}
        \label{fig:Tinyimagenet_resnet9_combined_Change_ModelInit}
	\end{subfigure}
 	\begin{subfigure}{0.24\linewidth}
		\centering
        BatchOrder\\
    	\includegraphics[width=1.0\linewidth]{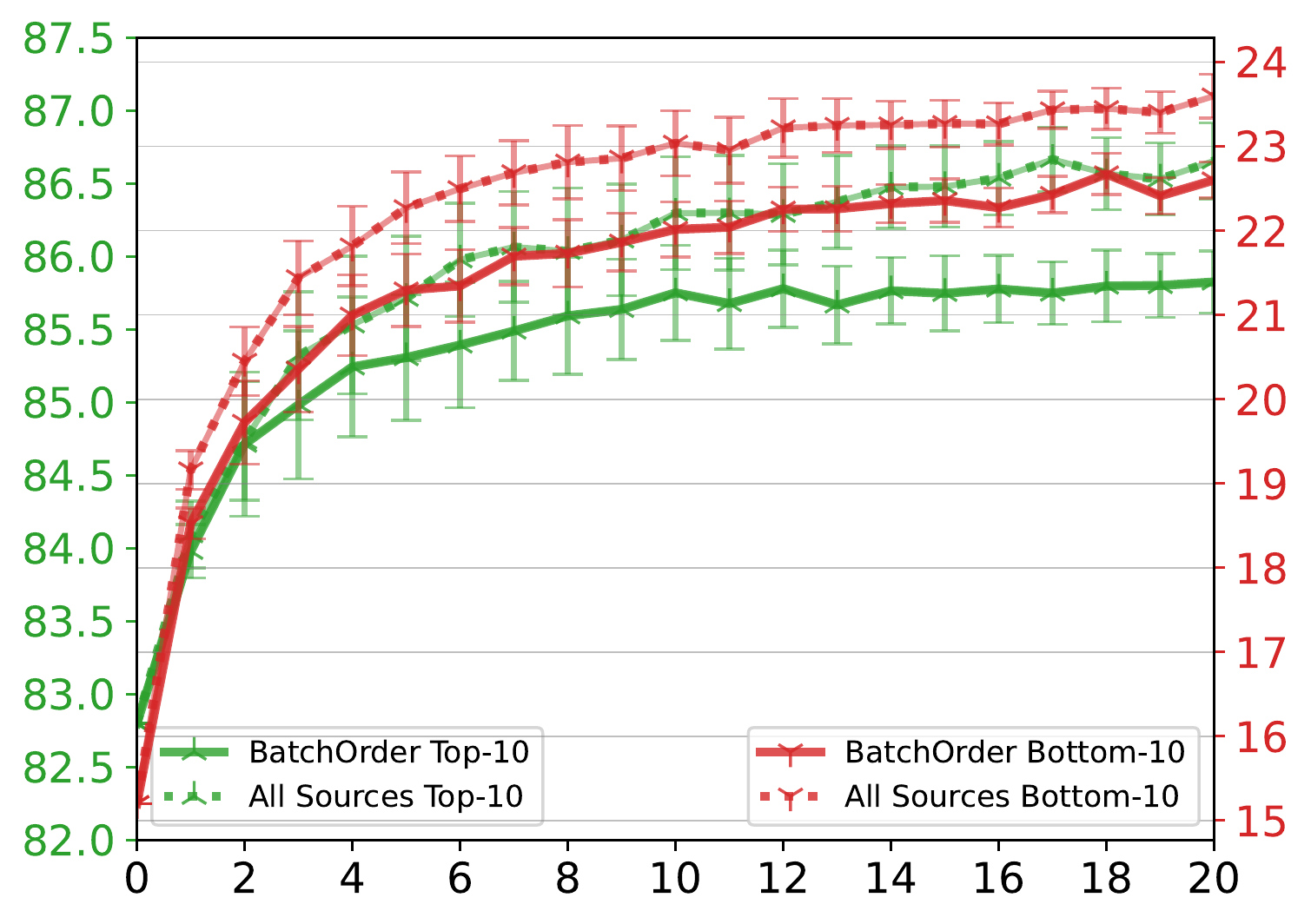}
        \label{fig:Tinyimagenet_resnet9_combined_Change_BatchOrder}
	\end{subfigure}
	\begin{subfigure}{0.24\linewidth}
		\centering
        DA\\
    	\includegraphics[width=1.0\linewidth]{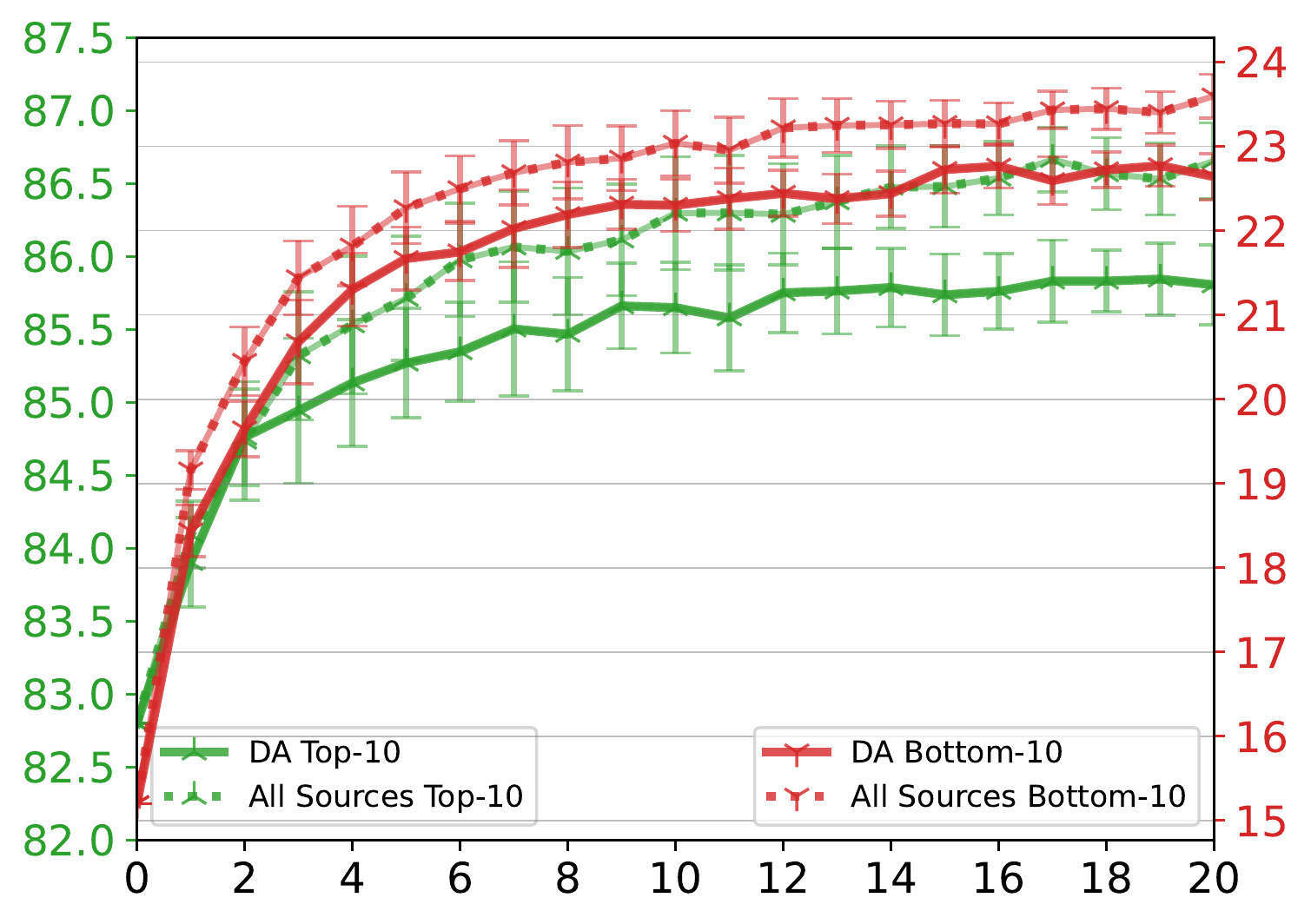}
        \label{fig:Tinyimagenet_resnet9_combined_Change_DA}
	\end{subfigure}
 	\begin{subfigure}{0.24\linewidth}
		\centering
        Init \& BatchOrder\\
    	\includegraphics[width=1.0\linewidth]{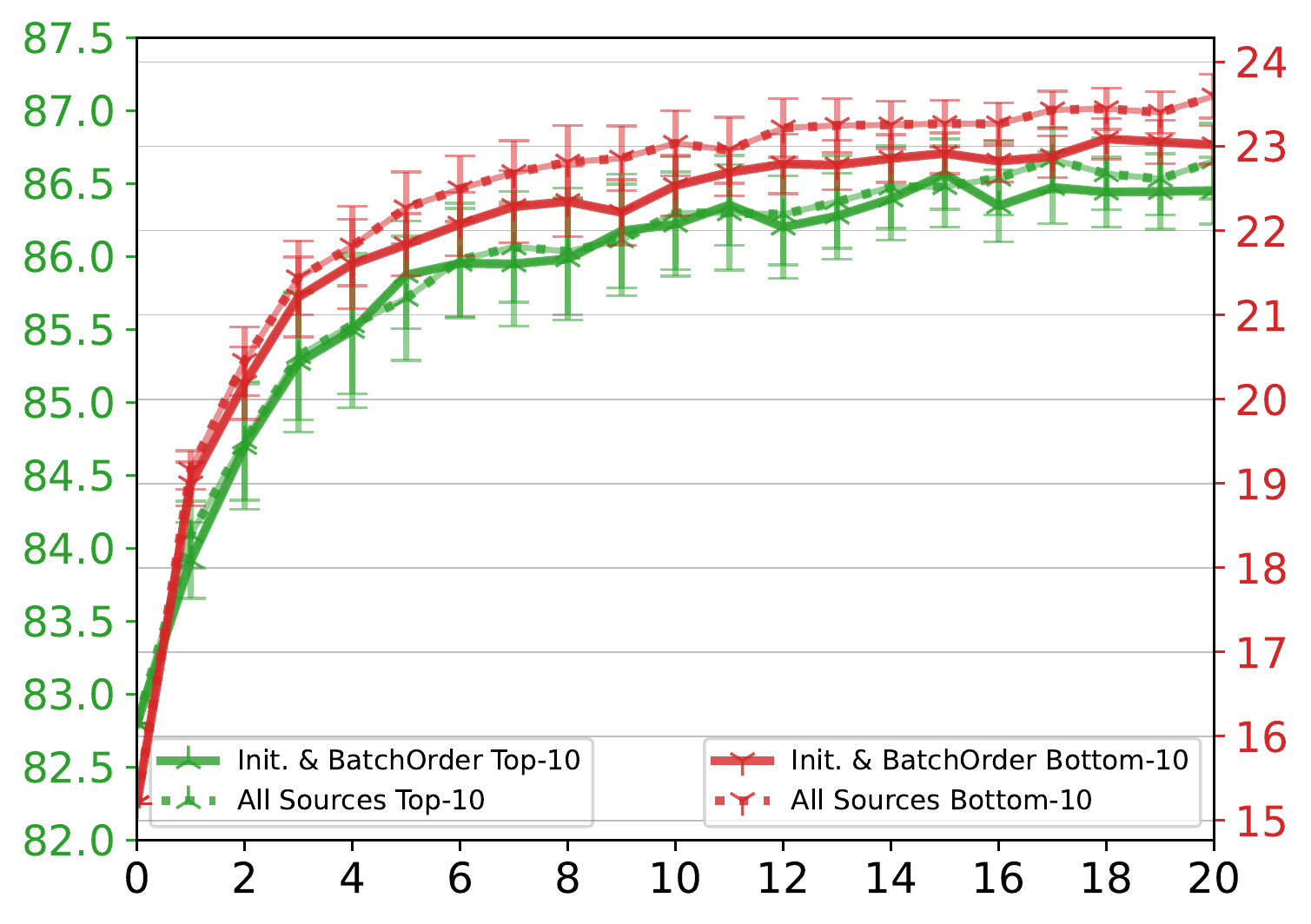}
        \label{fig:Tinyimagenet_resnet9_combined_Change_ModelInit_BatchOrder}
	\end{subfigure}
    \end{minipage}

    \centering
    \underline{ResNet18}\\
    \vspace{0.1cm}
    
    \begin{minipage}{0.01\linewidth}
        \rotatebox{90}{\hspace{0.2cm} ensemble/base}
    \end{minipage}
    \begin{minipage}{0.98\linewidth}
	\begin{subfigure}{0.24\linewidth}
		\centering
        Init\\
    	\includegraphics[width=1.0\linewidth]{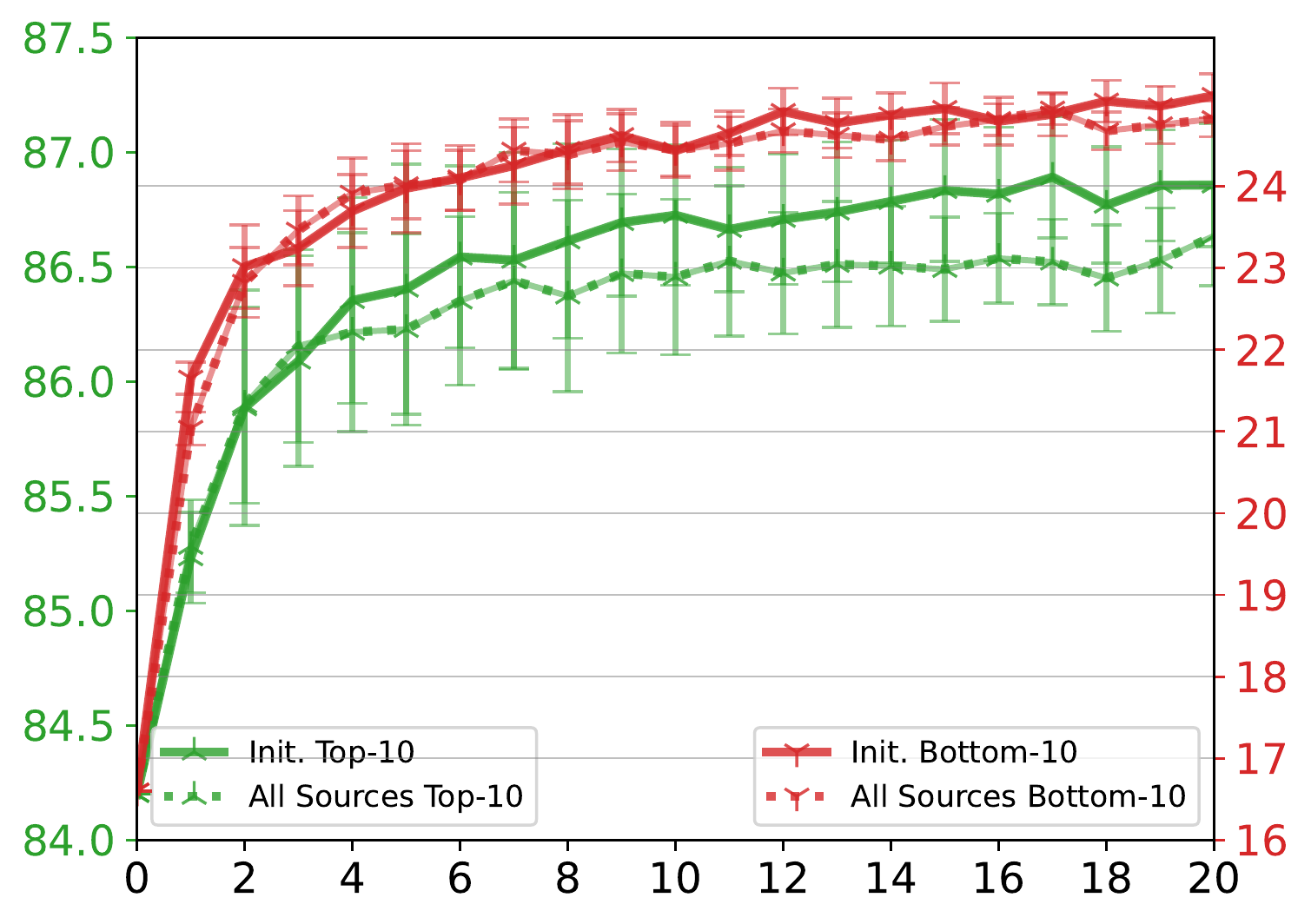}
        \label{fig:Tinyimagenet_resnet18_combined_Change_ModelInit}
	\end{subfigure}
 	\begin{subfigure}{0.24\linewidth}
		\centering
        BatchOrder\\
    	\includegraphics[width=1.0\linewidth]{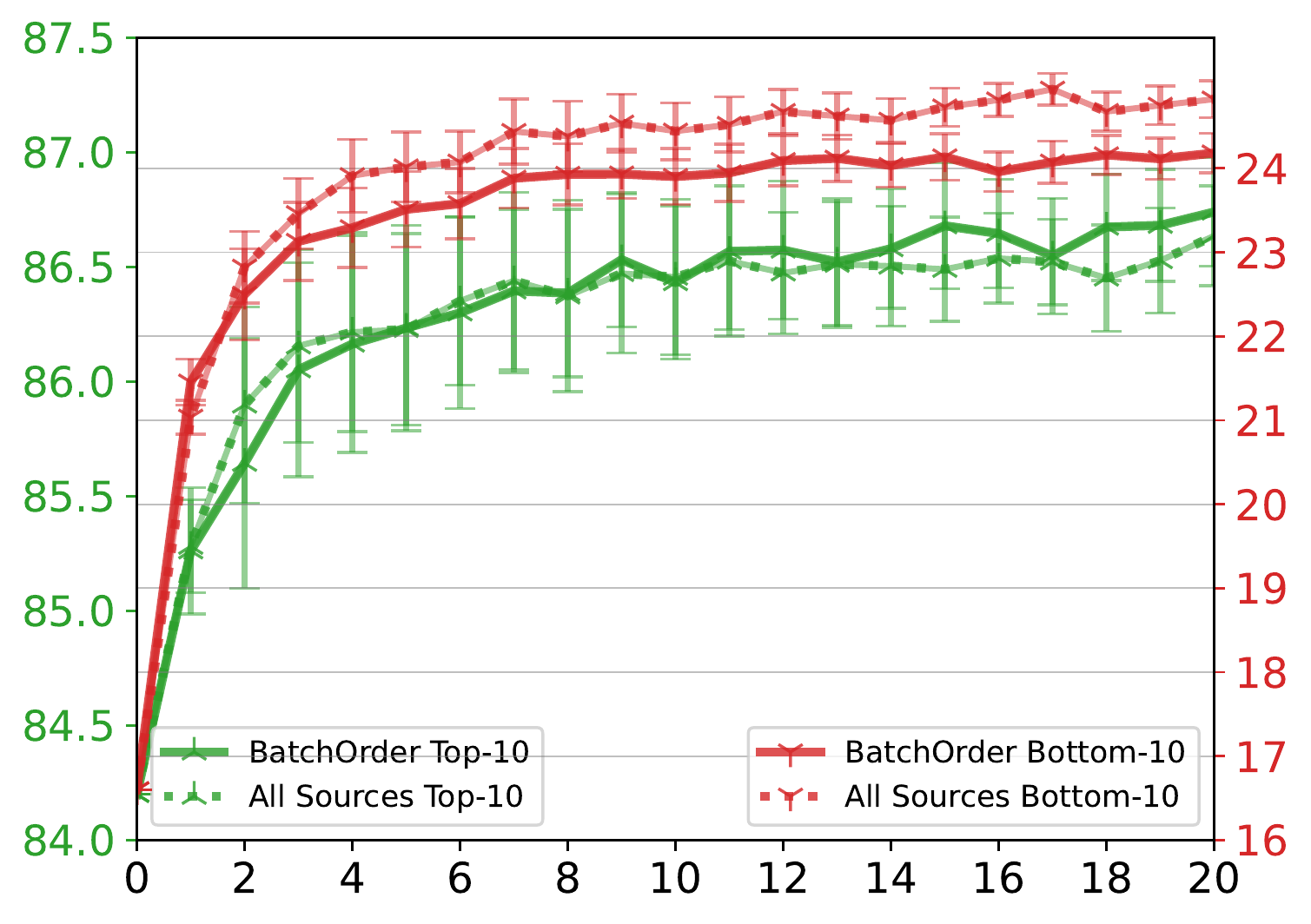}
        \label{fig:Tinyimagenet_resnet18_combined_Change_BatchOrder}
	\end{subfigure}
	\begin{subfigure}{0.24\linewidth}
		\centering
        DA\\
    	\includegraphics[width=1.0\linewidth]{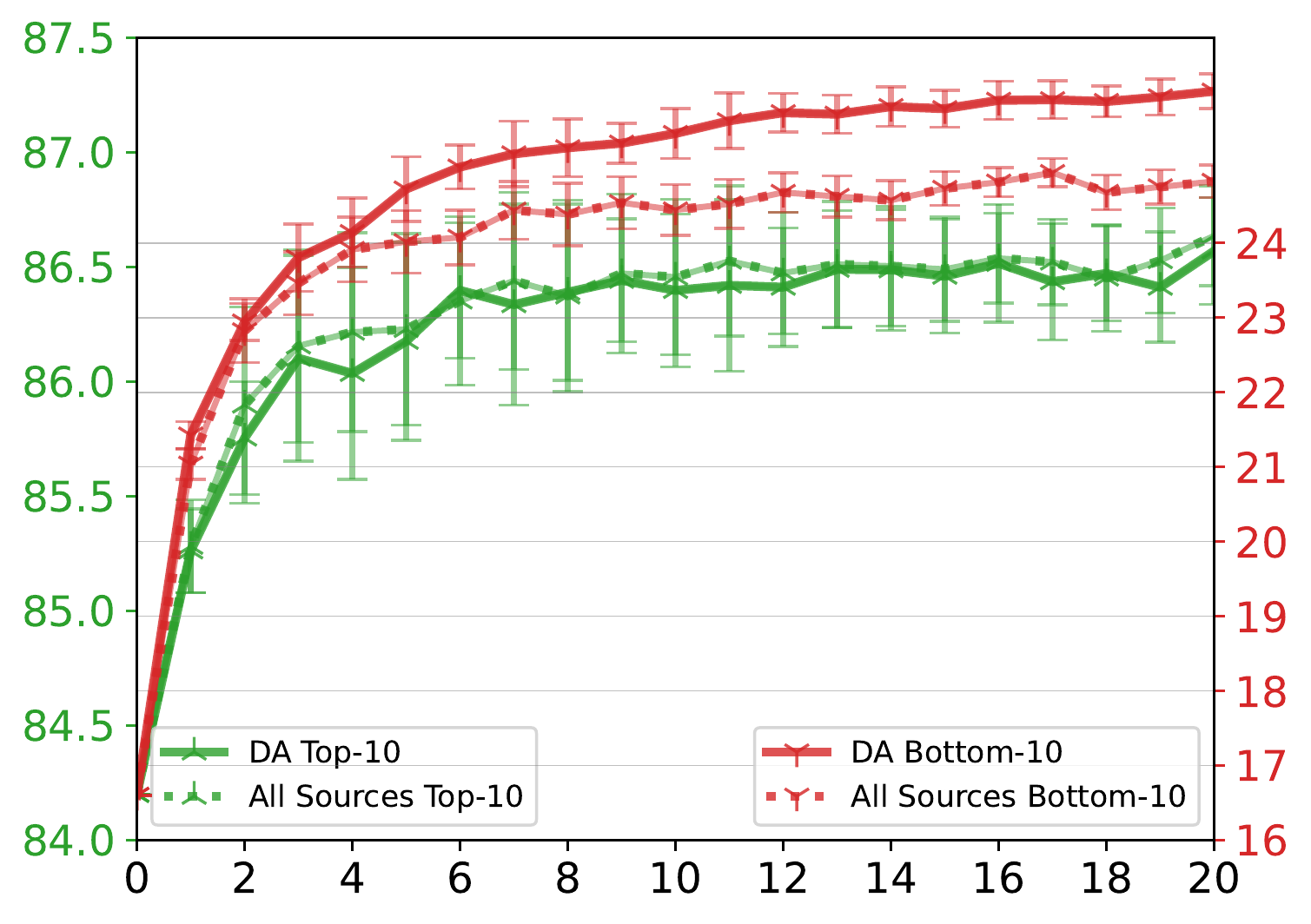}
        \label{fig:Tinyimagenet_resnet18_combined_Change_DA}
	\end{subfigure}
 	\begin{subfigure}{0.24\linewidth}
		\centering
        Init \& BatchOrder\\
    	\includegraphics[width=1.0\linewidth]{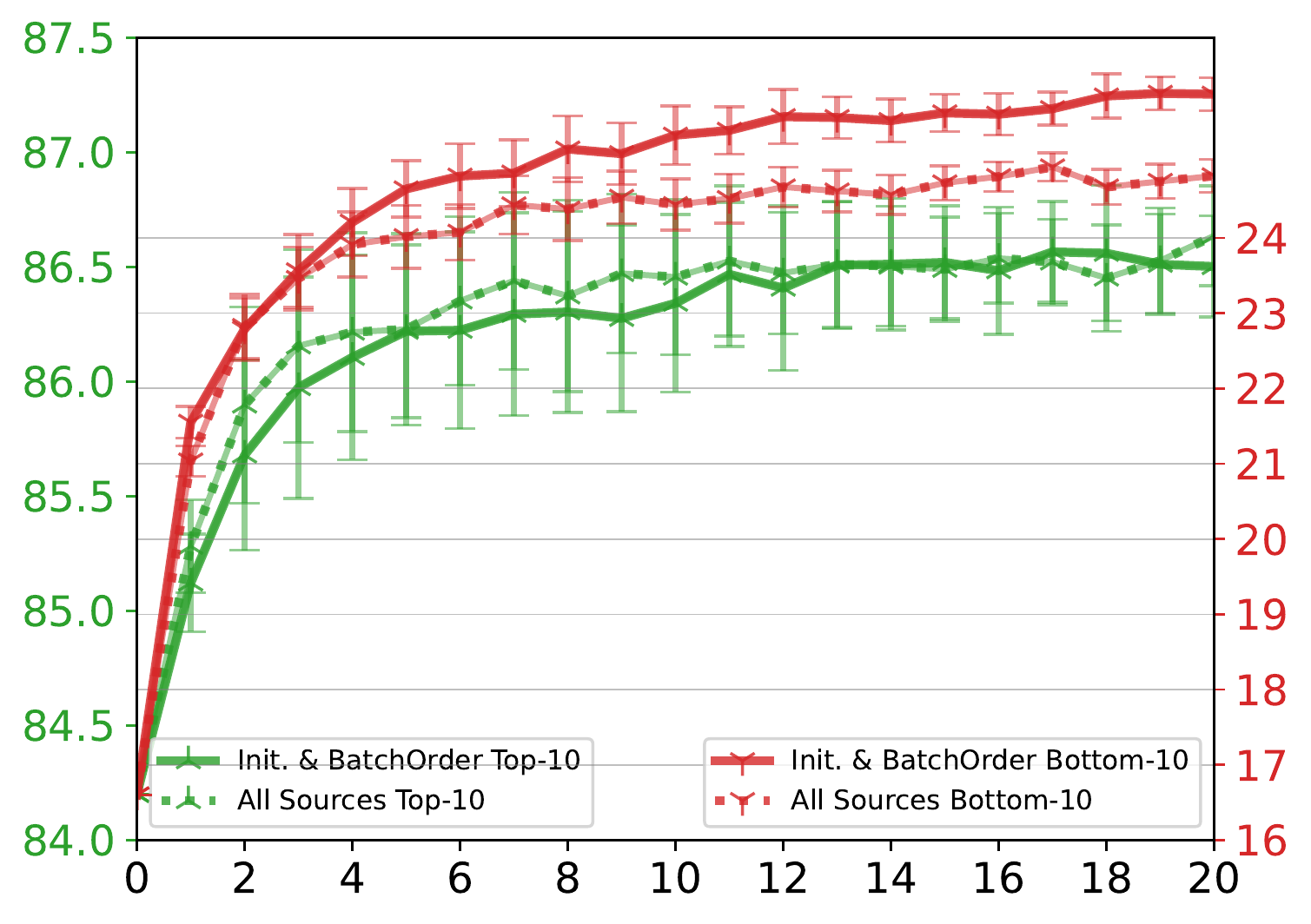}
        \label{fig:Tinyimagenet_resnet18_combined_Change_ModelInit_BatchOrder}
	\end{subfigure}
    \end{minipage}

    \centering
    \underline{ResNet34}\\
    \vspace{0.1cm}
    
    \begin{minipage}{0.01\linewidth}
        \rotatebox{90}{\hspace{0.2cm} ensemble/base}
    \end{minipage}
    \begin{minipage}{0.98\linewidth}
	\begin{subfigure}{0.24\linewidth}
		\centering
        Init\\
    	\includegraphics[width=1.0\linewidth]{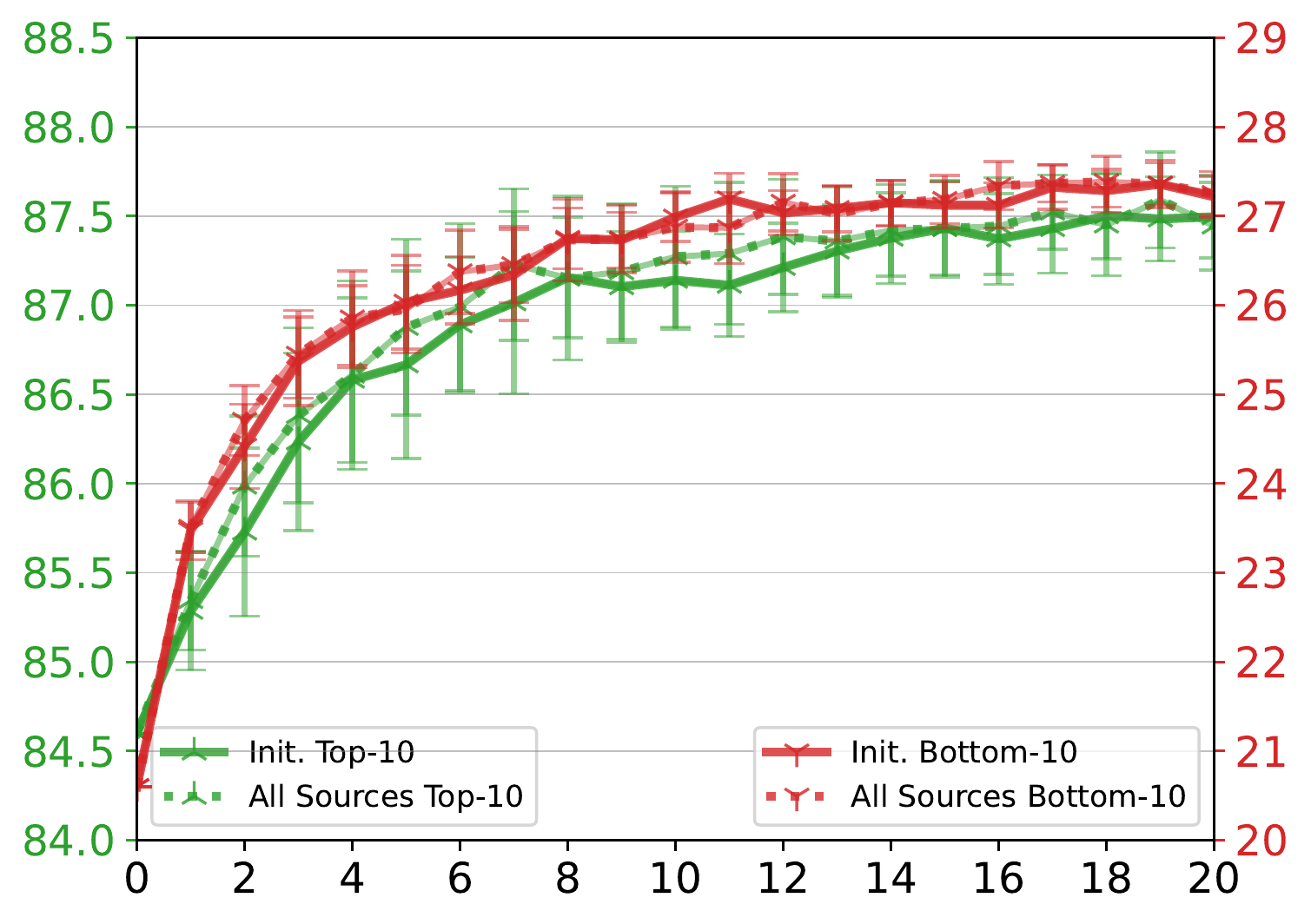}
        \label{fig:Tinyimagenet_resnet34_combined_Change_ModelInit}
	\end{subfigure}
 	\begin{subfigure}{0.24\linewidth}
		\centering
        BatchOrder\\
    	\includegraphics[width=1.0\linewidth]{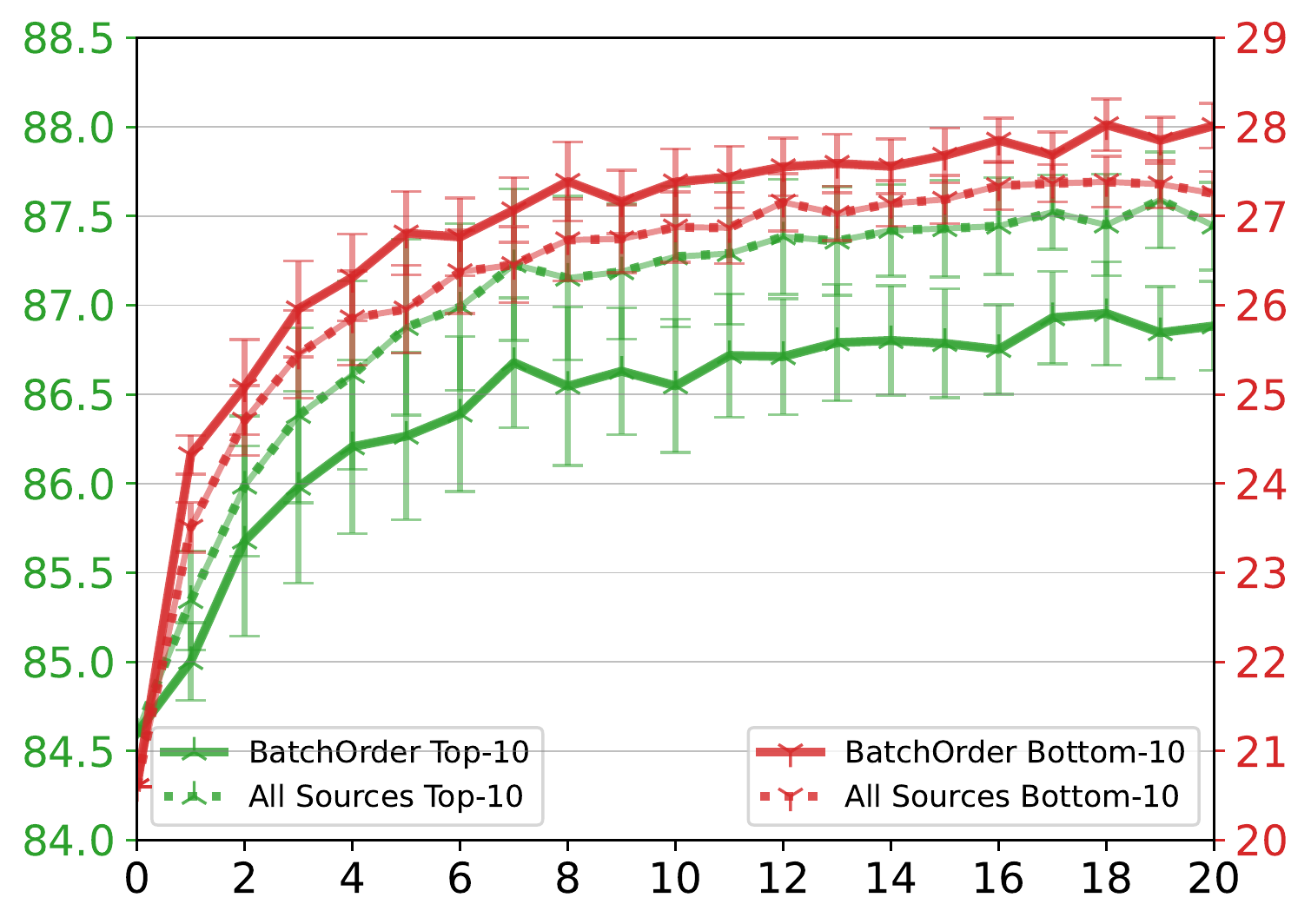}
        \label{fig:Tinyimagenet_resnet34_combined_Change_BatchOrder}
	\end{subfigure}
	\begin{subfigure}{0.24\linewidth}
		\centering
        DA\\
    	\includegraphics[width=1.0\linewidth]{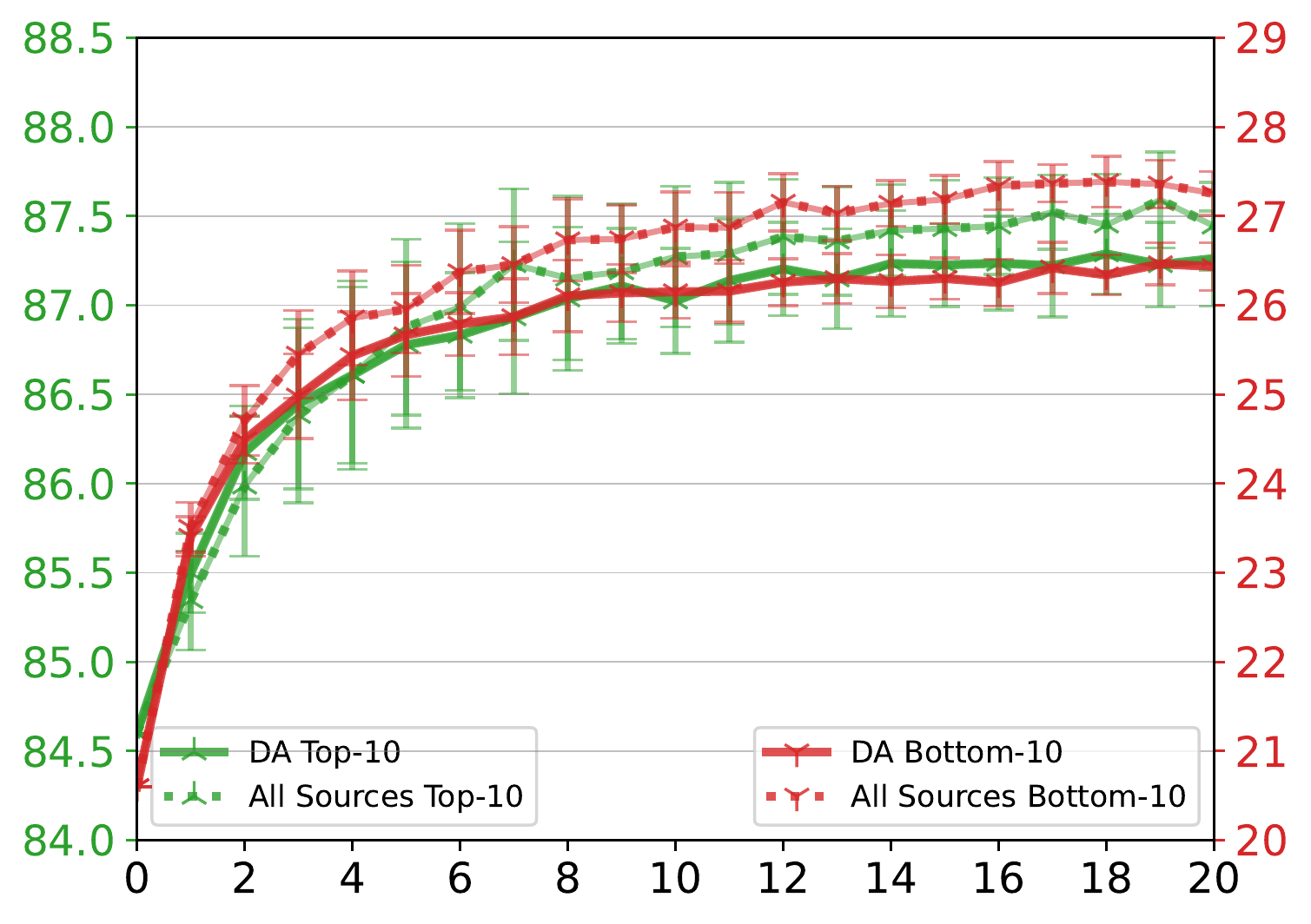}
        \label{fig:Tinyimagenet_resnet34_combined_Change_DA}
	\end{subfigure}
 	\begin{subfigure}{0.24\linewidth}
		\centering
        Init \& BatchOrder\\
    	\includegraphics[width=1.0\linewidth]{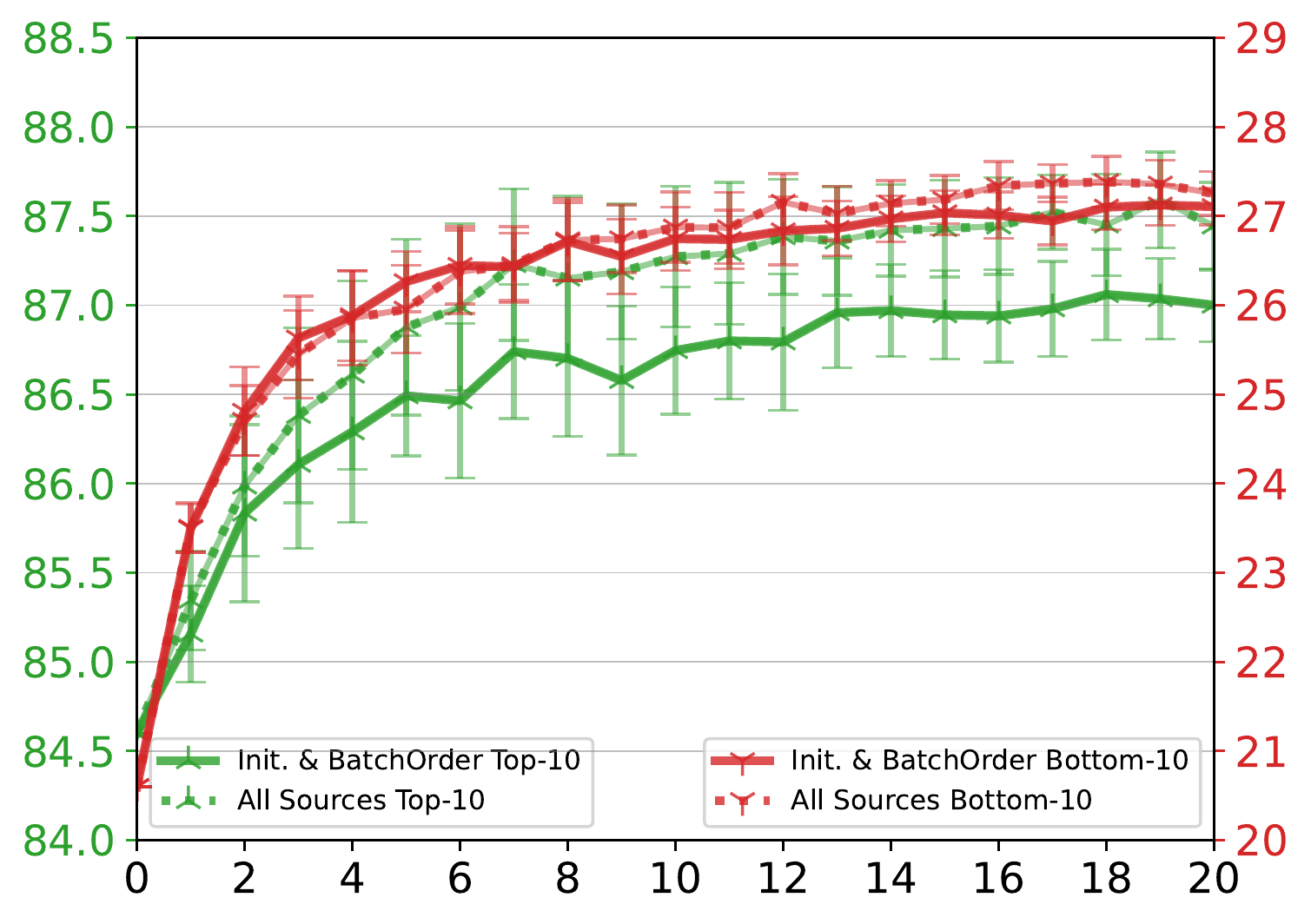}
        \label{fig:Tinyimagenet_resnet34_combined_Change_ModelInit_BatchOrder}
	\end{subfigure}
    \end{minipage}

    \centering
    \underline{ResNet50}\\
    \vspace{0.1cm}
    
    \begin{minipage}{0.01\linewidth}
        \rotatebox{90}{\hspace{0.2cm} ensemble/base}
    \end{minipage}
    \begin{minipage}{0.98\linewidth}
	\begin{subfigure}{0.24\linewidth}
		\centering
        Init\\
    	\includegraphics[width=1.0\linewidth]{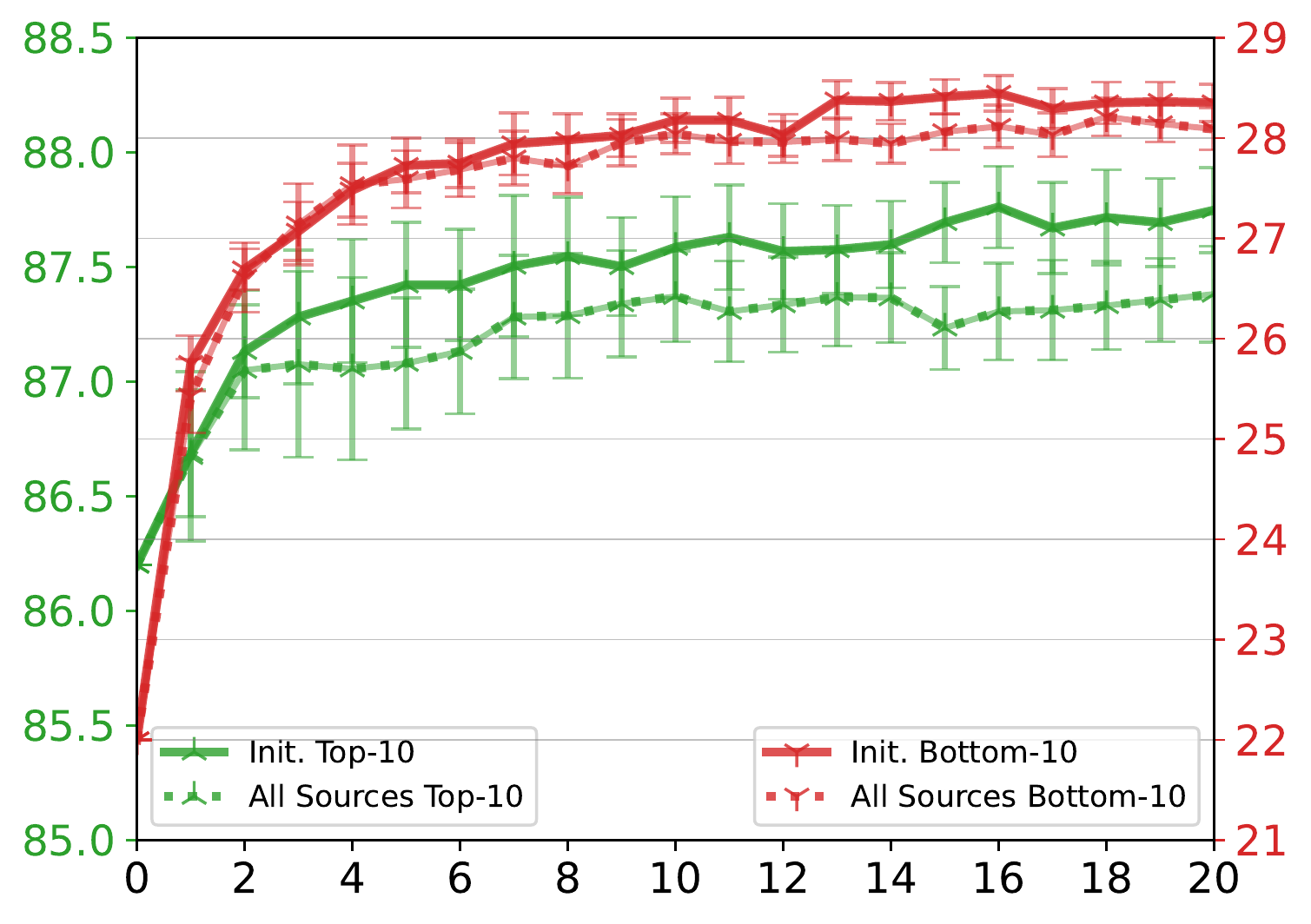}
        \label{fig:Tinyimagenet_resnet50_combined_Change_ModelInit}
	\end{subfigure}
 	\begin{subfigure}{0.24\linewidth}
		\centering
        BatchOrder\\
    	\includegraphics[width=1.0\linewidth]{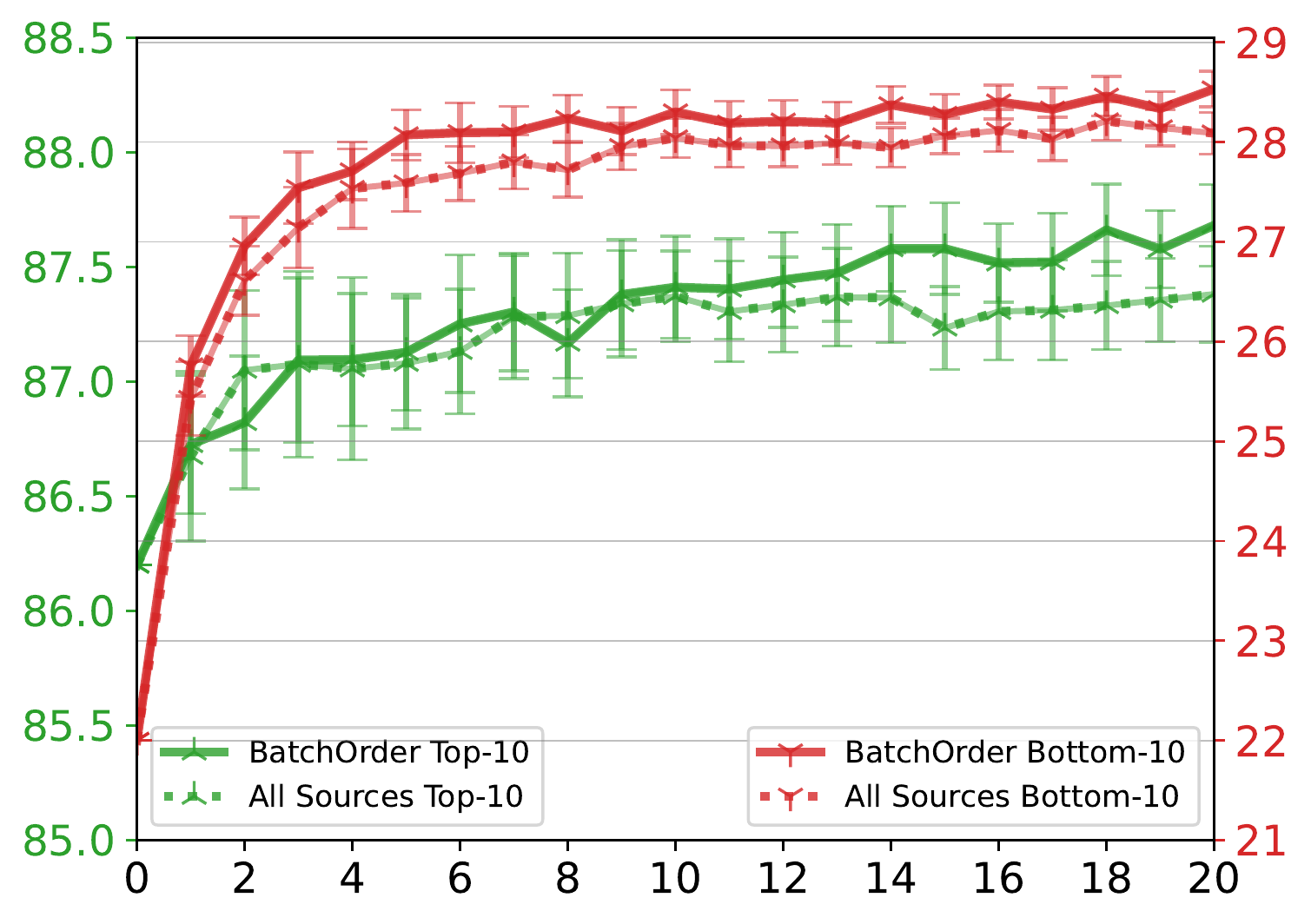}
        \label{fig:Tinyimagenet_resnet50_combined_Change_BatchOrder}
	\end{subfigure}
	\begin{subfigure}{0.24\linewidth}
		\centering
        DA\\
    	\includegraphics[width=1.0\linewidth]{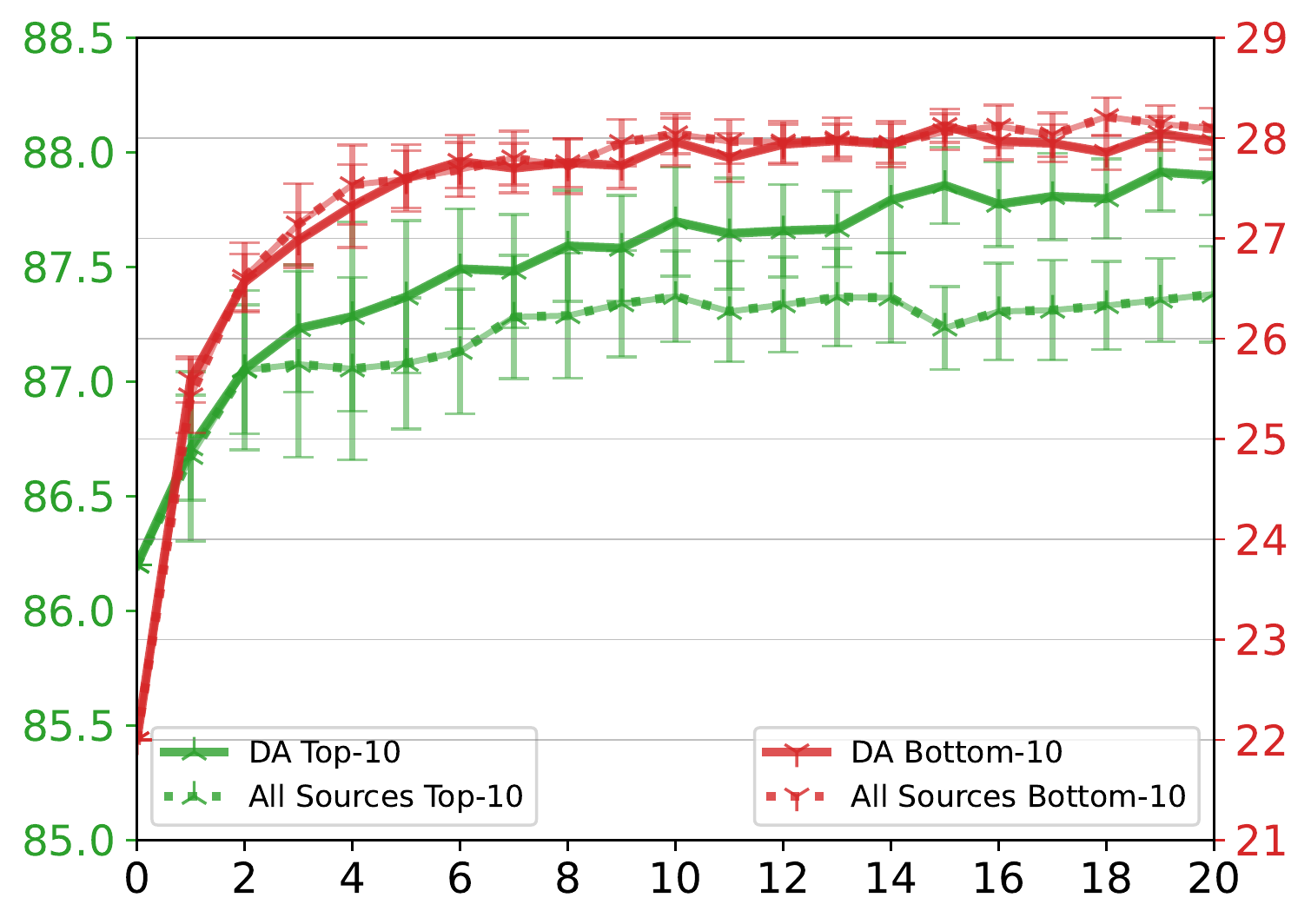}
        \label{fig:Tinyimagenet_resnet50_combined_Change_DA}
	\end{subfigure}
 	\begin{subfigure}{0.24\linewidth}
		\centering
        Init \& BatchOrder\\
    	\includegraphics[width=1.0\linewidth]{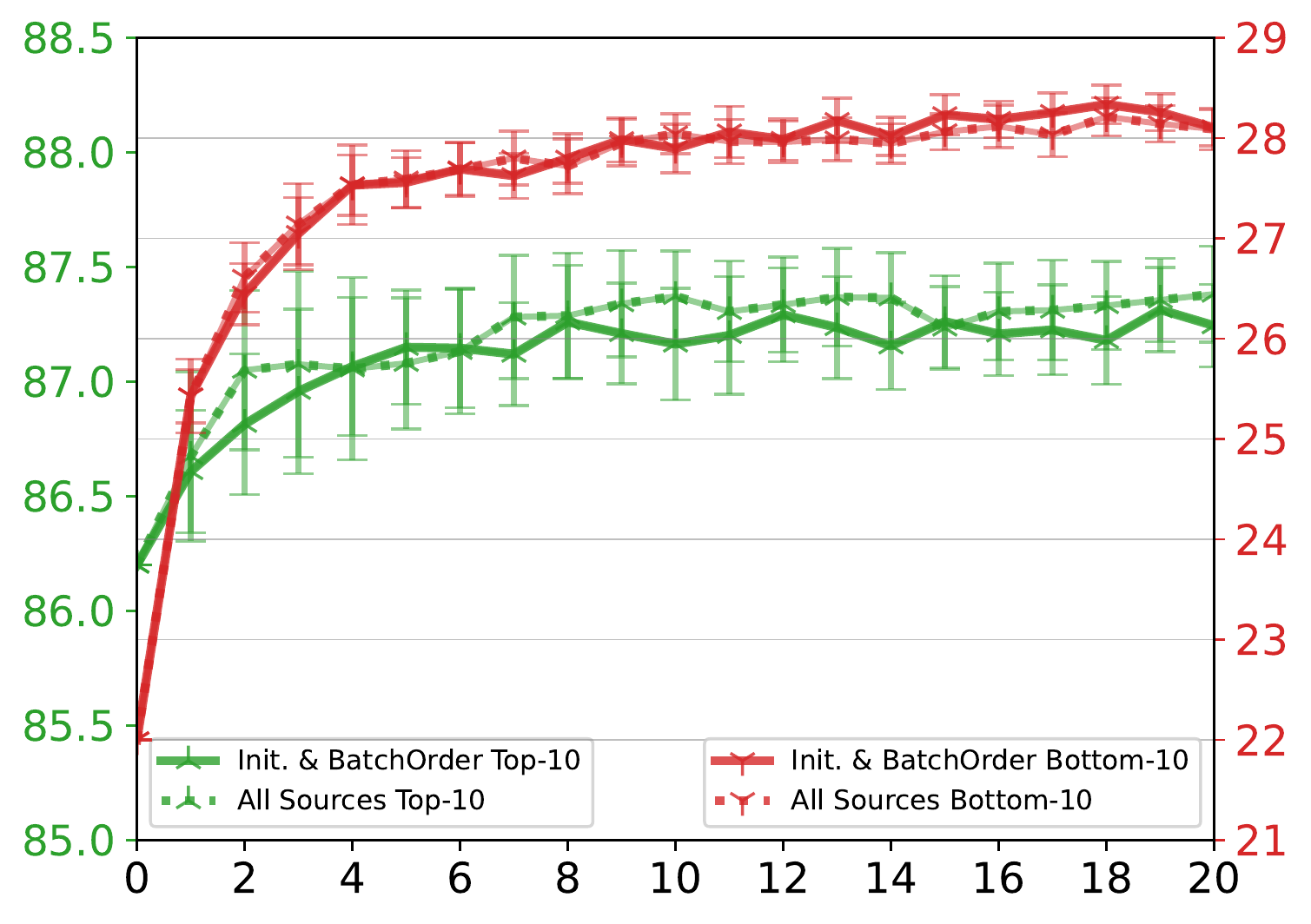}
        \label{fig:Tinyimagenet_resnet50_combined_Change_ModelInit_BatchOrder}
	\end{subfigure}
    \end{minipage}
\end{figure*}
    \begin{figure*}[ht!] \ContinuedFloat
    \centering
    \underline{VGG16}\\
    \vspace{0.1cm}
    
    \begin{minipage}{0.01\linewidth}
        \rotatebox{90}{\hspace{0.2cm} ensemble/base}
    \end{minipage}
    \begin{minipage}{0.98\linewidth}
	\begin{subfigure}{0.24\linewidth}
		\centering
        Init\\
    	\includegraphics[width=1.0\linewidth]{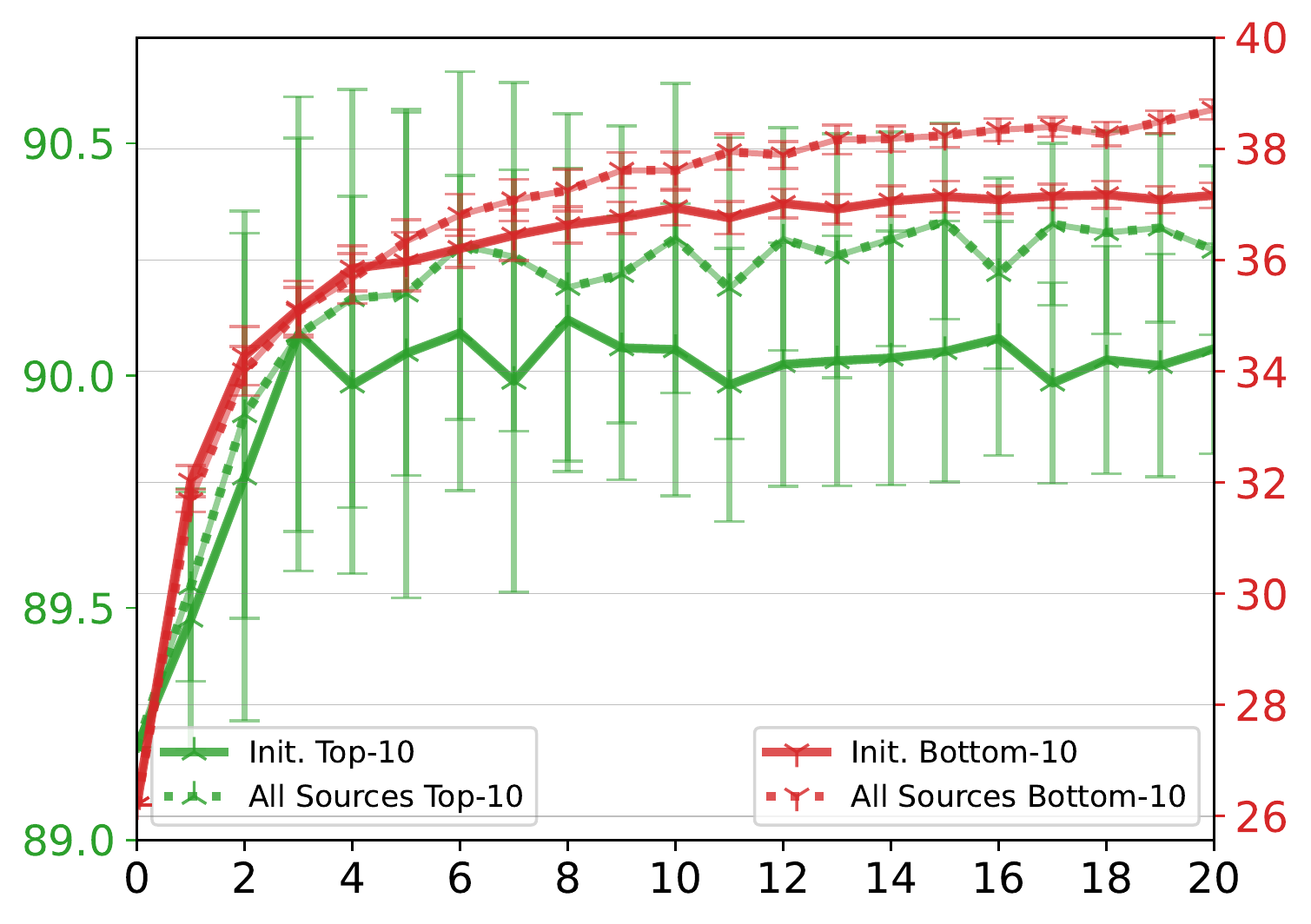}
        \label{fig:Tinyimagenet_vgg16_combined_Change_ModelInit}
	\end{subfigure}
 	\begin{subfigure}{0.24\linewidth}
		\centering
        BatchOrder\\
    	\includegraphics[width=1.0\linewidth]{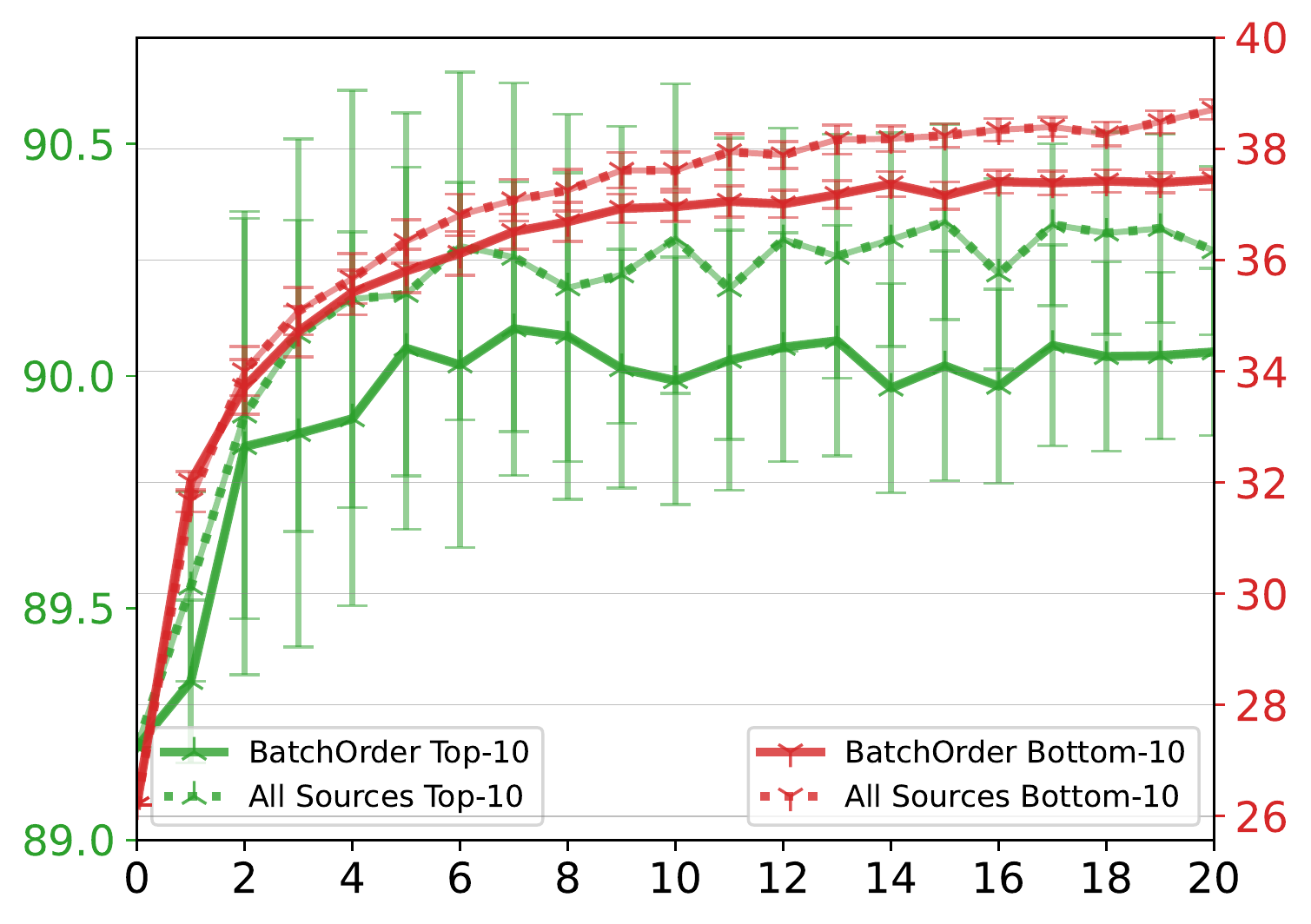}
        \label{fig:Tinyimagenet_vgg16_combined_Change_BatchOrder}
	\end{subfigure}
	\begin{subfigure}{0.24\linewidth}
		\centering
        DA\\
    	\includegraphics[width=1.0\linewidth]{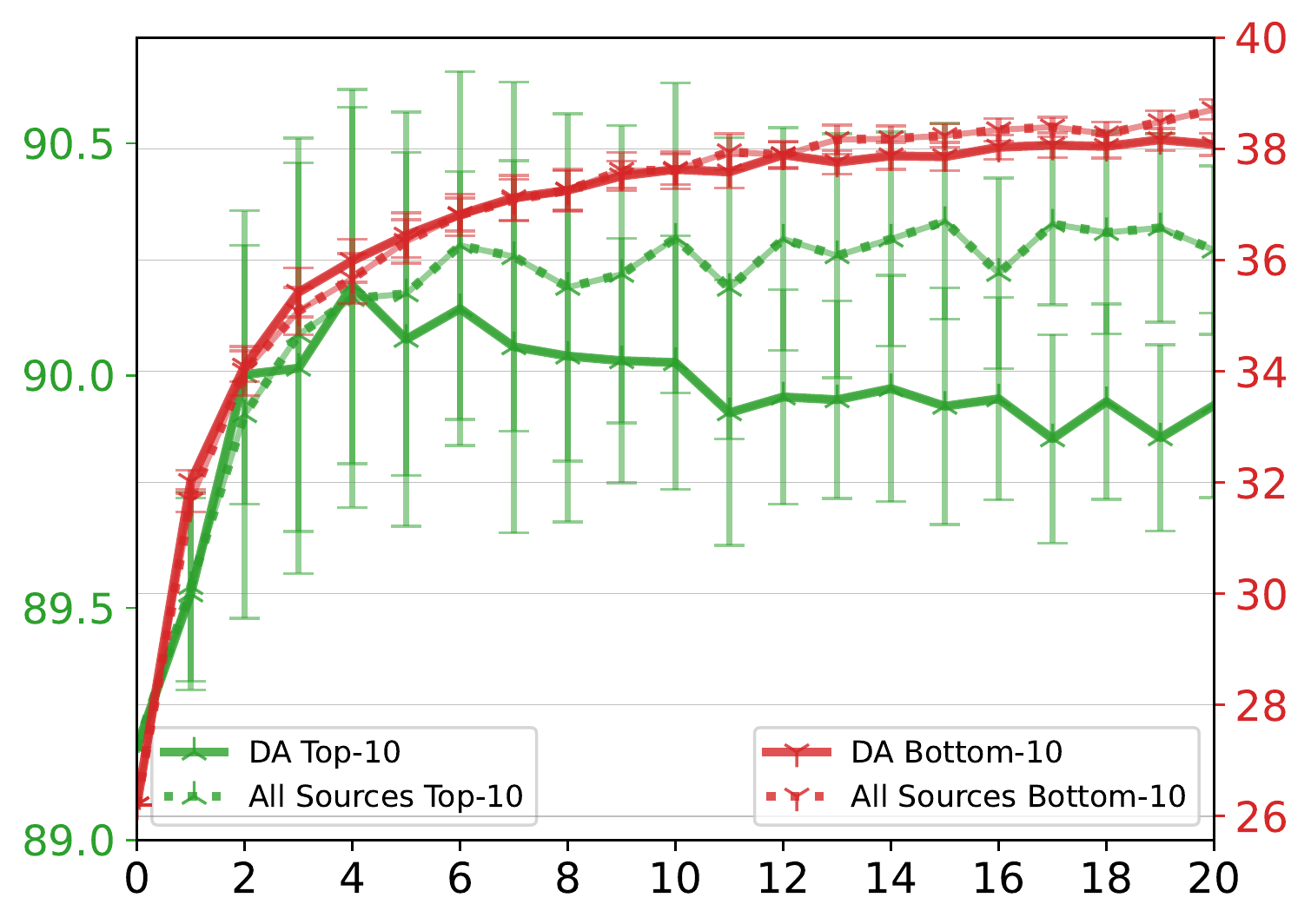}
        \label{fig:Tinyimagenet_vgg16_combined_Change_DA}
	\end{subfigure}
 	\begin{subfigure}{0.24\linewidth}
		\centering
        Init \& BatchOrder\\
    	\includegraphics[width=1.0\linewidth]{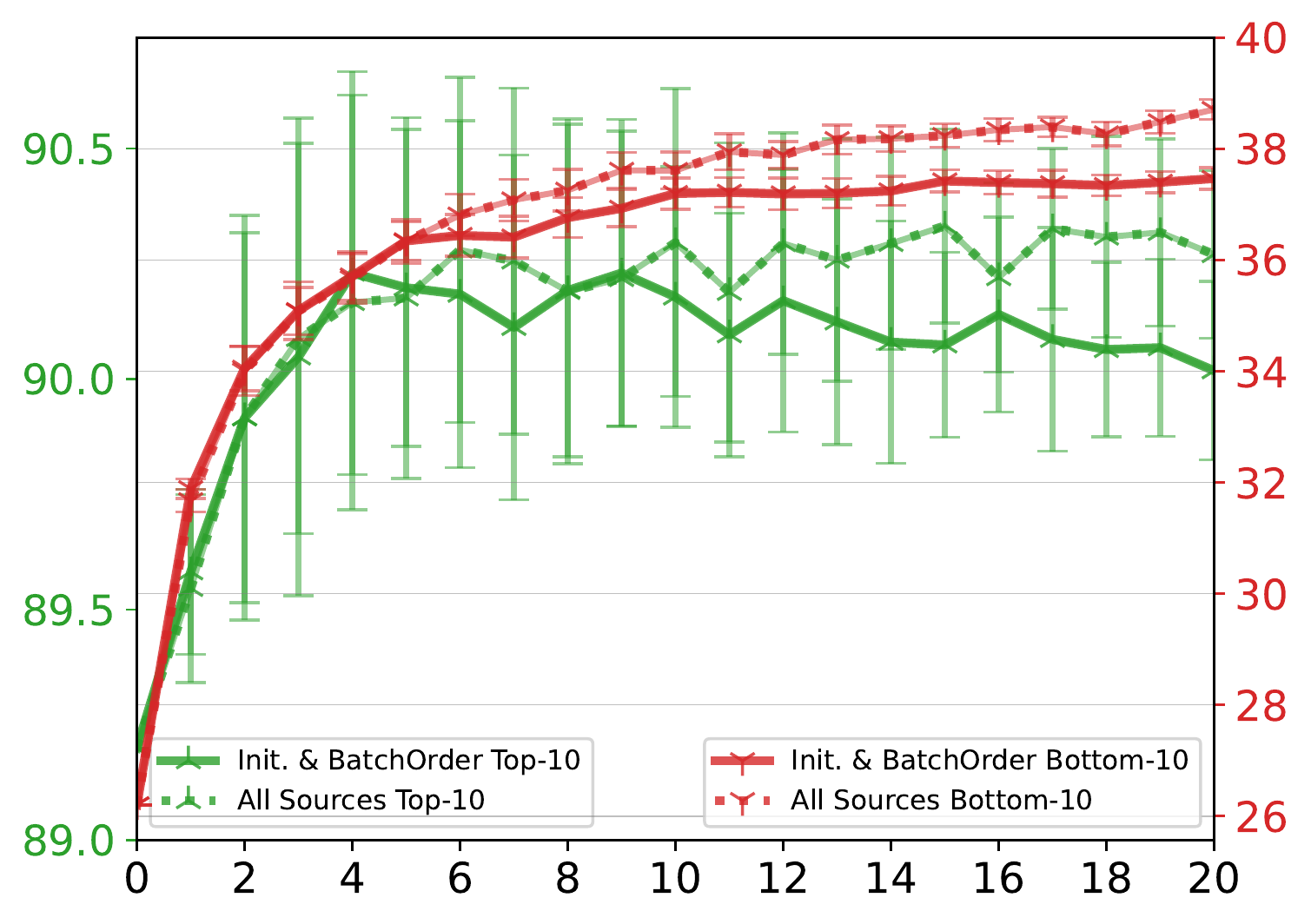}
        \label{fig:Tinyimagenet_vgg16_combined_Change_ModelInit_BatchOrder}
	\end{subfigure}
    \end{minipage}
    
    \centering
    \underline{ViT}\\
    \vspace{0.1cm}

    \begin{minipage}{0.01\linewidth}
        \rotatebox{90}{ensemble/base}
    \end{minipage}
    \begin{minipage}{0.98\linewidth}
	\begin{subfigure}{0.24\linewidth}
		\centering
        Init\\
    	\includegraphics[width=1.0\linewidth]{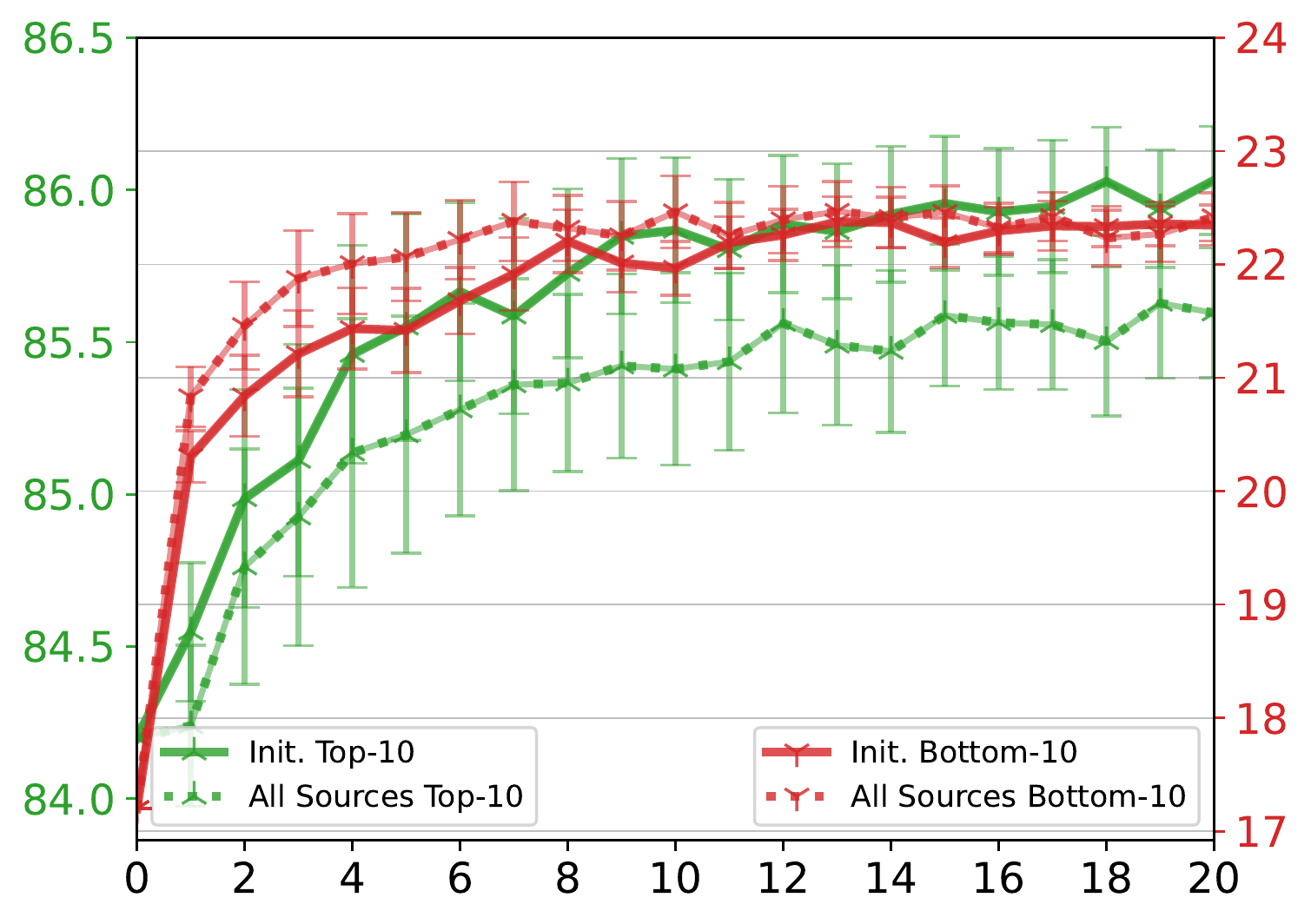}
        \\[-0.7em]
    	{\hspace{0.2cm} \small models in ensemble}
        \label{fig:Tinyimagenet_vit_combined_Change_ModelInit}
	\end{subfigure}
 	\begin{subfigure}{0.24\linewidth}
		\centering
        BatchOrder\\
    	\includegraphics[width=1.0\linewidth]{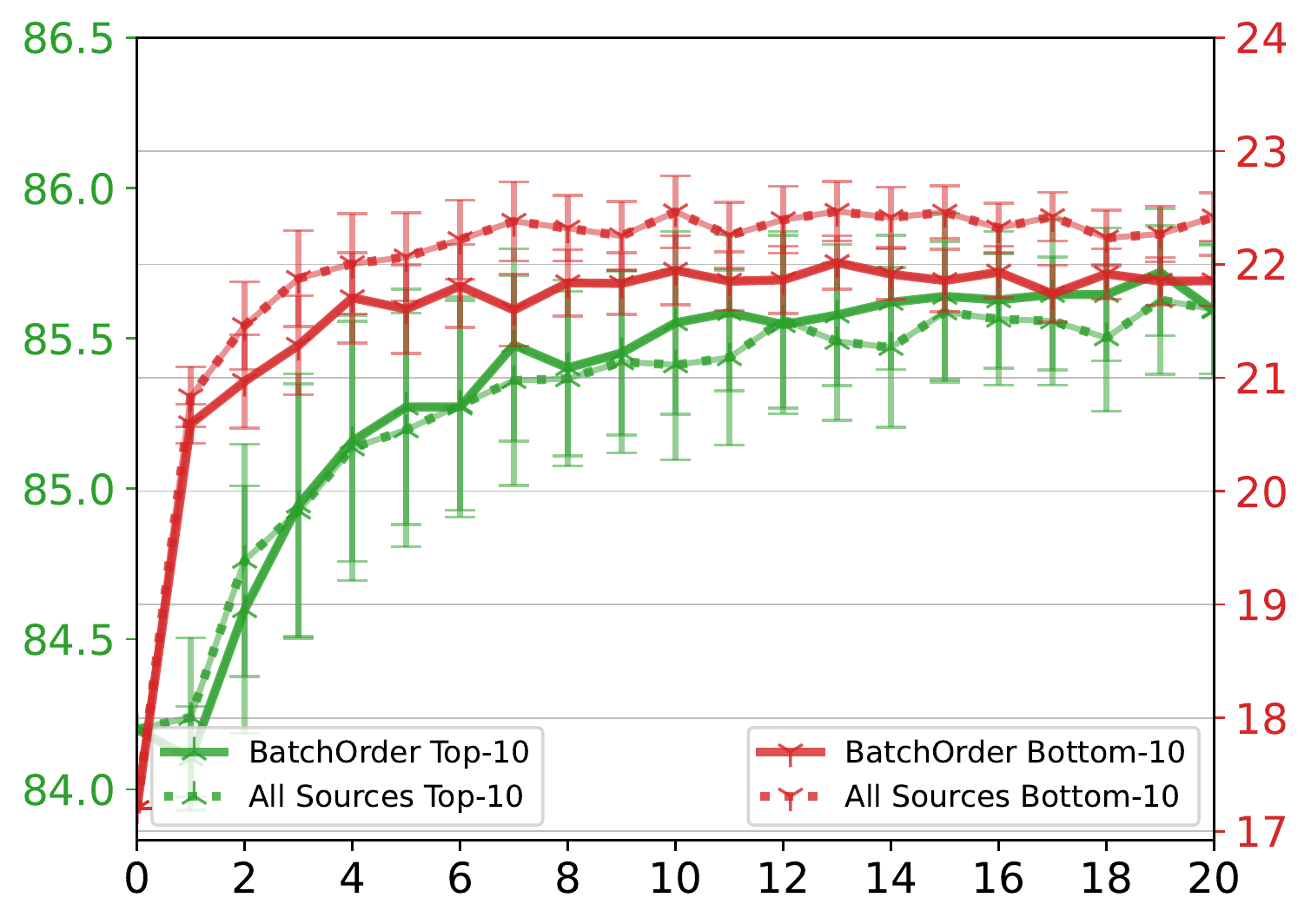}
        \\[-0.7em]
    	{\hspace{0.2cm} \small models in ensemble}
        \label{fig:Tinyimagenet_vit_combined_Change_BatchOrder}
	\end{subfigure}
	\begin{subfigure}{0.24\linewidth}
		\centering
        DA\\
    	\includegraphics[width=1.0\linewidth]{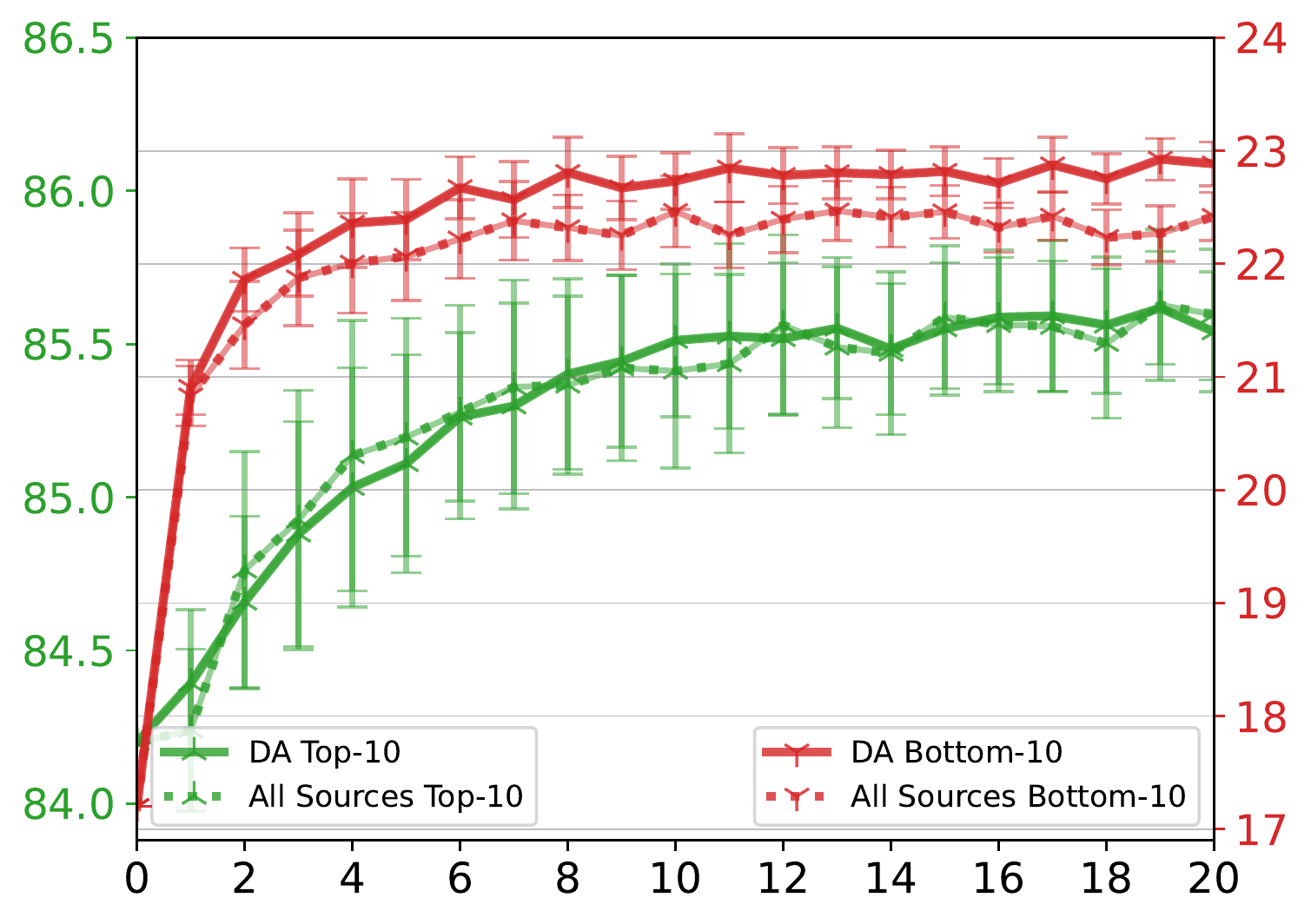}
        \\[-0.7em]
    	{\hspace{0.2cm} \small models in ensemble}
        \label{fig:Tinyimagenet_vit_combined_Change_DA}
	\end{subfigure}
 	\begin{subfigure}{0.24\linewidth}
		\centering
        Init \& BatchOrder\\
    	\includegraphics[width=1.0\linewidth]{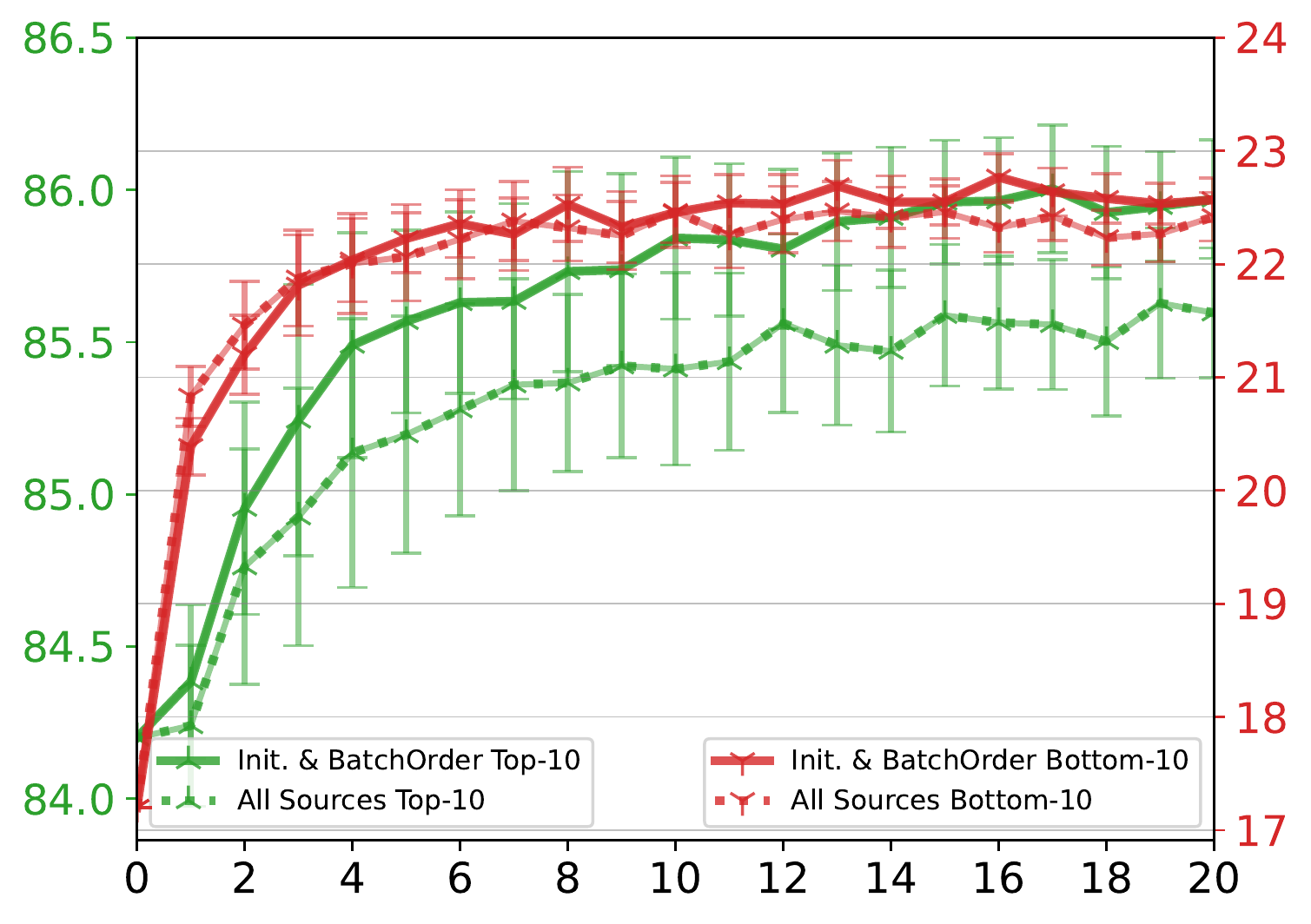}
        \\[-0.7em]
    	{\hspace{0.2cm} \small models in ensemble}
        \label{fig:Tinyimagenet_vit_combined_Change_ModelInit_BatchOrder}
	\end{subfigure}
    	
    \end{minipage}

	\caption{ Accuracy for Top-K and Bottom-K across models added to ensemble on TinyImageNet}
    \label{fig:TinyImageNet_dualaxis}
\end{figure*}

\begin{figure*}[ht]
    \centering
    \begin{minipage}{0.32\linewidth}
    \centering
    \underline{\small K=5}
    \includegraphics[width=\linewidth]{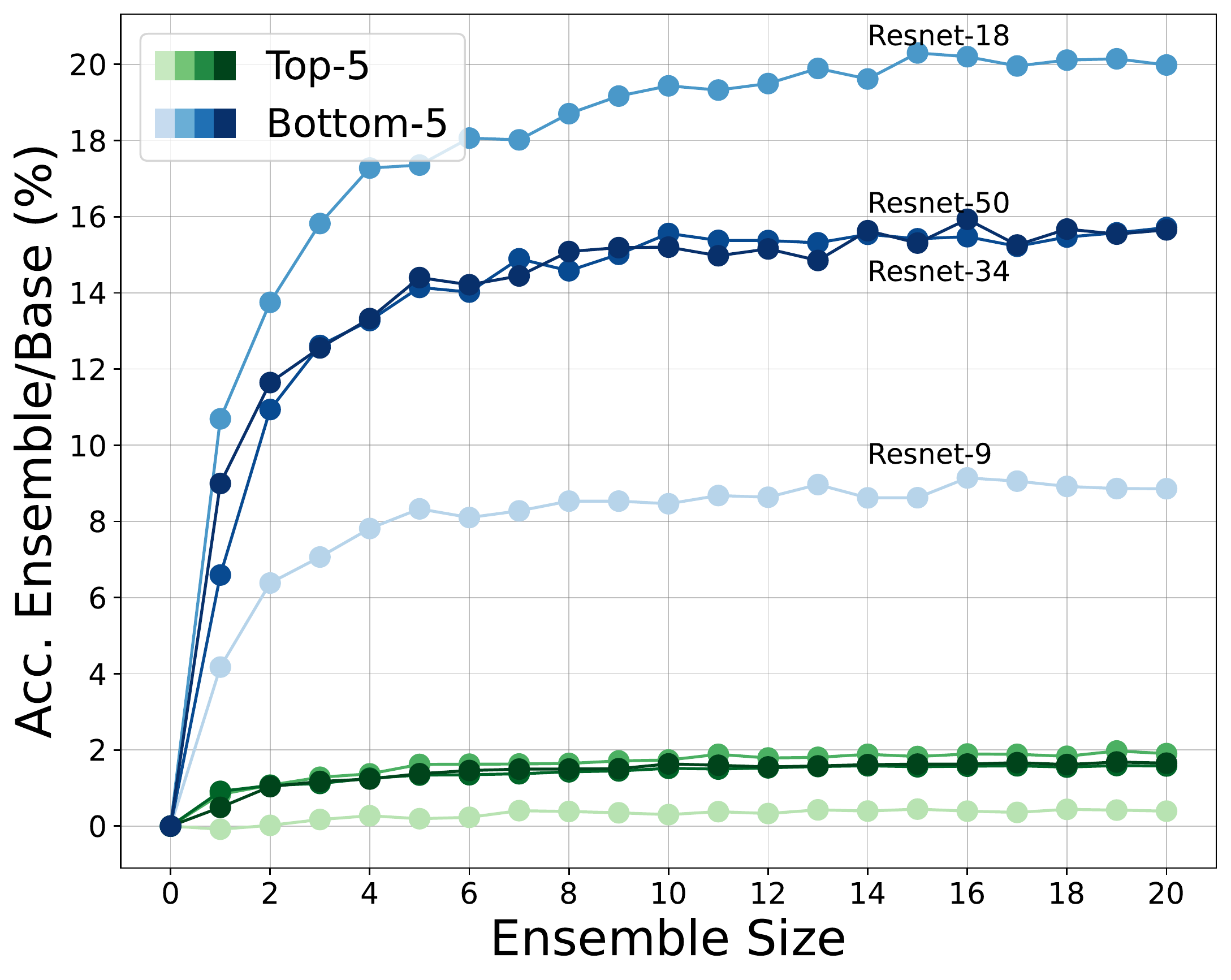}
    \end{minipage}
    \begin{minipage}{0.32\linewidth}
    \centering
    \underline{\small K=10}
    \includegraphics[width=\linewidth]{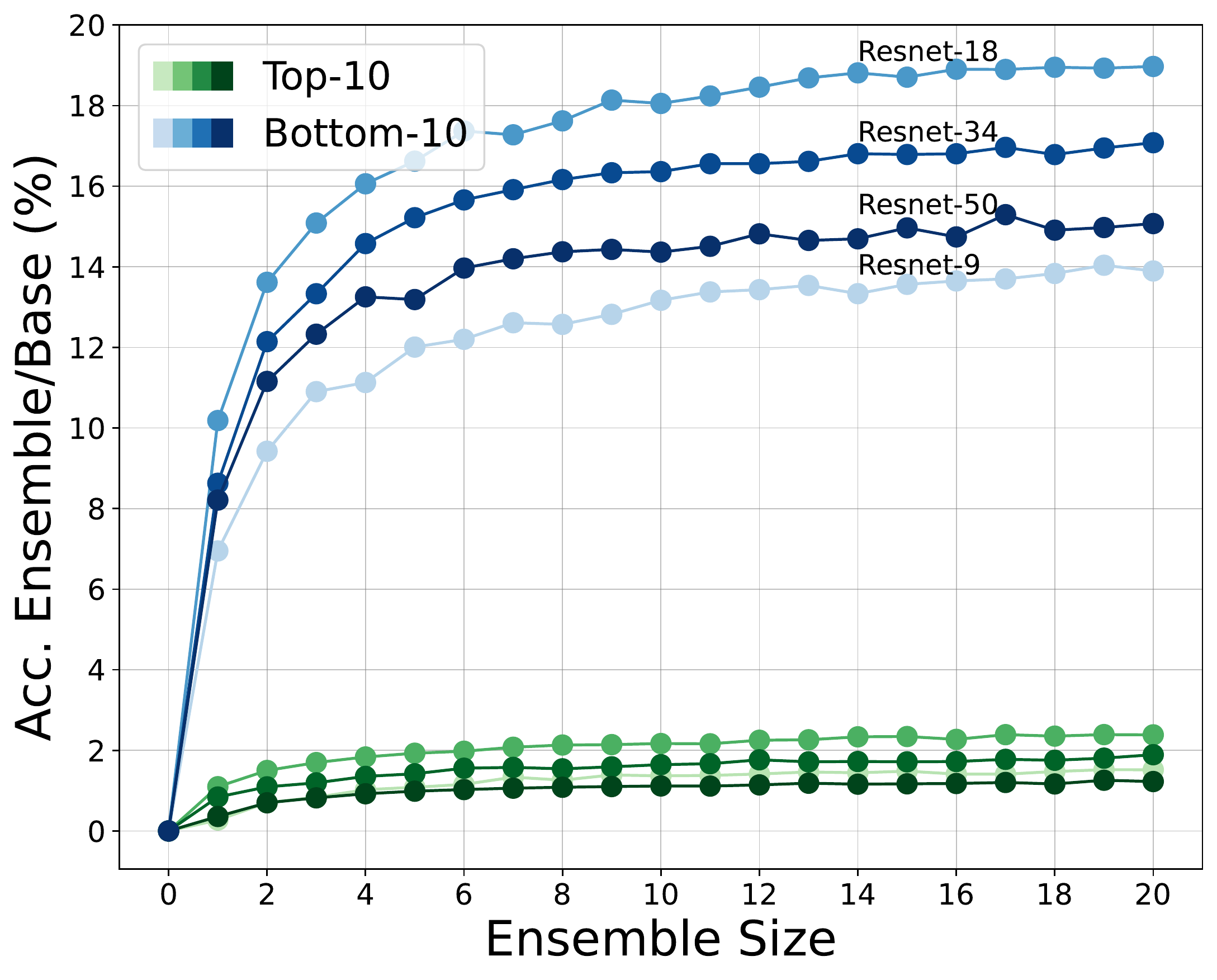}
    \end{minipage}
    \begin{minipage}{0.32\linewidth}
    \centering
    \underline{\small K=20}
    \includegraphics[width=\linewidth]{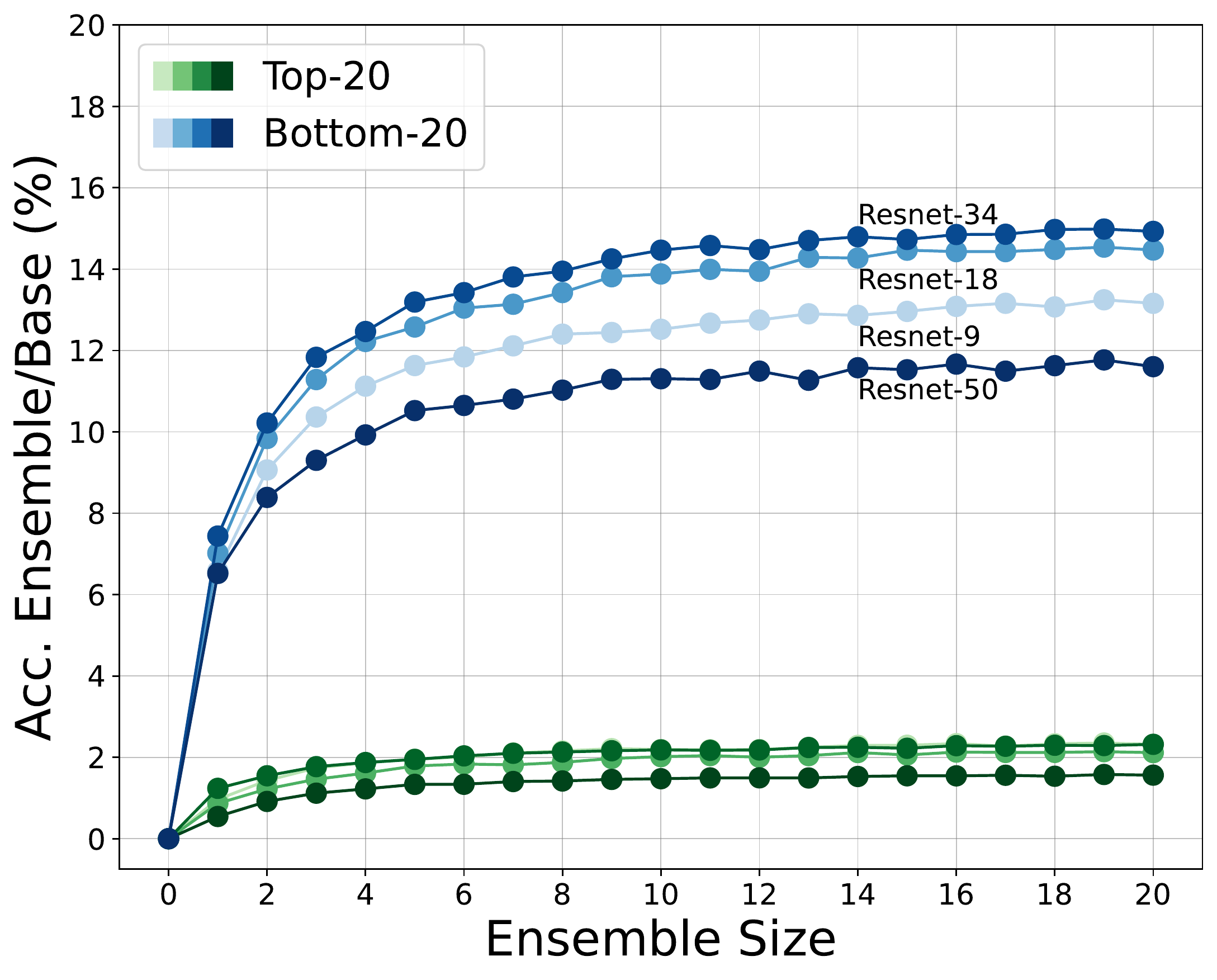}
    \end{minipage}\\
	\vspace{-0.3cm}
    \caption{Top/Bottom-K performance for K = \{5,10,20\} in CIFAR100}
    \vspace{-0.5cm}
    \label{fig:C100_K}
\end{figure*}

\begin{figure*}[t!] 
    \section{Controlling for the Sources of Stochasticity in Homogeneous Ensembles}
    \vspace{0.2cm}
    \subsection{CIFAR100 Accuracy \% difference for ResNet 18, 34, 50}
    \vspace{0.2cm}

    \centering
	\underline{ResNet family on CIFAR100}\\
    \begin{minipage}{0.01\linewidth}
        \rotatebox{90}{\hspace{1cm}  \% accuracy diff.}
    \end{minipage}
    \hspace{0.1cm}
    \begin{minipage}{0.97\linewidth}
	\begin{subfigure}{0.32\linewidth}
		\centering
    	\includegraphics[width=1.0\linewidth]{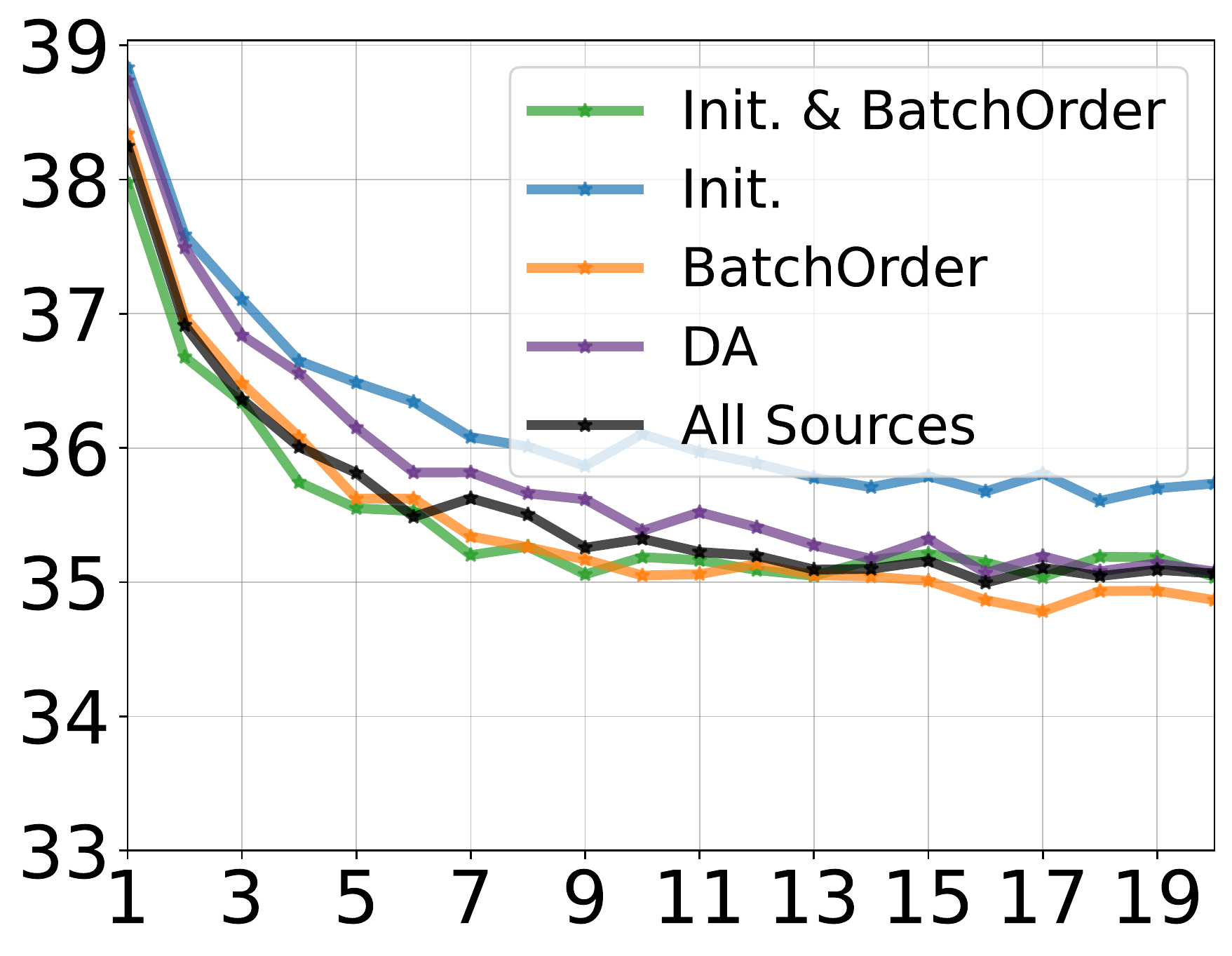}\\[-0.7em]
    	models in ensemble
    	\caption{ResNet18, 20 models}
        \label{fig:CIFAR100_res18_20_diff}
	\end{subfigure}
 	\begin{subfigure}{0.32\linewidth}
		\centering
    	\includegraphics[width=1.0\linewidth]{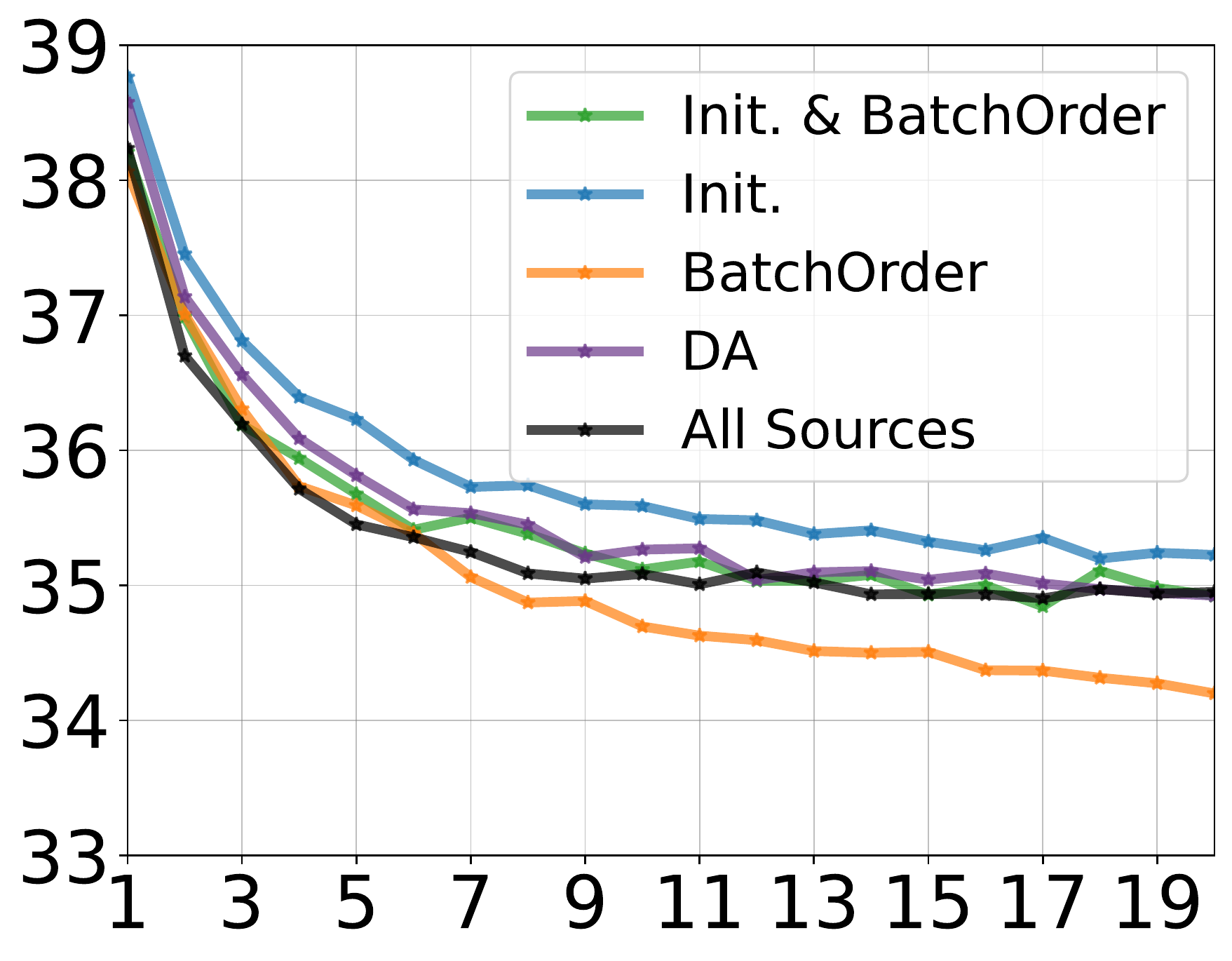}\\[-0.7em]
    	models in ensemble
    	\caption{ResNet34, 20 models}
        \label{fig:CIFAR100_res32_20_diff}
	\end{subfigure}
 	\begin{subfigure}{0.32\linewidth}
		\centering
    	\includegraphics[width=1.0\linewidth]{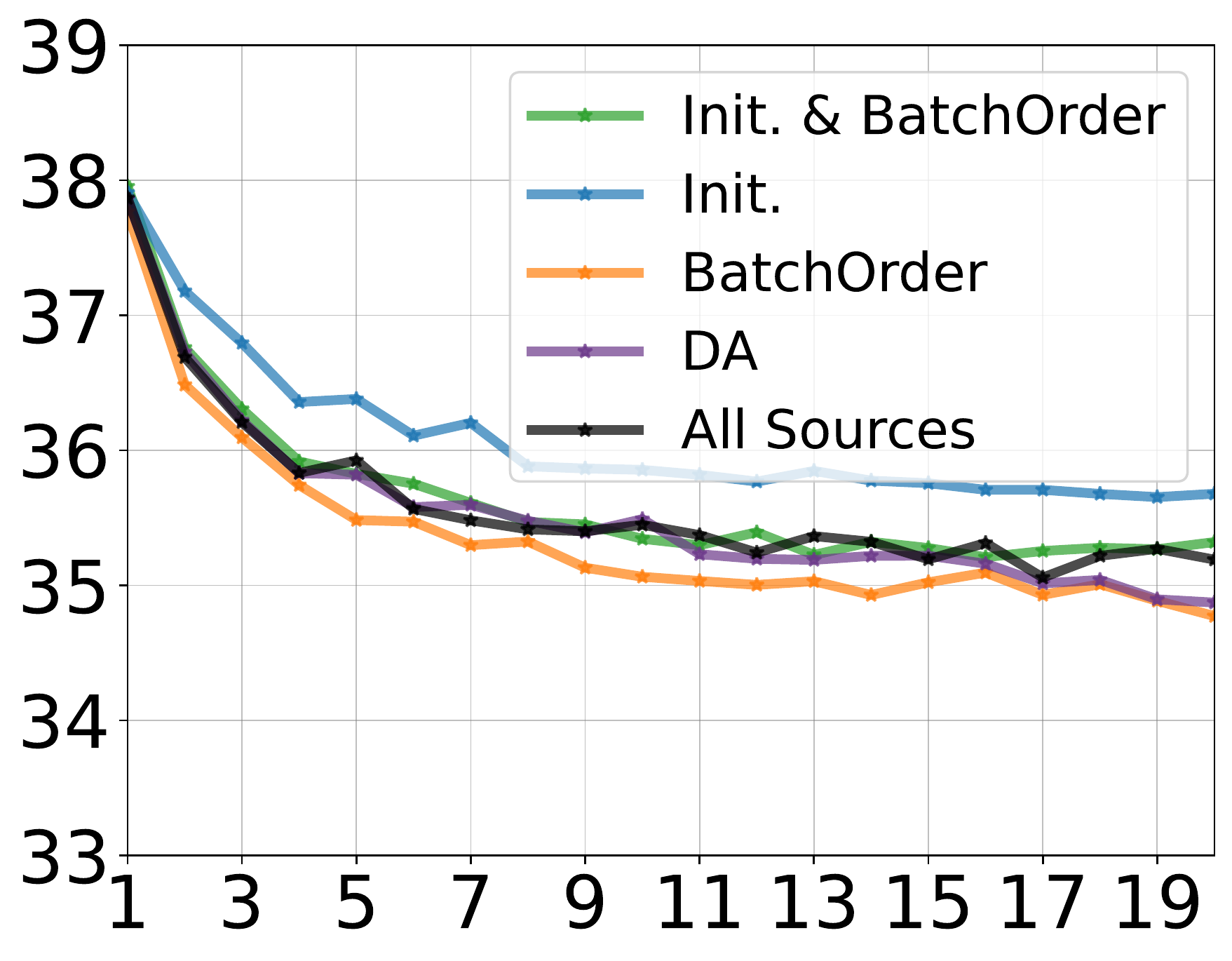}\\[-0.7em]
    	models in ensemble
    	\caption{ResNet50, 20 models}
        \label{fig:CIFAR100_res50_20_diff}
	\end{subfigure}\\
    \end{minipage}
	\caption{\small Accuracy \% difference between top and bottom 10 classes, for ResNet18, 34, and 50 for CIFAR100.
	}
    \label{fig:resnet_family_diff_c100}
\end{figure*}

\begin{figure*}[t!]
    \subsection{CIFAR100 and TinyImageNet Results}
    \vspace{0.2cm}
    \centering
    \underline{CIFAR100 Top-K}\\  
    \vspace{0.1cm}
     \begin{minipage}{0.01\linewidth}
        \rotatebox{90}{test accuracy \%}
    \end{minipage}
    \hspace{0.1cm}
    \begin{minipage}{0.97\linewidth}
    
    \begin{subfigure}{0.32\linewidth}
        \centering
        ResNet9\\
        \includegraphics[width=\linewidth]{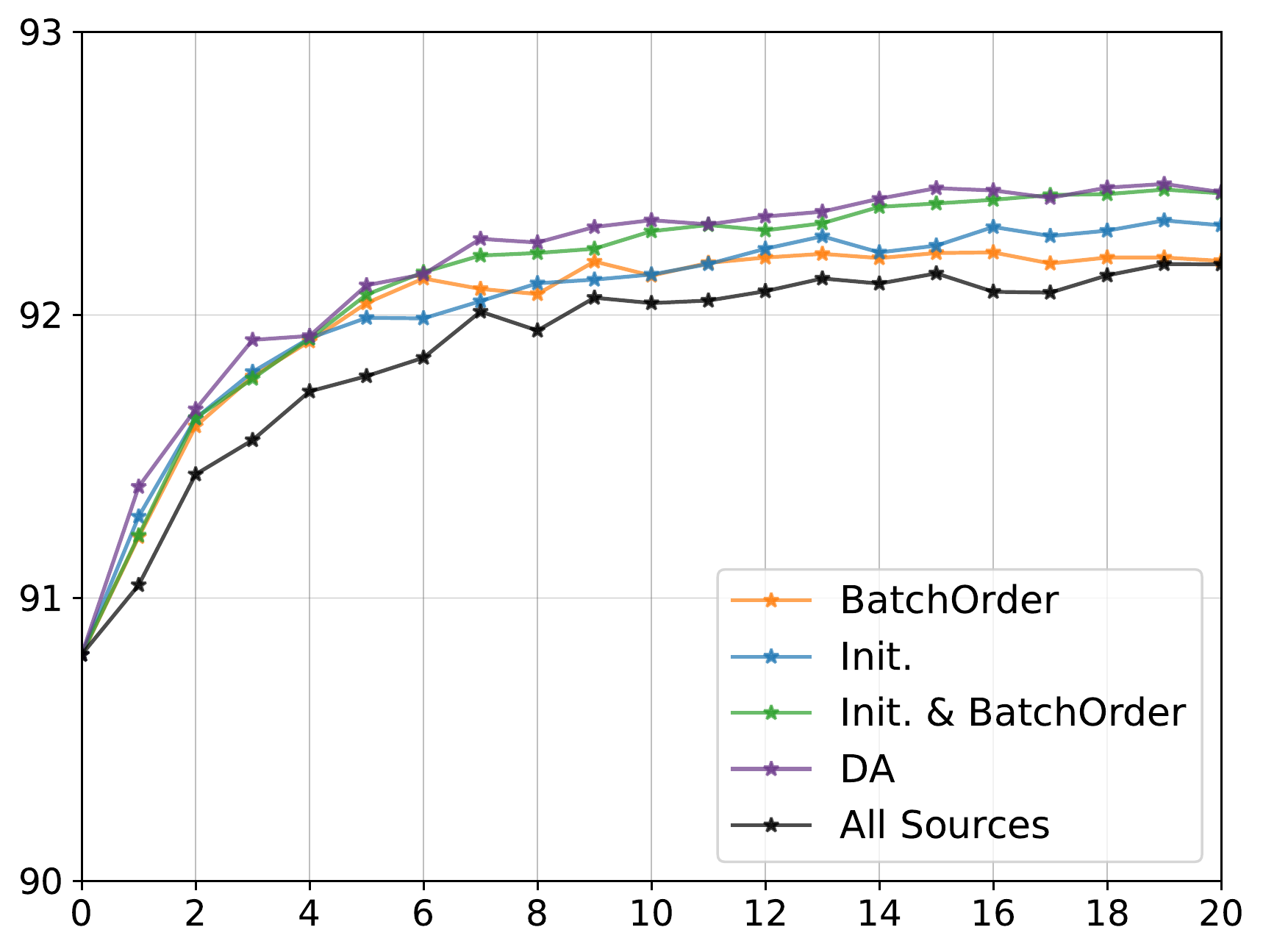}
    \end{subfigure}
    \begin{subfigure}{0.32\linewidth}
        \centering
        VGG16\\
        \includegraphics[width=\linewidth]{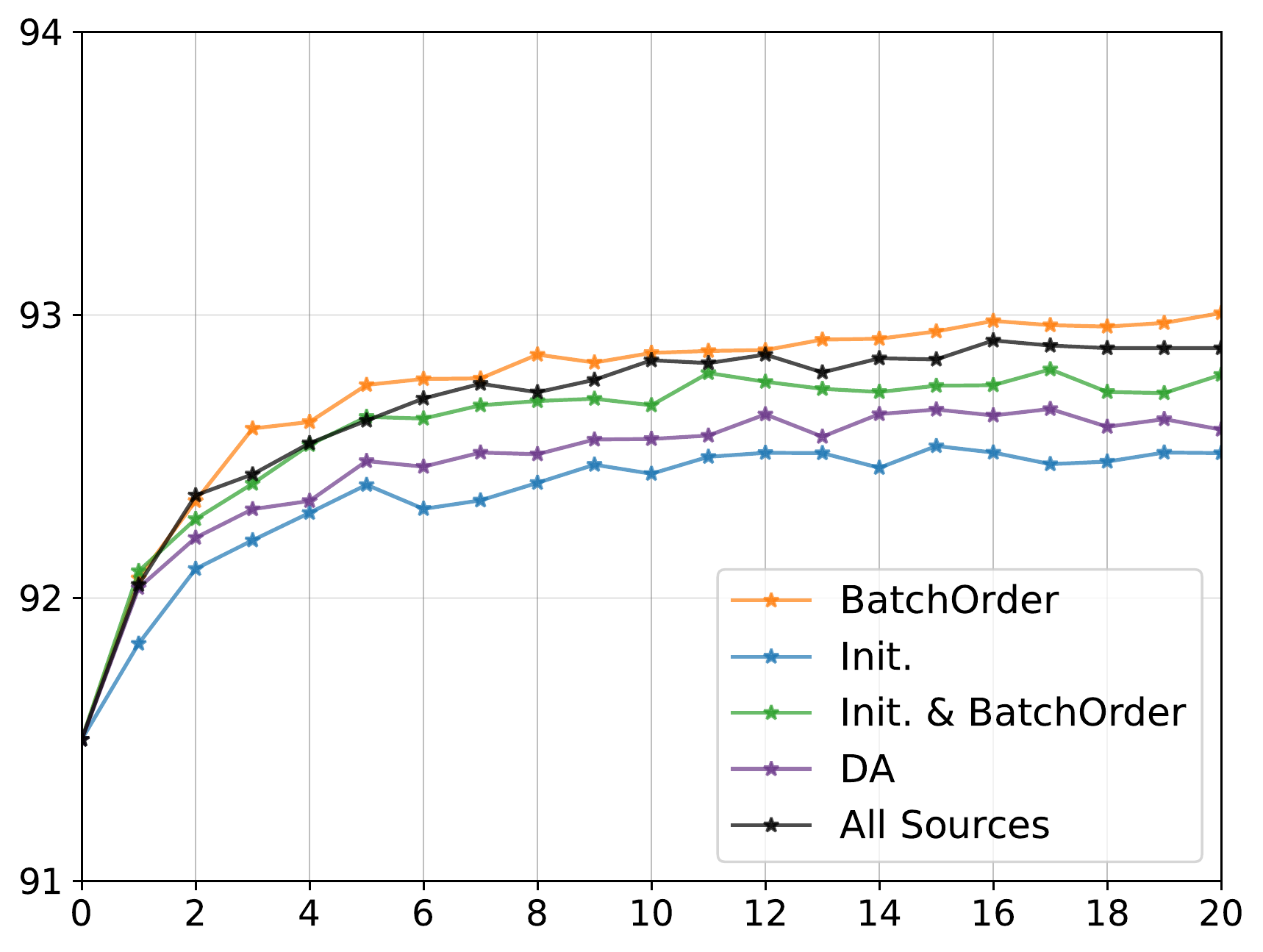}
    \end{subfigure}
    \begin{subfigure}{0.32\linewidth}
        \centering
        MLP-Mixer\\
        \includegraphics[width=\linewidth]{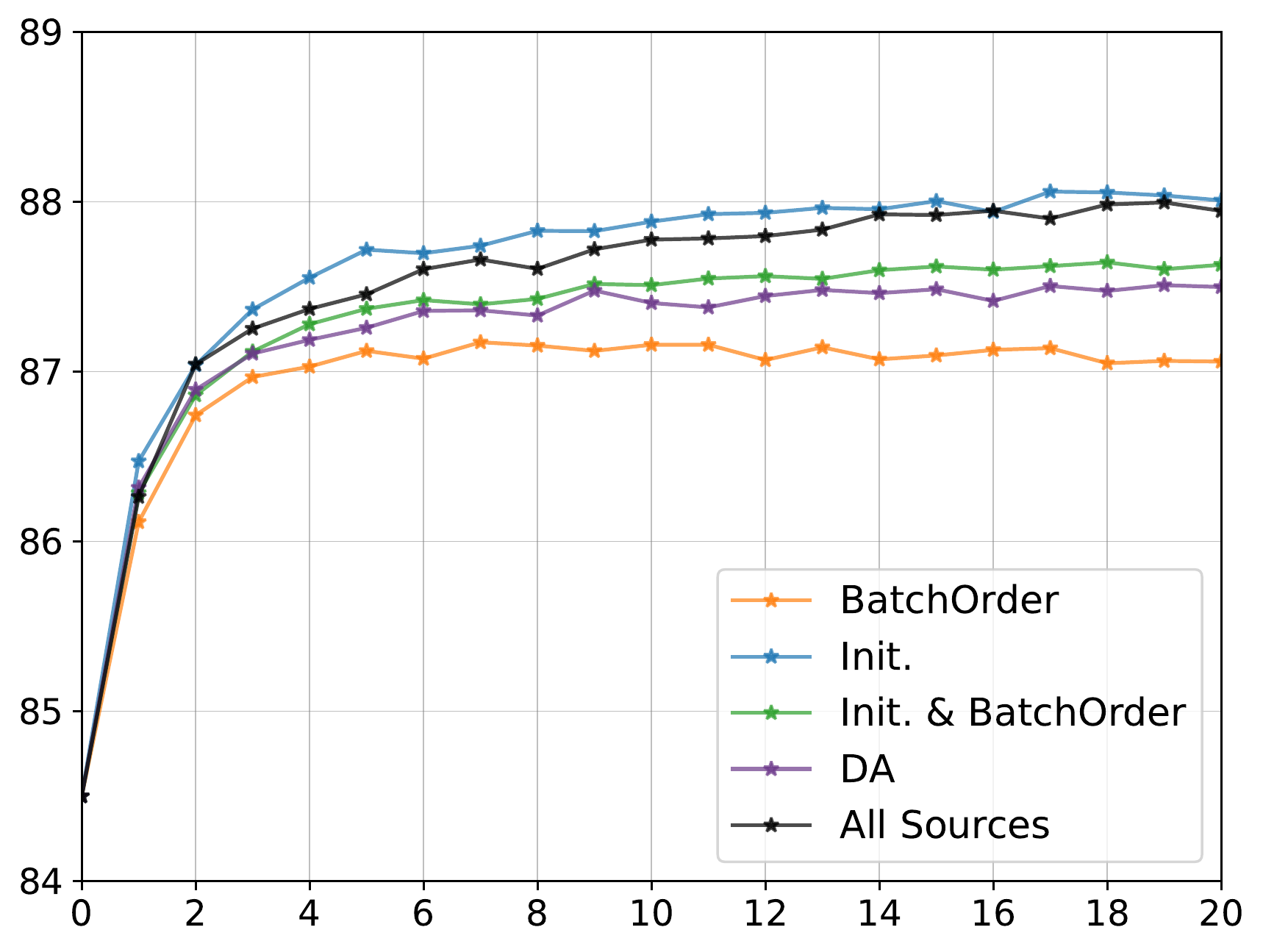}  
    \end{subfigure}
    \end{minipage}
    
    \centering
    \underline{CIFAR100 Bottom-K}\\
    \vspace{0.1cm}
    \begin{minipage}{0.01\linewidth}
        \rotatebox{90}{test accuracy \%}
    \end{minipage}
    \hspace{0.1cm}
    \begin{minipage}{0.97\linewidth}
    \begin{subfigure}{0.32\linewidth}
        \centering
        ResNet9\\
        \includegraphics[width=\linewidth]{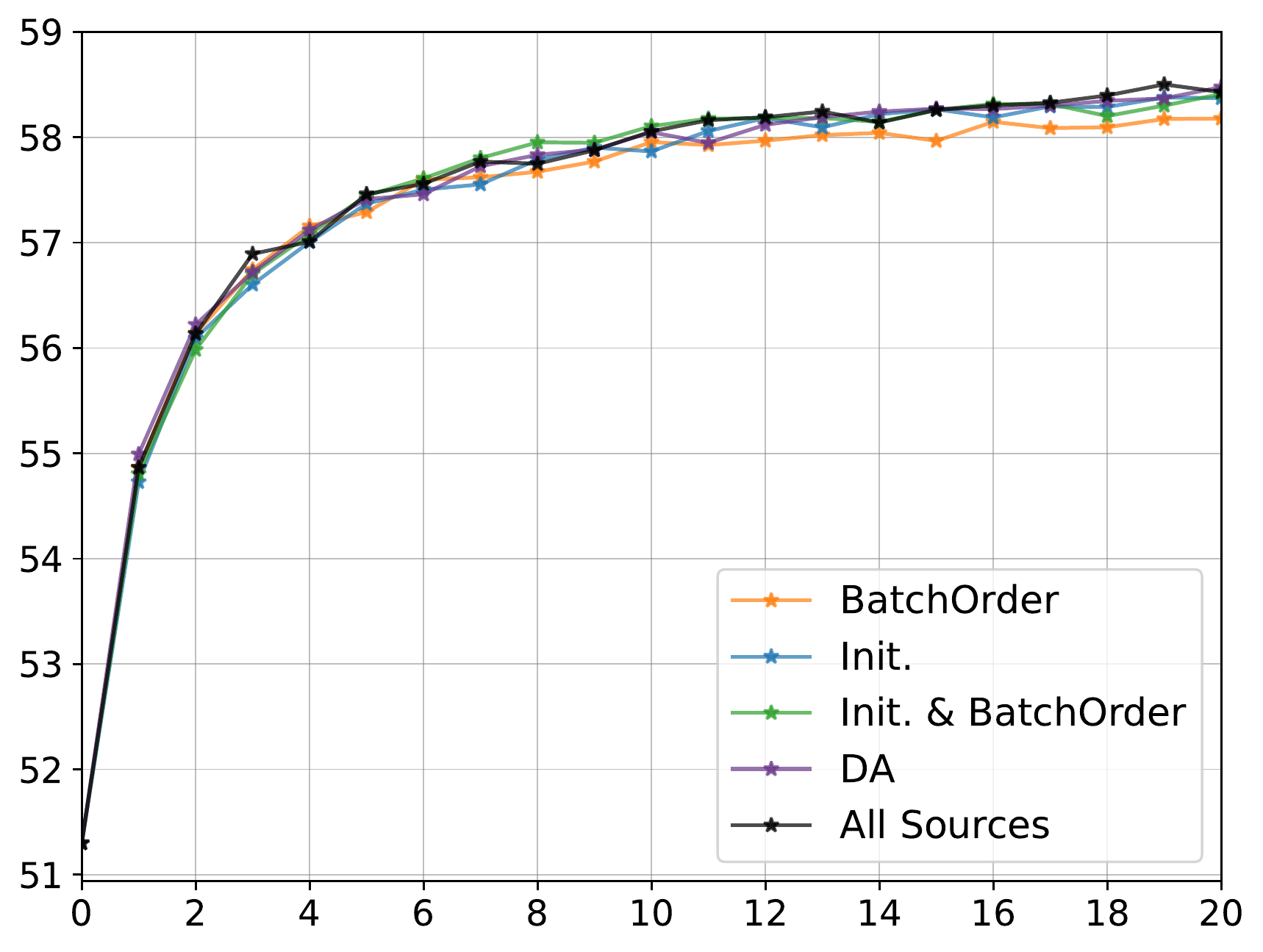}
    \end{subfigure}
    \begin{subfigure}{0.32\linewidth}
        \centering
        VGG16\\
        \includegraphics[width=\linewidth]{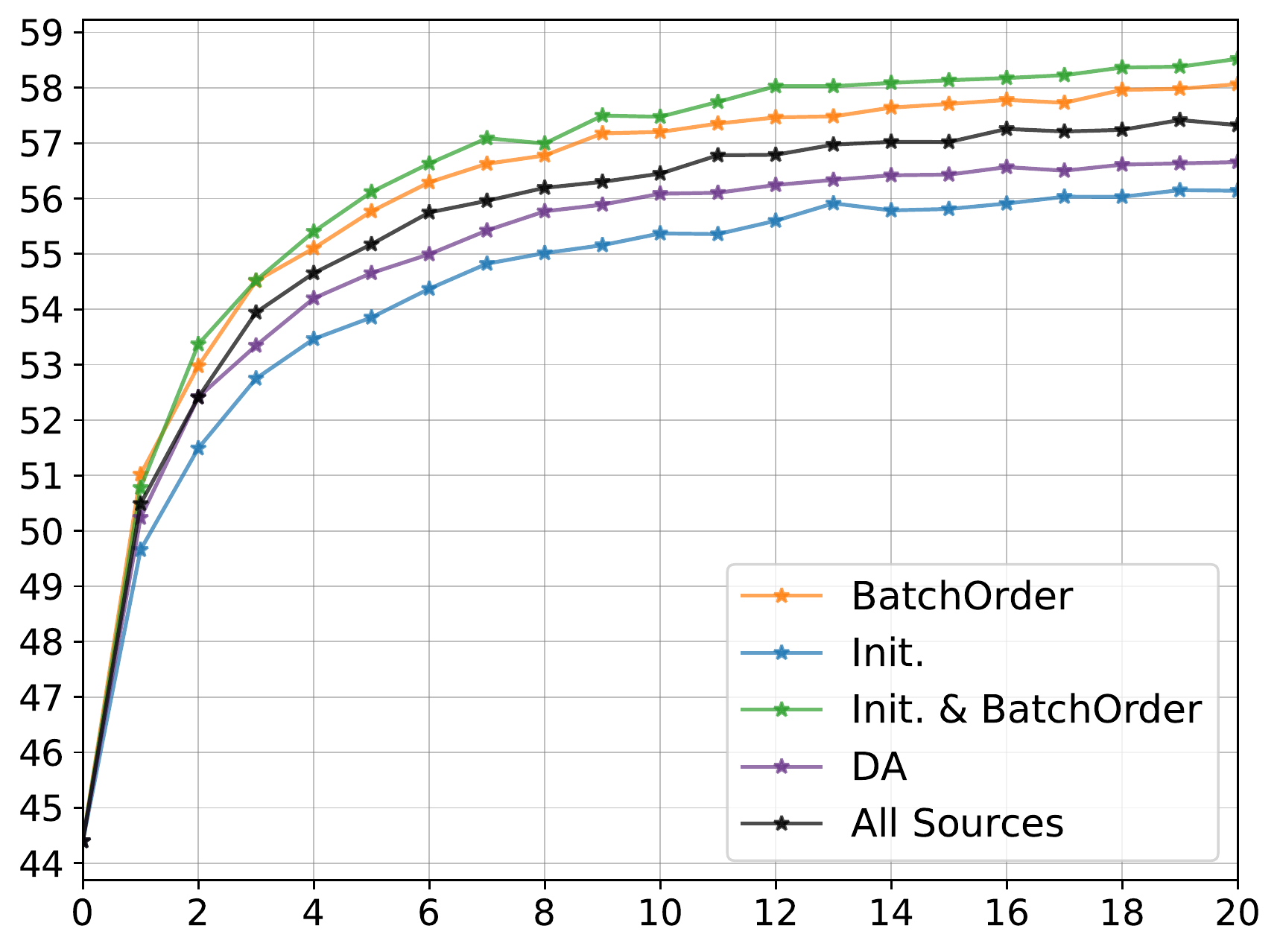}
    \end{subfigure}
    \begin{subfigure}{0.32\linewidth}
        \centering
        MLP-Mixer\\
        \includegraphics[width=\linewidth]{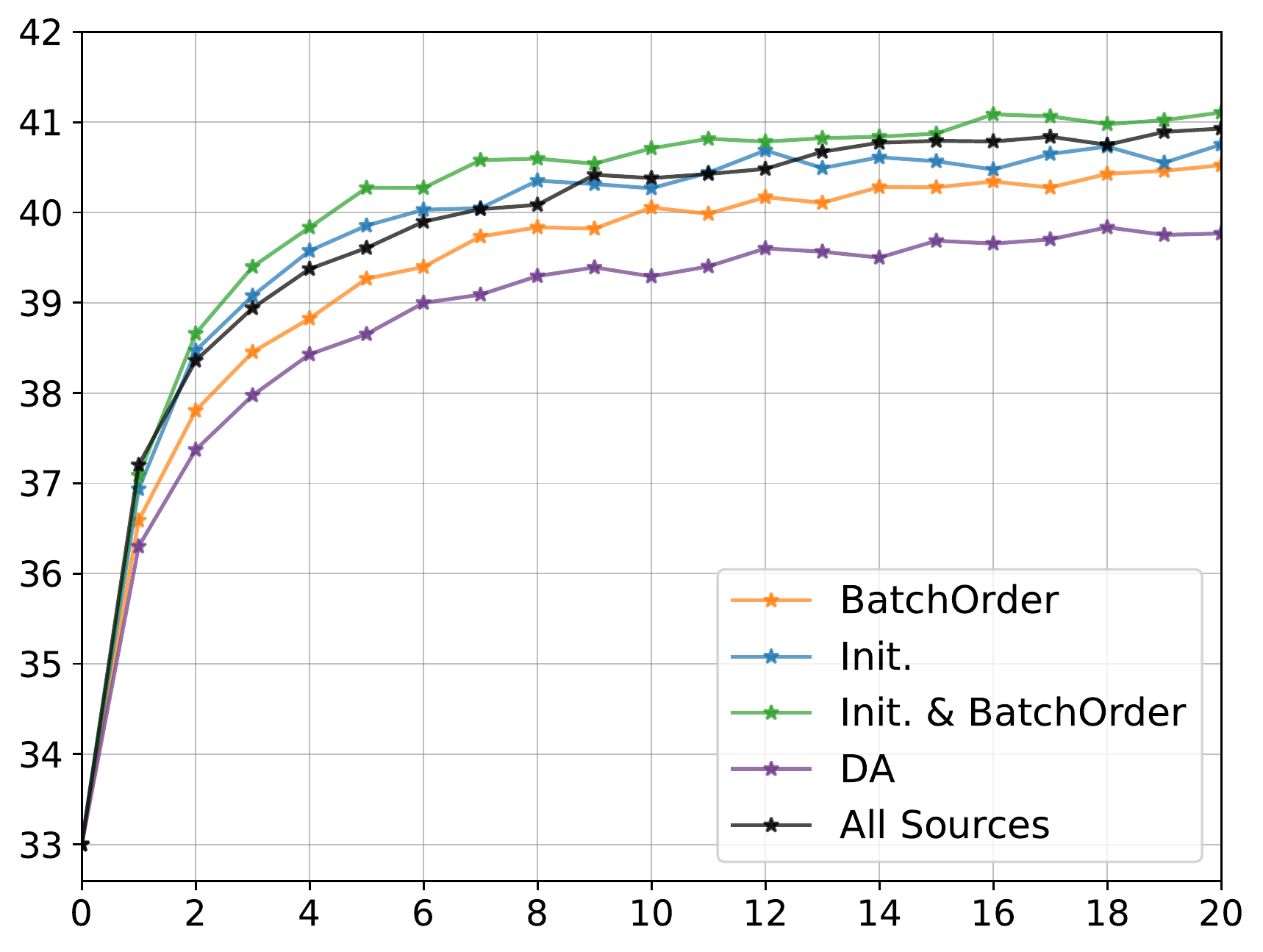}
    \end{subfigure}
    \end{minipage}
\end{figure*}
\begin{figure*}[!ht]
    \ContinuedFloat
    \centering
    \underline{TinyImageNet Top-K}\\
    \vspace{0.1cm}
    \begin{minipage}{0.01\linewidth}
        \rotatebox{90}{test accuracy \%}
    \end{minipage}
    \hspace{0.1cm}
    \begin{minipage}{0.97\linewidth}
    \begin{subfigure}{0.32\linewidth}
        \centering
        ResNet50\\
        \includegraphics[width=\linewidth]{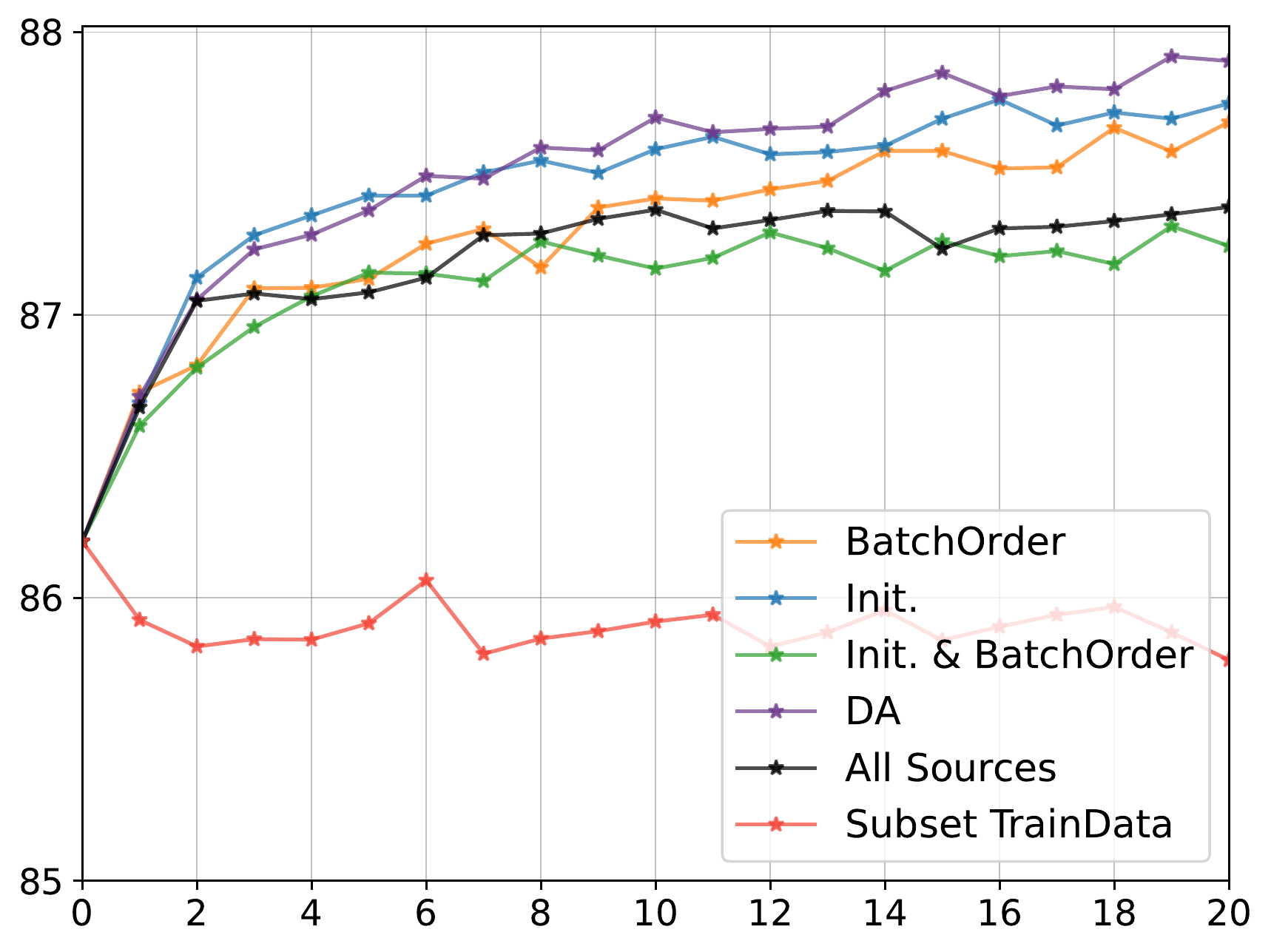}
    \end{subfigure}
    \begin{subfigure}{0.32\linewidth}
        \centering
        VGG16\\
        \includegraphics[width=\linewidth]{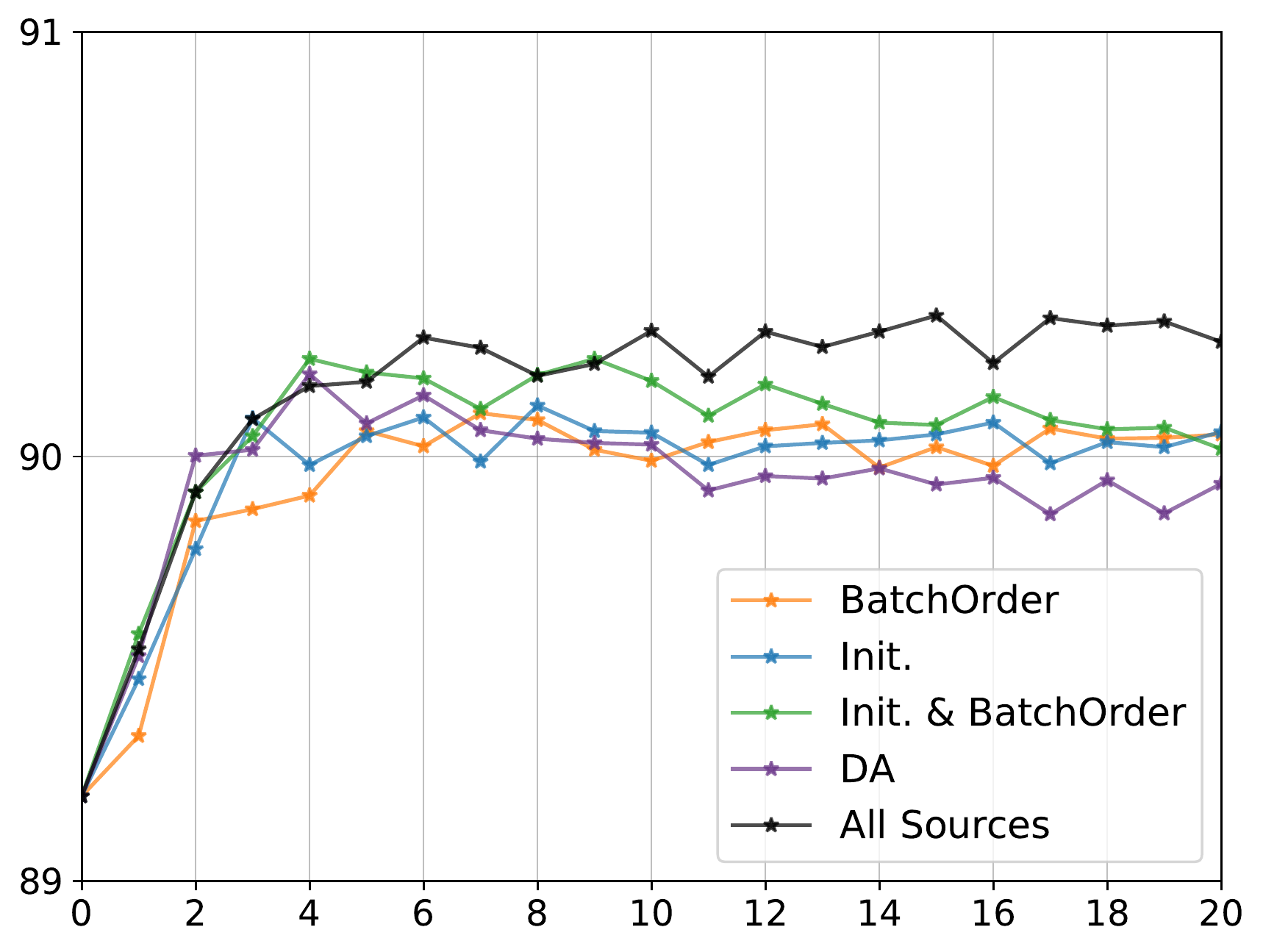}
    \end{subfigure}
    \begin{subfigure}{0.32\linewidth}
        \centering
        ViT\\
        \includegraphics[width=\linewidth]{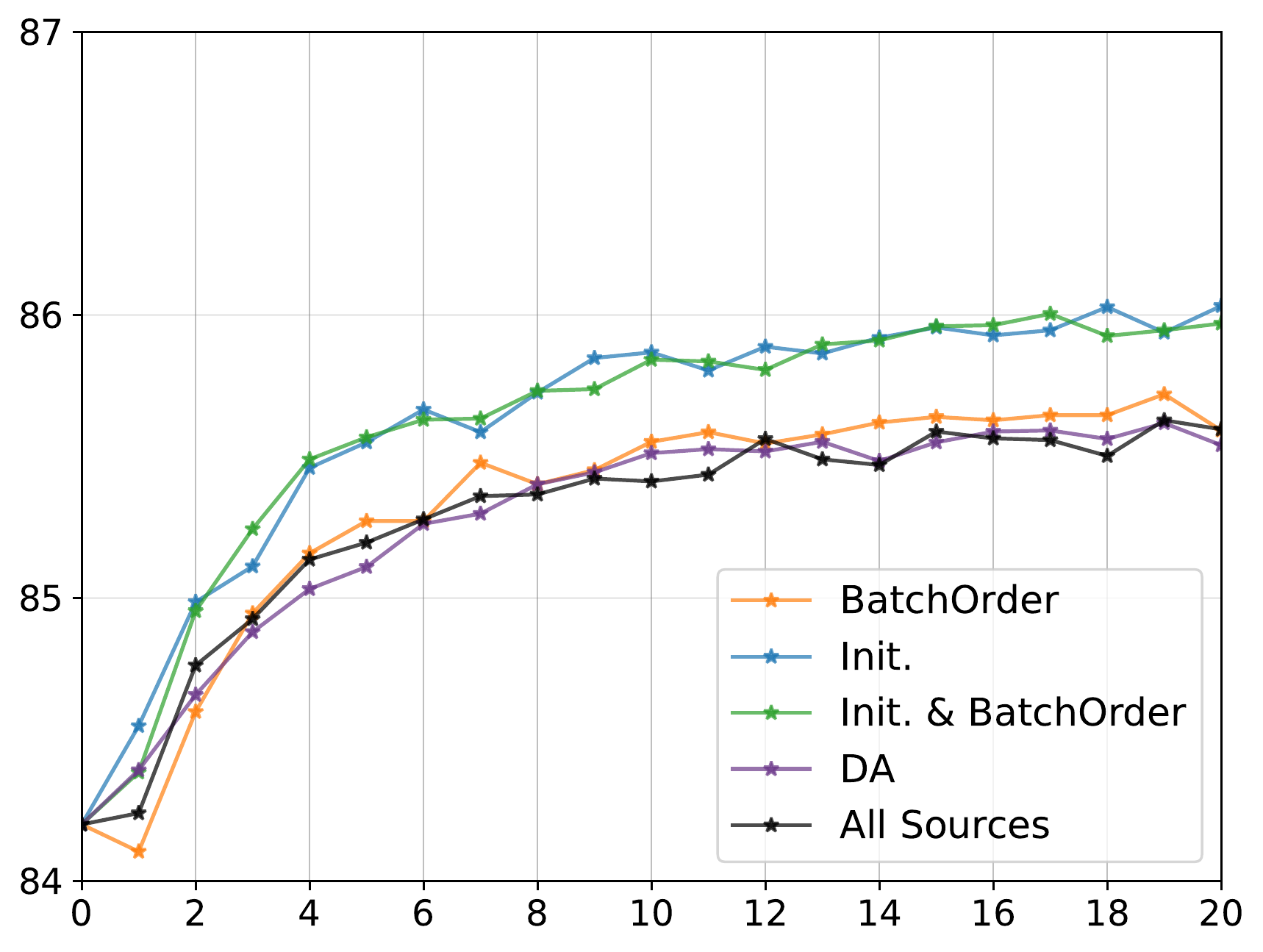}
    \end{subfigure}
    \end{minipage}
    
    \centering
    \underline{TinyImageNet Bottom-K}\\
    \vspace{0.1cm}
    \begin{minipage}{0.01\linewidth}
        \rotatebox{90}{test accuracy \%}
    \end{minipage}
    \hspace{0.1cm}
    \begin{minipage}{0.97\linewidth}
    \begin{subfigure}{0.32\linewidth}
        \centering
        ResNet50\\
        \includegraphics[width=\linewidth]{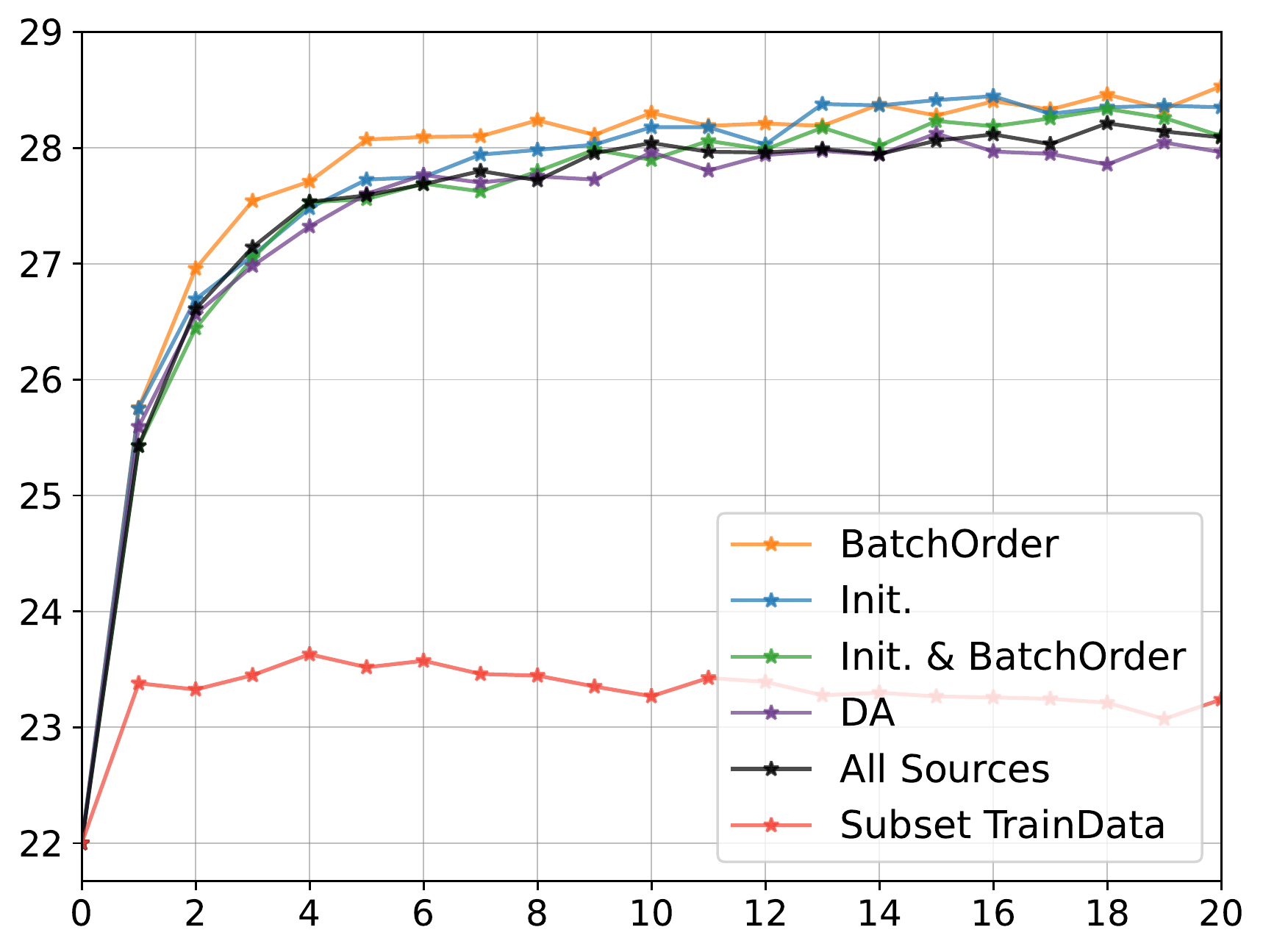}
        \\[-0.7em]
        {\small models in ensemble}
    \end{subfigure}
    \begin{subfigure}{0.32\linewidth}
        \centering
        VGG16\\
        \includegraphics[width=\linewidth]{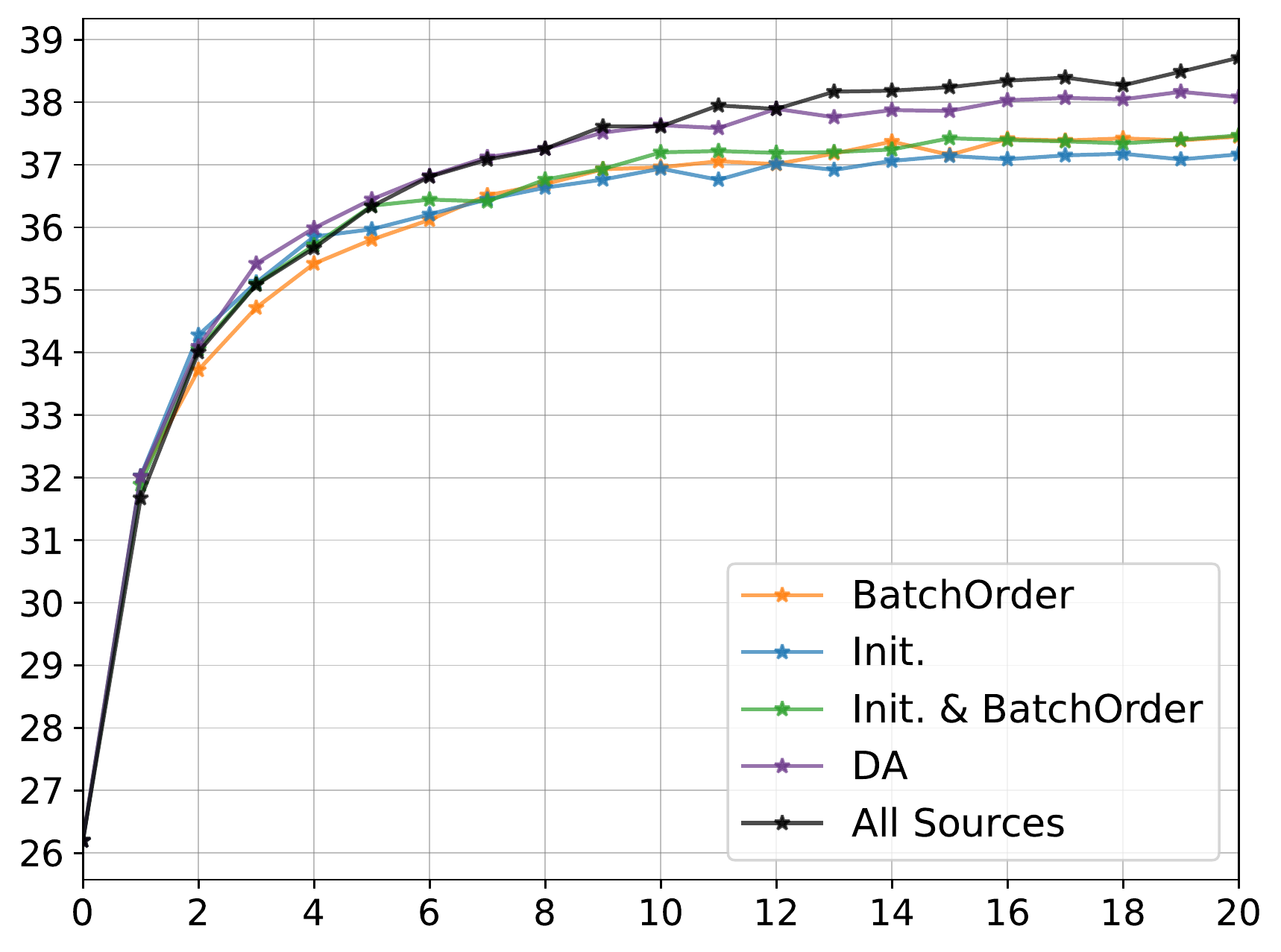}
        \\[-0.7em]
        {\small models in ensemble}
    \end{subfigure}
    \begin{subfigure}{0.32\linewidth}
        \centering
        ViT\\
        \includegraphics[width=\linewidth]{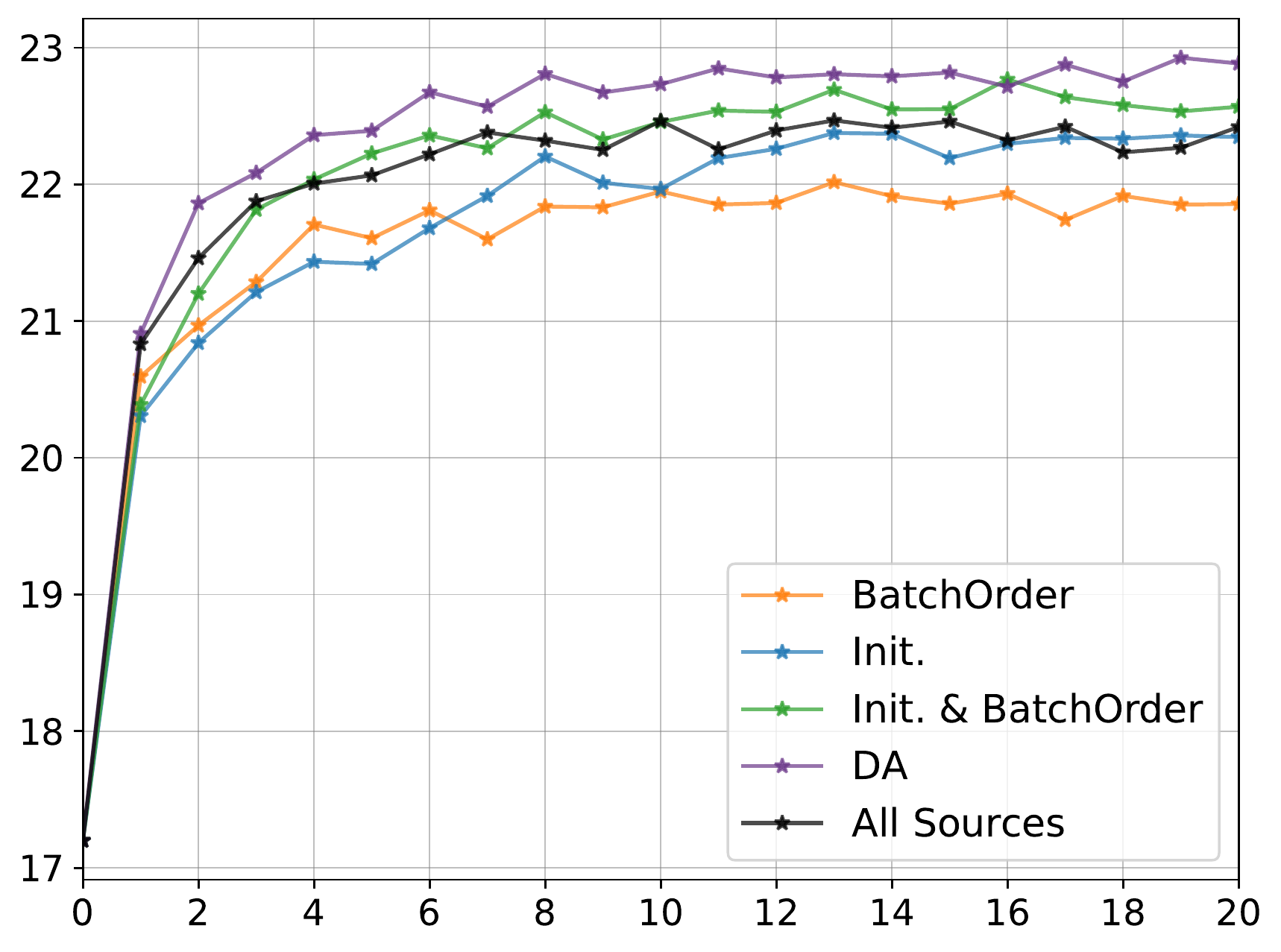}
        \\[-0.7em]
        {\small models in ensemble}
    \end{subfigure}
    \end{minipage}

	\caption{ Average Test Accuracy on CIFAR100 and TinyImageNet for Top-K and Bottom-K ($K=10$) Performing Classes
	}
    \label{fig:CIFAR100_Tinyimagenet_BestK_WorstK}
\end{figure*}

\begin{figure*}[!t]
    \subsection{Dominant sources of stochasticity}
    \vspace{0.2cm}
    \centering
    \underline{CIFAR100 Accuracy \% difference}\\
    \vspace{0.1cm}
    \begin{minipage}{0.01\linewidth}
        \rotatebox{90}{\% accuracy diff.}
    \end{minipage}
    \hspace{0.1cm}
    \begin{minipage}{0.97\linewidth}
    \begin{subfigure}{0.32\linewidth}
        \centering
        ResNet9\\
        \includegraphics[width=\linewidth]{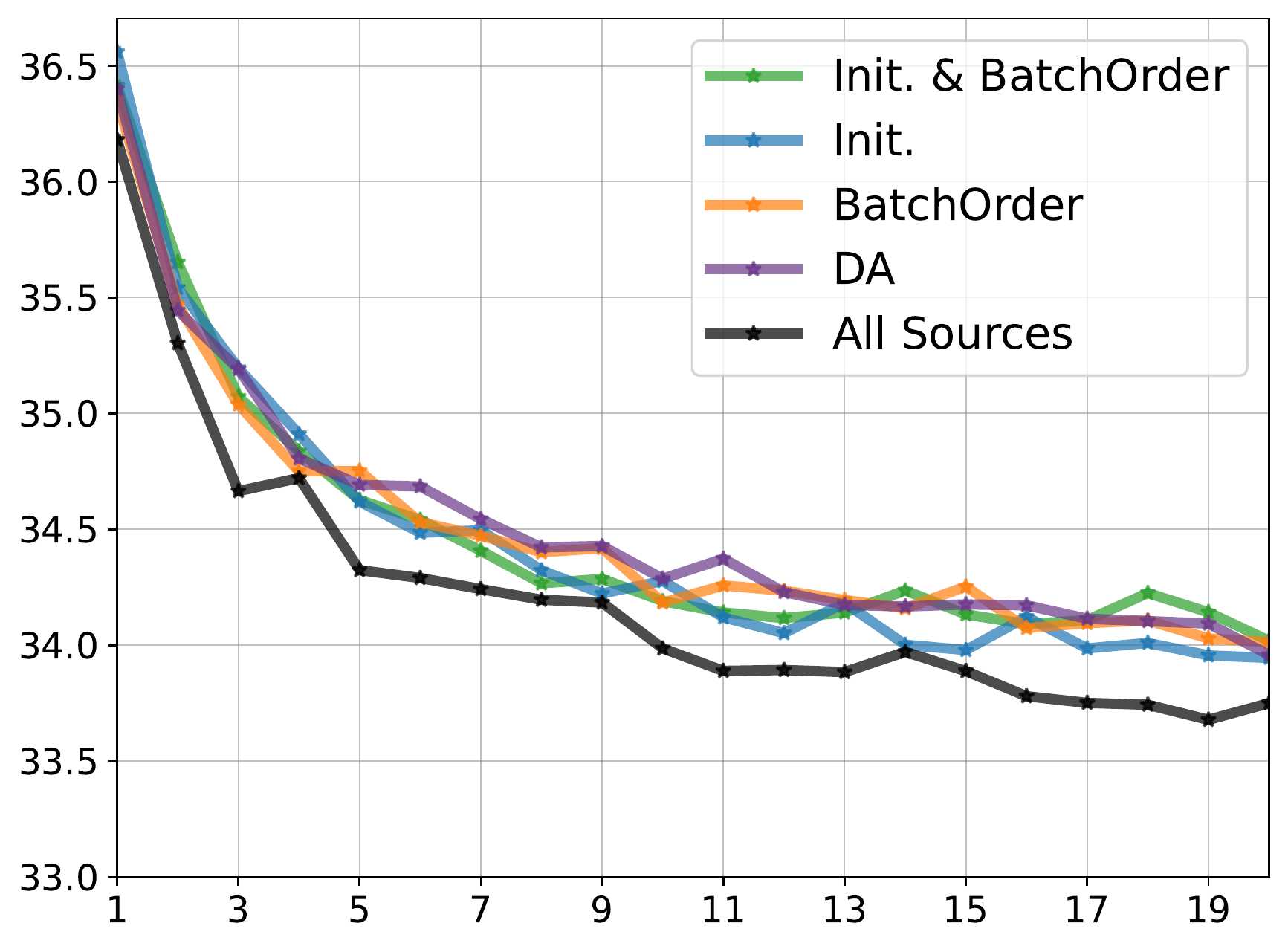}\\[-0.7em]
        models in ensemble
    \end{subfigure}
    \begin{subfigure}{0.32\linewidth}
        \centering
        VGG16\\
        \includegraphics[width=\linewidth]{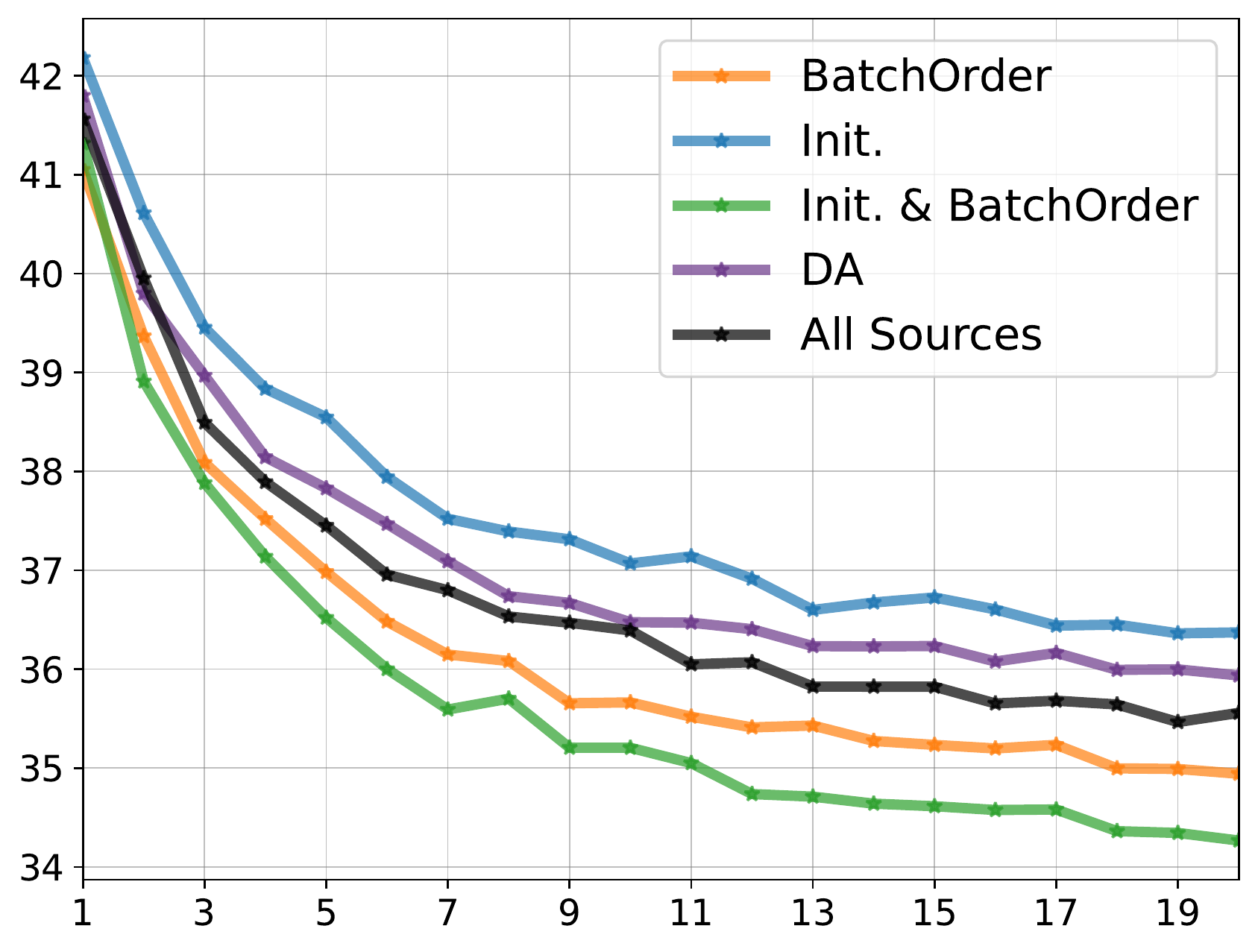}\\[-0.7em]
        models in ensemble
    \end{subfigure}
    \begin{subfigure}{0.32\linewidth}
        \centering
        MLP-Mixer\\
        \includegraphics[width=\linewidth]{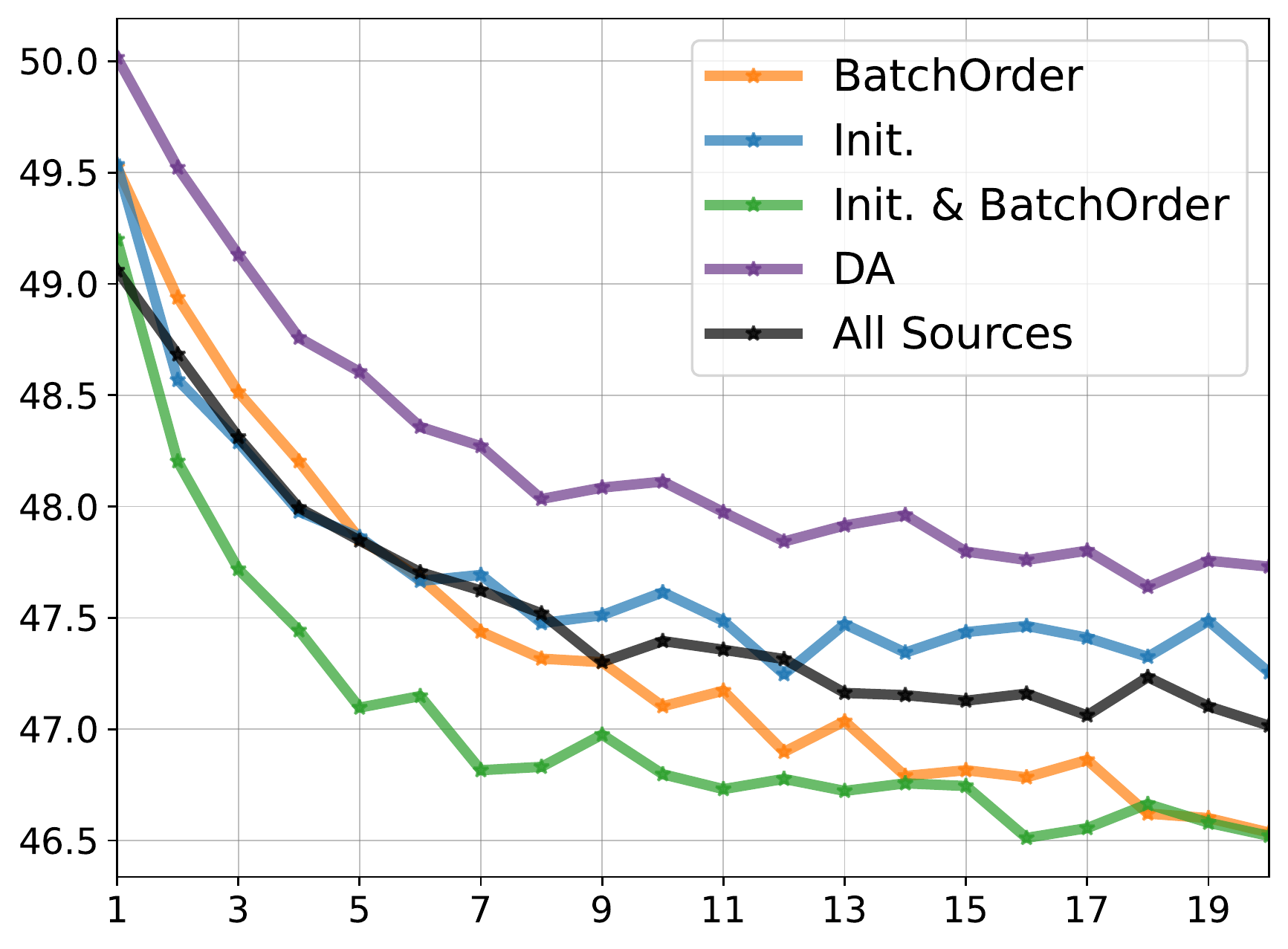}\\[-0.7em]
        models in ensemble
    \end{subfigure}
    \end{minipage}
    
    \centering
    \vspace{0.2cm}
    \underline{TinyImageNet Accuracy \% difference}\\
    \vspace{0.1cm}
    \begin{minipage}{0.01\linewidth}
        \rotatebox{90}{\% accuracy diff.}
    \end{minipage}
    \hspace{0.1cm}
    \begin{minipage}{0.97\linewidth}
  	\begin{subfigure}{0.32\linewidth}
		\centering
        ResNet9\\
    	\includegraphics[width=\linewidth]{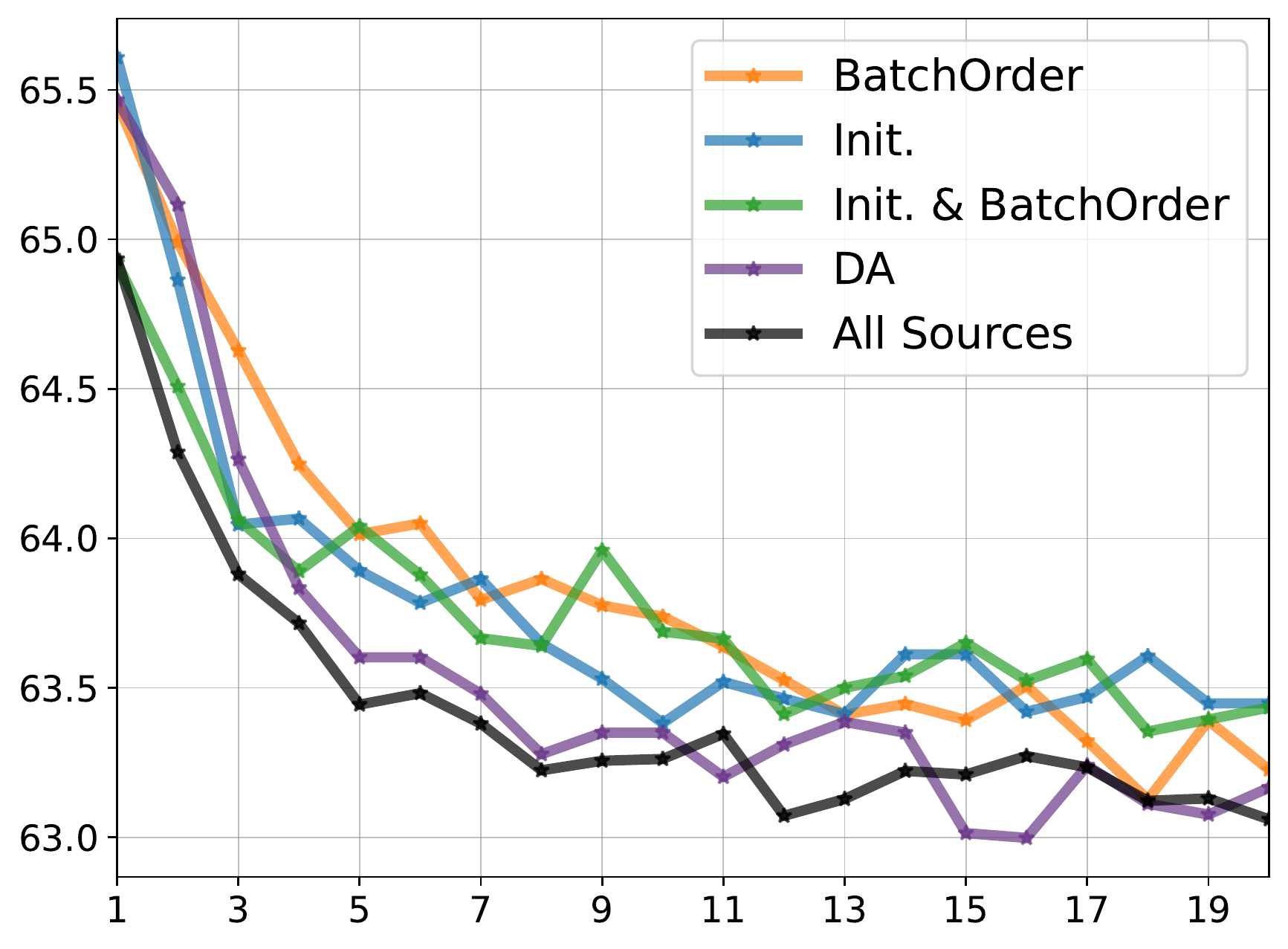}\\[-0.7em]
    	models in ensemble
        \label{fig:Tinyimagenet_res9_diff}
	\end{subfigure}
 	\begin{subfigure}{0.32\linewidth}
		\centering
        VGG16\\
    	\includegraphics[width=\linewidth]{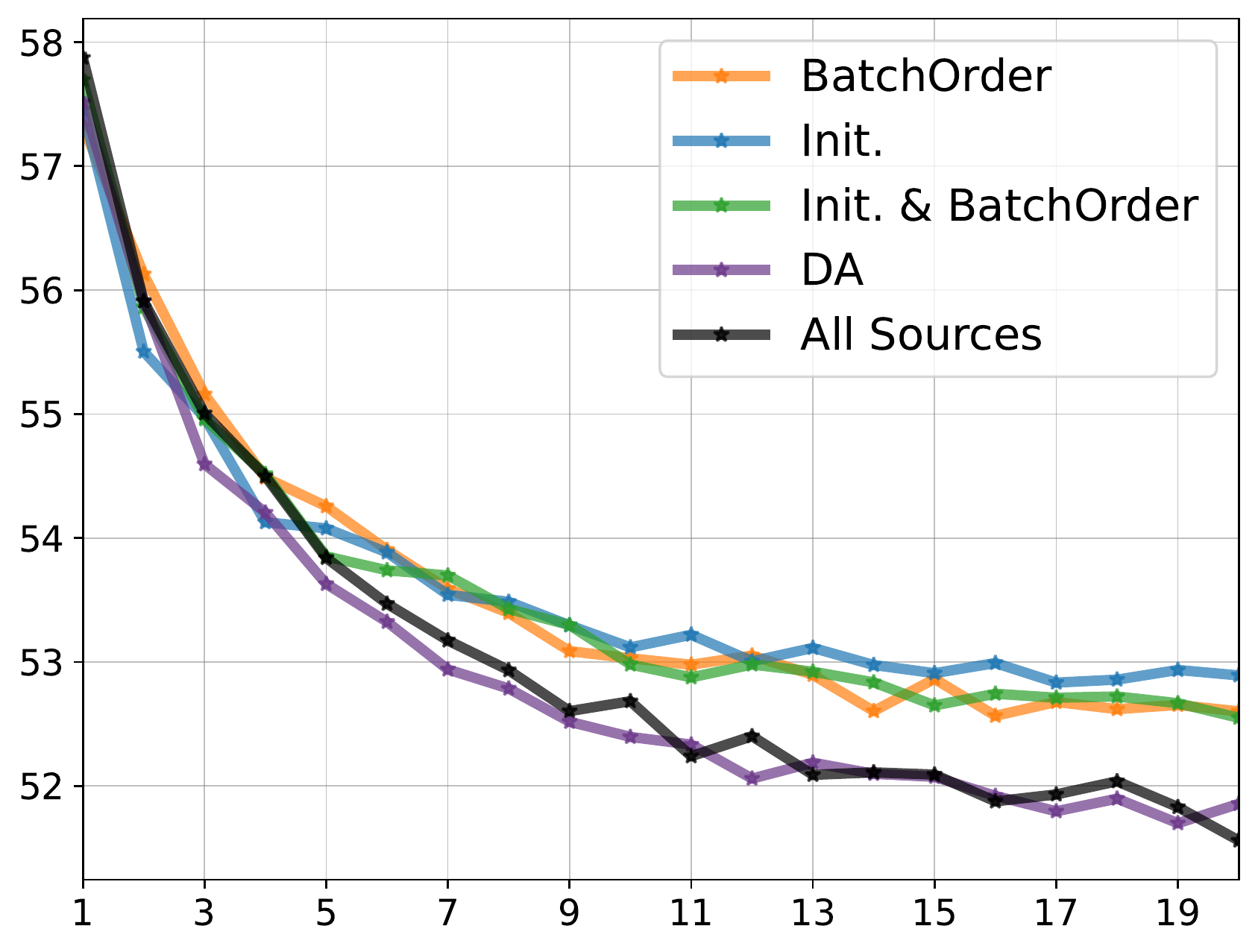}\\[-0.7em]
    	models in ensemble
        \label{fig:Tinyimagenet_vgg16_diff}
	\end{subfigure}
 	\begin{subfigure}{0.32\linewidth}
		\centering
        ViT\\
    	\includegraphics[width=\linewidth]{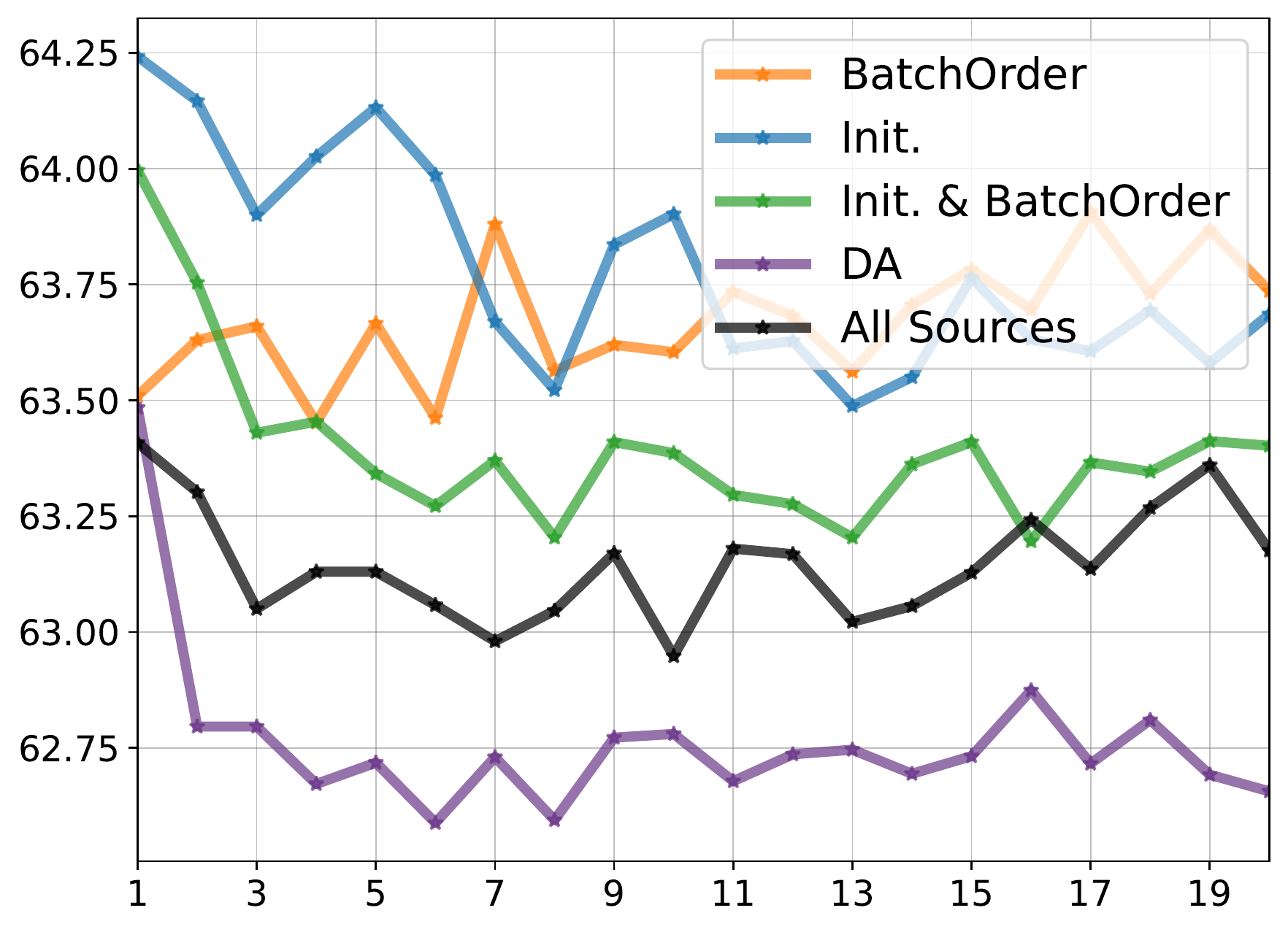}\\[-0.7em]
    	models in ensemble
        \label{fig:Tinyimagenet_vit_diff}
	\end{subfigure}
    \end{minipage}
    
	\caption{ Accuracy \% difference between top and bottom 10 classes for ResNet-9, VGG16, and MLPMixer trained on CIFAR100 and TinyImageNet
	}
    \label{fig:training_dynamics_extra}
\end{figure*}

\begin{figure*}[!t]
    \section{Results and Discussion}
    \vspace{0.2cm}
    \subsection{Comparison between ResNet Architectures}
    \vspace{0.2cm}
    \centering
    \underline{Top-K}\\
    \vspace{0.1cm}
    \begin{minipage}{0.01\linewidth}
        \rotatebox{90}{test accuracy \%}
    \end{minipage}
    \hspace{0.1cm}
    \begin{minipage}{0.97\linewidth}
    \begin{subfigure}{0.24\linewidth}
        \centering
        ResNet9\\
    	\includegraphics[width=1.0\linewidth]{figures/CIFAR100/ensemble_plots/resnet9_20/resnet9_Best_K.pdf}
        \\[-0.7em]
        models in ensemble
    \end{subfigure}
    \begin{subfigure}{0.24\linewidth}
        \centering
        ResNet18\\
    	\includegraphics[width=1.0\linewidth]{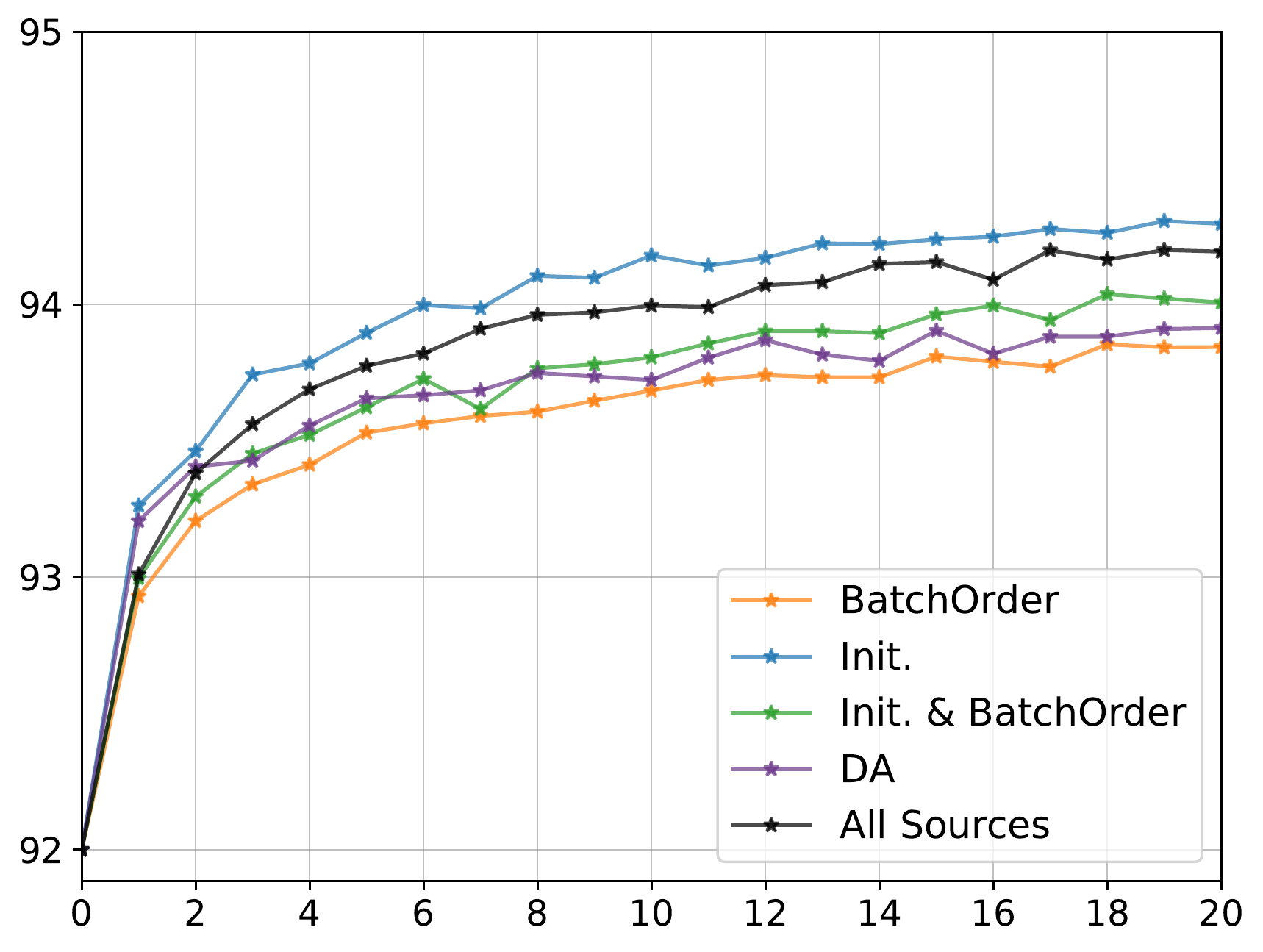}
        \\[-0.7em]
        models in ensemble
    \end{subfigure}
    \begin{subfigure}{0.24\linewidth}
        \centering
        ResNet34\\
        \includegraphics[width=1.0\linewidth]{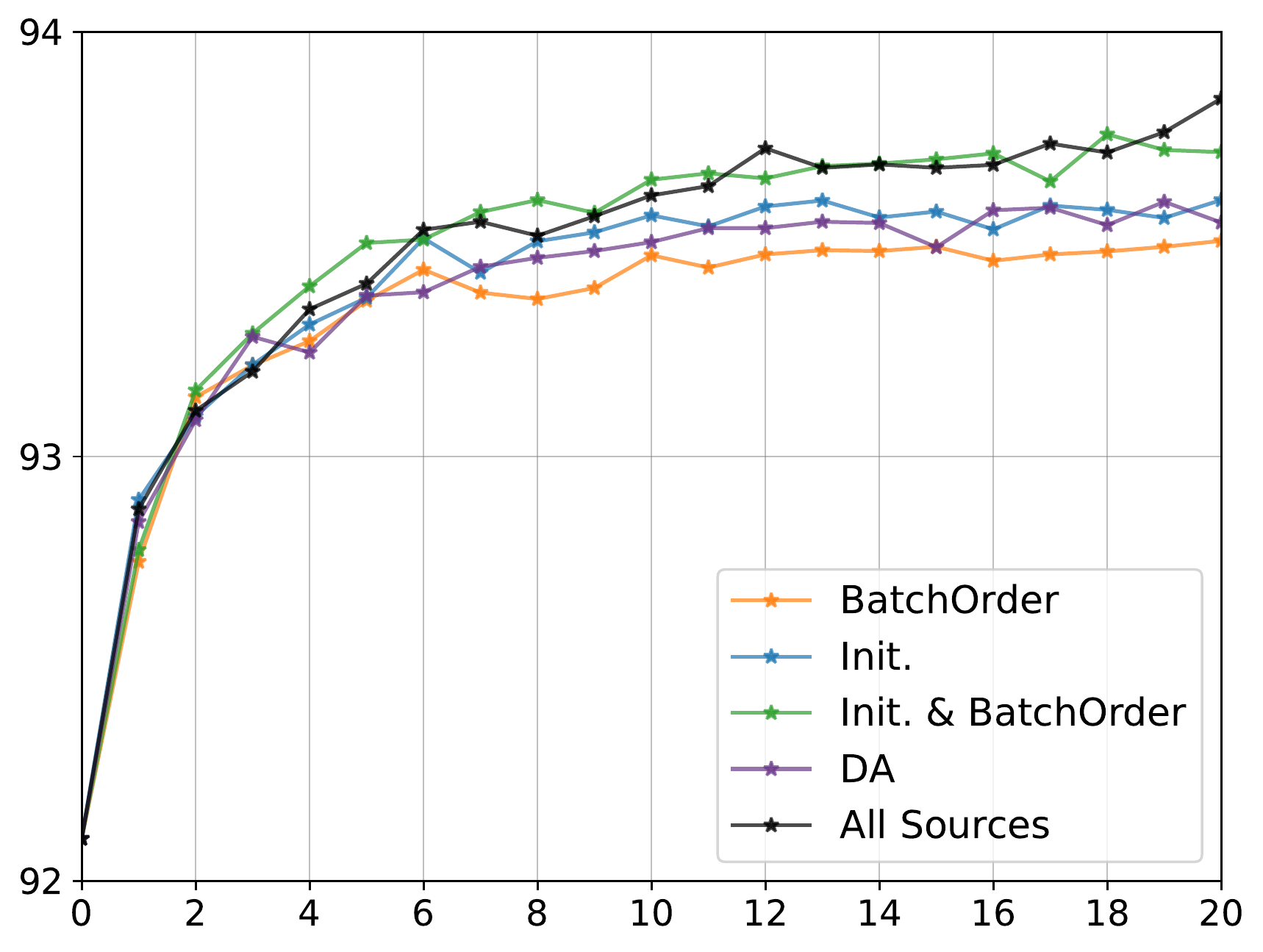}  
        \\[-0.7em]
        models in ensemble
    \end{subfigure}
    \begin{subfigure}{0.24\linewidth}
        \centering
        ResNet50\\
        \includegraphics[width=1.0\linewidth]{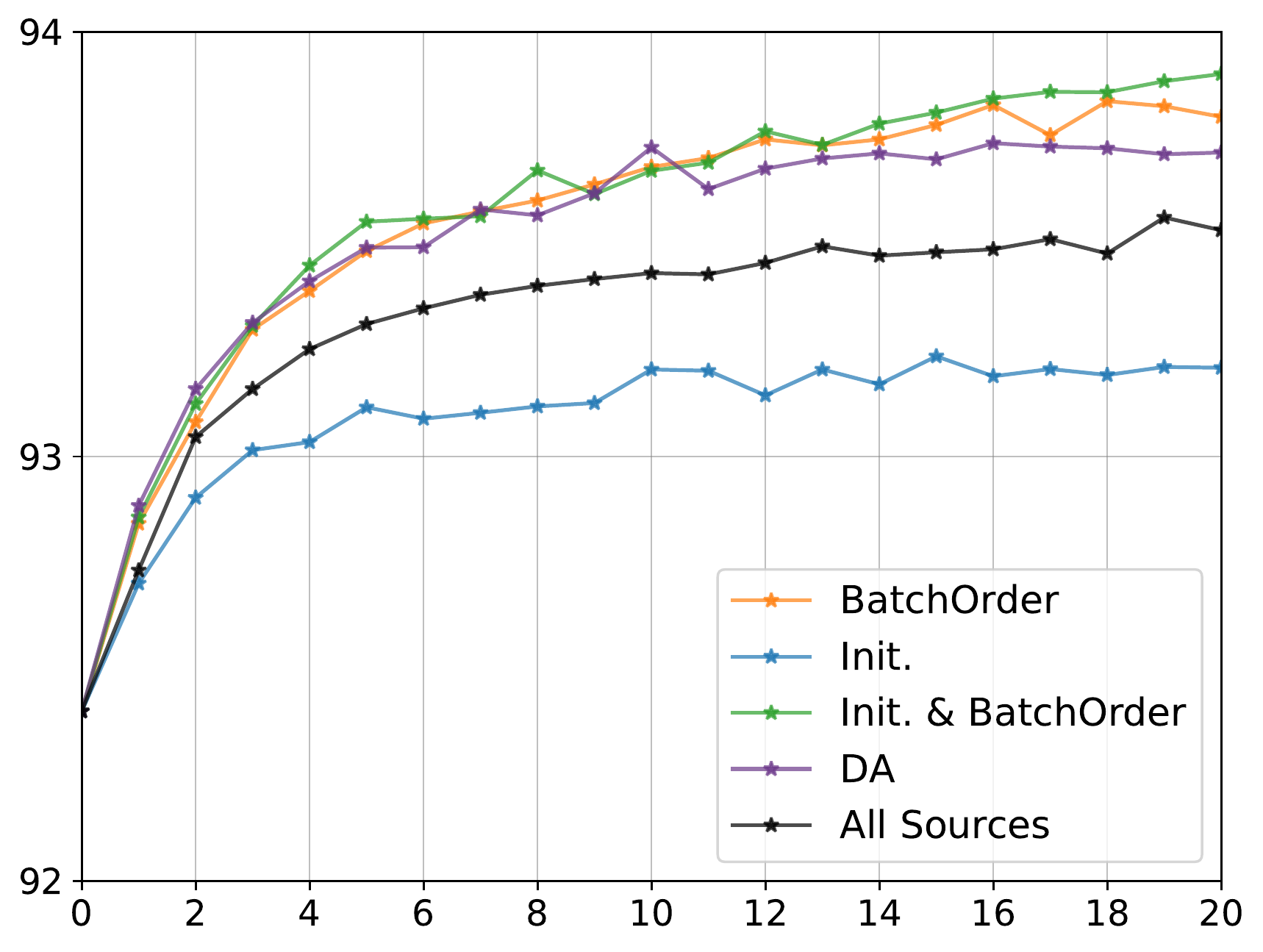}
        \\[-0.7em]
        models in ensemble
    \end{subfigure}
    \end{minipage}
    \vspace{0.3cm}
    
    \centering
    \underline{Bottom-K}\\
    \vspace{0.1cm}
    \begin{minipage}{0.01\linewidth}
        \rotatebox{90}{test accuracy \%}
    \end{minipage}
    \hspace{0.1cm}
    \begin{minipage}{0.97\linewidth}
    \begin{subfigure}{0.24\linewidth}
        \centering
        ResNet9\\
    	\includegraphics[width=1.0\linewidth]{figures/CIFAR100/ensemble_plots/resnet9_20/resnet9_Worst_K.pdf}
        \\[-0.7em]
        models in ensemble
    \end{subfigure}
    \begin{subfigure}{0.24\linewidth}
        \centering
        ResNet18\\
    	\includegraphics[width=1.0\linewidth]{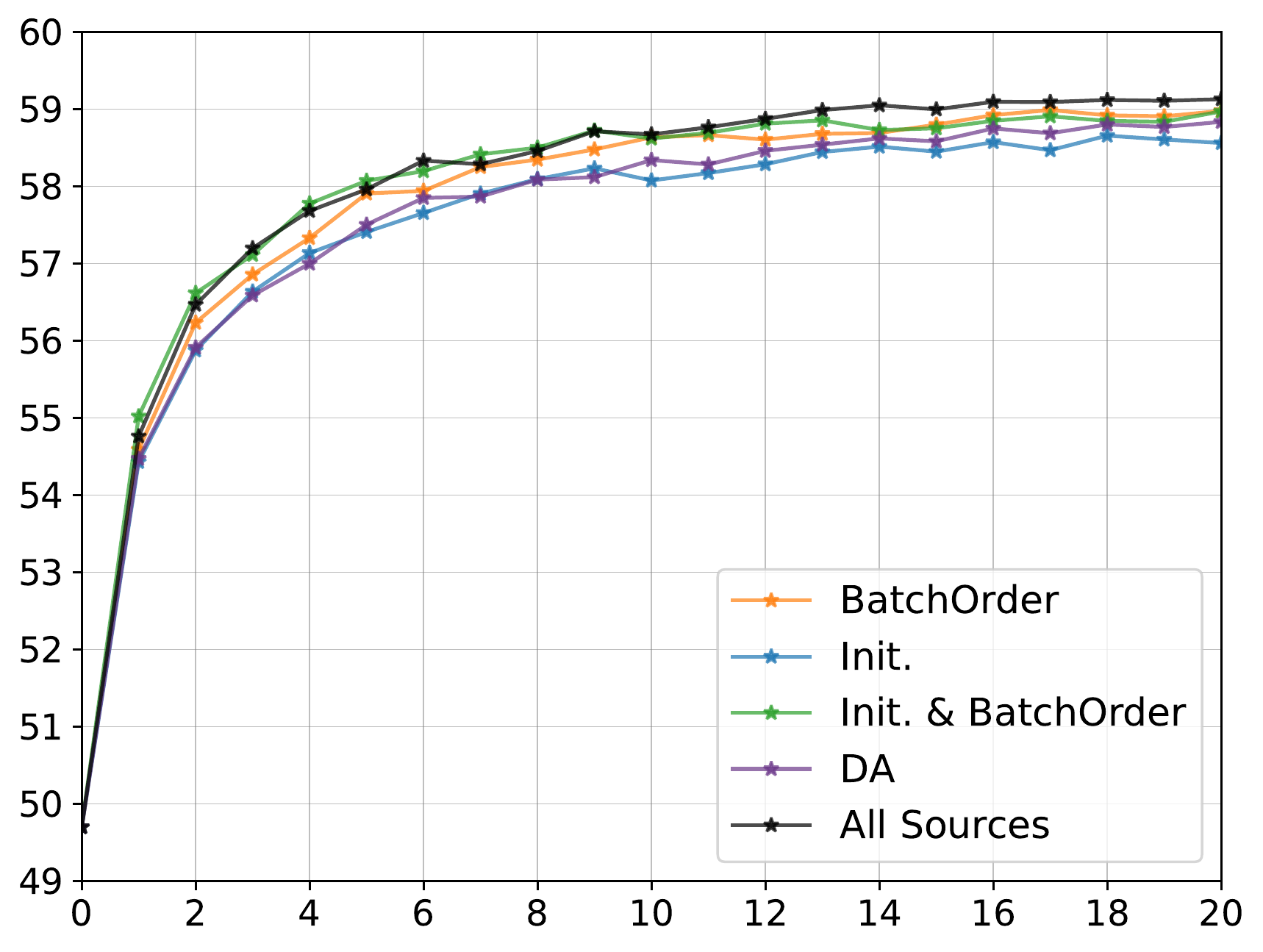}
        \\[-0.7em]
        models in ensemble
    \end{subfigure}
    \begin{subfigure}{0.24\linewidth}
        \centering
        ResNet34\\
        \includegraphics[width=1.0\linewidth]{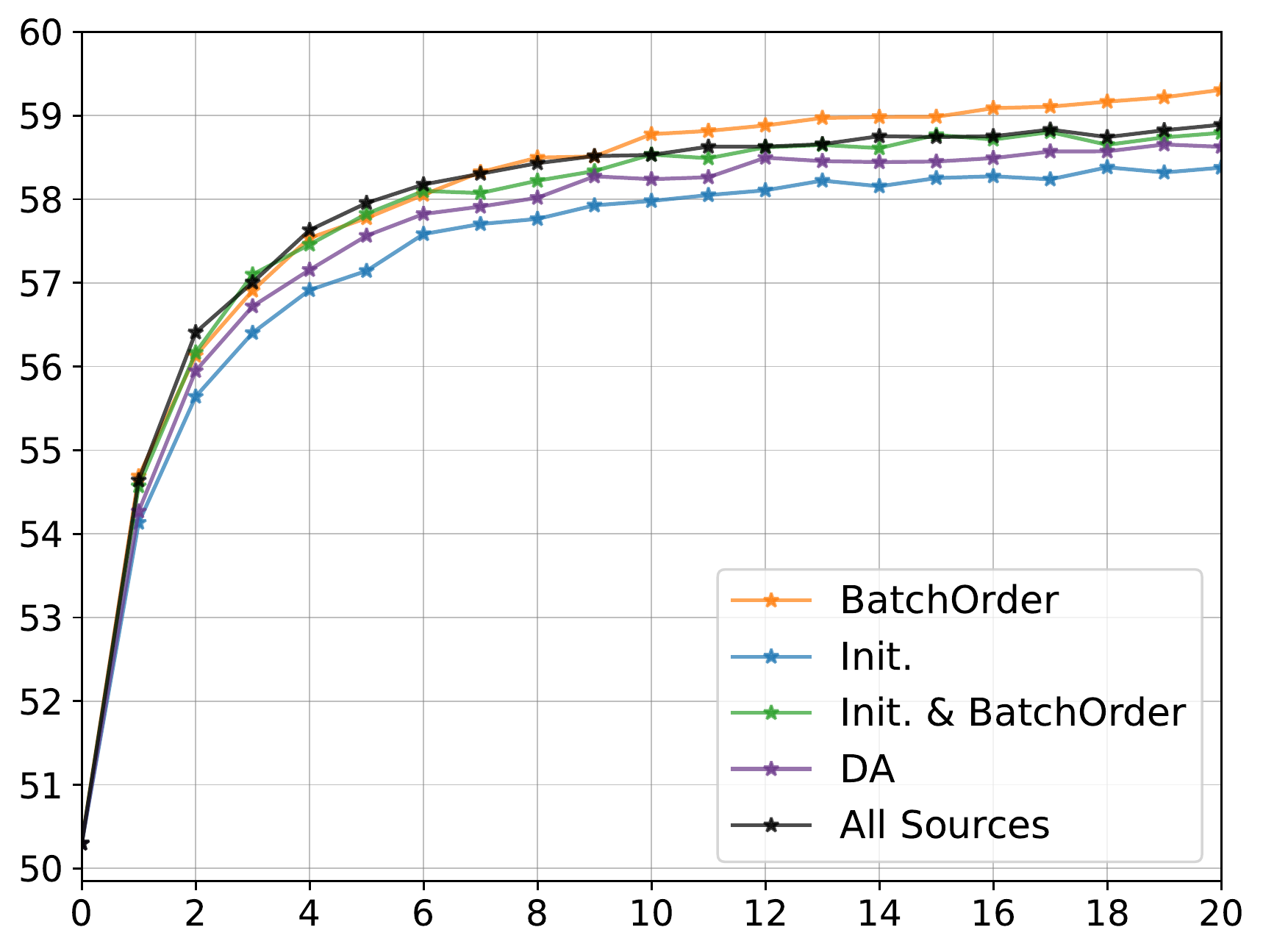}  
        \\[-0.7em]
        models in ensemble
    \end{subfigure}
    \begin{subfigure}{0.24\linewidth}
        \centering
        ResNet50\\
        \includegraphics[width=1.0\linewidth]{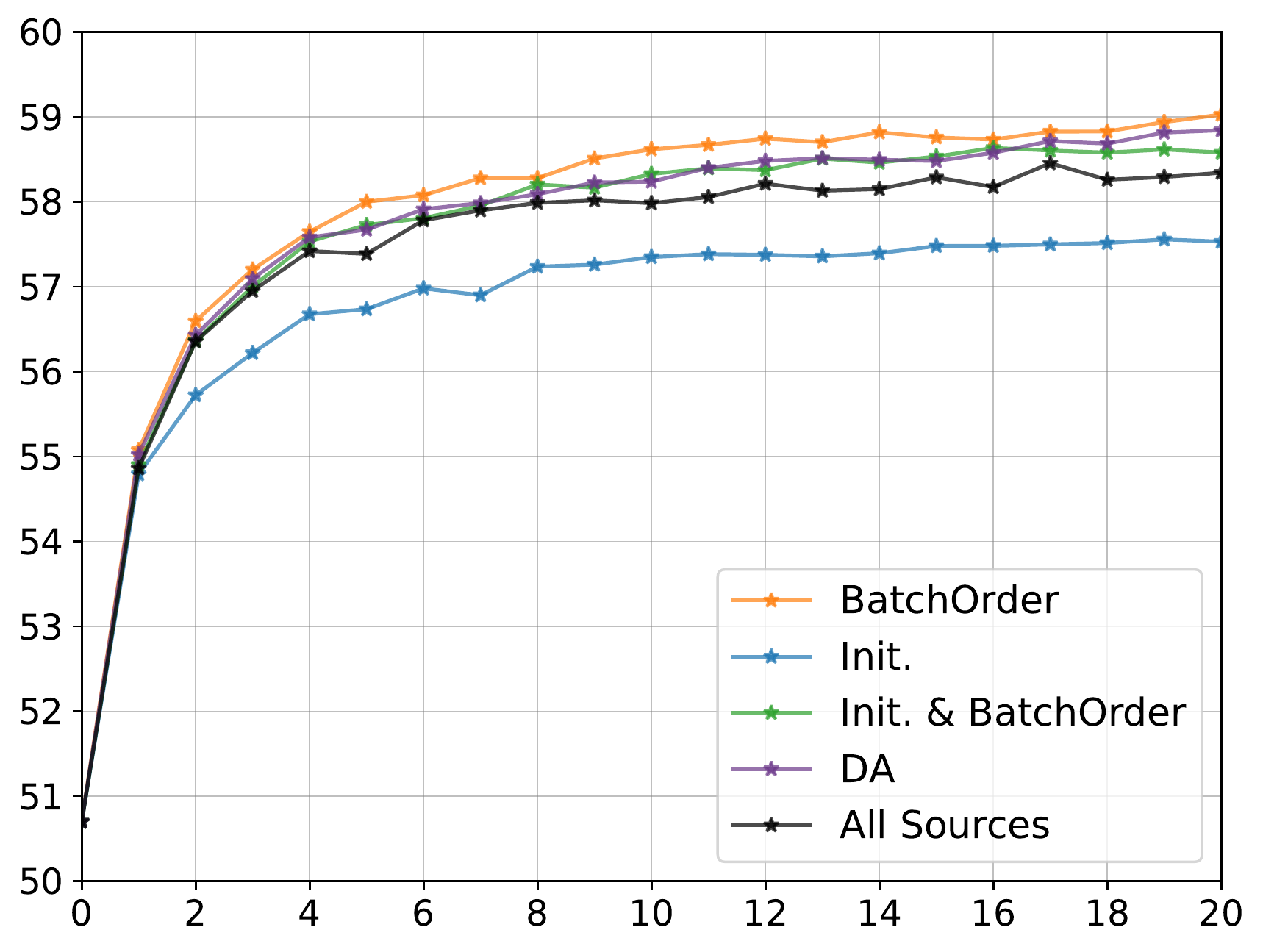}
        \\[-0.7em]
        models in ensemble
    \end{subfigure}
    \end{minipage}
    
	\caption{ Average Test Accuracy on CIFAR100 for Top-K and Bottom-K classes across different sizes of ResNets.
	}
    \label{fig:CIFAR100_resnets}
\end{figure*}

\begin{figure*}[ht]
    \vspace{0.2cm}
    \subsection{Benefits of even larger homogeneous ensembles}
    \vspace{0.2cm}
	\centering
    \begin{minipage}{0.01\linewidth}
        \rotatebox{90}{\hspace{1cm} test accuracy \%}
    \end{minipage}
    \hspace{0.07cm}
    \begin{minipage}{0.3\linewidth}
	\begin{subfigure}{=\linewidth}
		\centering
    	\includegraphics[width=1.0\linewidth]{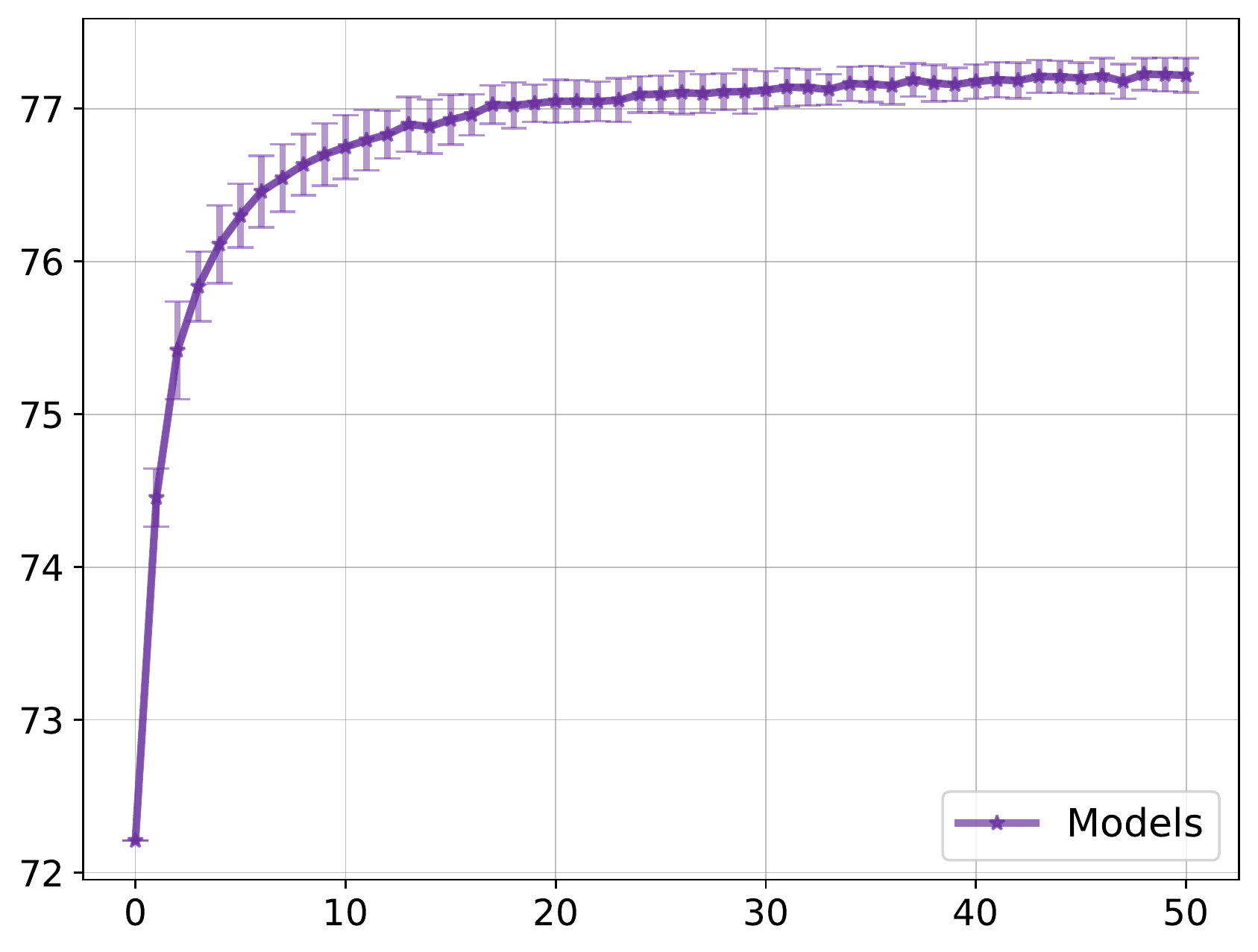}
        \\[-0.7em]
        models in ensemble
        \caption{ResNet9 on CIFAR100}
        \label{fig:ensemble_size_comp_res9}
	\end{subfigure}
    \end{minipage}
     \begin{minipage}{0.01\linewidth}
        \rotatebox{90}{\hspace{1cm} test accuracy \%}
    \end{minipage}
    \hspace{0.07cm}
    \begin{minipage}{0.3\linewidth}
	\begin{subfigure}{=\linewidth}
		\centering
    	\includegraphics[width=1.0\linewidth]{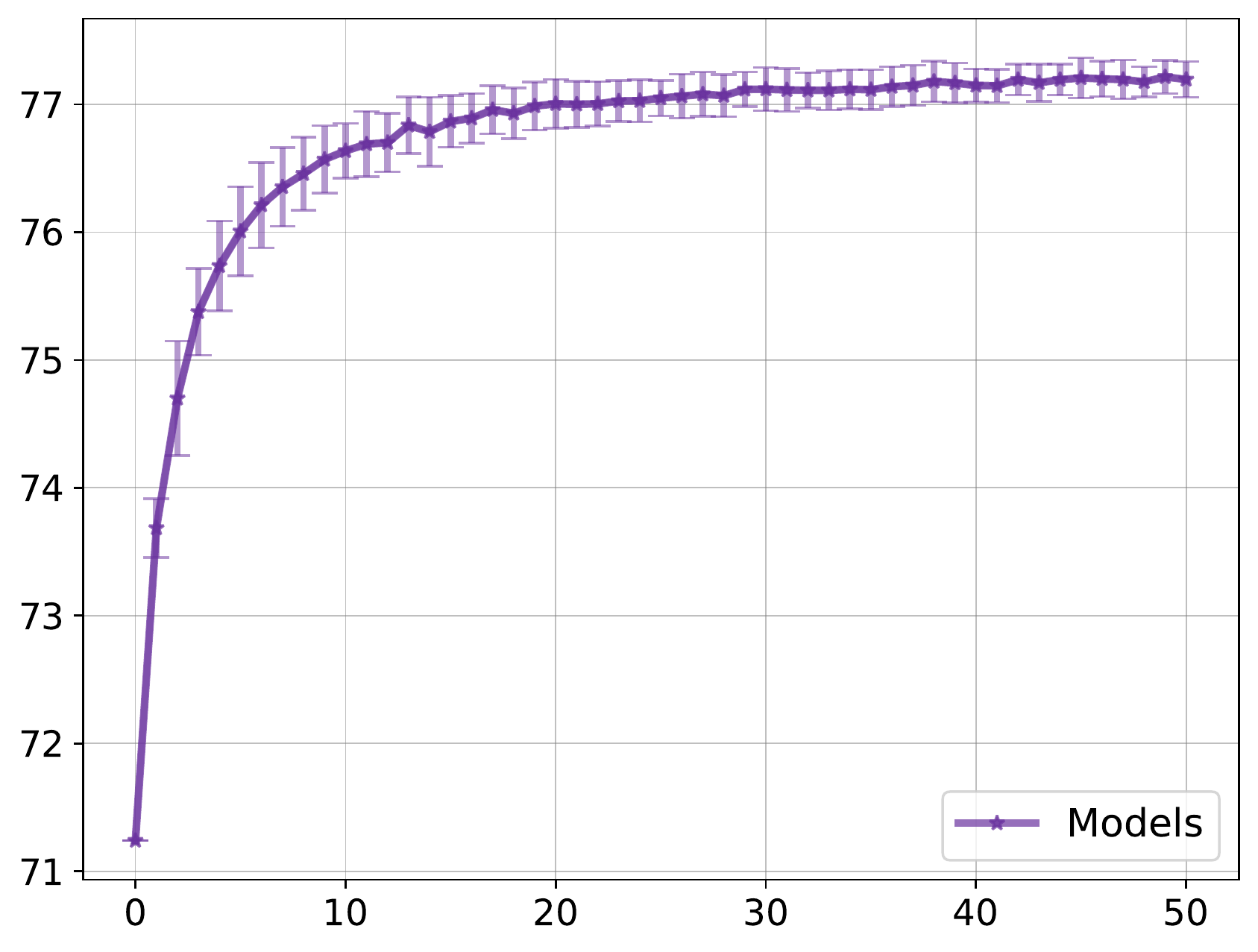}
        \\[-0.7em]
        models in ensemble
        \caption{VGG16 on CIFAR100}
        \label{fig:ensemble_size_comp_vgg16}
	\end{subfigure}
    \end{minipage}
     \begin{minipage}{0.01\linewidth}
        \rotatebox{90}{\hspace{1cm} test accuracy \%}
    \end{minipage}
    \hspace{0.07cm}
    \begin{minipage}{0.3\linewidth}
	\begin{subfigure}{=\linewidth}
		\centering
    	\includegraphics[width=1.0\linewidth]{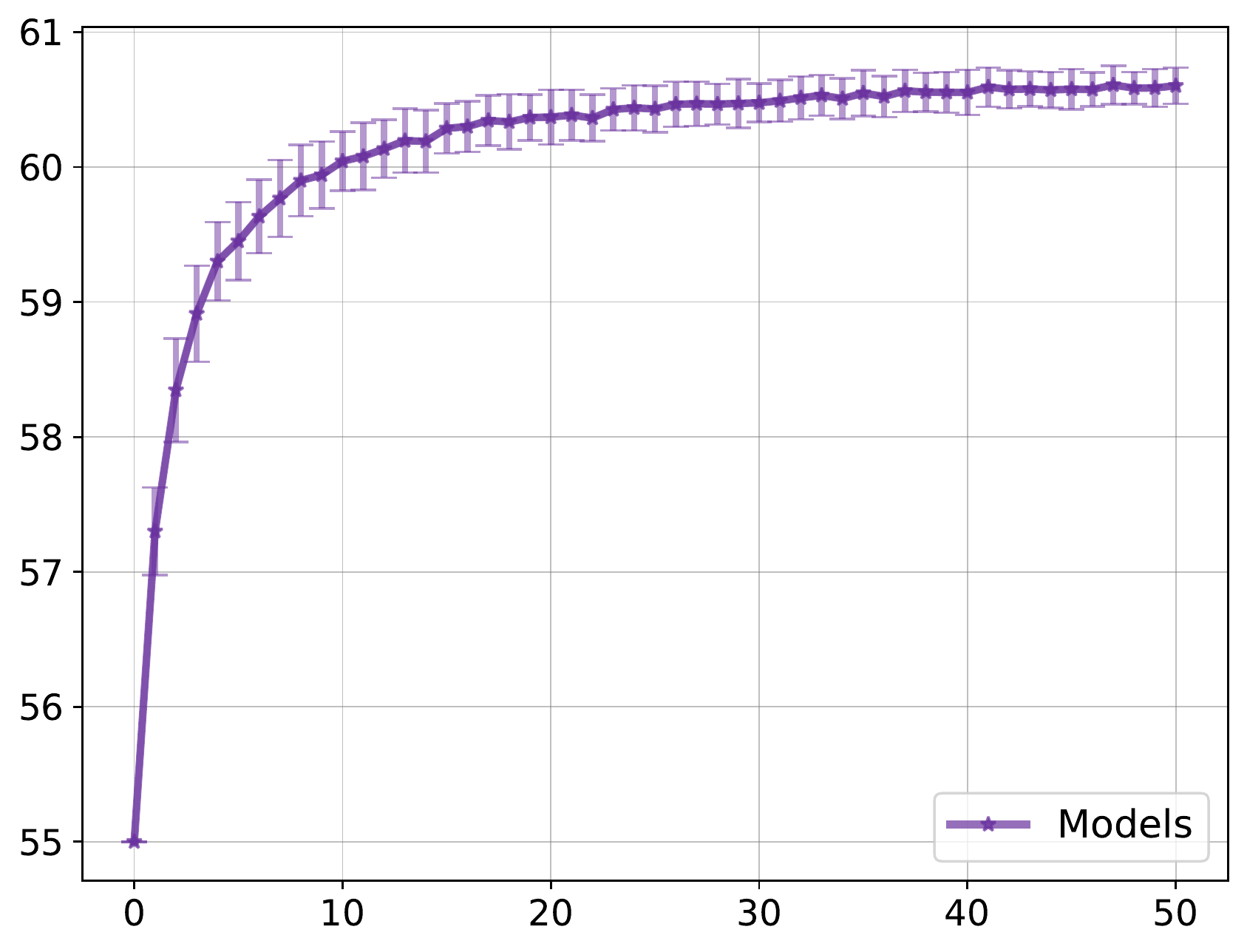}
        \\[-0.7em]
        models in ensemble
        \caption{ResNet50 on TinyImageNet}
        \label{fig:ensemble_size_comp_res50}
	\end{subfigure}
    \end{minipage}
	\caption{Average Accuracy per size of homogeneous ensemble. The average accuracy for each model added is calculated by averaging 100 random samples from a population of 50 models. We can see that the average accuracy starts to slowly plateau as the ensemble grows to 50 models. 
	}
	\label{fig:ensemble_size_comp}
\end{figure*}

\begin{figure*}
    \subsection{CIFAR-100}
    \centering
    \underline{Resent9 20 model ensemble}\\
    \vspace{0.1cm}

    \begin{minipage}{0.01\linewidth}
        \rotatebox{90}{\hspace{0.2cm} ensemble/base }
    \end{minipage}
    \begin{minipage}{0.98\linewidth}
	\begin{subfigure}{0.19\linewidth}
		\centering
        Init\\
    	\includegraphics[width=1.0\linewidth]{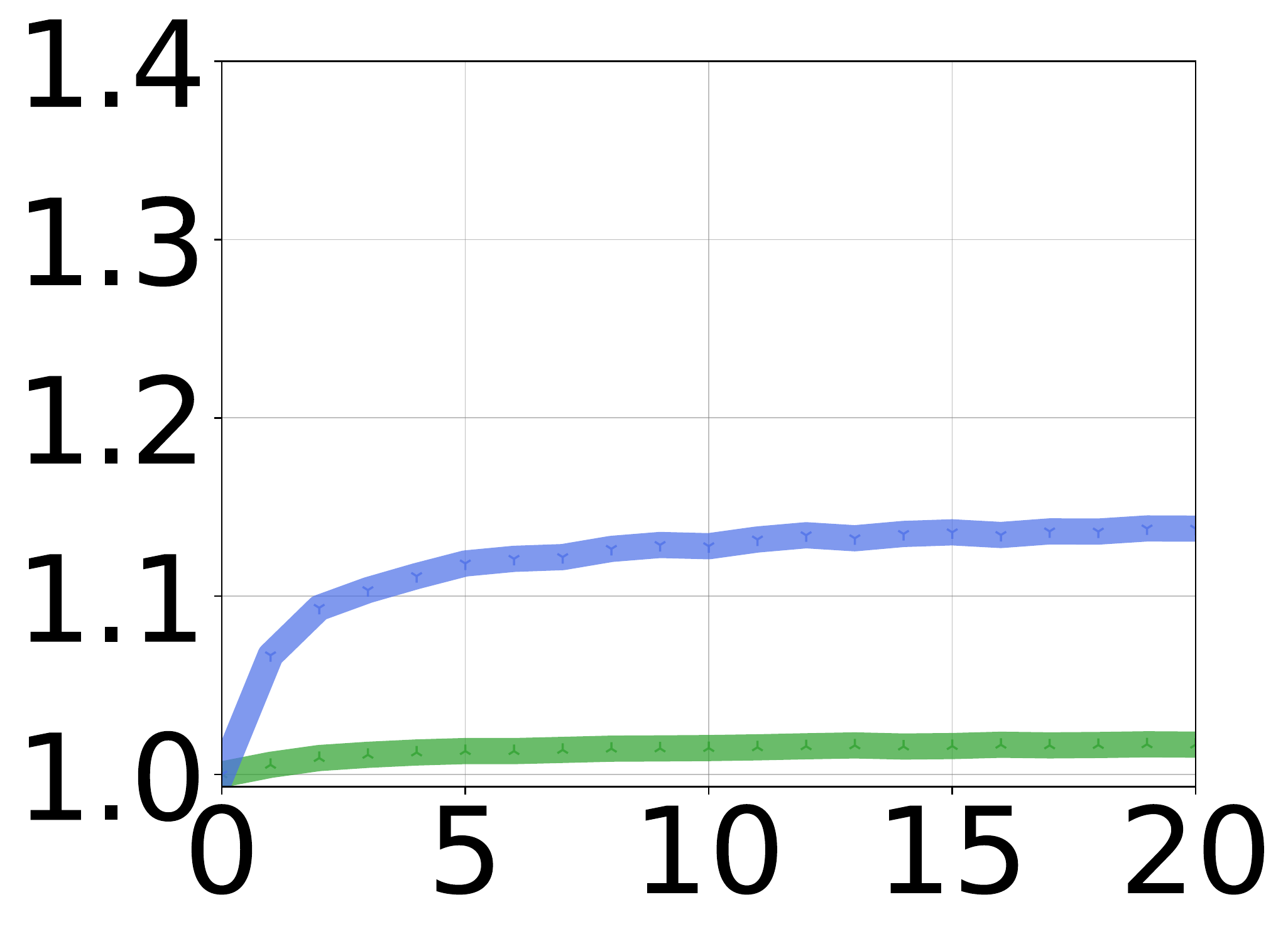}
	\end{subfigure}
 	\begin{subfigure}{0.19\linewidth}
		\centering
        BatchOrder\\
    	\includegraphics[width=1.0\linewidth]{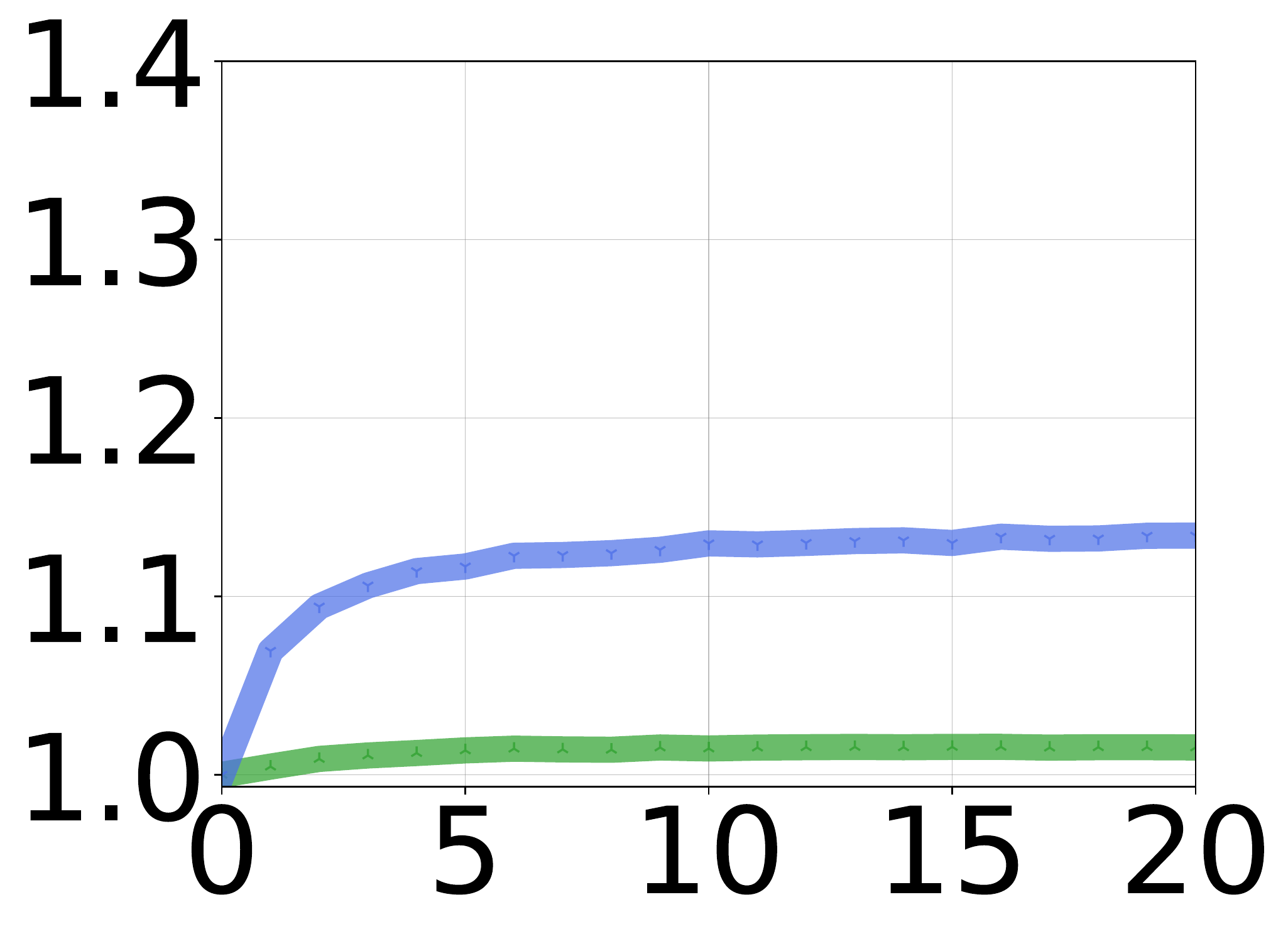}
	\end{subfigure}
	\begin{subfigure}{0.19\linewidth}
		\centering
        DA\\
    	\includegraphics[width=1.0\linewidth]{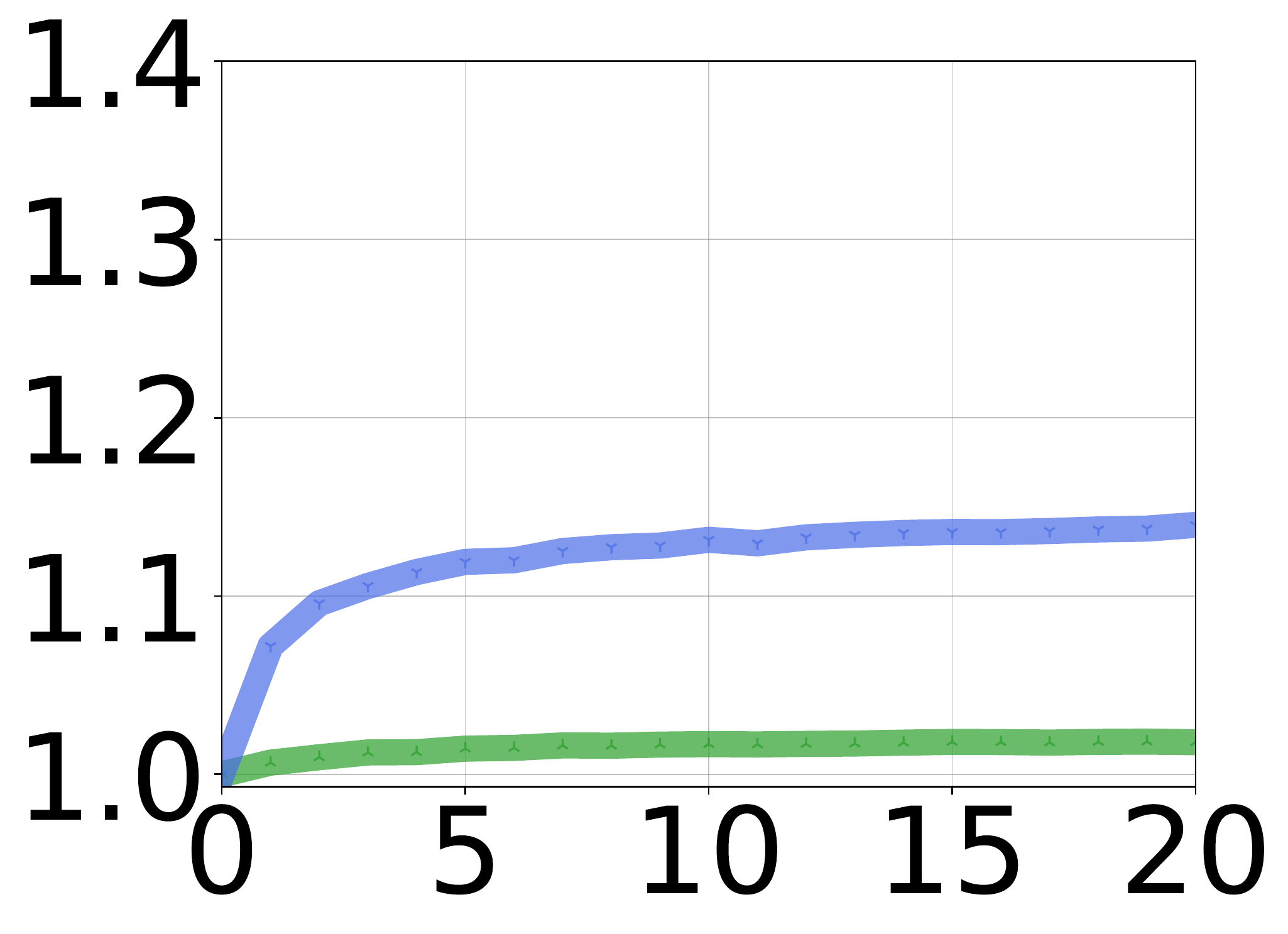}
	\end{subfigure}
 	\begin{subfigure}{0.19\linewidth}
		\centering
        Init \& BatchOrder\\
    	\includegraphics[width=1.0\linewidth]{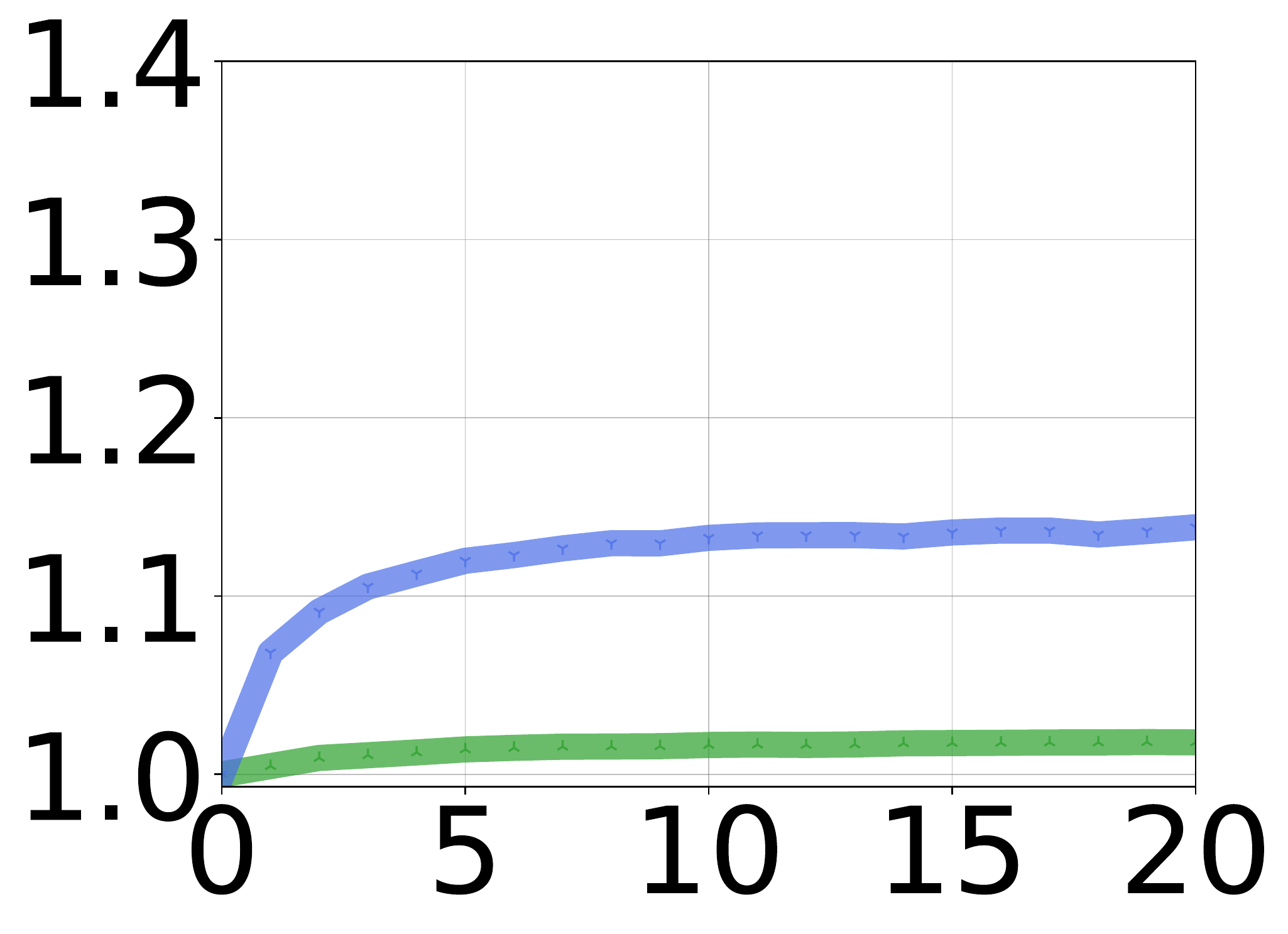}
	\end{subfigure}
   	\begin{subfigure}{0.19\linewidth}
		\centering
        All Sources\\
    	\includegraphics[width=1.0\linewidth]{figures/CIFAR100/ratio_plots/resnet9_20/resnet9_RANDOM_ratio.pdf}
	\end{subfigure}
    \end{minipage}

    \centering
    \underline{Resent9 50 model ensemble}\\
    \vspace{0.1cm}

    \begin{minipage}{0.01\linewidth}
        \rotatebox{90}{\hspace{0.2cm} ensemble/base }
    \end{minipage}
    \begin{minipage}{0.98\linewidth}
	\begin{subfigure}{0.19\linewidth}
		\centering
        Init\\
    	\includegraphics[width=1.0\linewidth]{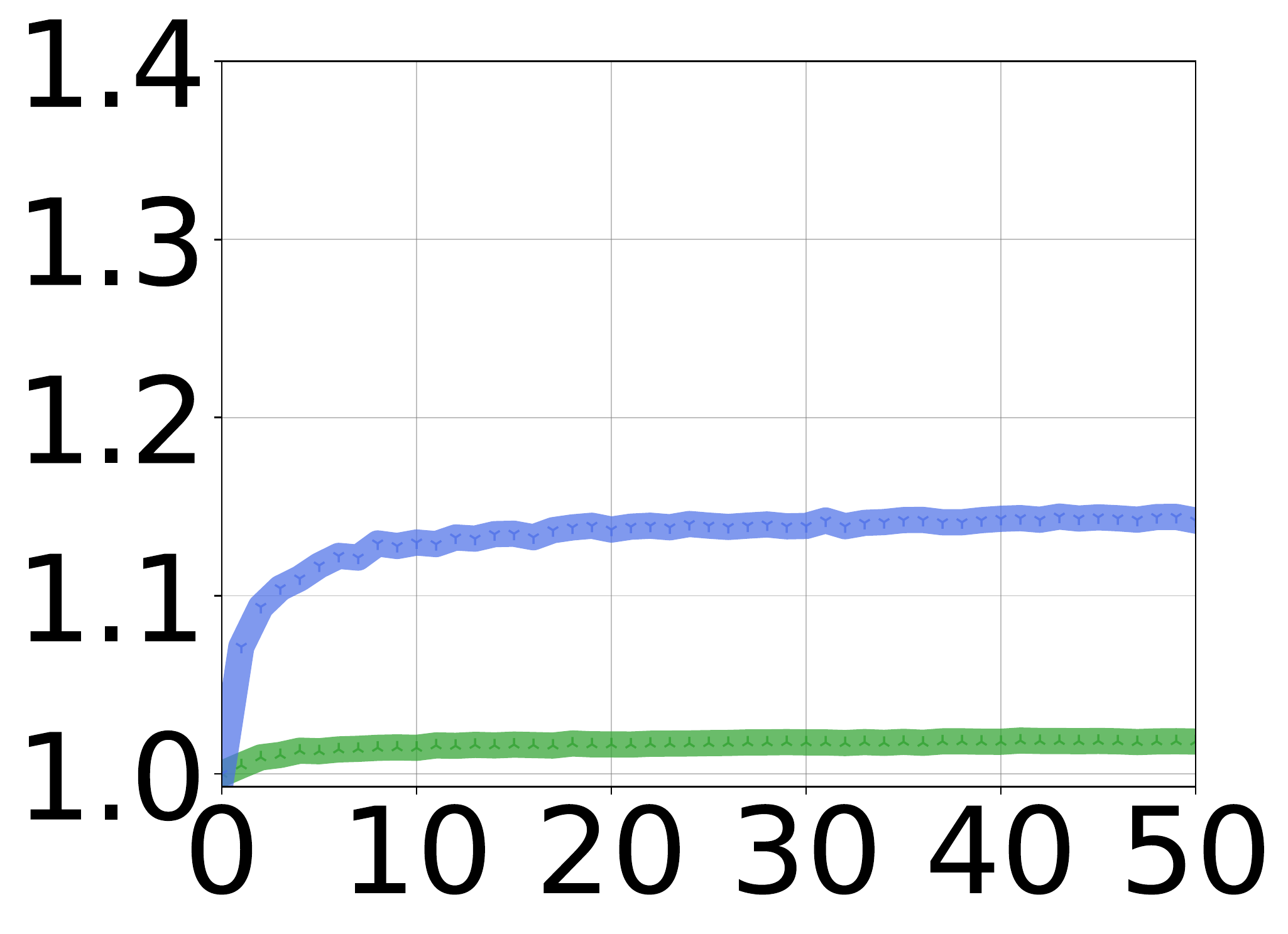}
	\end{subfigure}
 	\begin{subfigure}{0.19\linewidth}
		\centering
        BatchOrder\\
    	\includegraphics[width=1.0\linewidth]{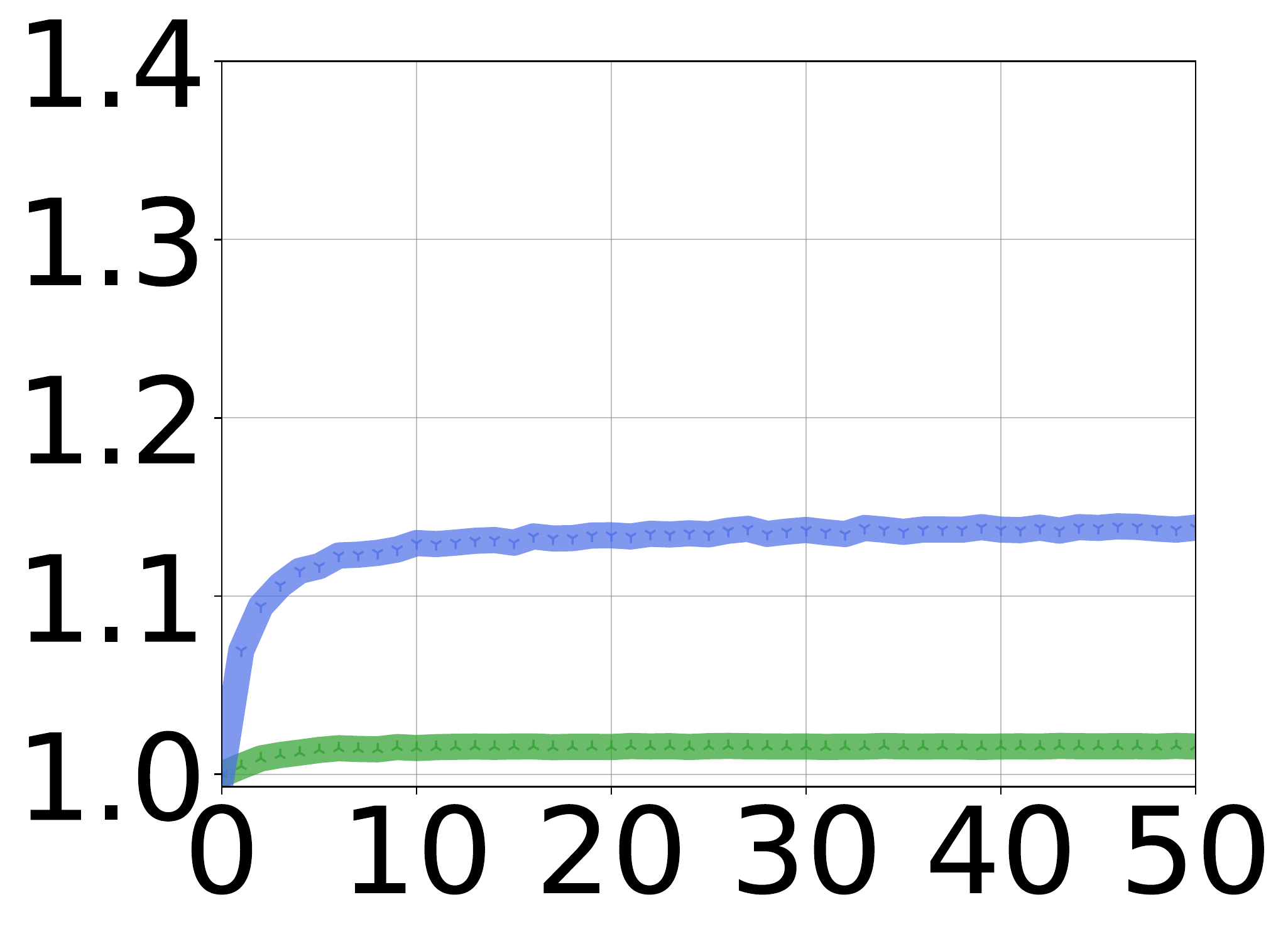}
	\end{subfigure}
	\begin{subfigure}{0.19\linewidth}
		\centering
        DA\\
    	\includegraphics[width=1.0\linewidth]{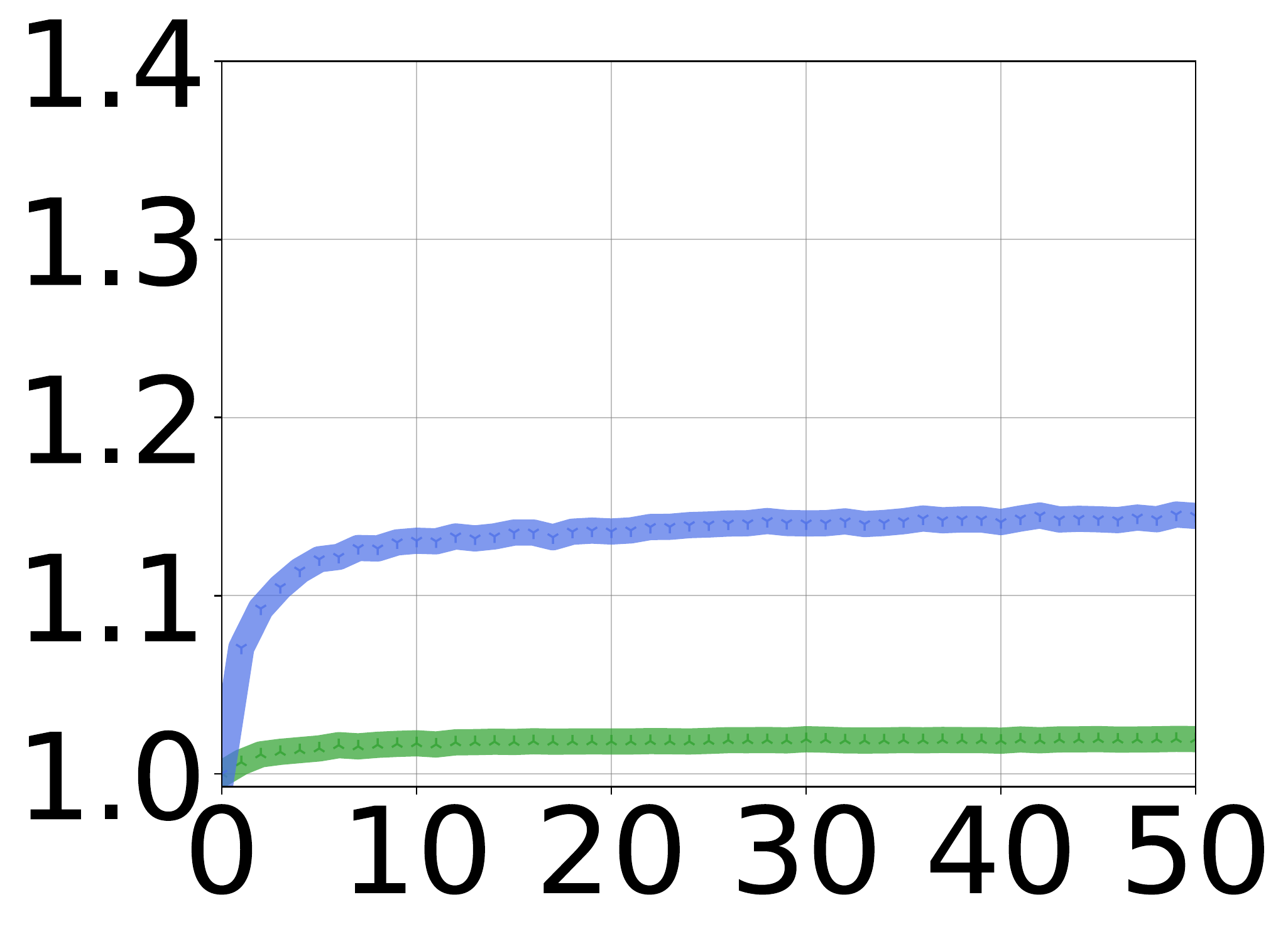}
	\end{subfigure}
 	\begin{subfigure}{0.19\linewidth}
		\centering
        Init \& BatchOrder\\
    	\includegraphics[width=1.0\linewidth]{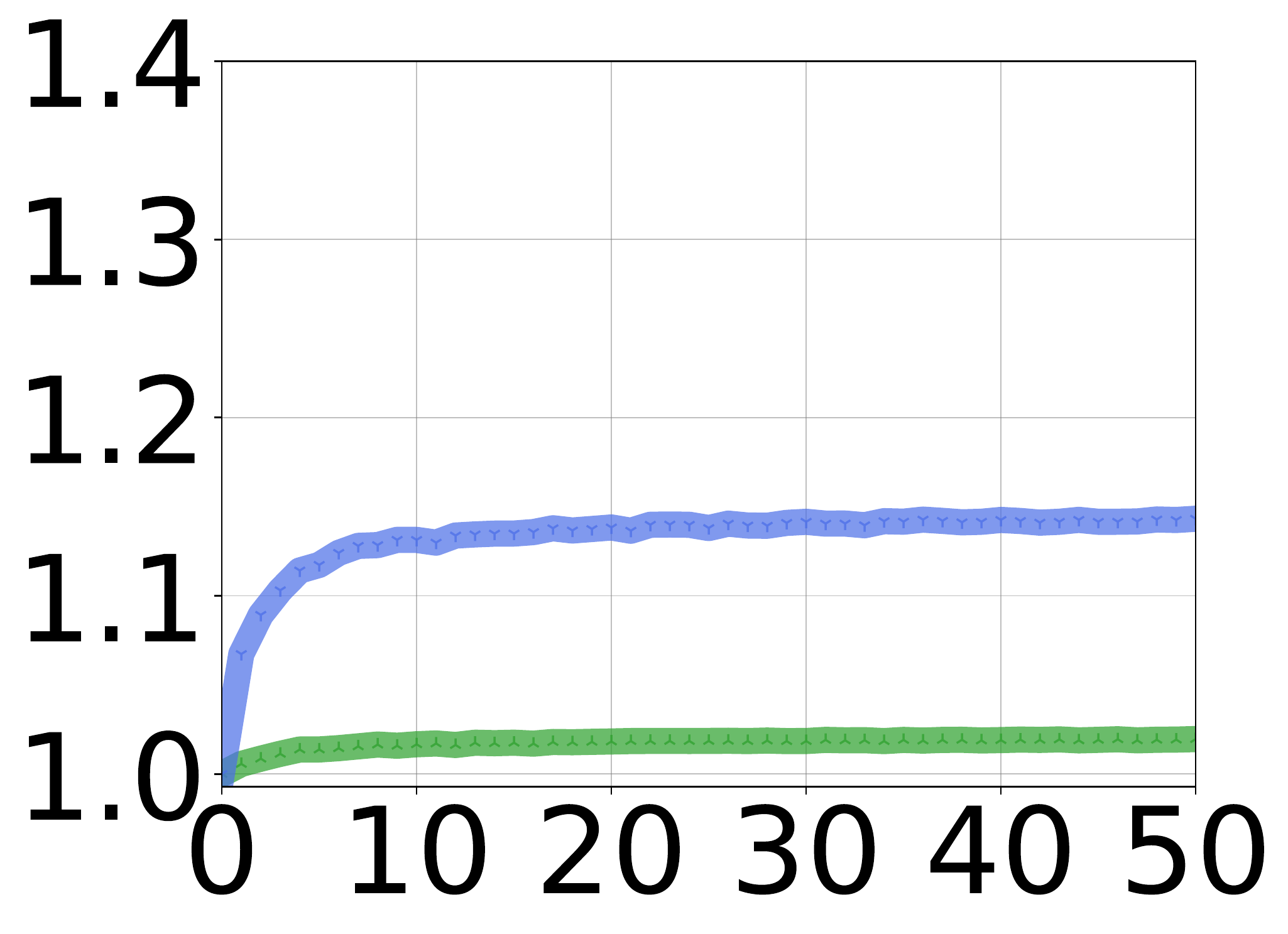}
	\end{subfigure}
   	\begin{subfigure}{0.19\linewidth}
		\centering
        All Sources\\
    	\includegraphics[width=1.0\linewidth]{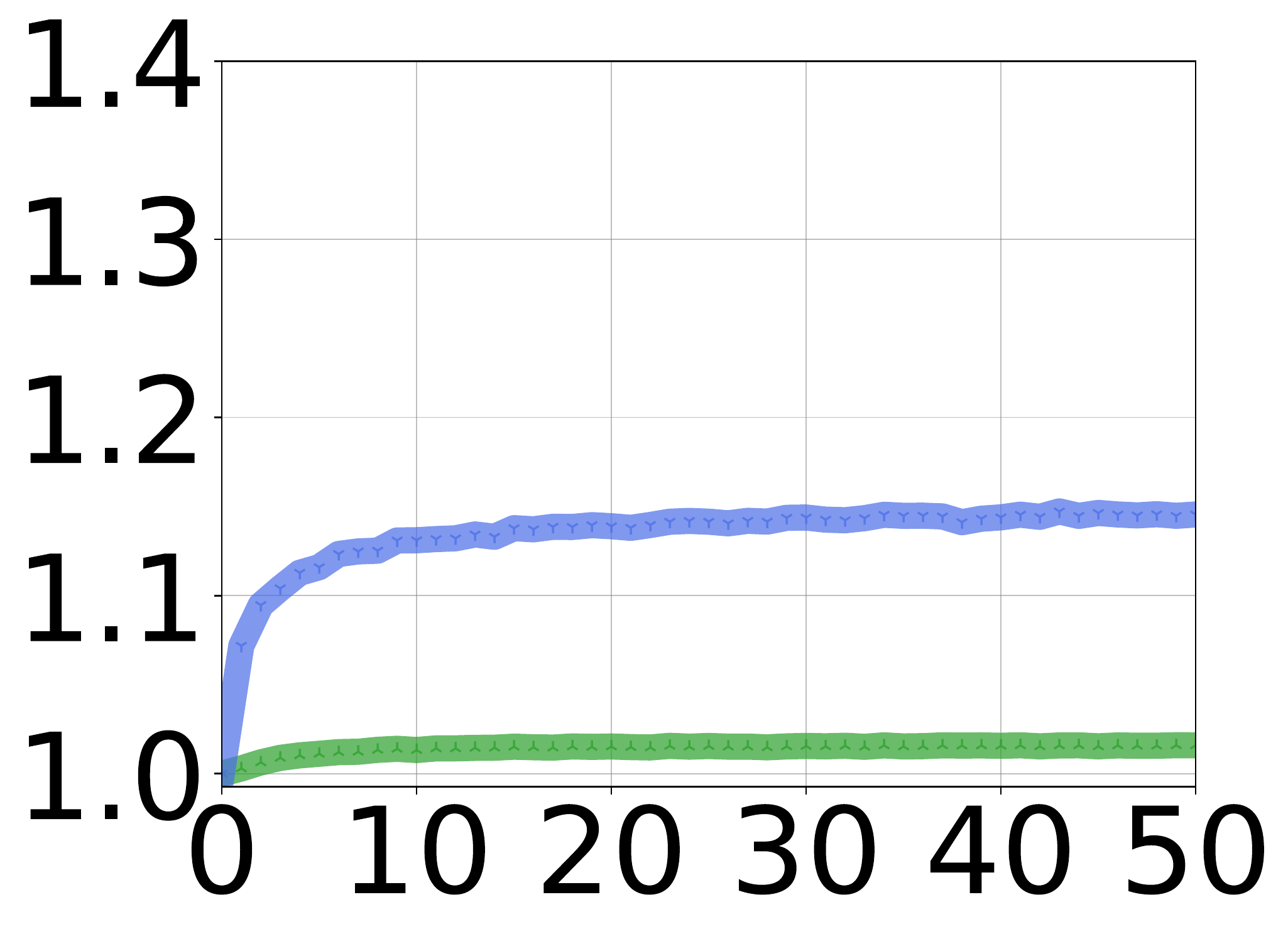}
	\end{subfigure}
    \end{minipage}

    \centering
    \underline{VGG16 20 model ensemble}\\
    \vspace{0.1cm}

    \begin{minipage}{0.01\linewidth}
        \rotatebox{90}{\hspace{0.2cm} ensemble/base}
    \end{minipage}
    \begin{minipage}{0.98\linewidth}
	\begin{subfigure}{0.19\linewidth}
		\centering
        Init\\
    	\includegraphics[width=1.0\linewidth]{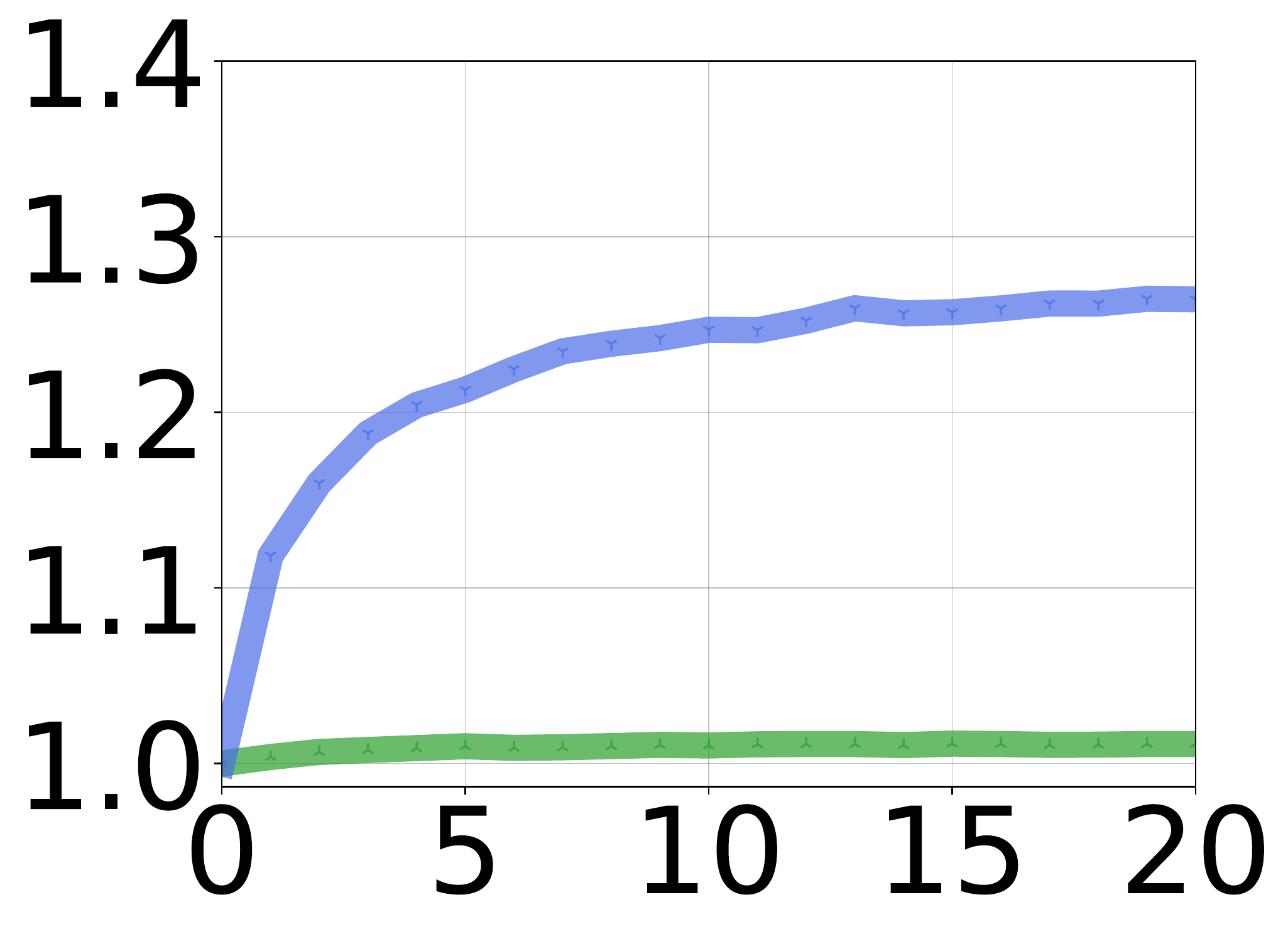}
	\end{subfigure}
 	\begin{subfigure}{0.19\linewidth}
		\centering
        BatchOrder\\
    	\includegraphics[width=1.0\linewidth]{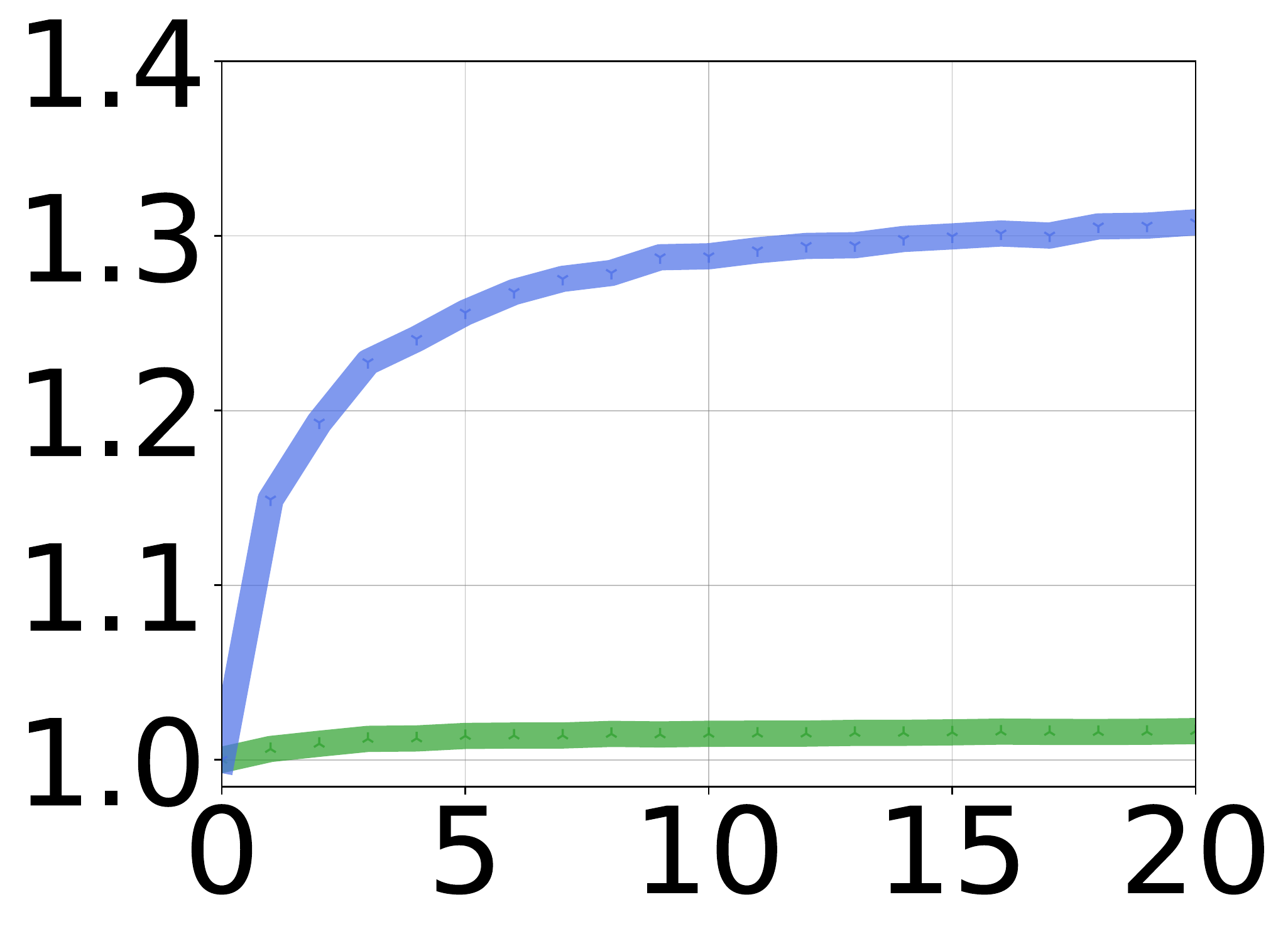}
	\end{subfigure}
	\begin{subfigure}{0.19\linewidth}
		\centering
        DA\\
    	\includegraphics[width=1.0\linewidth]{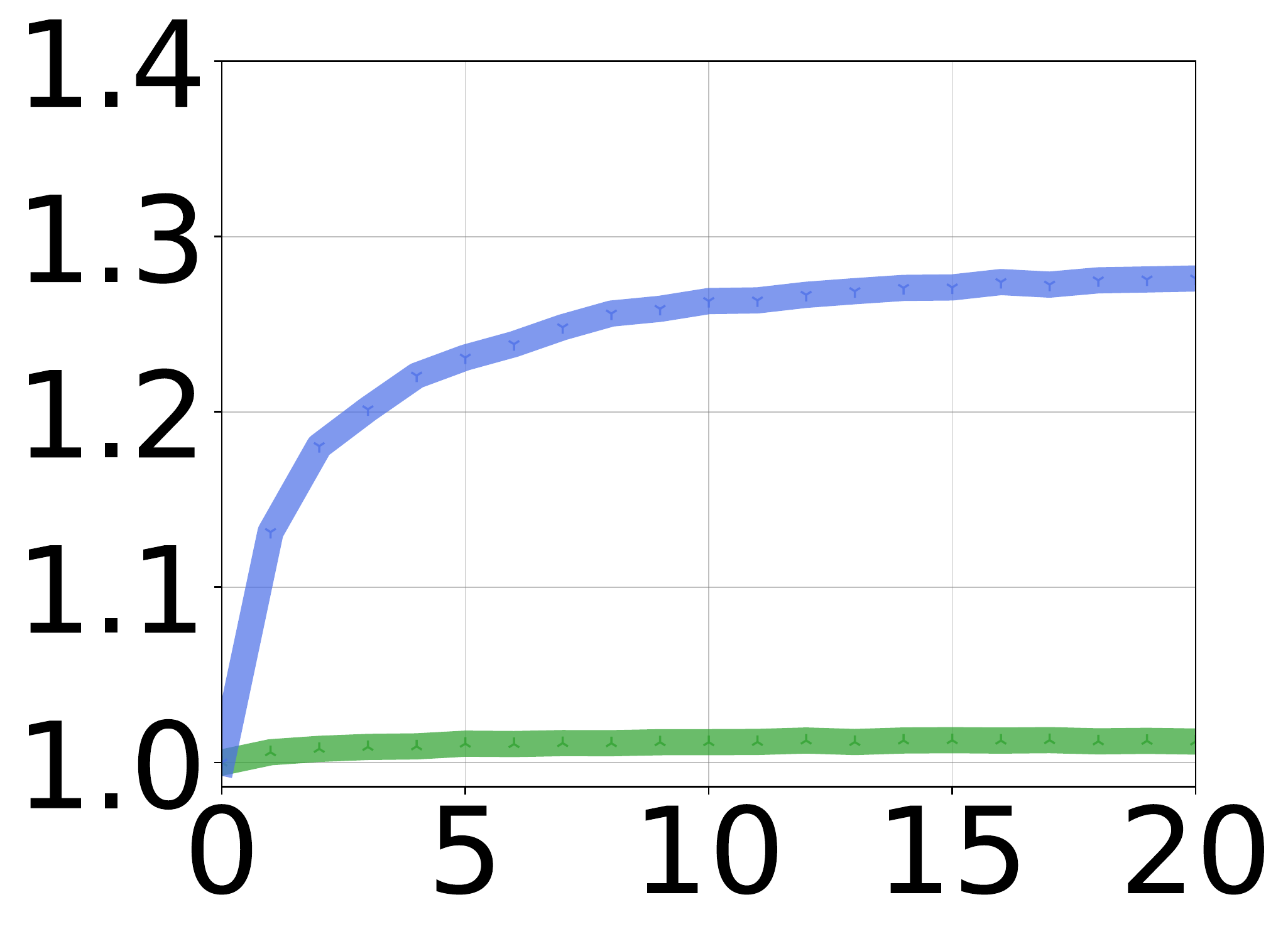}
	\end{subfigure}
 	\begin{subfigure}{0.19\linewidth}
		\centering
        Init \& BatchOrder
    	\includegraphics[width=1.0\linewidth]{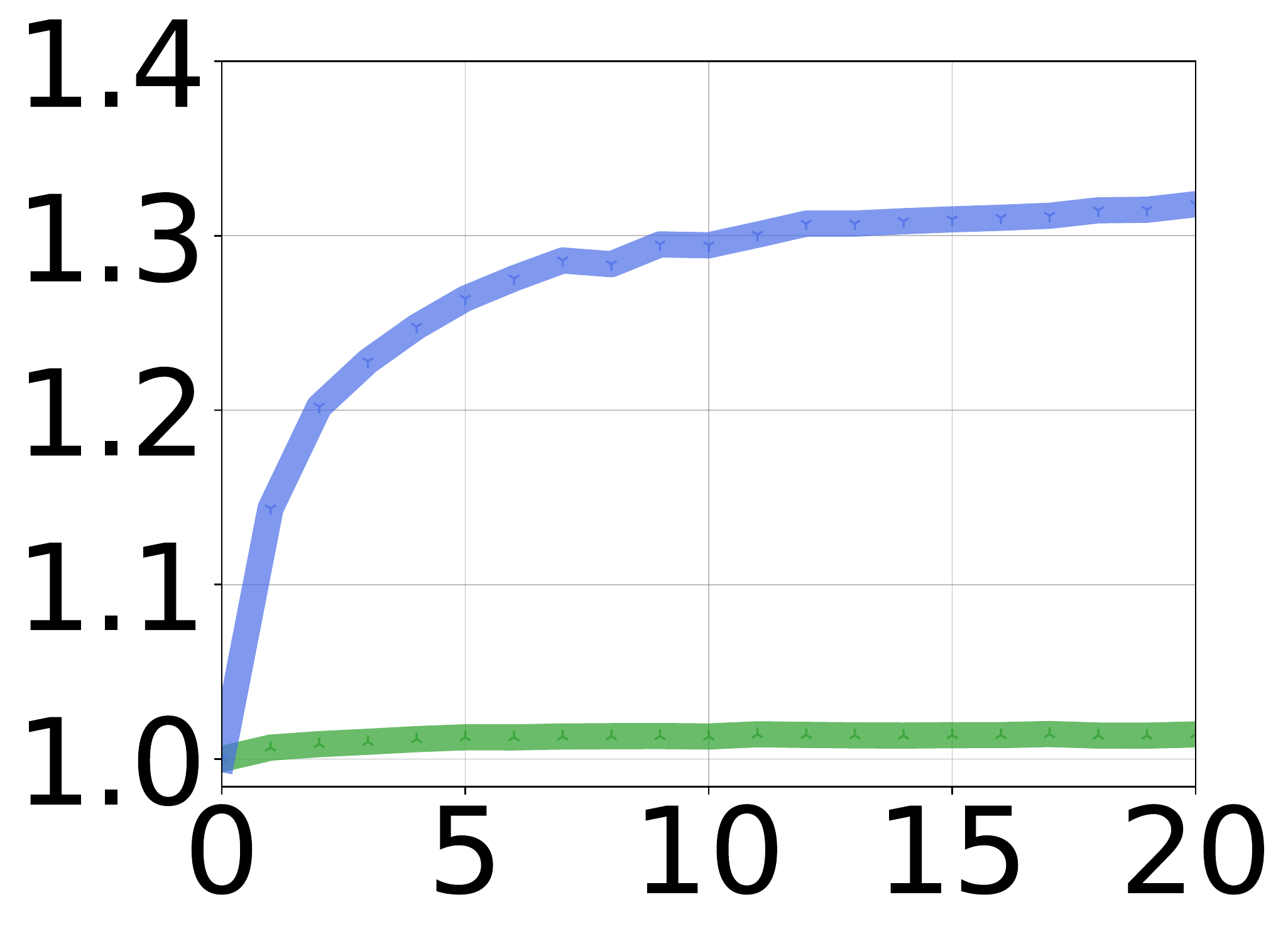}
	\end{subfigure}
    \begin{subfigure}{0.19\linewidth}
		\centering
        All Sources
    	\includegraphics[width=1.0\linewidth]{figures/CIFAR100/ratio_plots/vgg16_20/vgg16_RANDOM_ratio.pdf}
	\end{subfigure}
    \end{minipage}

    \centering
    \underline{VGG16 50 model ensemble}\\
    \vspace{0.1cm}
    \begin{minipage}{0.01\linewidth}
        \rotatebox{90}{\hspace{0.2cm} ensemble/base}
    \end{minipage}
    \begin{minipage}{0.98\linewidth}
	\begin{subfigure}{0.19\linewidth}
		\centering
        Init\\
    	\includegraphics[width=1.0\linewidth]{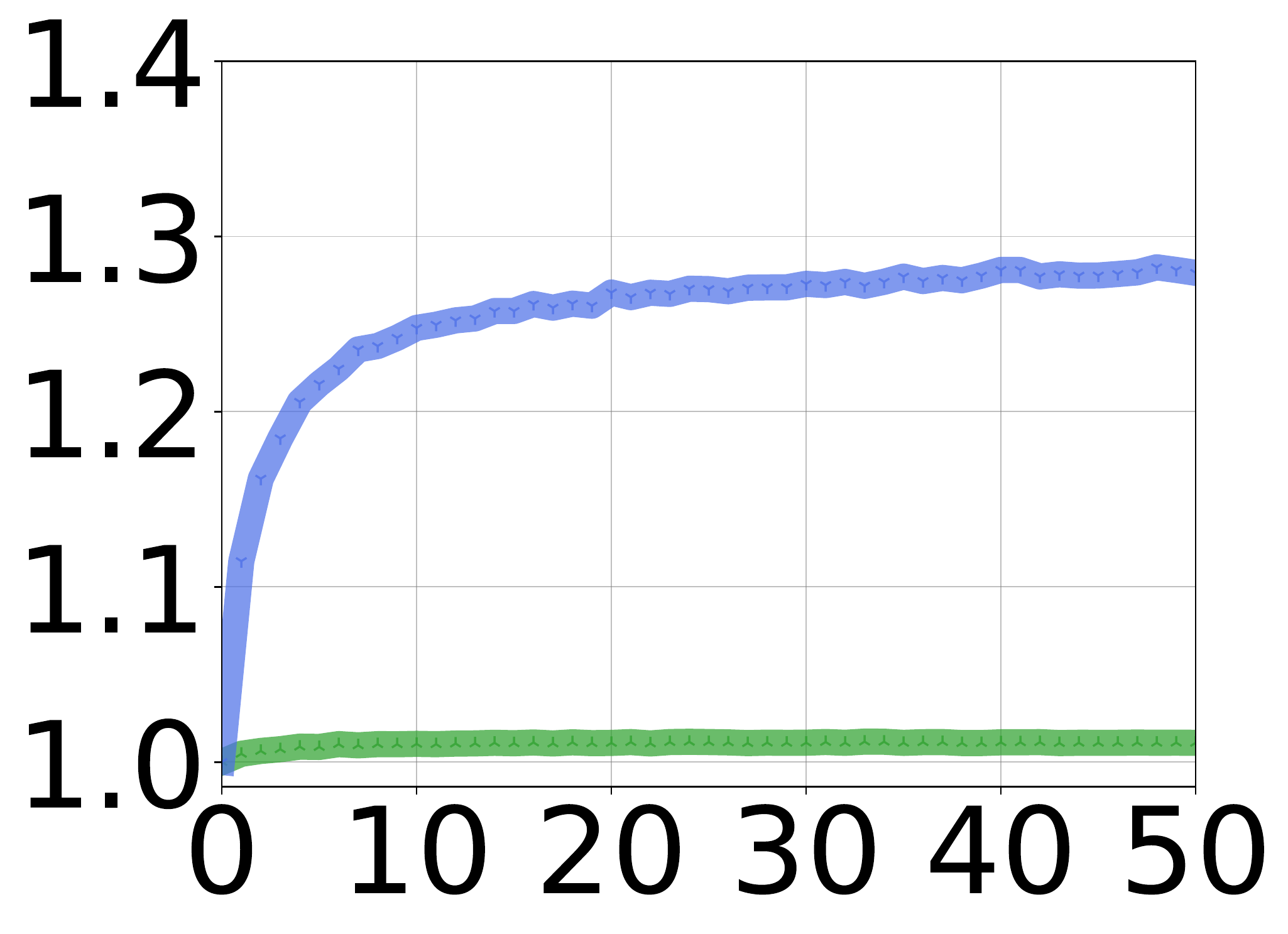}
	\end{subfigure}
 	\begin{subfigure}{0.19\linewidth}
		\centering
        BatchOrder\\
    	\includegraphics[width=1.0\linewidth]{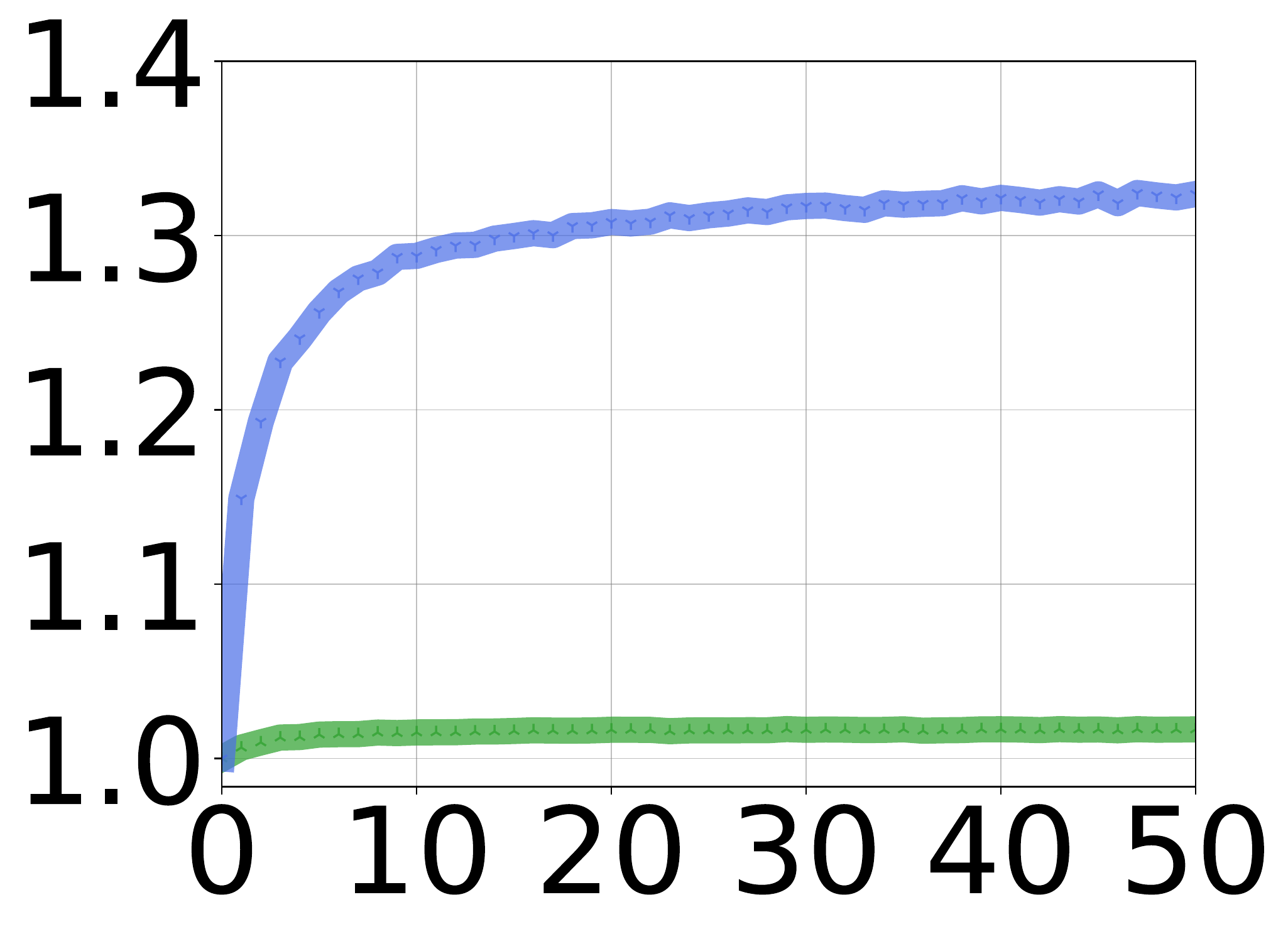}
	\end{subfigure}
	\begin{subfigure}{0.19\linewidth}
		\centering
        DA\\
    	\includegraphics[width=1.0\linewidth]{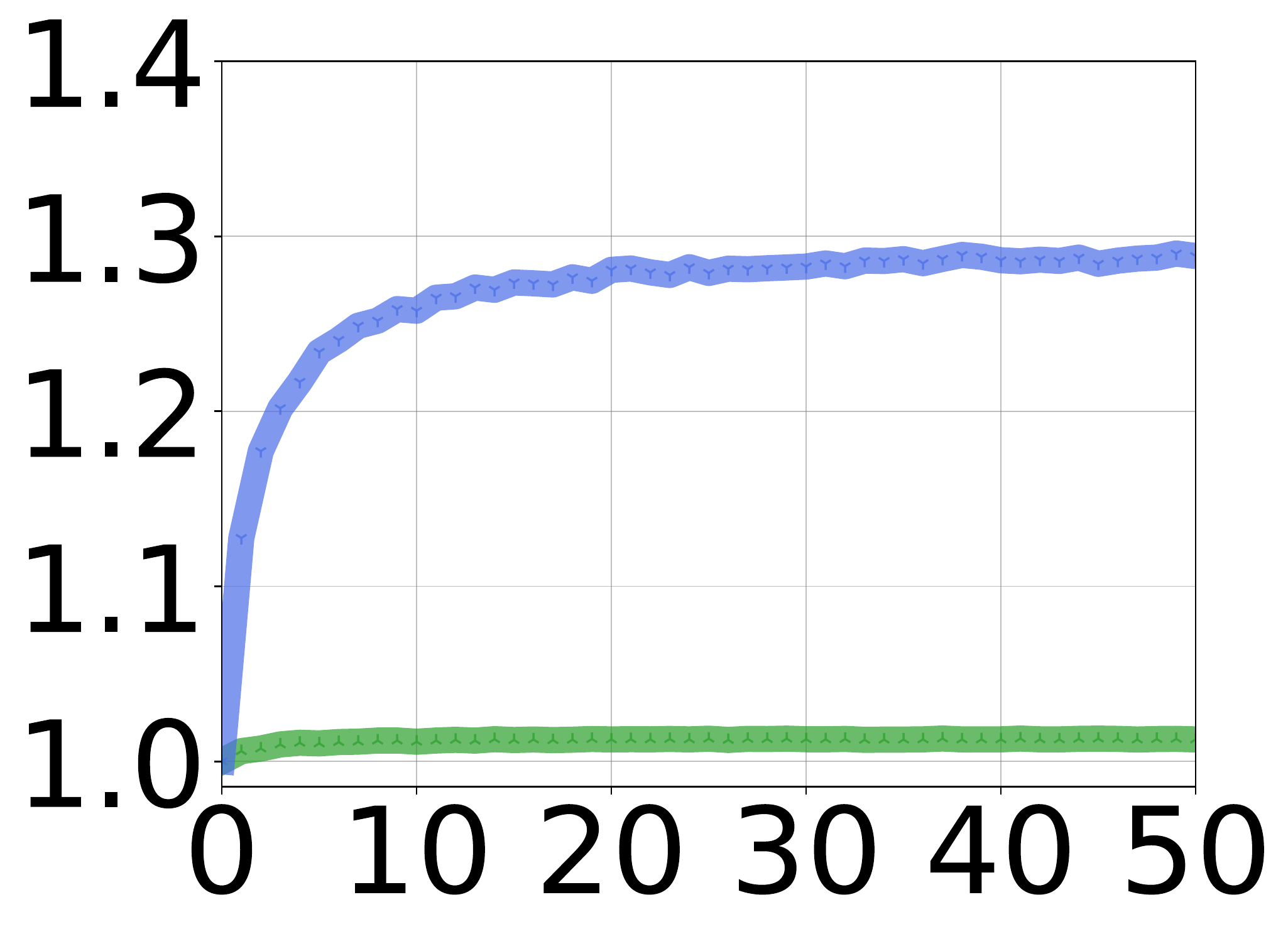}
	\end{subfigure}
 	\begin{subfigure}{0.19\linewidth}
		\centering
        Init \& BatchOrder\\
    	\includegraphics[width=1.0\linewidth]{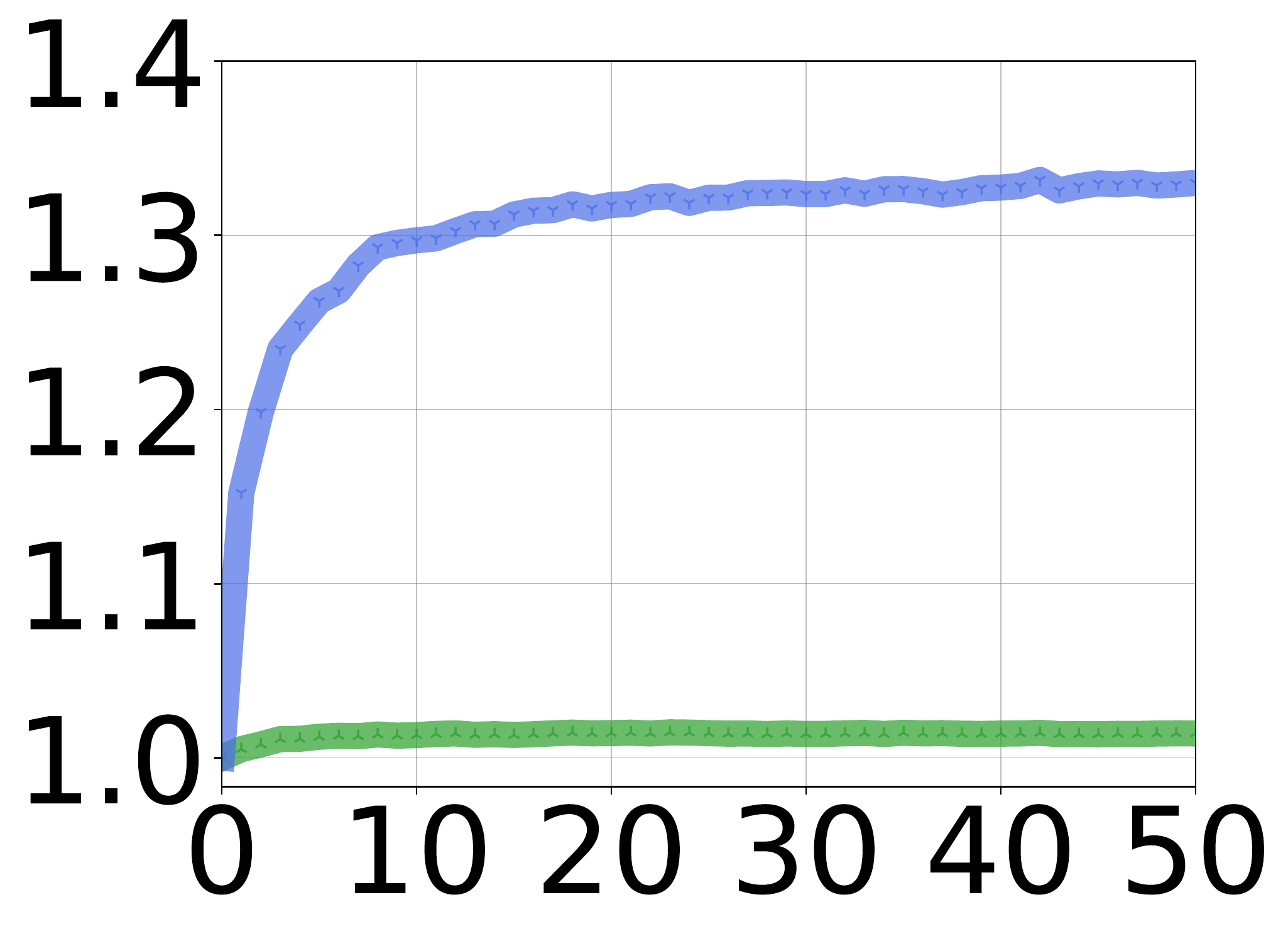}
	\end{subfigure}
    	\begin{subfigure}{0.19\linewidth}
		\centering
        All Sources\\
    	\includegraphics[width=1.0\linewidth]{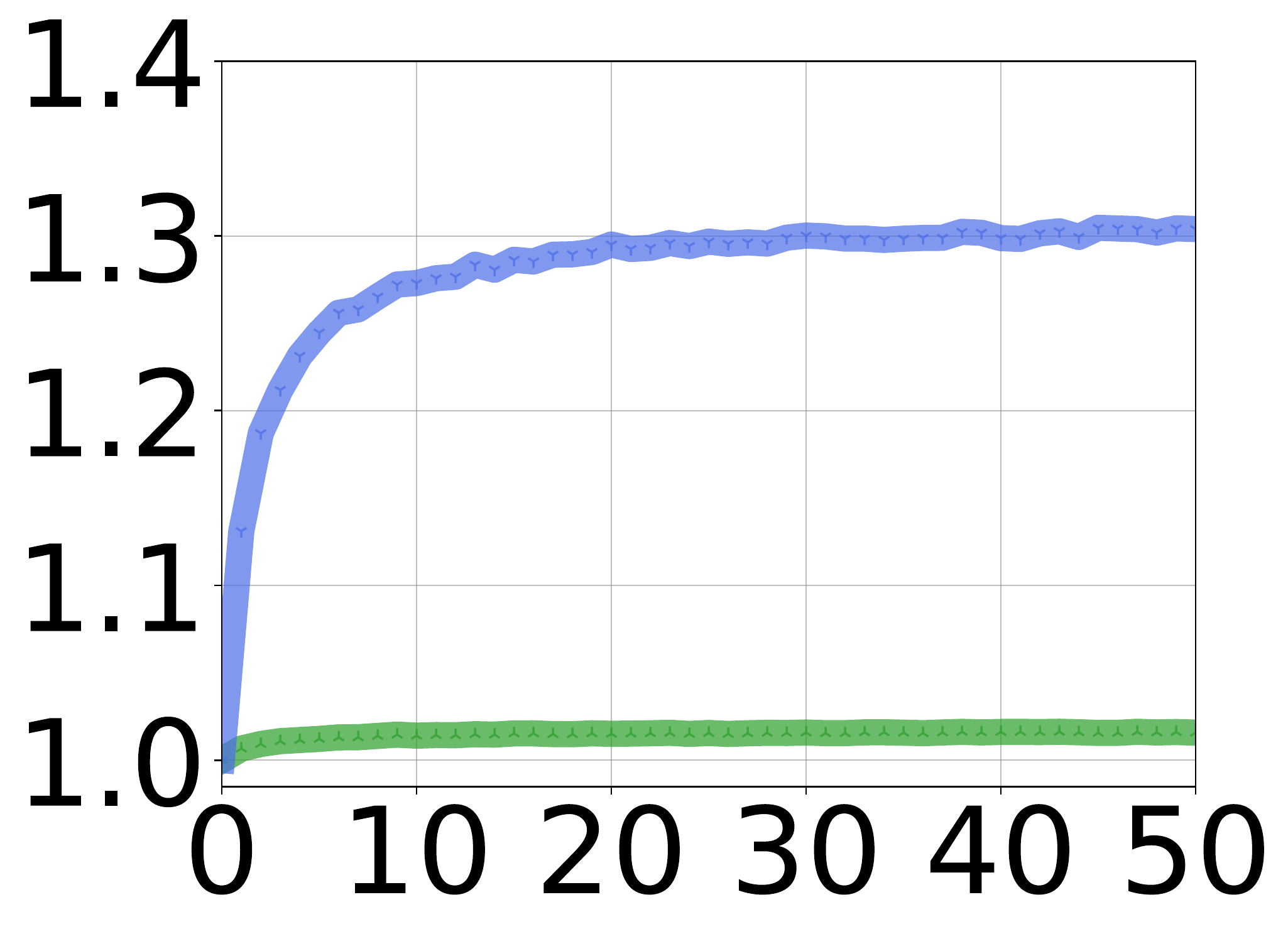}
	\end{subfigure}
    \end{minipage}

    \centering
    \underline{MLP-Mixer 20 model ensemble}\\
    \vspace{0.1cm}
    \begin{minipage}{0.01\linewidth}
        \rotatebox{90}{ensemble/base}
    \end{minipage}
    \begin{minipage}{0.98\linewidth}
	\begin{subfigure}{0.19\linewidth}
		\centering
        Init\\
    	\includegraphics[width=1.0\linewidth]{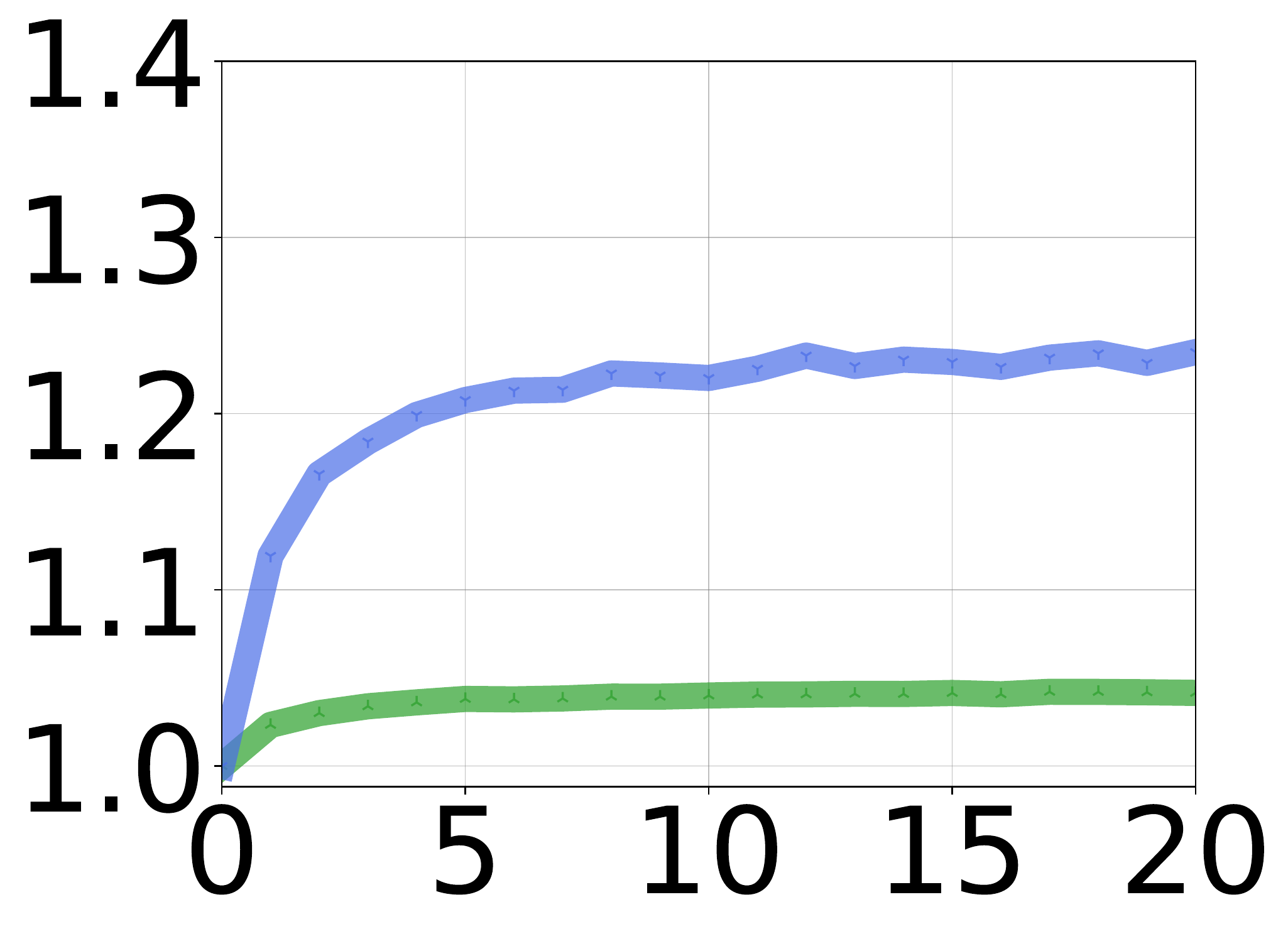}
        \\[-0.7em]
    	{\hspace{0.2cm} \small models in ensemble}
        \label{fig:c100_mlp_mixer_ratio_Change_ModelInit}
	\end{subfigure}
 	\begin{subfigure}{0.19\linewidth}
		\centering
        BatchOrder\\
    	\includegraphics[width=1.0\linewidth]{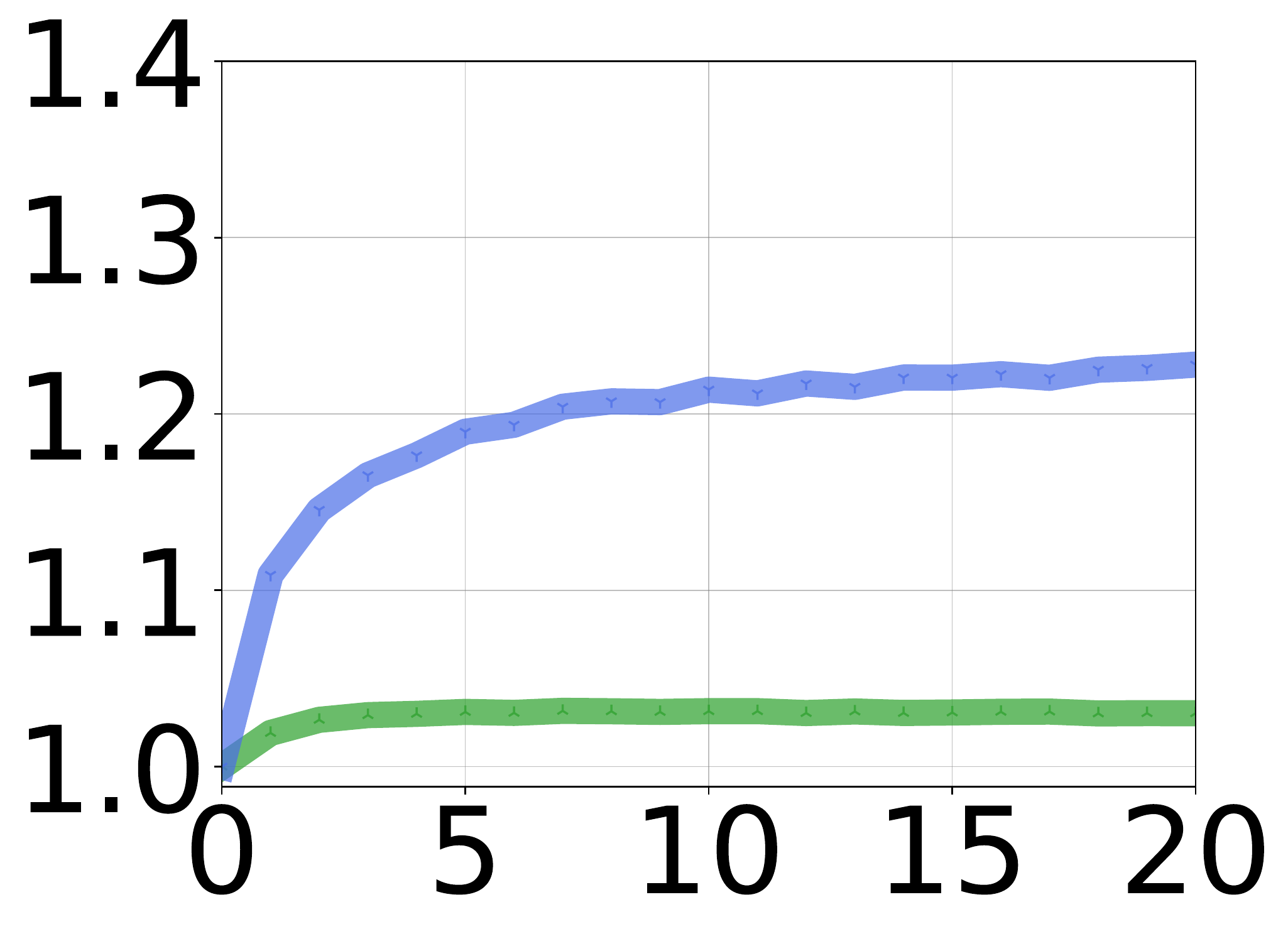}
        \\[-0.7em]
    	{\hspace{0.2cm} \small models in ensemble}
        \label{fig:c100_mlp_mixer_ratio_Change_BatchOrder}
	\end{subfigure}
	\begin{subfigure}{0.19\linewidth}
		\centering
        DA\\
    	\includegraphics[width=1.0\linewidth]{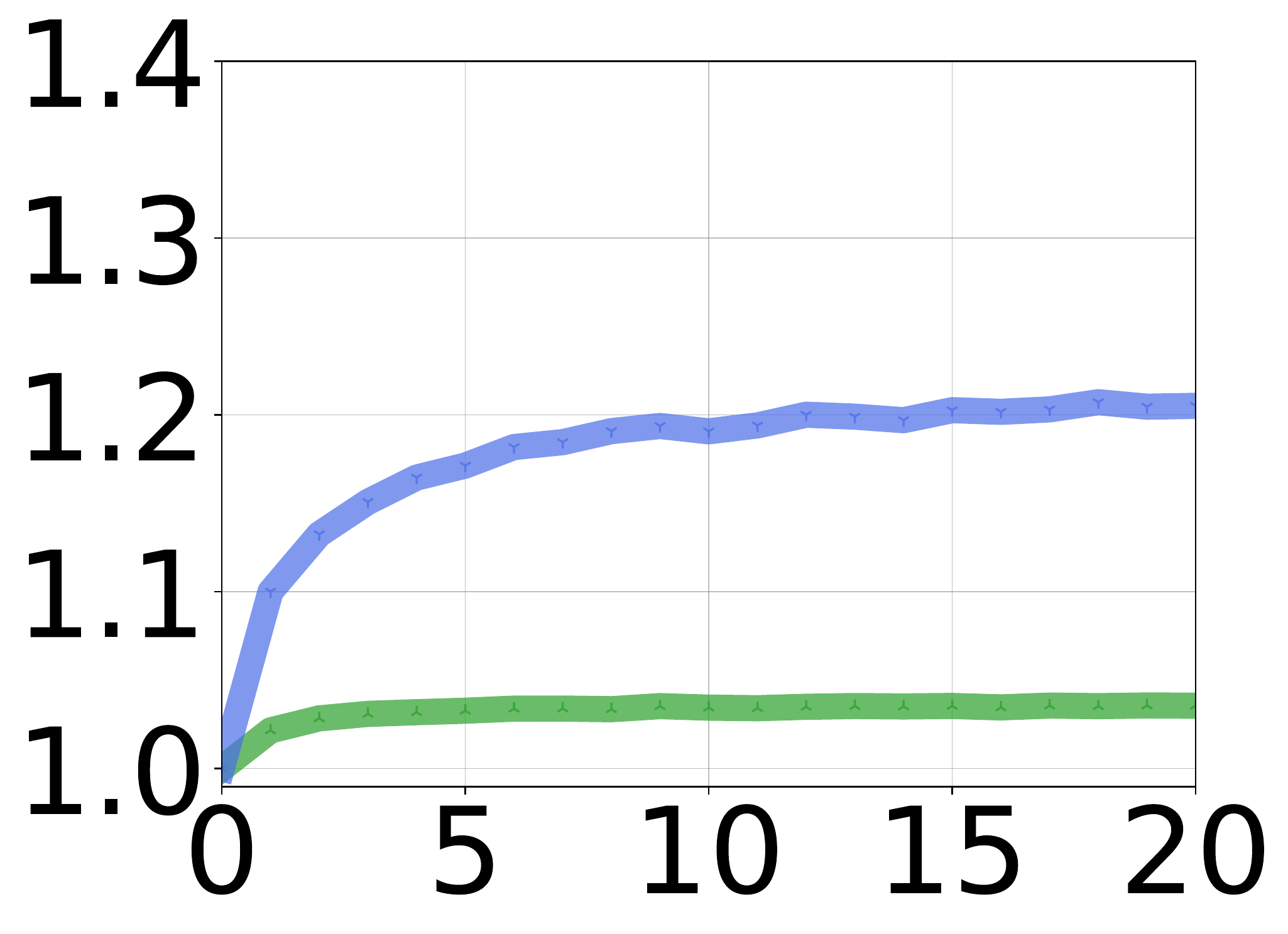}
        \\[-0.7em]
    	{\hspace{0.2cm} \small models in ensemble}
        \label{fig:c100_mlp_mixer_ratio_Change_DA}
	\end{subfigure}
 	\begin{subfigure}{0.19\linewidth}
		\centering
        Init \& BatchOrder\\
    	\includegraphics[width=1.0\linewidth]{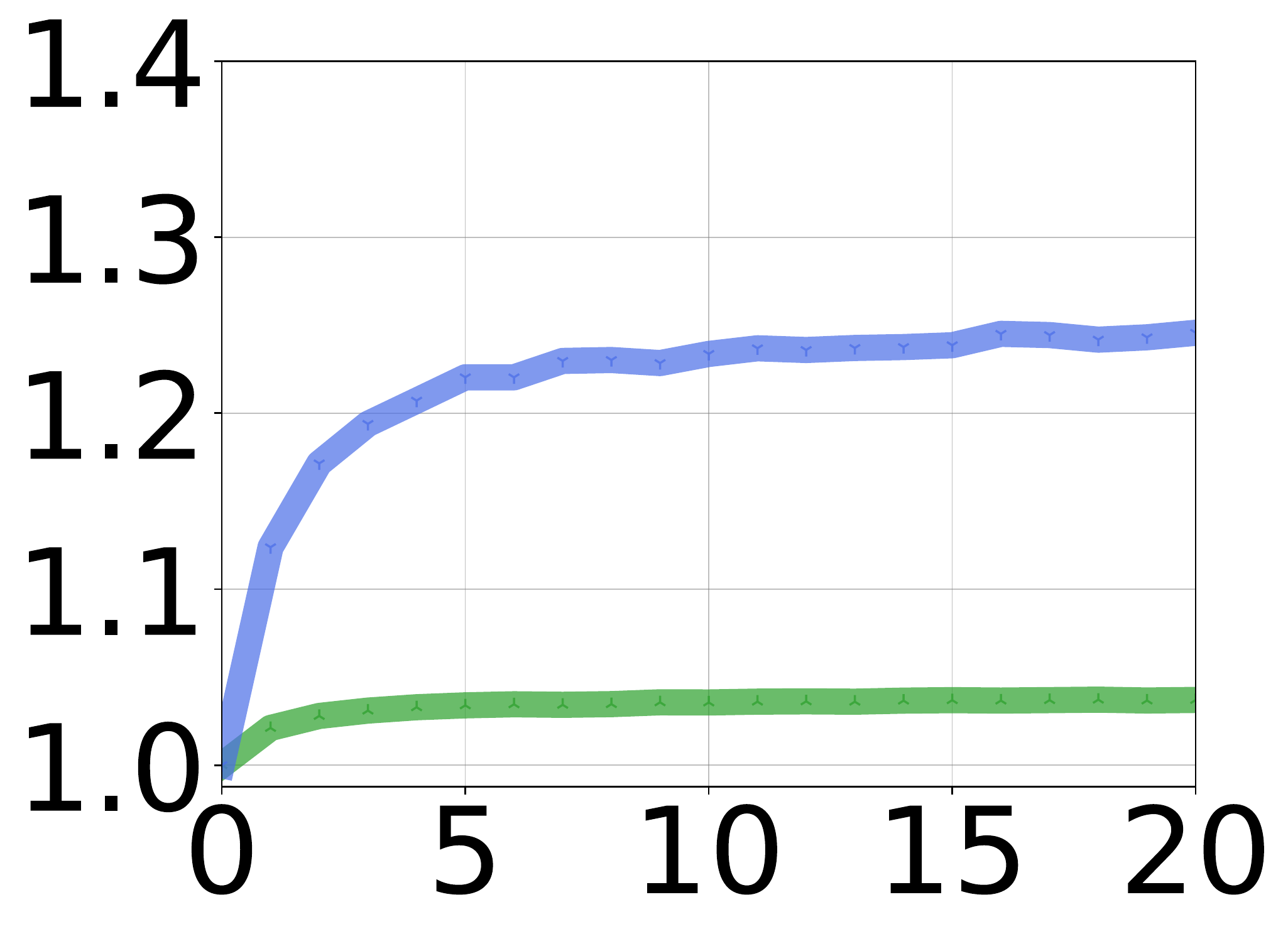}
        \\[-0.7em]
    	{\hspace{0.2cm} \small models in ensemble}
        \label{fig:c100_mlp_mixer_ratio_Change_ModelInit_BatchOrder}
	\end{subfigure}
  	\begin{subfigure}{0.19\linewidth}
		\centering
        All Sources\\
    	\includegraphics[width=1.0\linewidth]{figures/CIFAR100/ratio_plots/mlp_mixer_20/mlp_mixer_RANDOM_ratio.pdf}
        \\[-0.7em]
    	{\hspace{0.2cm} \small models in ensemble}
        \label{fig:c100_mlp_mixer_ratio_Random}
	\end{subfigure}
    \end{minipage}

	\caption{ Ratio of Top \& Bottom K ensemble accuracy for different model architectures and ensemble sizes on CIFAR100
	}
    \label{fig:CIFAR100_ratio}
\end{figure*}

\begin{figure*}
    \subsection{TinyImageNet}
    \centering
    \underline{ResNet50 20 model ensemble}\\
    \vspace{0.1cm}
    
    \begin{minipage}{0.01\linewidth}
        \rotatebox{90}{\hspace{0.2cm} ensemble/base}
    \end{minipage}
    \begin{minipage}{0.98\linewidth}
	\begin{subfigure}{0.19\linewidth}
		\centering
        Init\\
    	\includegraphics[width=1.0\linewidth]{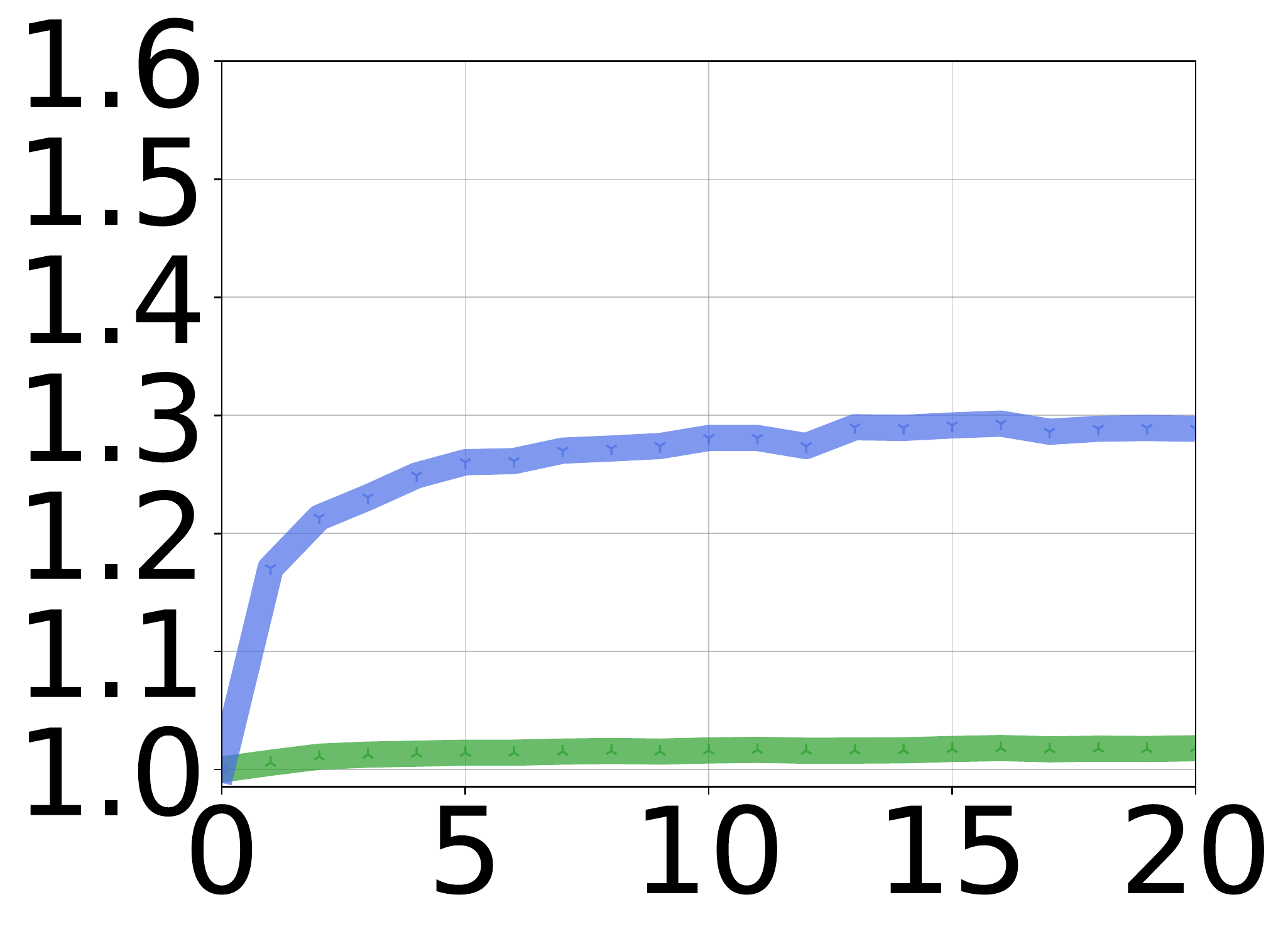}
        \label{fig:Tinyimagenet_resnet50_ratio_Change_ModelInit}
	\end{subfigure}
 	\begin{subfigure}{0.19\linewidth}
		\centering
        BatchOrder\\
    	\includegraphics[width=1.0\linewidth]{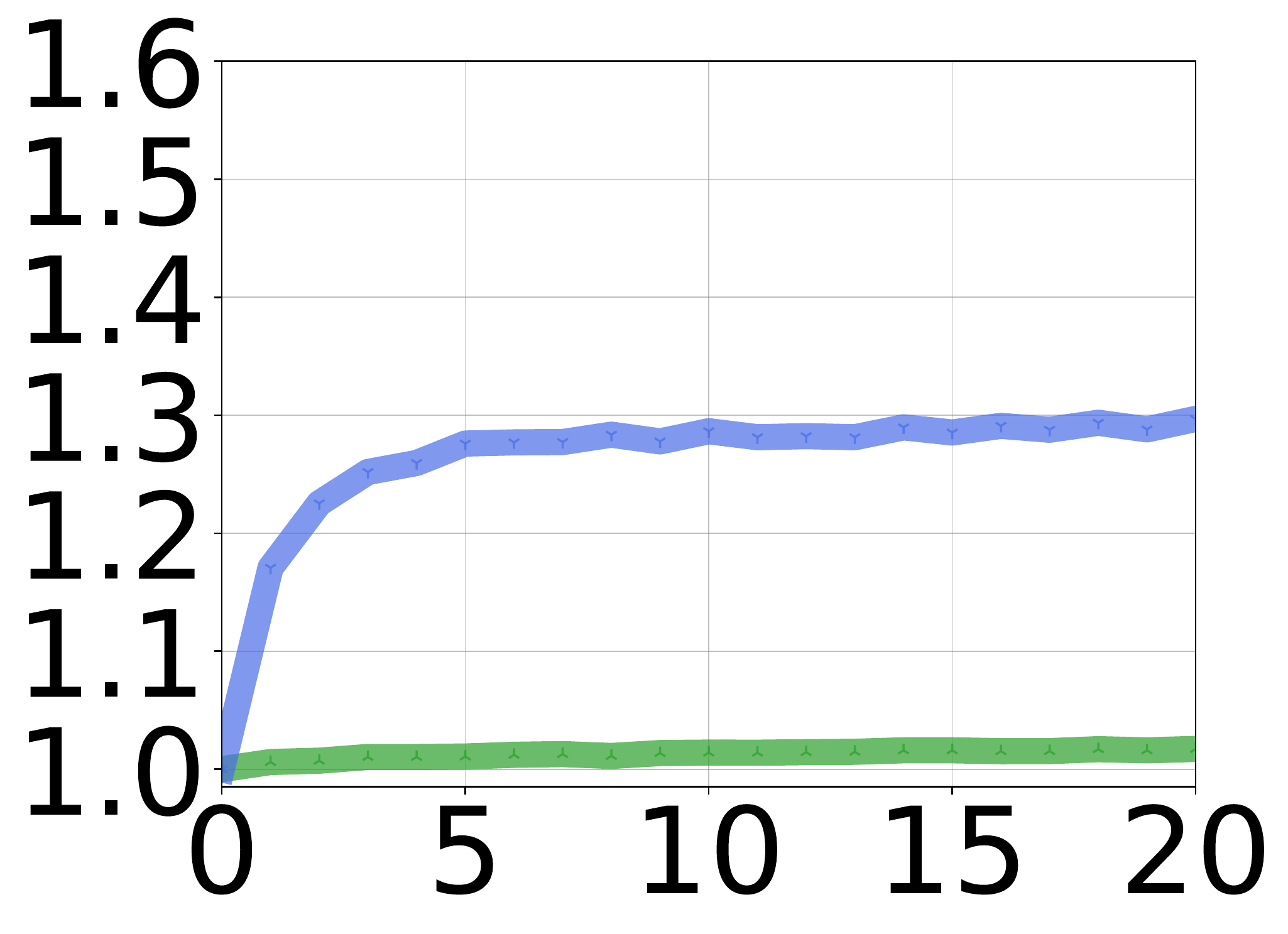}
        \label{fig:Tinyimagenet_resnet50_ratio_Change_BatchOrder}
	\end{subfigure}
	\begin{subfigure}{0.19\linewidth}
		\centering
        DA\\
    	\includegraphics[width=1.0\linewidth]{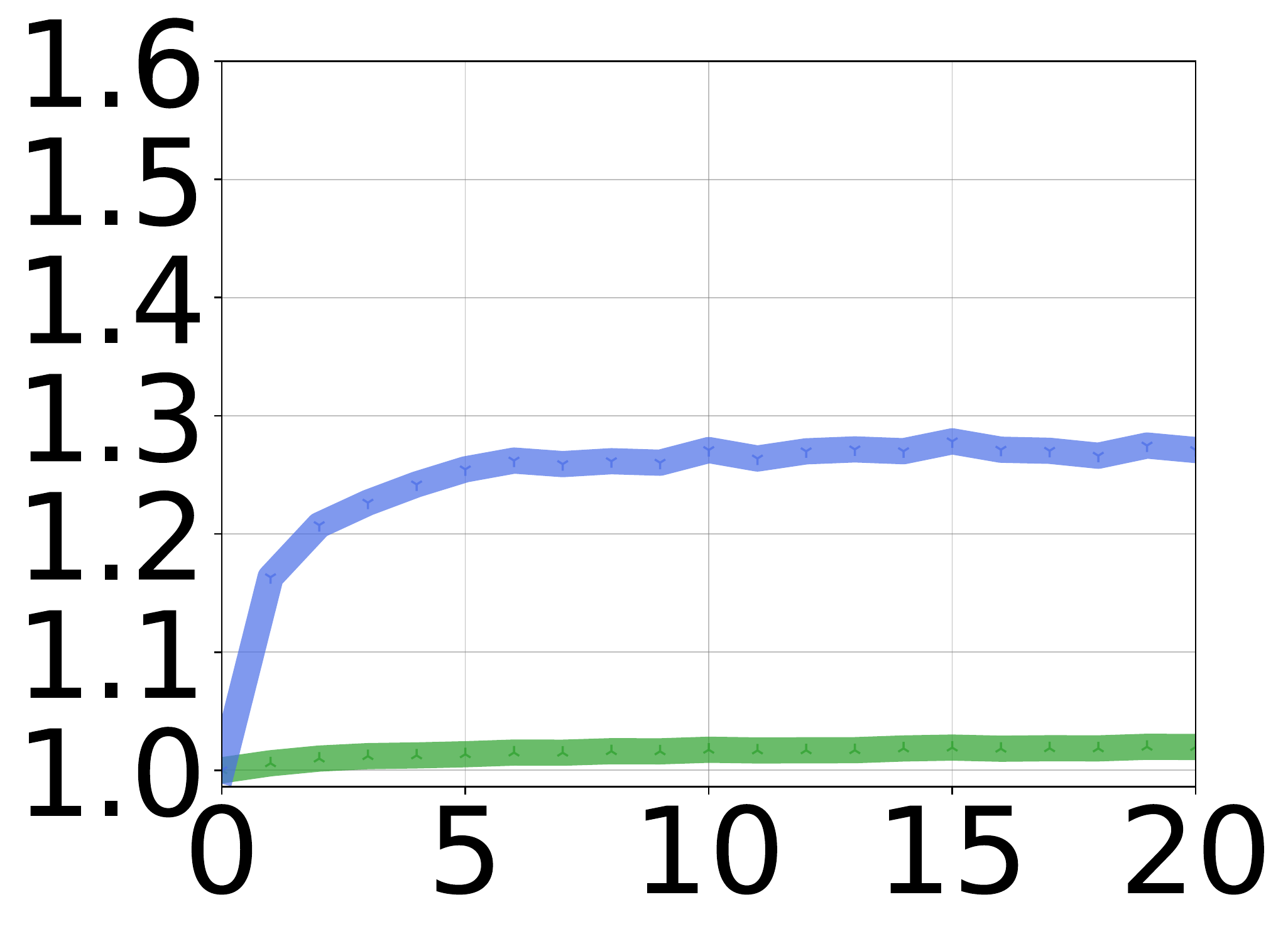}
        \label{fig:Tinyimagenet_resnet50_ratio_Change_DA}
	\end{subfigure}
 	\begin{subfigure}{0.19\linewidth}
		\centering
        Init \& BatchOrder\\
    	\includegraphics[width=1.0\linewidth]{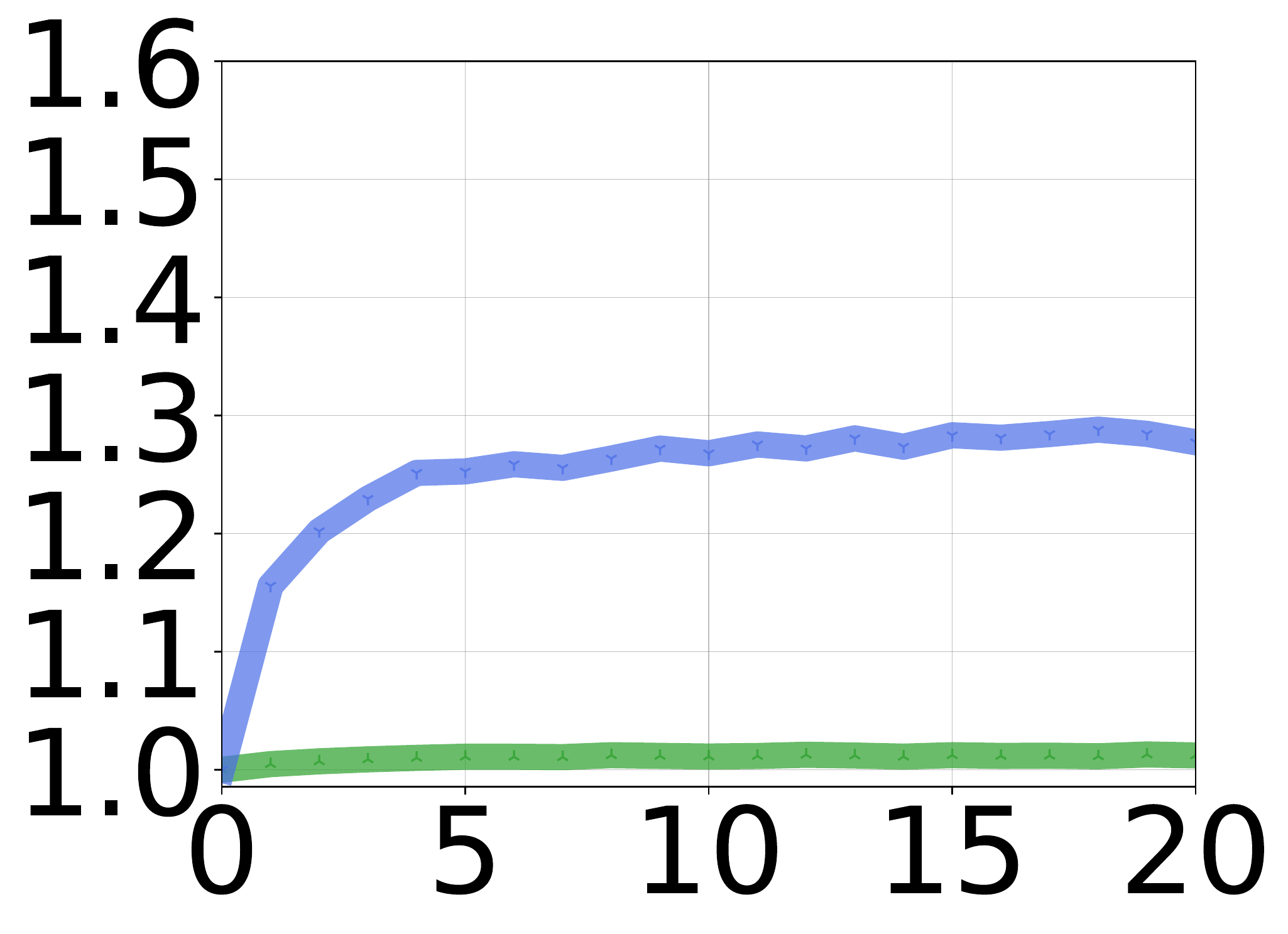}
        \label{fig:Tinyimagenet_resnet50_ratio_Change_ModelInit_BatchOrder}
	\end{subfigure}
    	\begin{subfigure}{0.19\linewidth}
		\centering
        All Sources\\
    	\includegraphics[width=1.0\linewidth]{figures/TINYIMAGENET/ratio_plots/resnet50_20/resnet50_RANDOM_ratio.pdf}
        \label{fig:Tinyimagenet_resnet50_ratio_Random}
	\end{subfigure}
    \end{minipage}
    
    \centering
    \underline{ResNet50 30 model ensemble}\\
    \vspace{0.1cm}

    \begin{minipage}{0.01\linewidth}
        \rotatebox{90}{\hspace{0.2cm} ensemble/base}
    \end{minipage}
    \begin{minipage}{0.98\linewidth}
	\begin{subfigure}{0.19\linewidth}
		\centering
        Init\\
    	\includegraphics[width=1.0\linewidth]{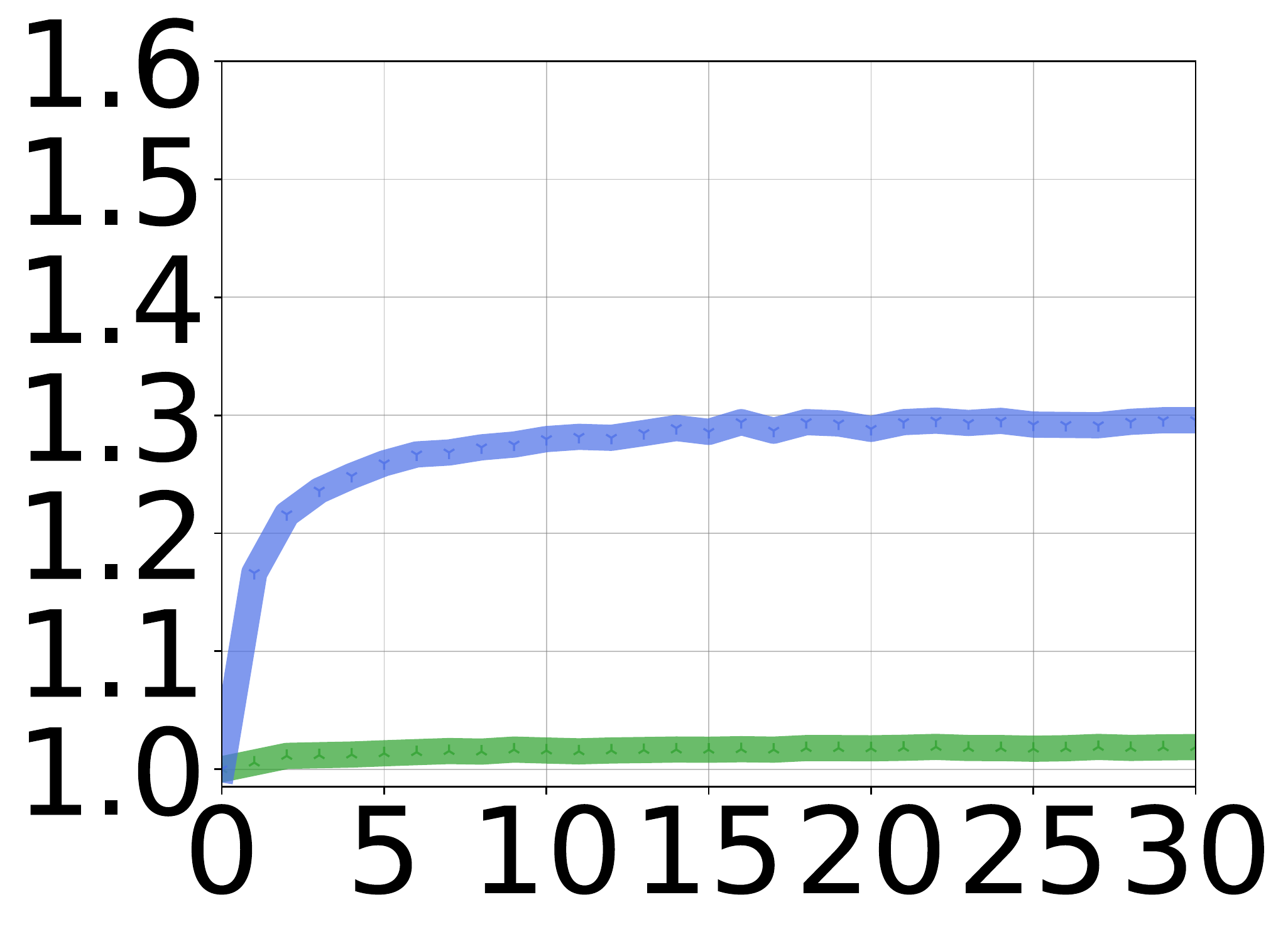}
        \label{fig:Tinyimagenet_resnet50_30_ratio_Change_ModelInit}
	\end{subfigure}
 	\begin{subfigure}{0.19\linewidth}
		\centering
        BatchOrder\\
    	\includegraphics[width=1.0\linewidth]{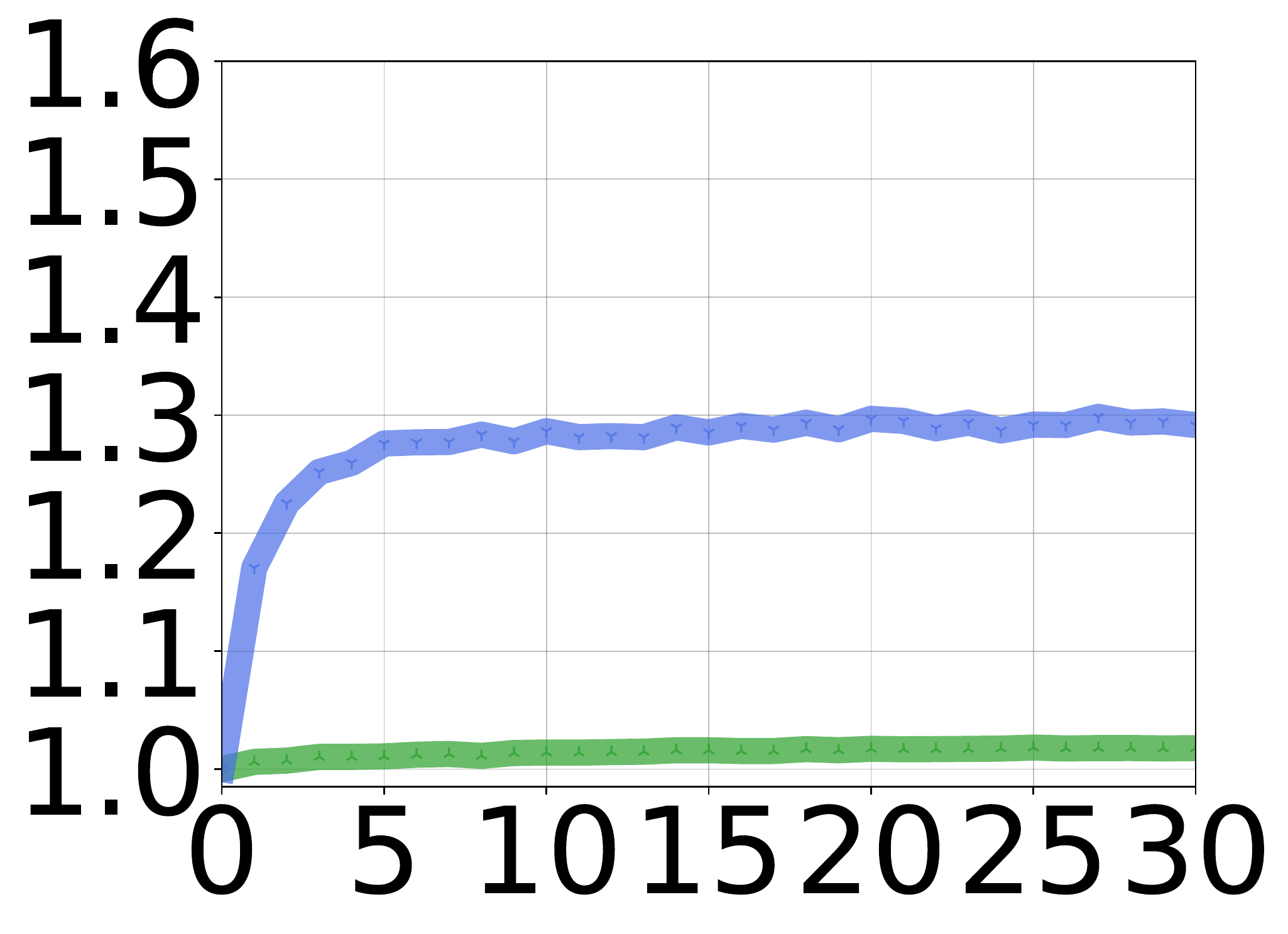}
        \label{fig:Tinyimagenet_resnet50_30_ratio_Change_BatchOrder}
	\end{subfigure}
	\begin{subfigure}{0.19\linewidth}
		\centering
        DA\\
    	\includegraphics[width=1.0\linewidth]{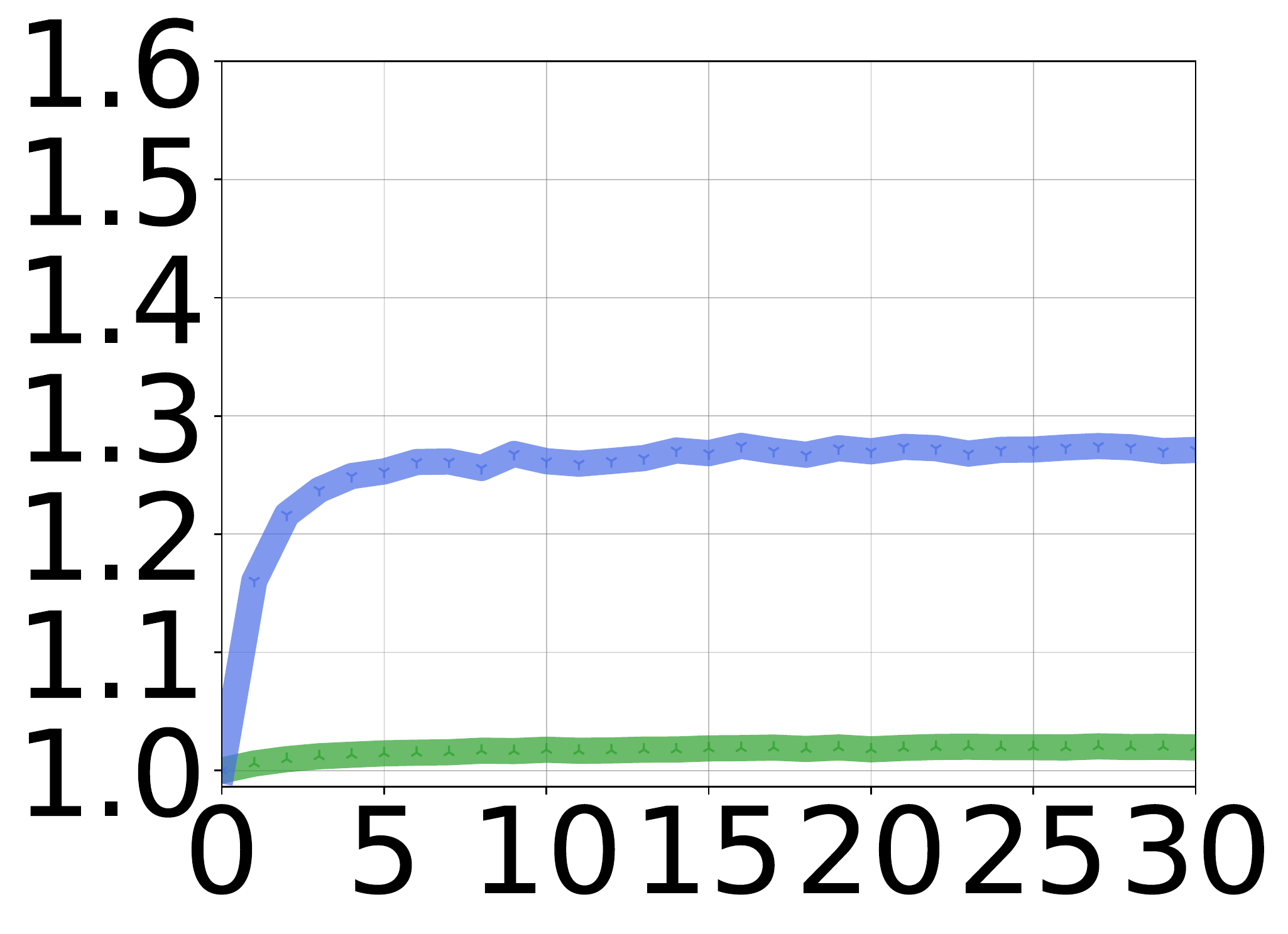}
        \label{fig:Tinyimagenet_resnet50_30_ratio_Change_DA}
	\end{subfigure}
 	\begin{subfigure}{0.19\linewidth}
		\centering
        Init \& BatchOrder\\
    	\includegraphics[width=1.0\linewidth]{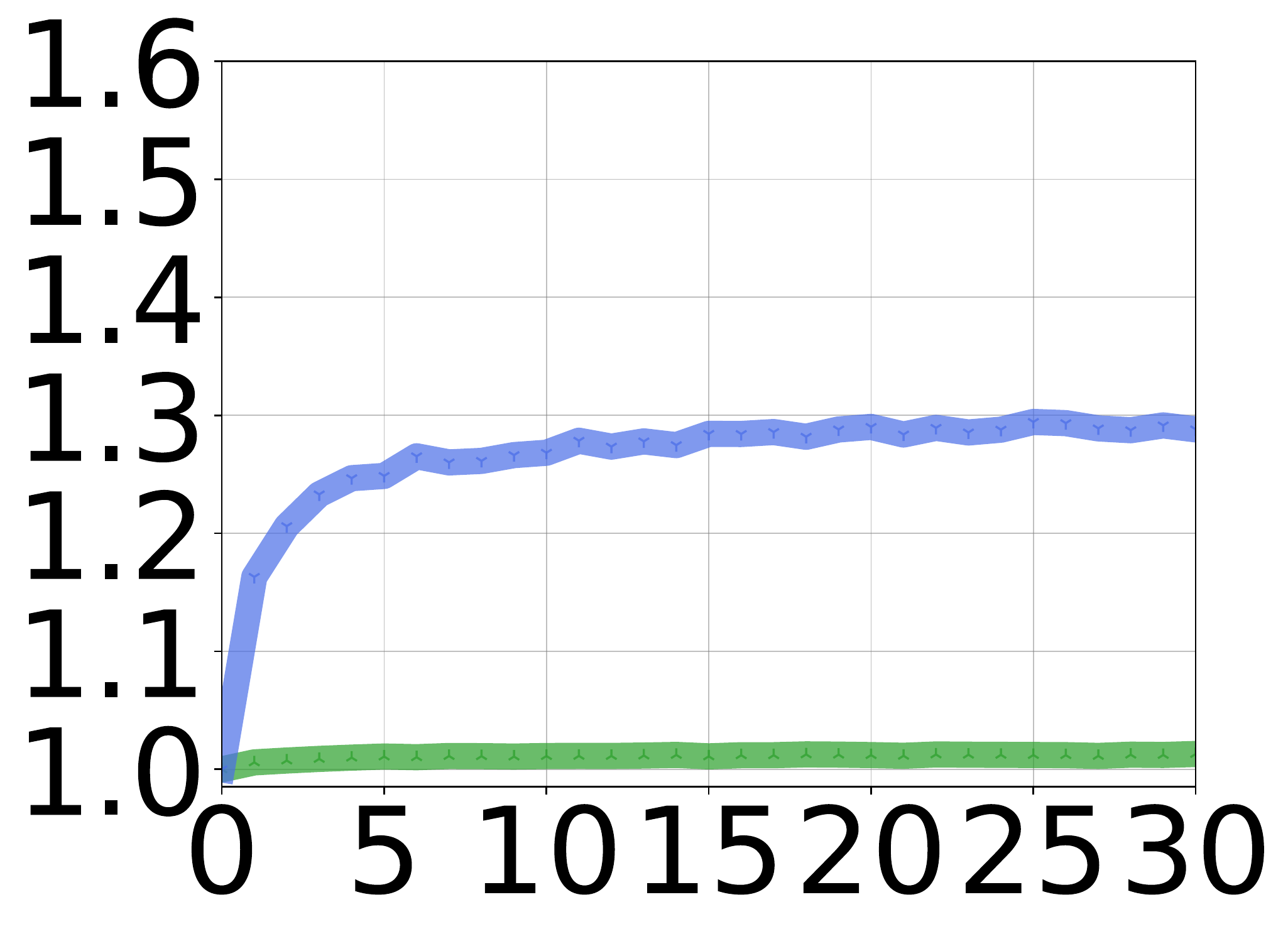}
        \label{fig:Tinyimagenet_resnet50_30_ratio_Change_ModelInit_BatchOrder}
	\end{subfigure}
    	\begin{subfigure}{0.19\linewidth}
		\centering
        All Sources\\
    	\includegraphics[width=1.0\linewidth]{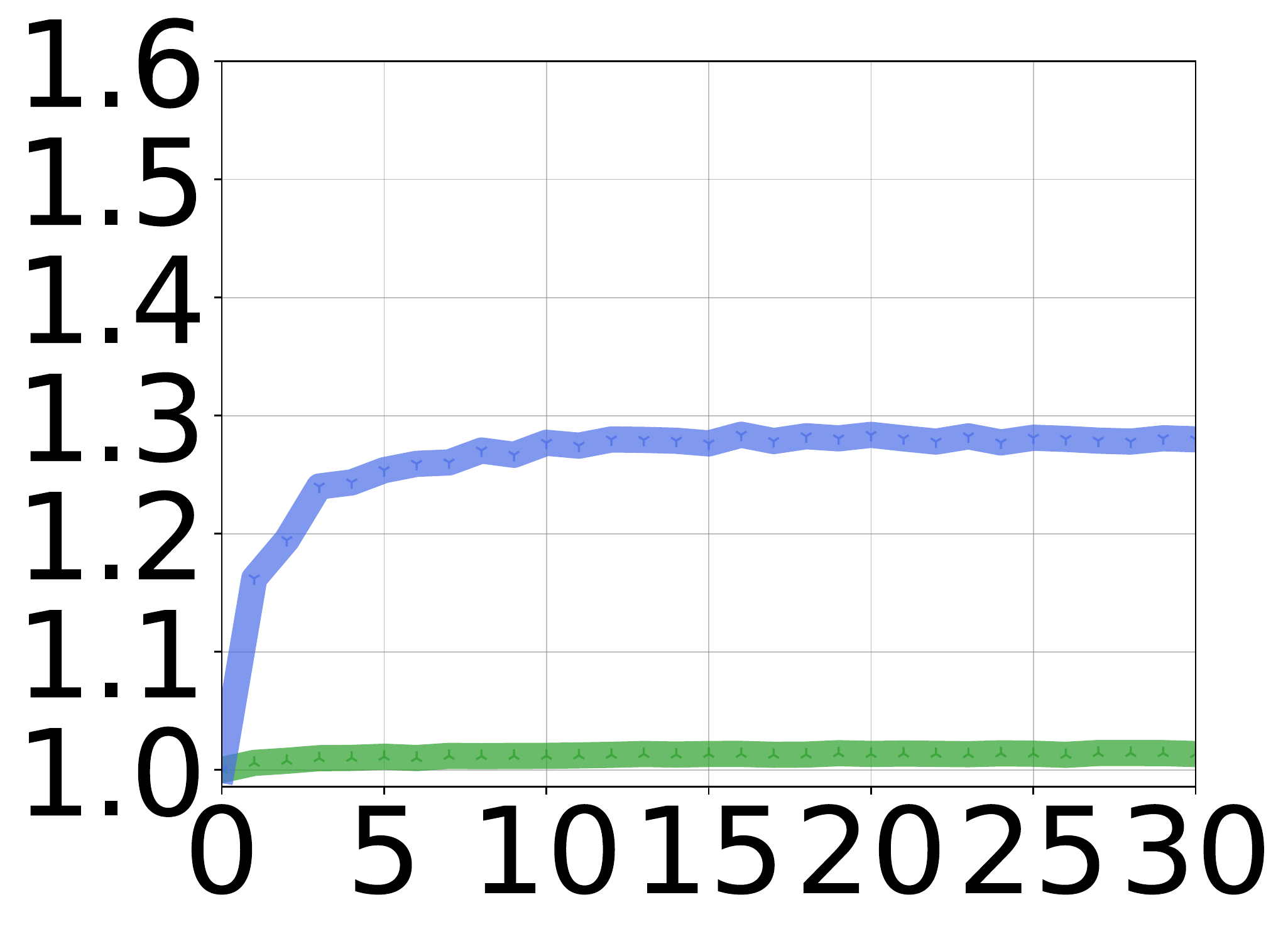}
        \label{fig:Tinyimagenet_resnet50_30_ratio_Random}
	\end{subfigure}
    \end{minipage}
    
    \centering
    \underline{ResNet50 50 model ensemble}\\
    \vspace{0.1cm}
    
    \begin{minipage}{0.01\linewidth}
        \rotatebox{90}{\hspace{0.2cm} ensemble/base}
    \end{minipage}
    \begin{minipage}{0.98\linewidth}
	\begin{subfigure}{0.19\linewidth}
		\centering
        Init\\
    	\includegraphics[width=1.0\linewidth]{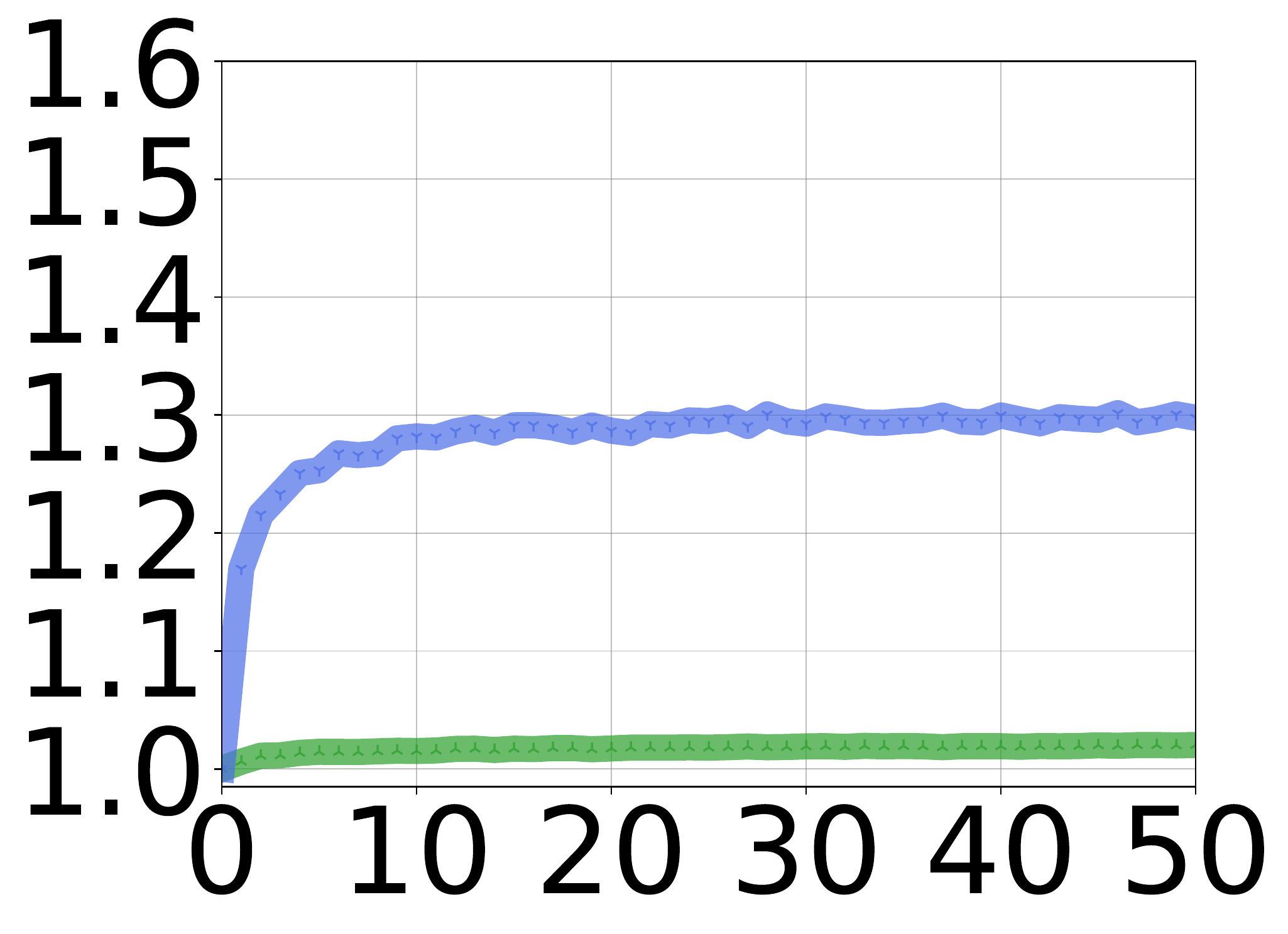}
        \label{fig:Tinyimagenet_resnet50_50_ratio_Change_ModelInit}
	\end{subfigure}
 	\begin{subfigure}{0.19\linewidth}
		\centering
        BatchOrder\\
    	\includegraphics[width=1.0\linewidth]{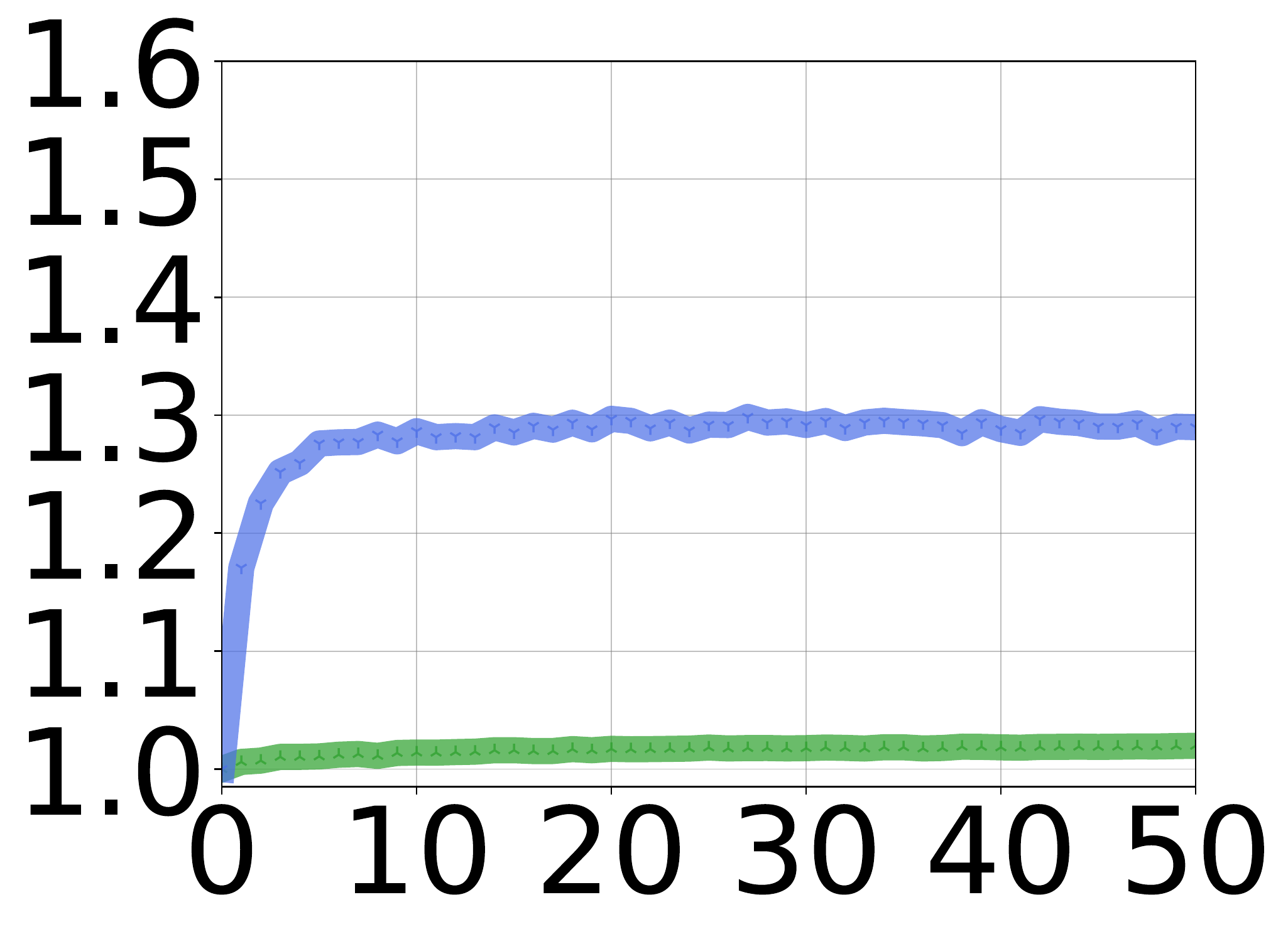}
        \label{fig:Tinyimagenet_resnet50_50_ratio_Change_BatchOrder}
	\end{subfigure}
	\begin{subfigure}{0.19\linewidth}
		\centering
        DA\\
    	\includegraphics[width=1.0\linewidth]{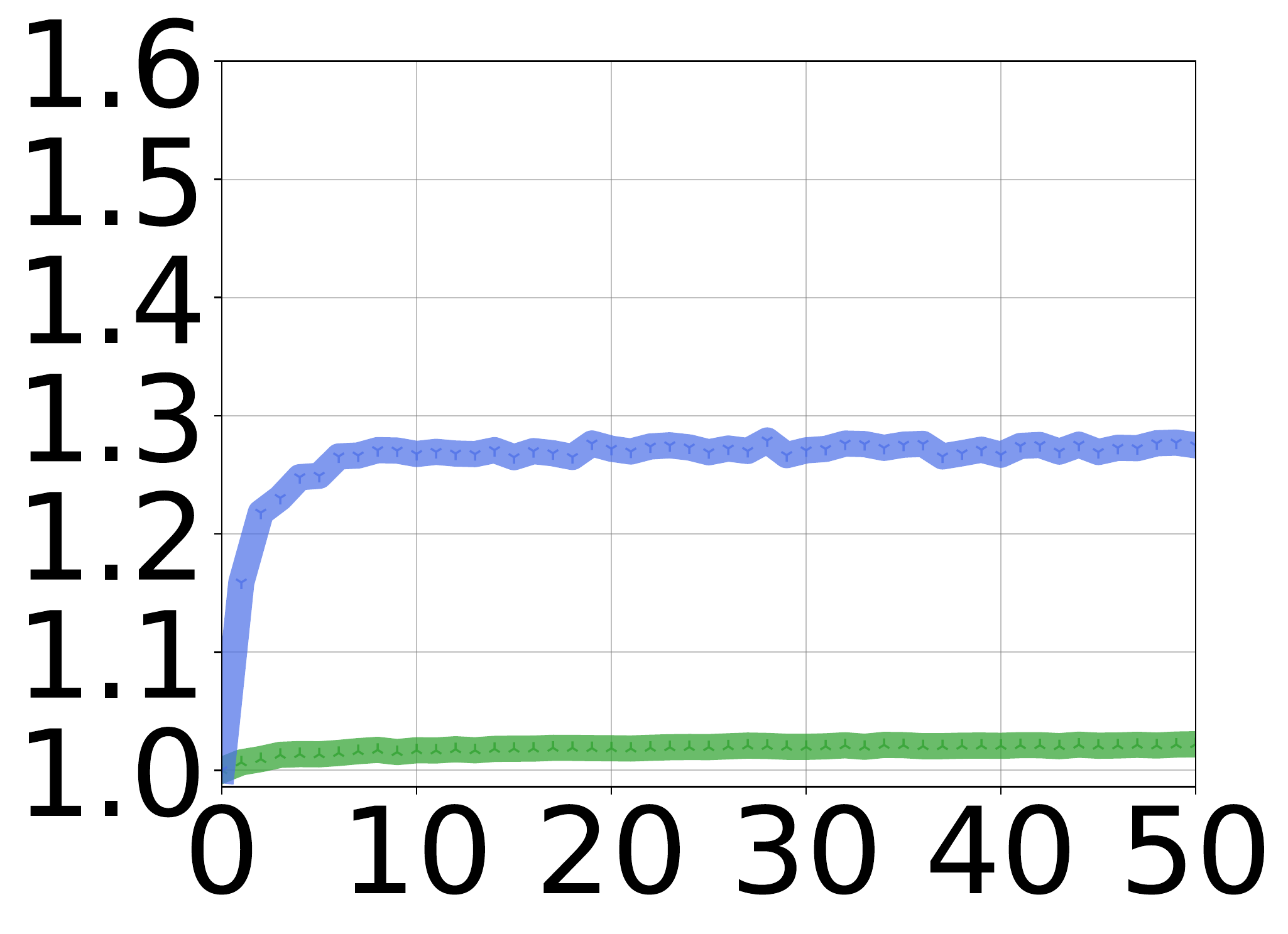}
        \label{fig:Tinyimagenet_resnet50_50_ratio_Change_DA}
	\end{subfigure}
 	\begin{subfigure}{0.19\linewidth}
		\centering
        Init \& BatchOrder\\
    	\includegraphics[width=1.0\linewidth]{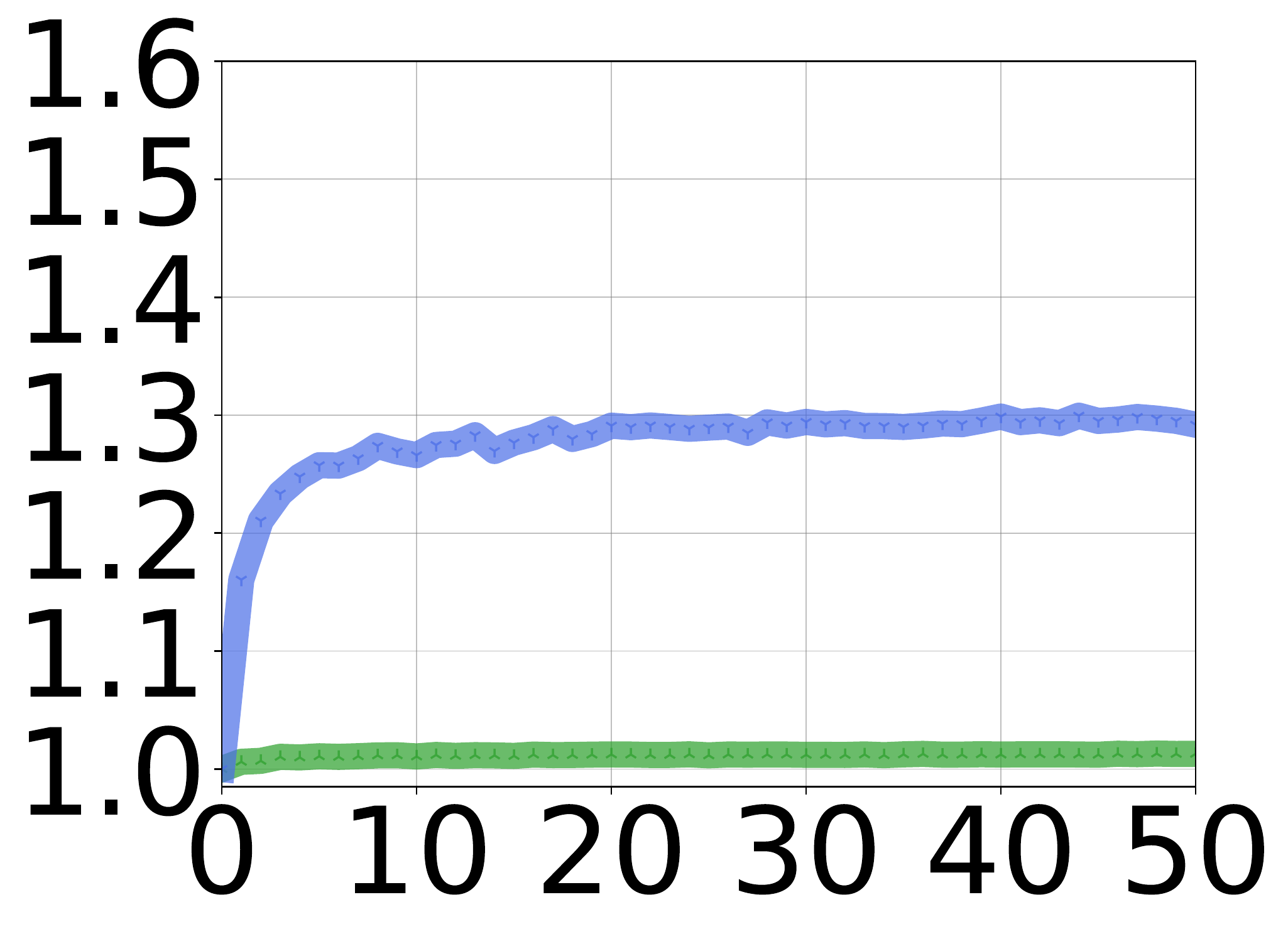}
        \label{fig:Tinyimagenet_resnet50_50_ratio_Change_ModelInit_BatchOrder}
	\end{subfigure}
    	\begin{subfigure}{0.19\linewidth}
		\centering
        All Sources\\
    	\includegraphics[width=1.0\linewidth]{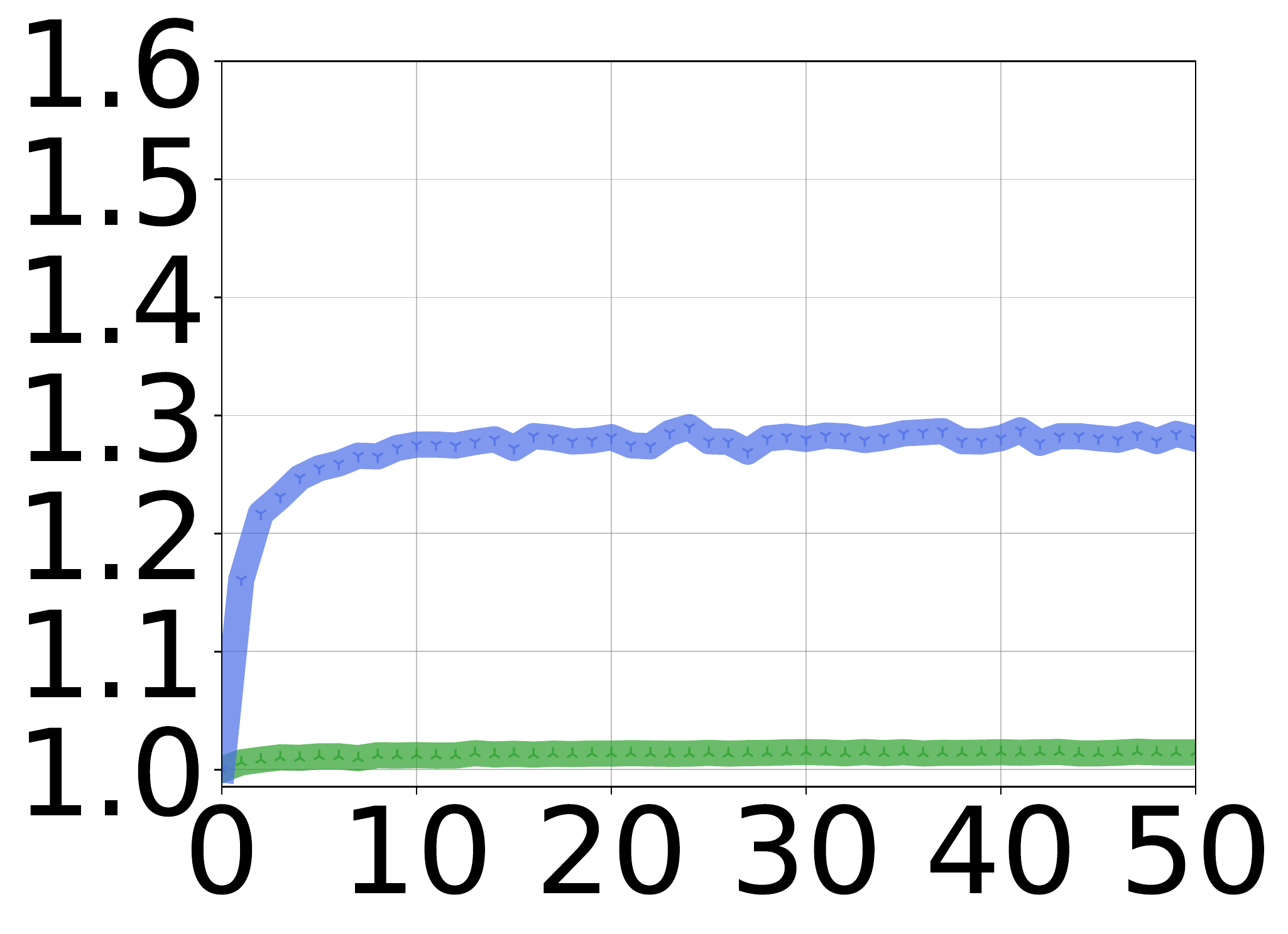}
        \label{fig:Tinyimagenet_resnet50_50_ratio_Random}
	\end{subfigure}
    \end{minipage}
\end{figure*}
\begin{figure*}
    \ContinuedFloat
    \centering
    \underline{ResNet34 20 model ensemble}\\
    \vspace{0.1cm}
    
    \begin{minipage}{0.01\linewidth}
        \rotatebox{90}{\hspace{0.2cm} ensemble/base}
    \end{minipage}
    \begin{minipage}{0.98\linewidth}
	\begin{subfigure}{0.19\linewidth}
		\centering
        Init\\
    	\includegraphics[width=1.0\linewidth]{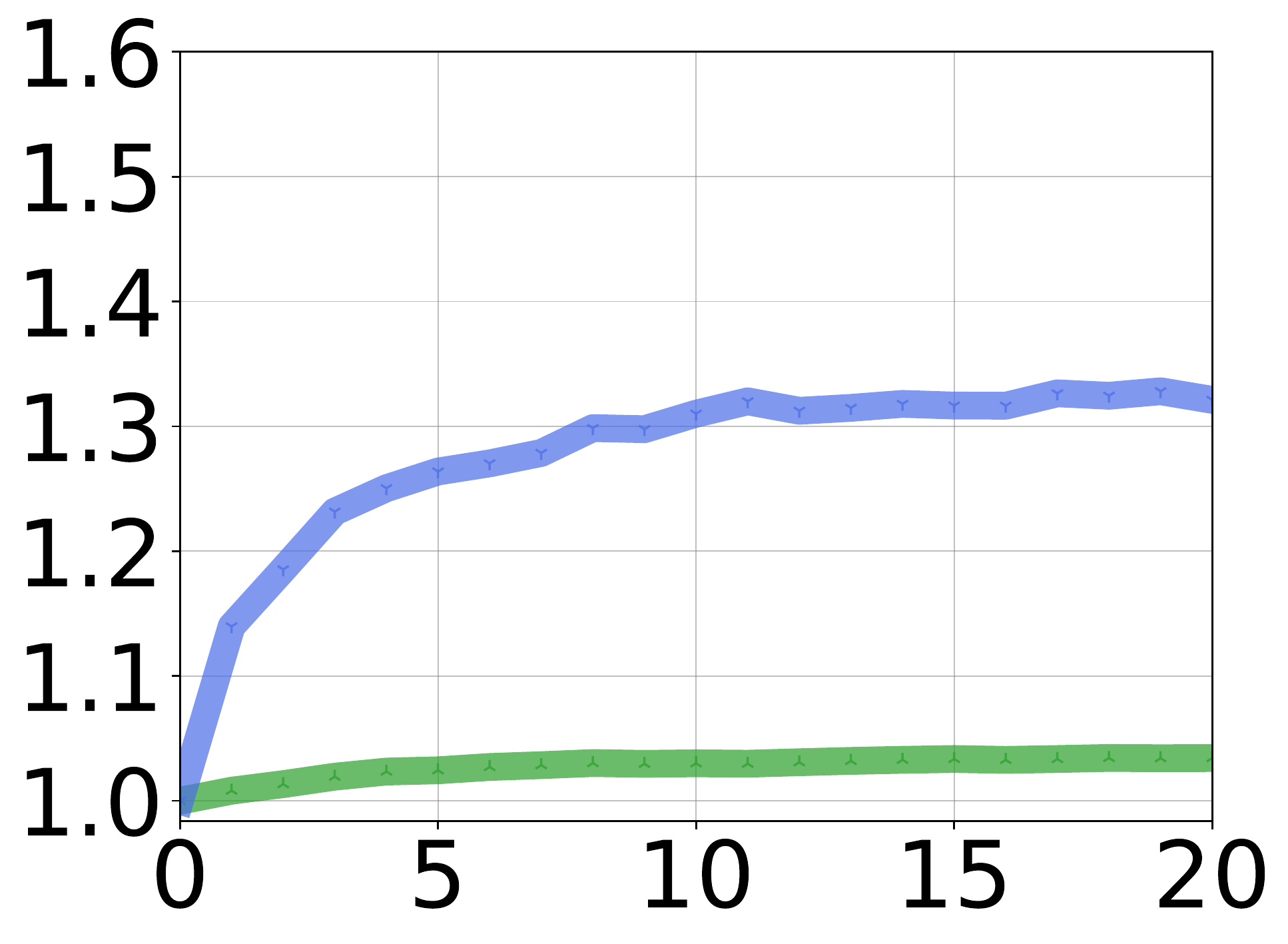}
        \label{fig:Tinyimagenet_resnet34_ratio_Change_ModelInit}
	\end{subfigure}
 	\begin{subfigure}{0.19\linewidth}
		\centering
        BatchOrder\\
    	\includegraphics[width=1.0\linewidth]{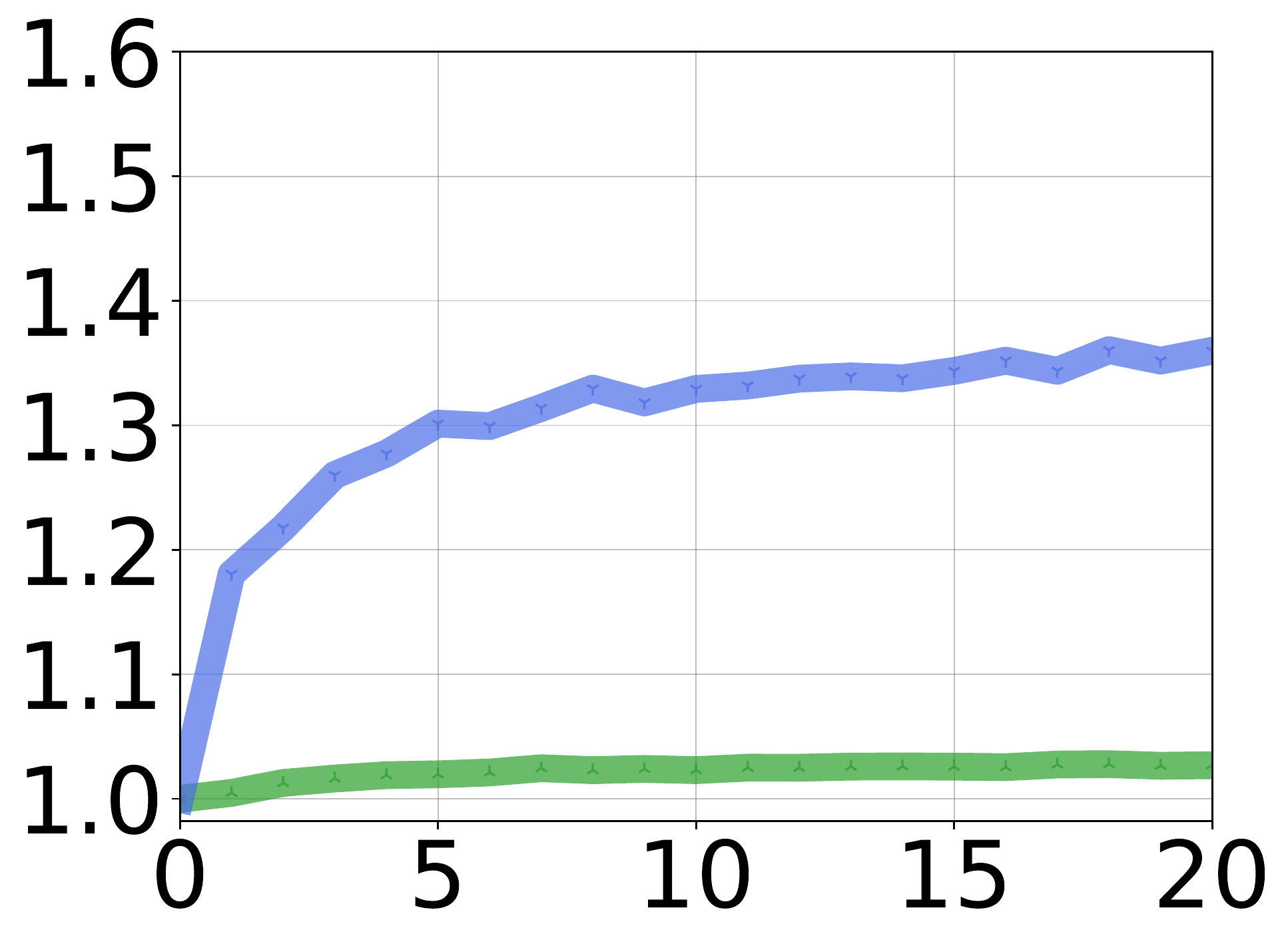}
        \label{fig:Tinyimagenet_resnet34_ratio_Change_BatchOrder}
	\end{subfigure}
	\begin{subfigure}{0.19\linewidth}
		\centering
        DA\\
    	\includegraphics[width=1.0\linewidth]{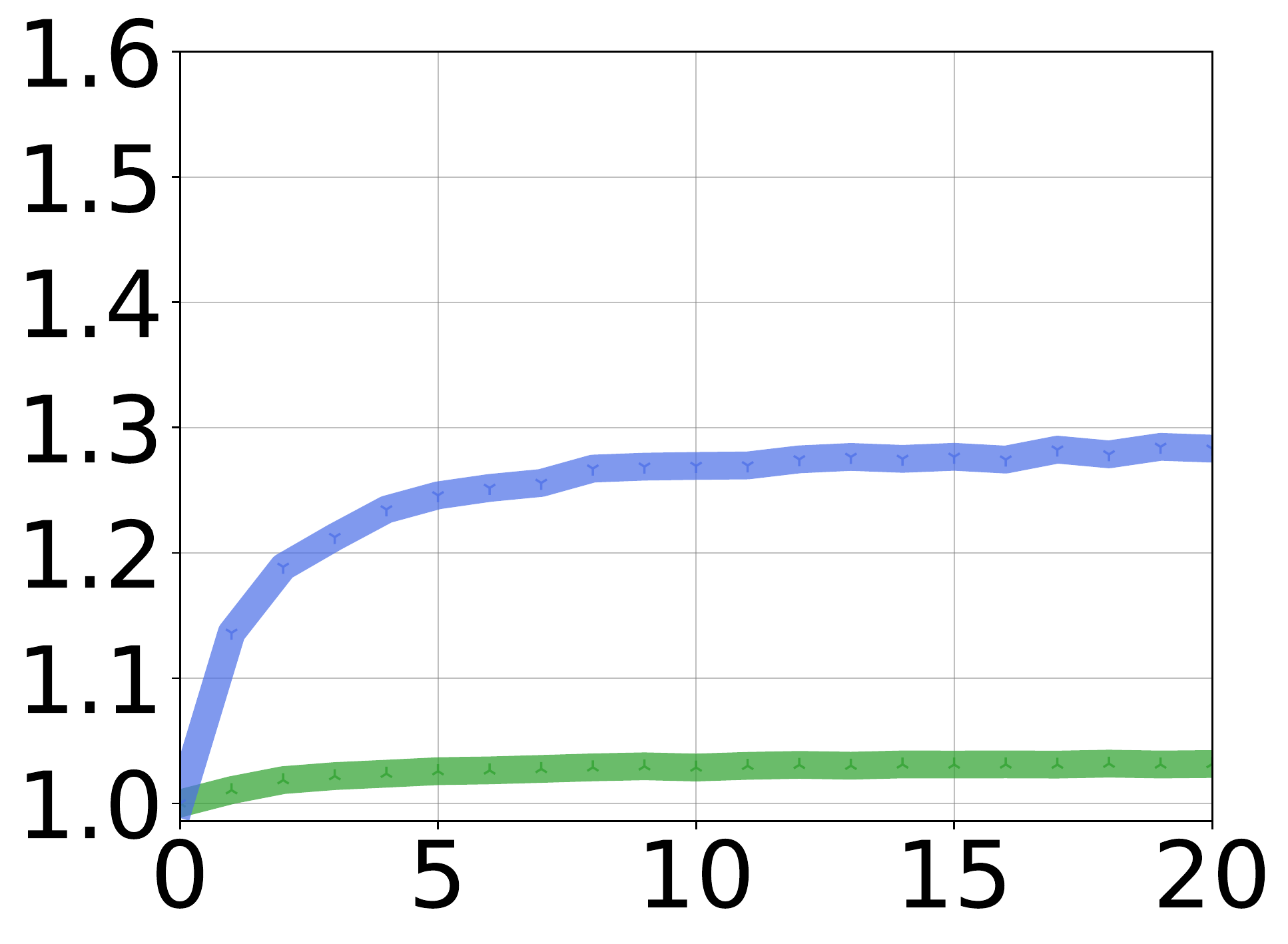}
        \label{fig:Tinyimagenet_resnet34_ratio_Change_DA}
	\end{subfigure}
 	\begin{subfigure}{0.19\linewidth}
		\centering
        Init \& BatchOrder\\
    	\includegraphics[width=1.0\linewidth]{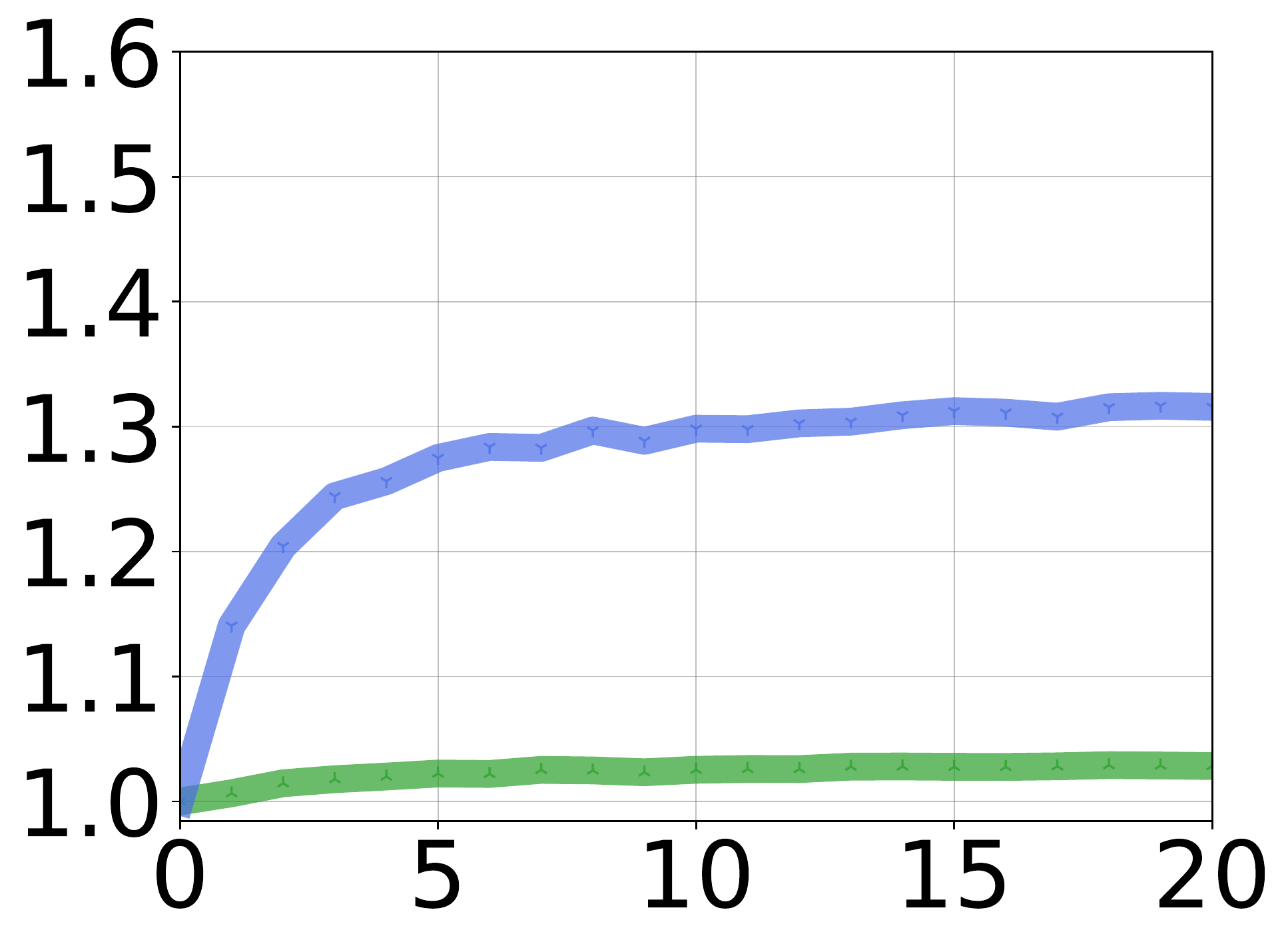}
        \label{fig:Tinyimagenet_resnet34_ratio_Change_ModelInit_BatchOrder}
	\end{subfigure}
    	\begin{subfigure}{0.19\linewidth}
		\centering
        All Sources\\
    	\includegraphics[width=1.0\linewidth]{figures/TINYIMAGENET/ratio_plots/resnet34_20/resnet34_RANDOM_ratio.pdf}
        \label{fig:Tinyimagenet_resnet34_ratio_Random}
	\end{subfigure}
    \end{minipage}
    
    \centering
    \underline{VGG16 20 model ensemble}\\
    \vspace{0.1cm}
    
    \begin{minipage}{0.01\linewidth}
        \rotatebox{90}{\hspace{0.2cm} ensemble/base}
    \end{minipage}
    \begin{minipage}{0.98\linewidth}
	\begin{subfigure}{0.19\linewidth}
		\centering
        Init\\
    	\includegraphics[width=1.0\linewidth]{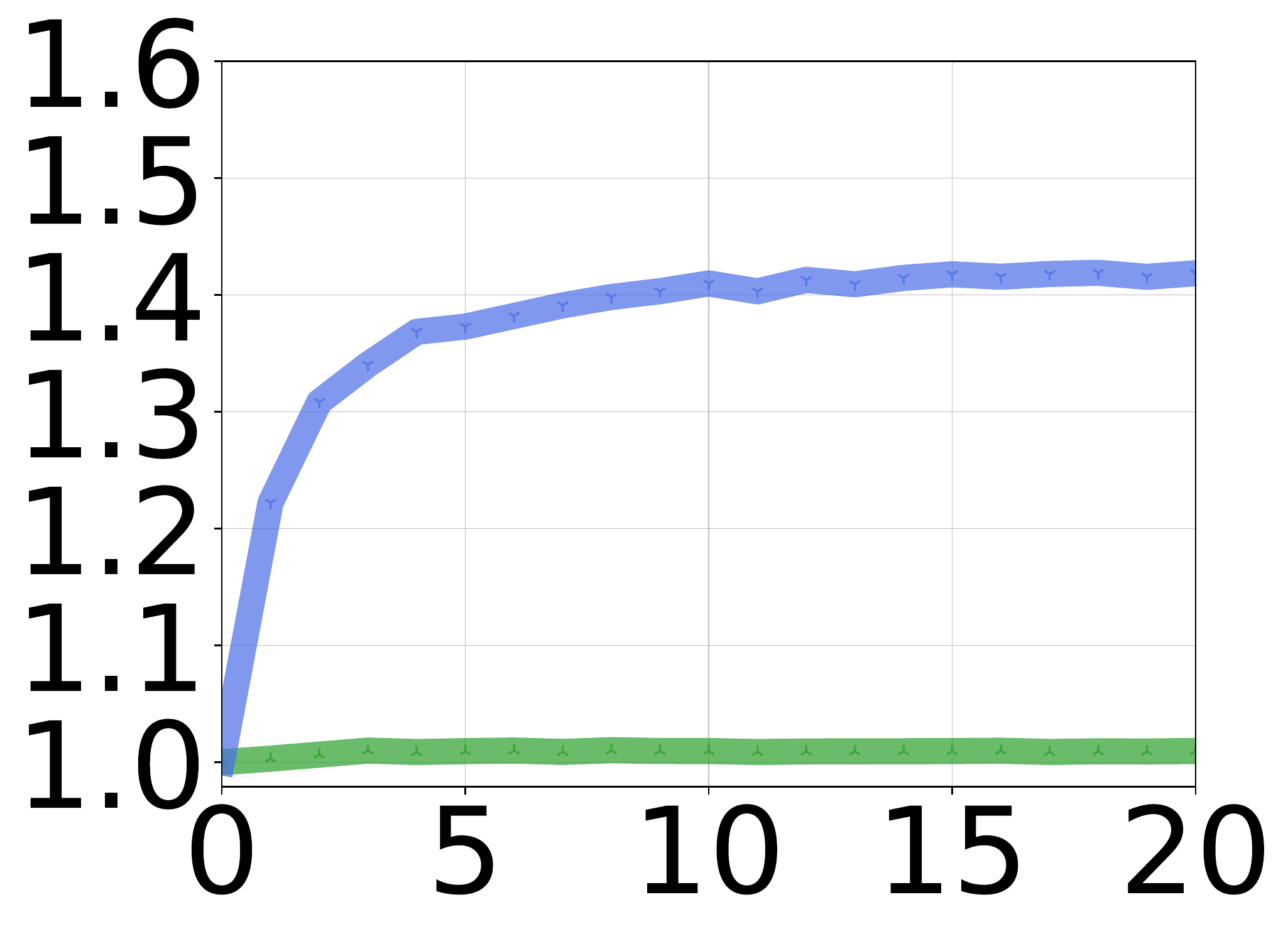}
        \label{fig:Tinyimagenet_vgg16_ratio_Change_ModelInit}
	\end{subfigure}
 	\begin{subfigure}{0.19\linewidth}
		\centering
        BatchOrder\\
    	\includegraphics[width=1.0\linewidth]{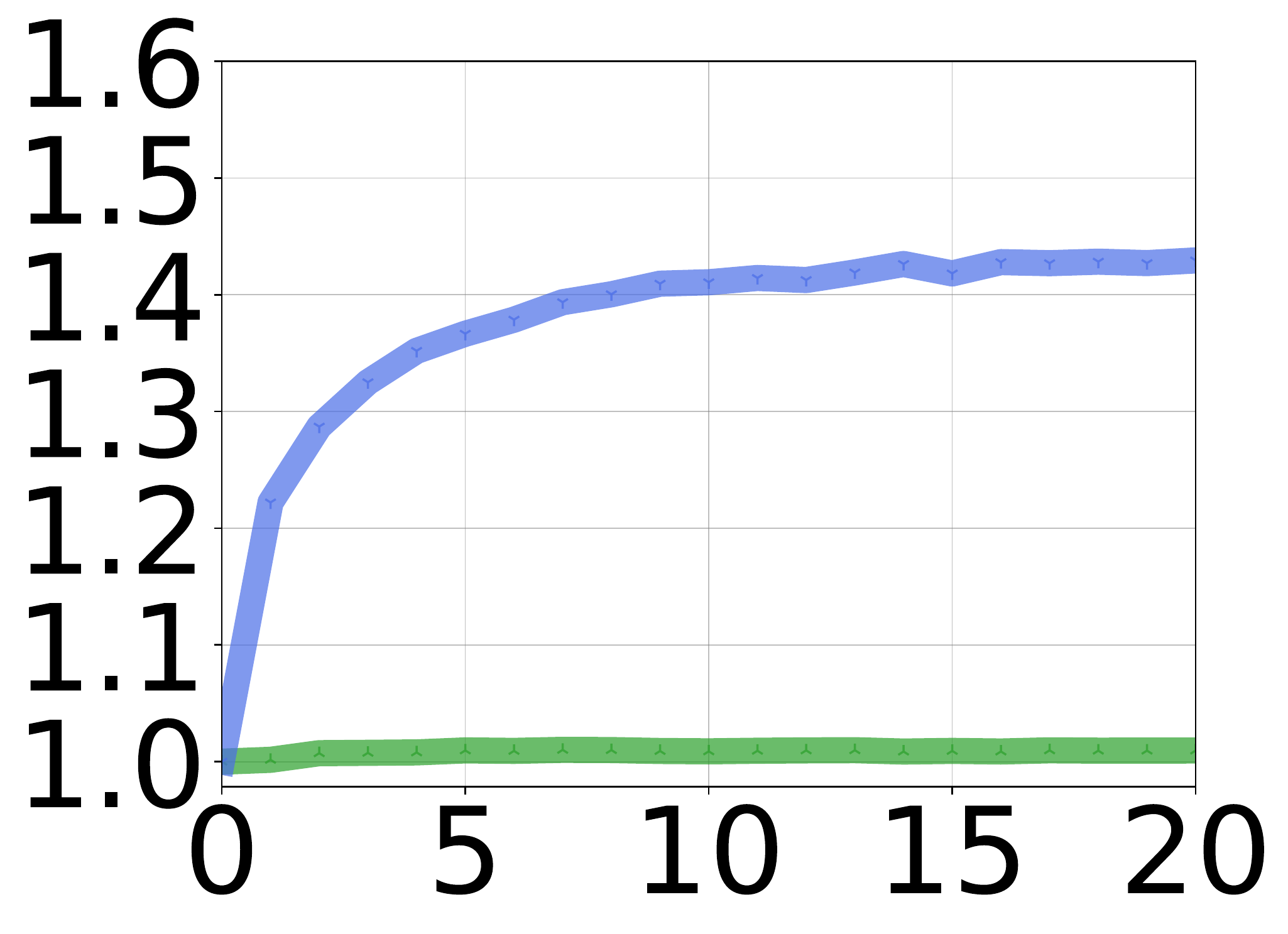}
        \label{fig:Tinyimagenet_vgg16_ratio_Change_BatchOrder}
	\end{subfigure}
	\begin{subfigure}{0.19\linewidth}
		\centering
        DA\\
    	\includegraphics[width=1.0\linewidth]{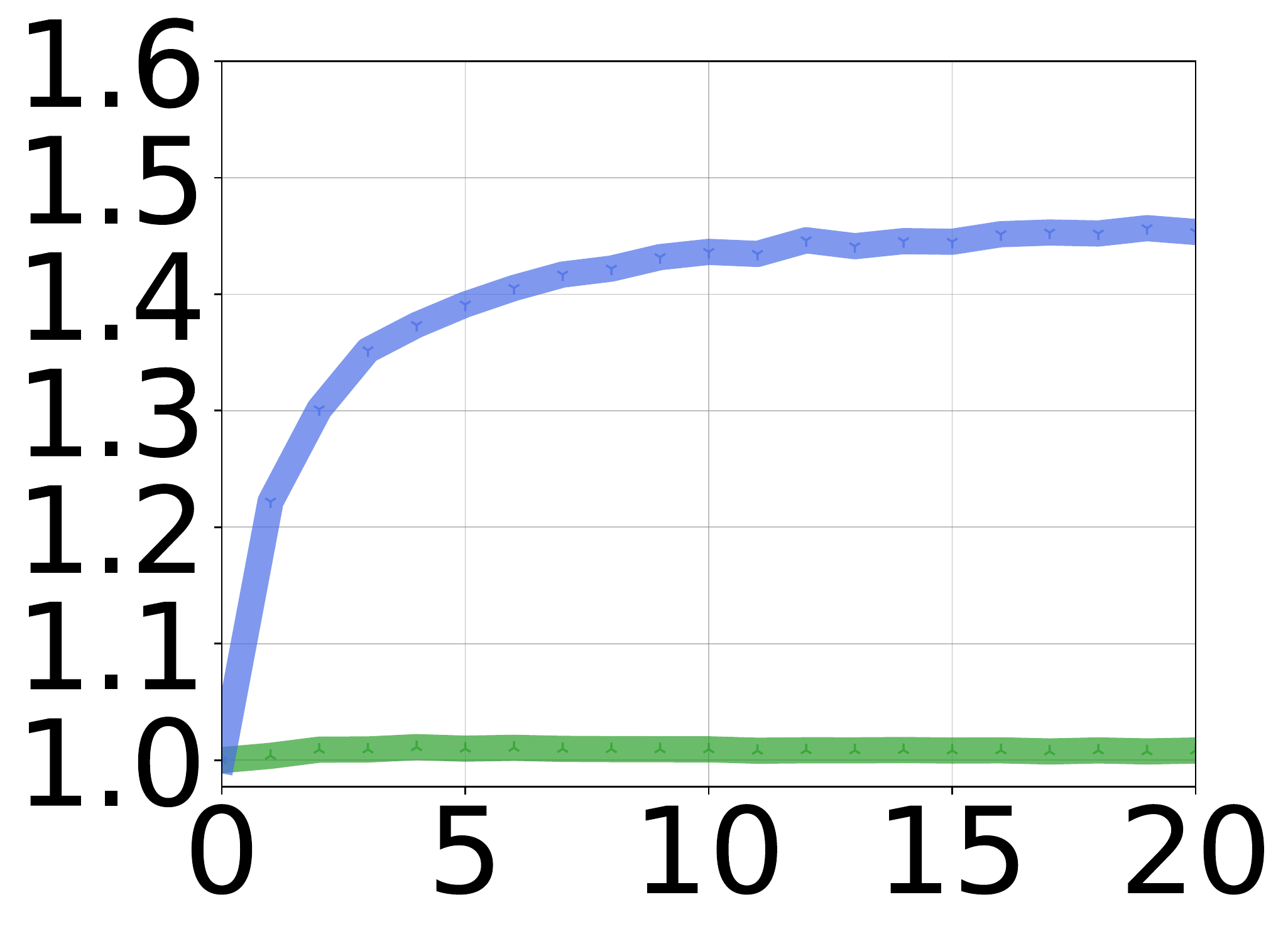}
        \label{fig:Tinyimagenet_vgg16_ratio_Change_DA}
	\end{subfigure}
 	\begin{subfigure}{0.19\linewidth}
		\centering
        Init \& BatchOrder\\
    	\includegraphics[width=1.0\linewidth]{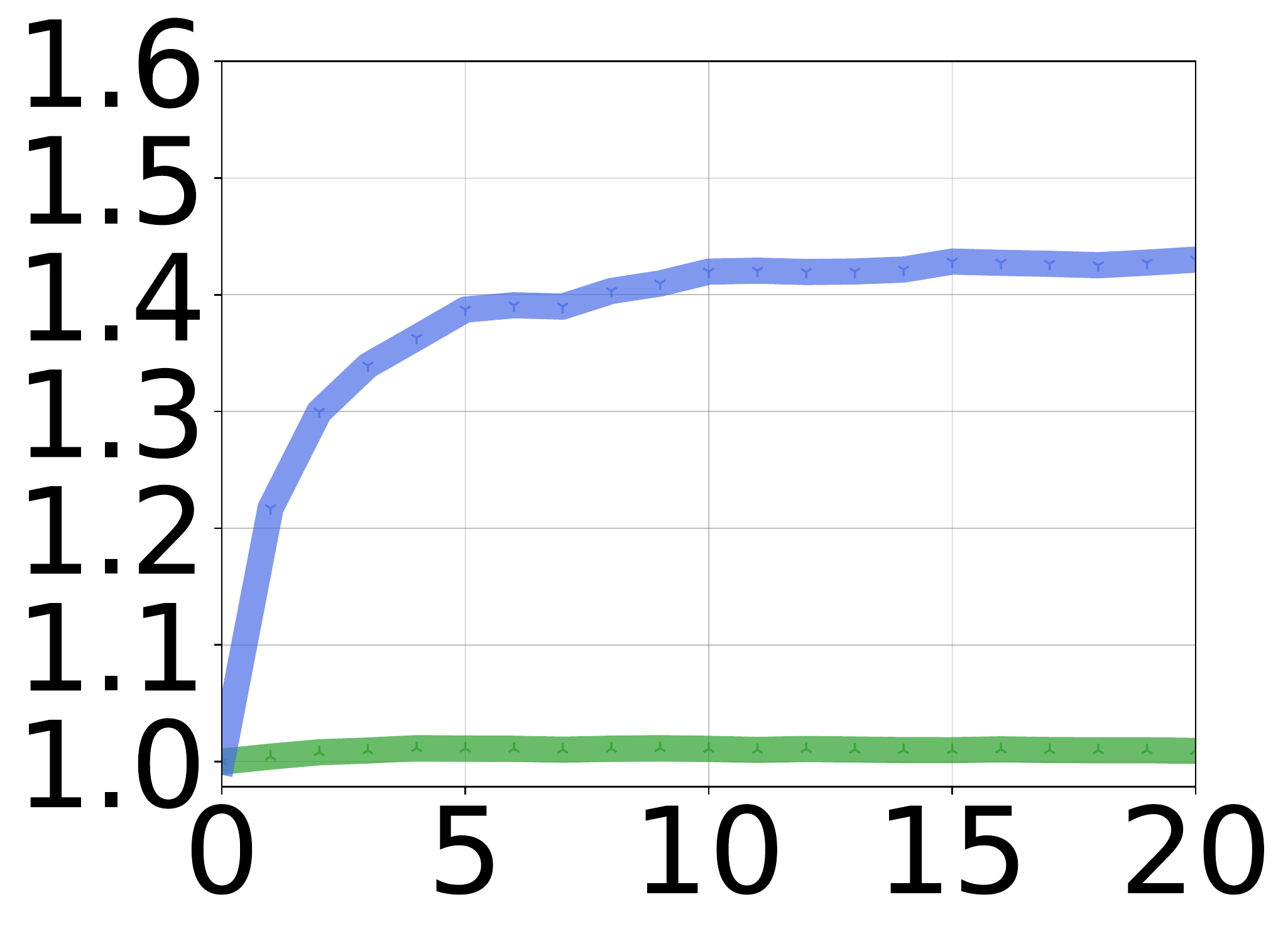}
        \label{fig:Tinyimagenet_vgg16_ratio_Change_ModelInit_BatchOrder}
	\end{subfigure}
    	\begin{subfigure}{0.19\linewidth}
		\centering
        All Sources\\
    	\includegraphics[width=1.0\linewidth]{figures/TINYIMAGENET/ratio_plots/vgg16_20/vgg16_RANDOM_ratio.pdf}
        \label{fig:Tinyimagenet_vgg16_ratio_Random}
	\end{subfigure}
    \end{minipage}
    
    \centering
    \underline{ViT 20 model ensemble}\\
    \vspace{0.1cm}

    \begin{minipage}{0.01\linewidth}
        \rotatebox{90}{ensemble/base}
    \end{minipage}
    \begin{minipage}{0.98\linewidth}
	\begin{subfigure}{0.19\linewidth}
		\centering
        Init\\
    	\includegraphics[width=1.0\linewidth]{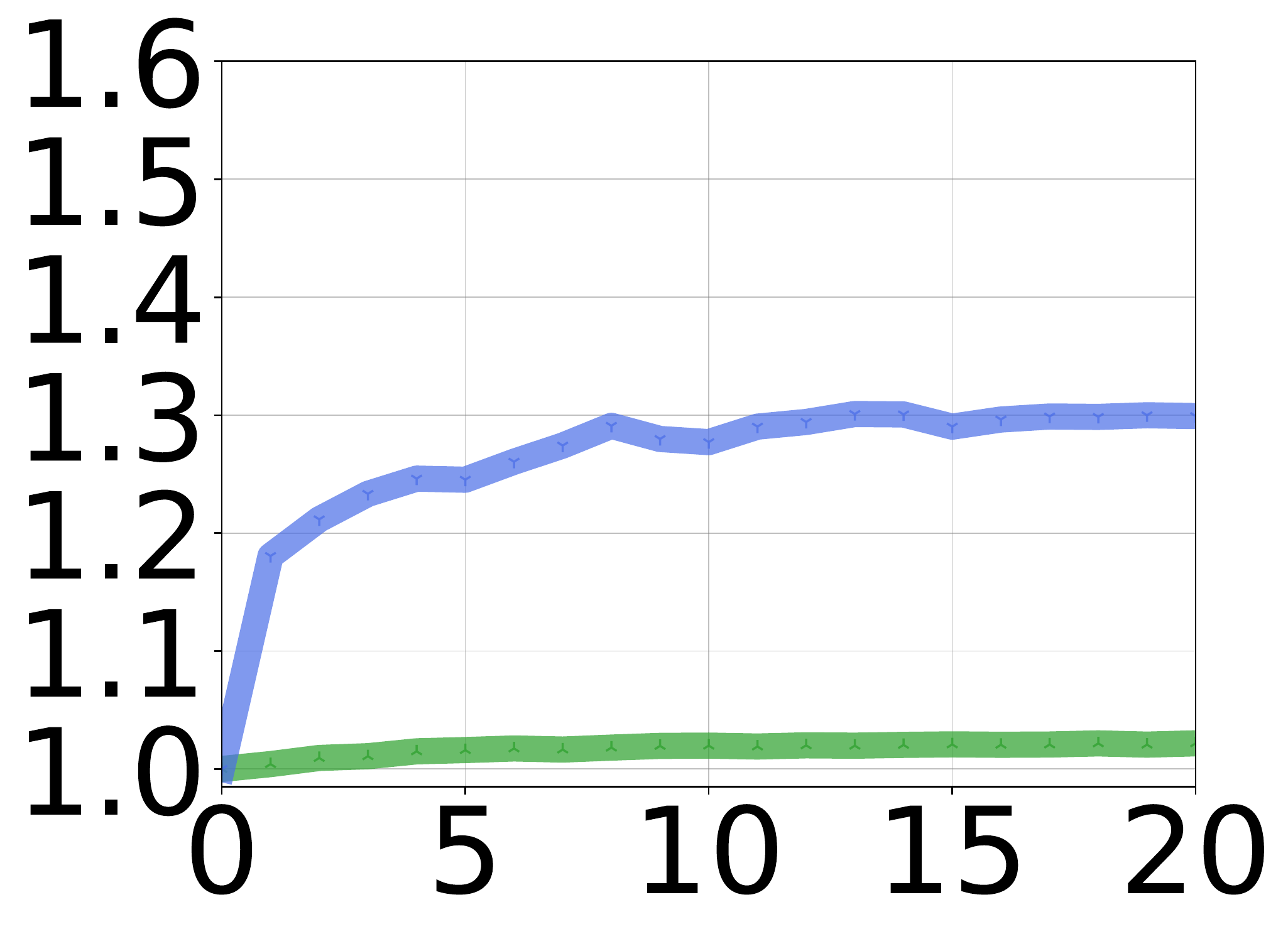}
        \\[-0.7em]
    	{\hspace{0.2cm} \small models in ensemble}
        \label{fig:Tinyimagenet_vit_ratio_Change_ModelInit}
	\end{subfigure}
 	\begin{subfigure}{0.19\linewidth}
		\centering
        BatchOrder\\
    	\includegraphics[width=1.0\linewidth]{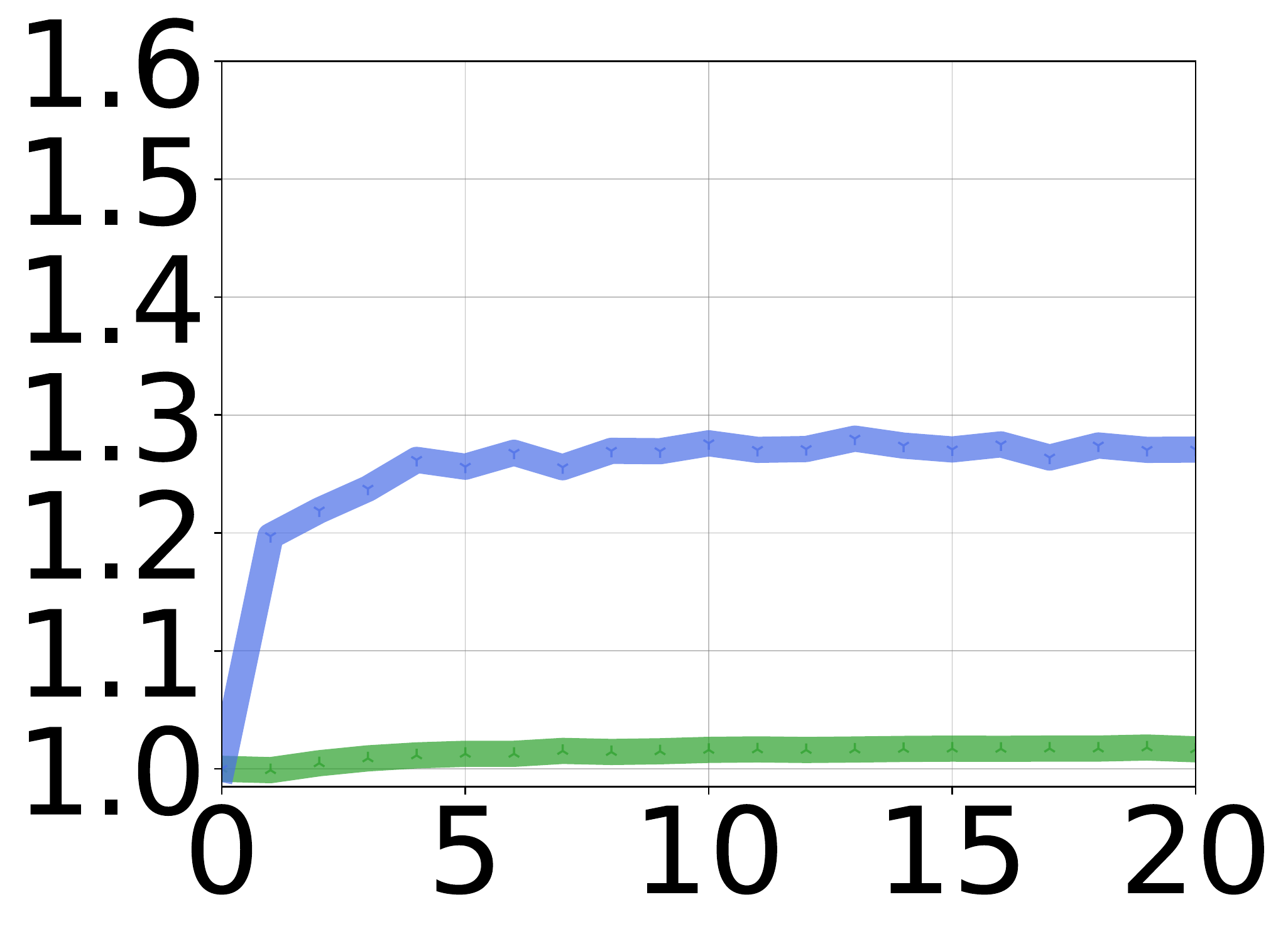}
        \\[-0.7em]
    	{\hspace{0.2cm} \small models in ensemble}
        \label{fig:Tinyimagenet_vit_ratio_Change_BatchOrder}
	\end{subfigure}
	\begin{subfigure}{0.19\linewidth}
		\centering
        DA\\
    	\includegraphics[width=1.0\linewidth]{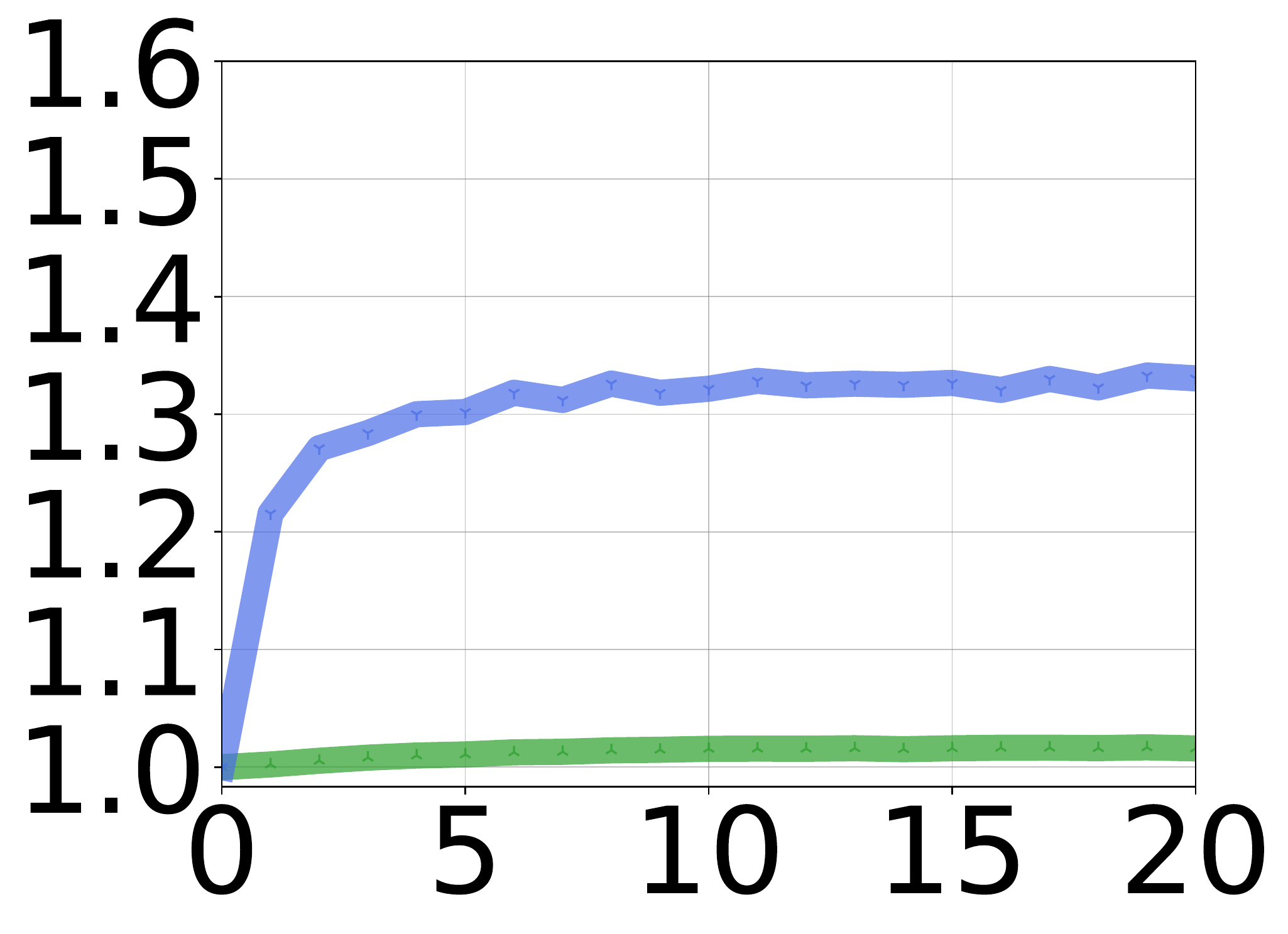}
        \\[-0.7em]
    	{\hspace{0.2cm} \small models in ensemble}
        \label{fig:Tinyimagenet_vit_ratio_Change_DA}
	\end{subfigure}
 	\begin{subfigure}{0.19\linewidth}
		\centering
        Init \& BatchOrder\\
    	\includegraphics[width=1.0\linewidth]{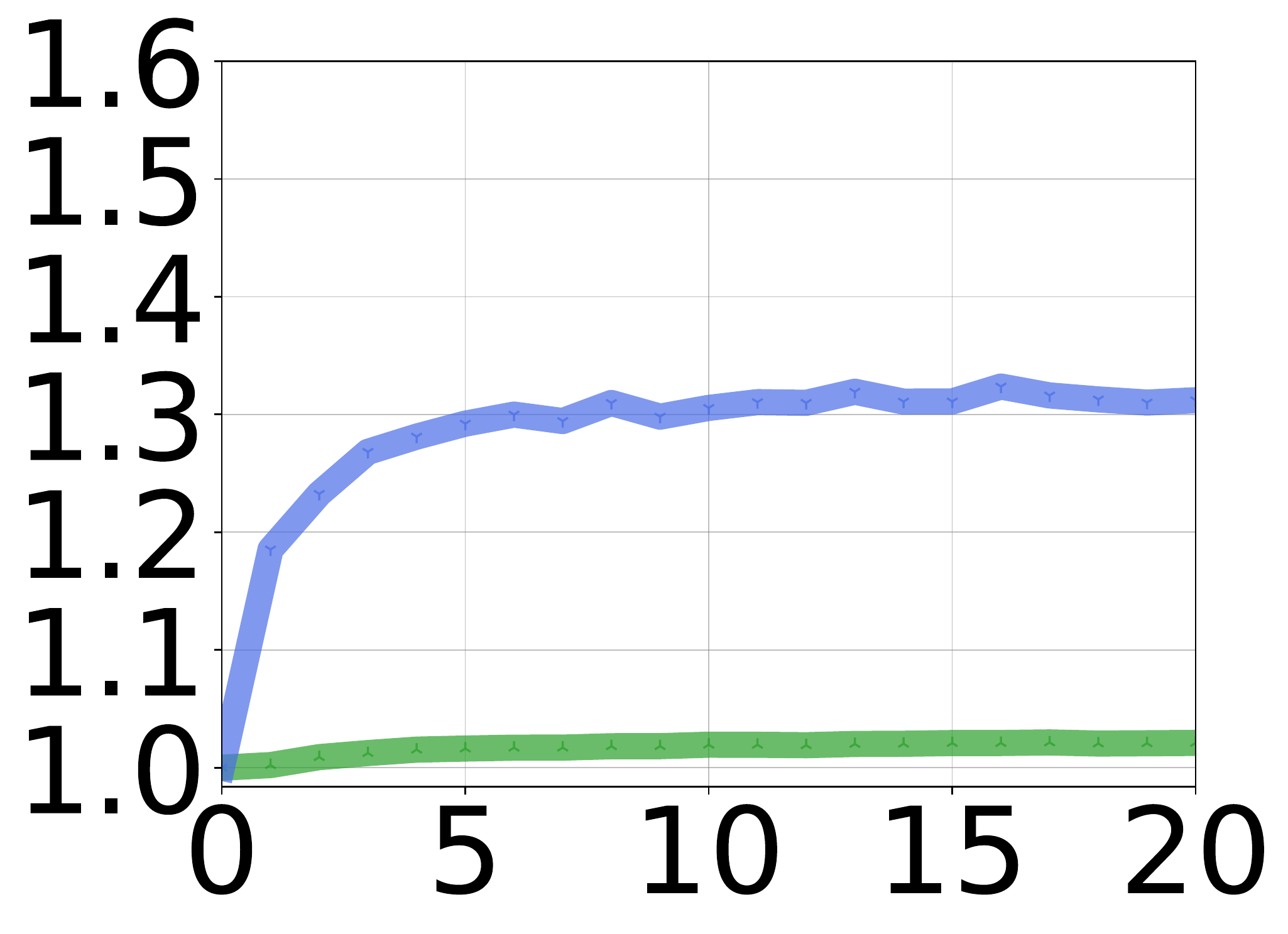}
        \\[-0.7em]
    	{\hspace{0.2cm} \small models in ensemble}
        \label{fig:Tinyimagenet_vit_ratio_Change_ModelInit_BatchOrder}
	\end{subfigure}
    	\begin{subfigure}{0.19\linewidth}
		\centering
        All Sources\\
    	\includegraphics[width=1.0\linewidth]{figures/TINYIMAGENET/ratio_plots/vit_20/vit_RANDOM_ratio.pdf}
        \\[-0.7em]
    	{\hspace{0.2cm} \small models in ensemble}
        \label{fig:Tinyimagenet_vit_ratio_Random}
	\end{subfigure}
    \end{minipage}

	\caption{ Ratio of Top \& Bottom K ensemble accuracy for different model architectures and ensemble sizes on TinyImageNet
	}
    \label{fig:TinyImageNet_ratio}
\end{figure*}

\begin{figure*}[ht]
    \section{FAIR Ensemble: Improved Robustness}
    \vspace{0.2cm}
        \centering
        \includegraphics[width=0.8\linewidth]{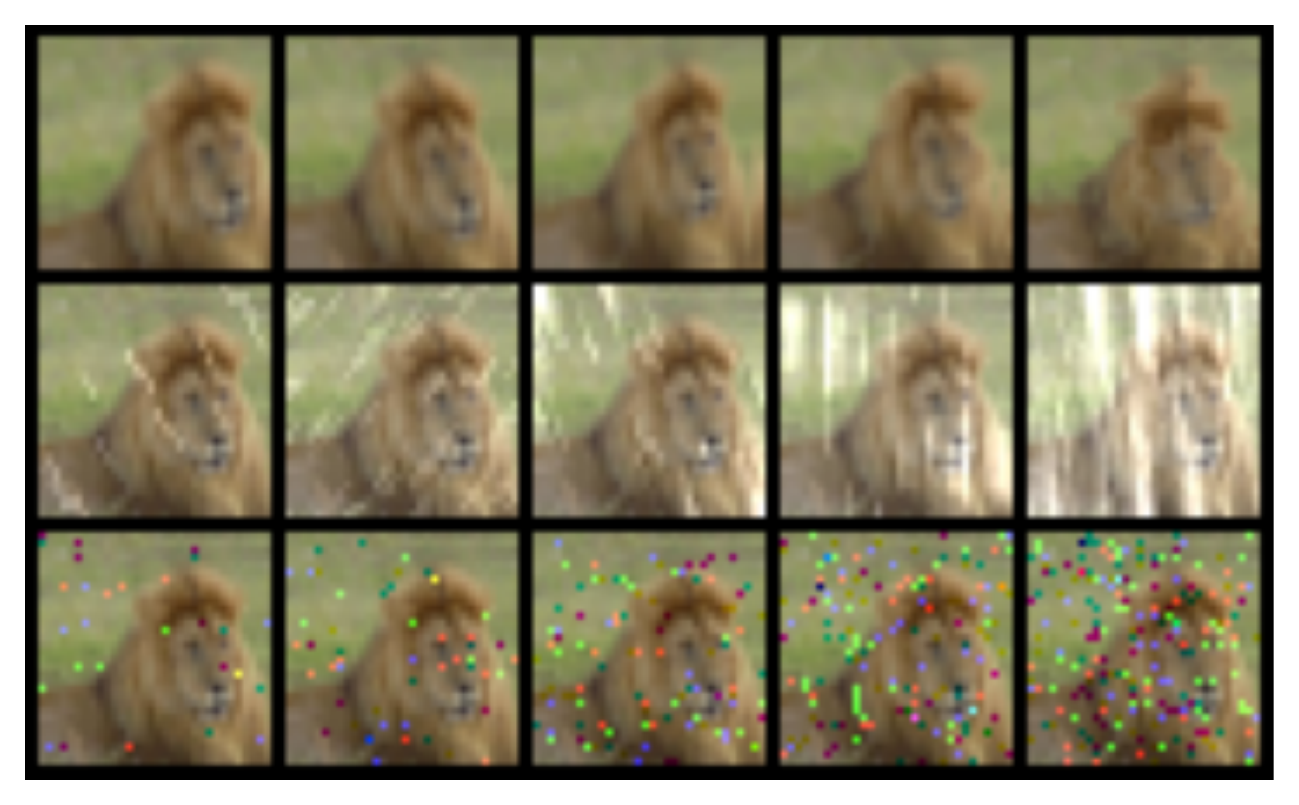}
        \caption{\small For each {\bf row} we depict three of the different types of corruptions from the CIFAR100-C dataset (\texttt{elastic\_transform}, \texttt{snow}, and \texttt{impulse\_noise} respectively), and for each {\bf column} we depict the corruption severity levels ($1\mapsto 5$). The image belongs to the \texttt{lion} class.}
        \label{fig:cifar100c_samples}

\end{figure*}

\begin{figure*}[ht]
	\centering
	\begin{subfigure}{0.32\linewidth}
		\centering
    	\includegraphics[width=1.0\linewidth]{figures/CIFAR100-C/Corrupt_Subset_Severity_Resnet9.pdf}
        \\[-0.7em]
    	{\hspace{0.5cm} \small models in ensemble}
    	\caption{ResNet-9}
        \label{fig:Res9_CIFAR100_AllSev}
	\end{subfigure}
	\begin{subfigure}{0.32\linewidth}
		\centering
    	\includegraphics[width=1.0\linewidth]{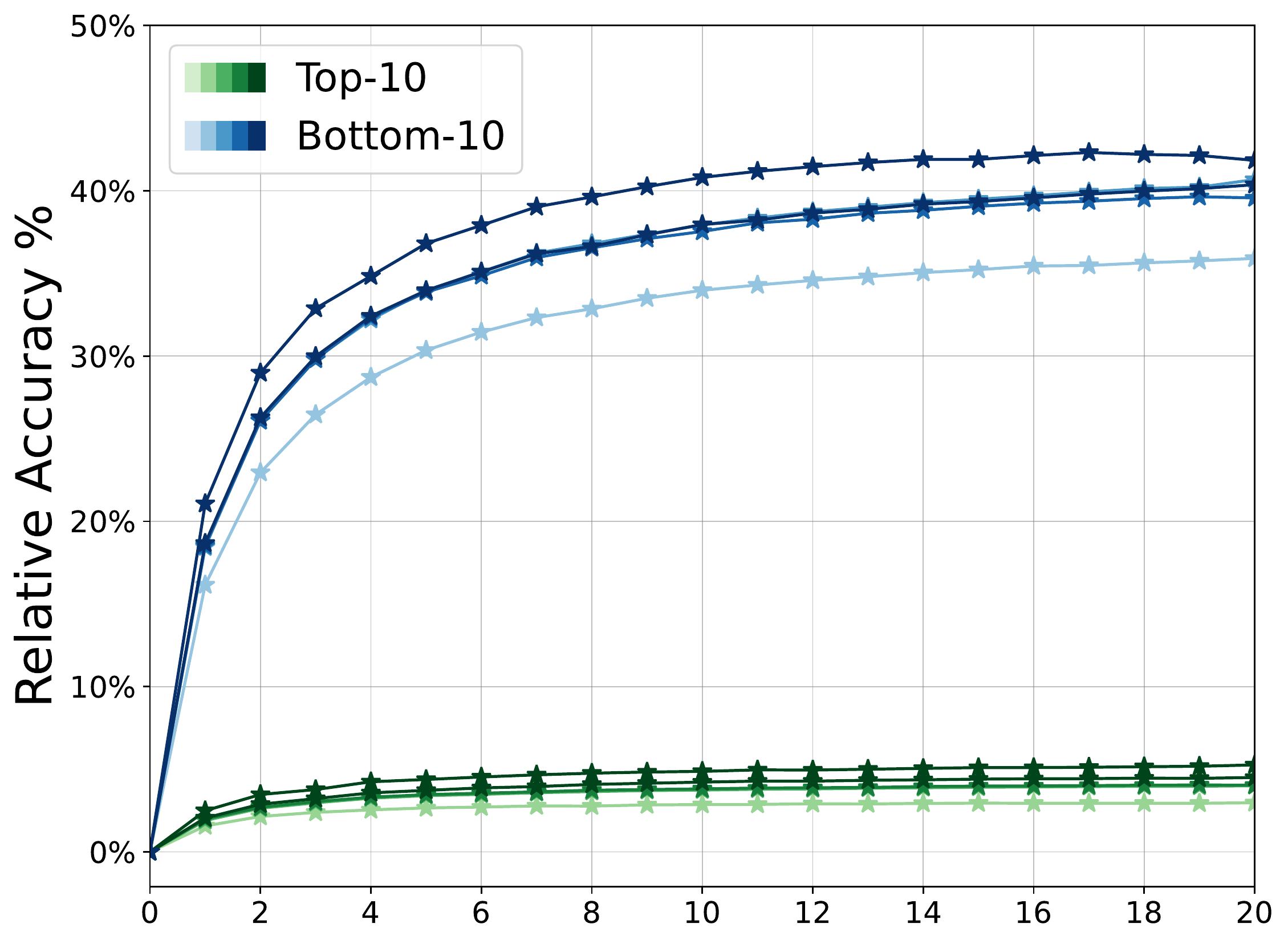}
        \\[-0.7em]
    	{\hspace{0.5cm} \small models in ensemble}
    	\caption{VGG-16}
        \label{fig:Vgg16_CIFAR100_AllSev}
	\end{subfigure}
	\begin{subfigure}{0.32\linewidth}
		\centering
    	\includegraphics[width=1.0\linewidth]{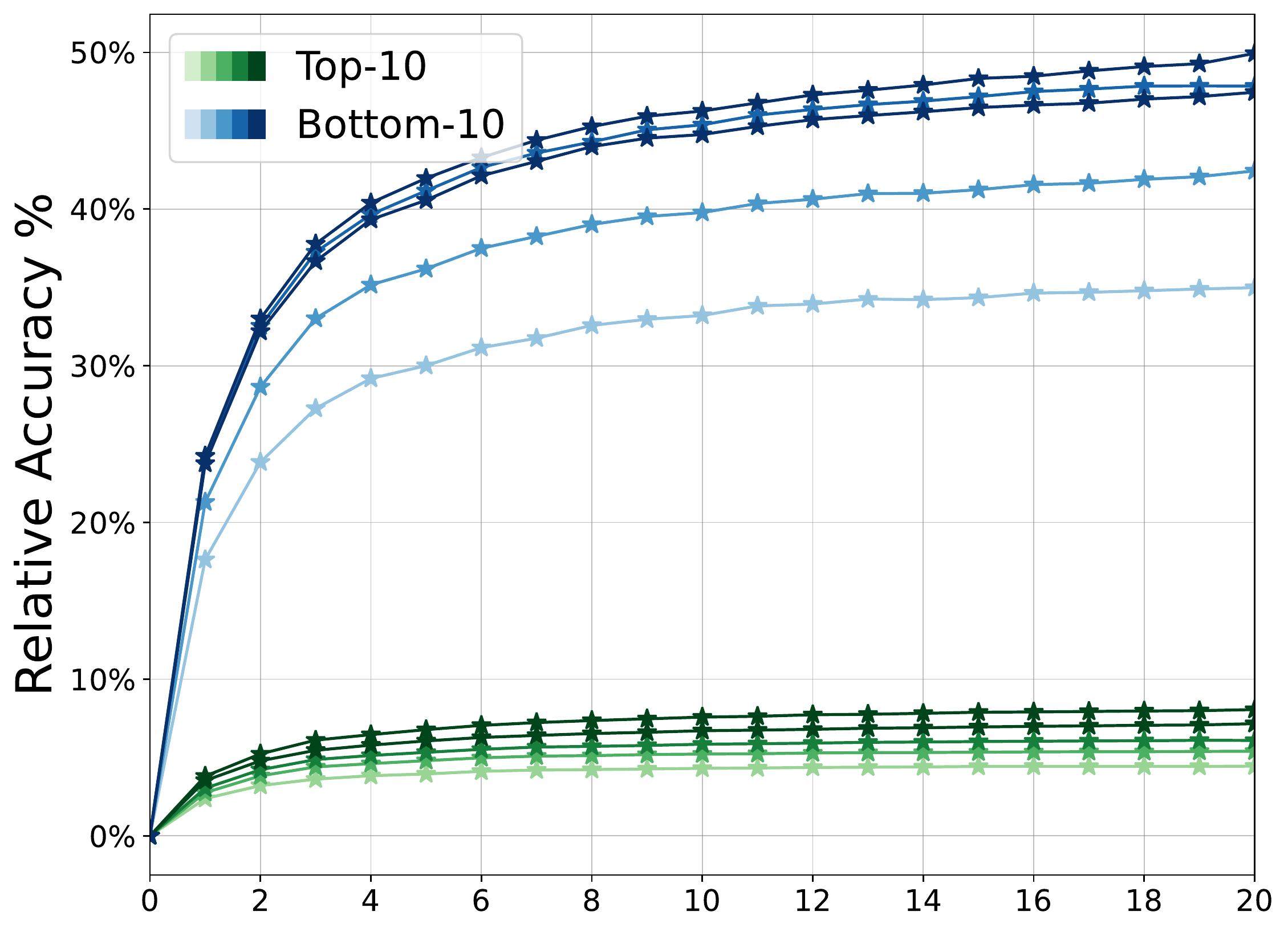}
        \\[-0.7em]
    	{\hspace{0.5cm} \small models in ensemble}
    	\caption{MLP-Mixer}
        \label{fig:MlpMixer_CIFAR100_AllSev}
	\end{subfigure}
	\caption{
	    Performance on CIFAR-100 Corrupt based on Severity Levels. 
	}
	\label{fig:Cifar100C_AllArch}
\end{figure*}

\begin{figure*}[ht]
    \section{Difference in Churn Between Models Explains Ensemble Fairness}
    \vspace{0.2cm}
    \centering
    Churn(\%) vs Relative Test Accuracy Improvement(\%)\\
    \begin{minipage}{0.49\linewidth}
    \centering
    \underline{CIFAR}\\
    \includegraphics[width=\linewidth]{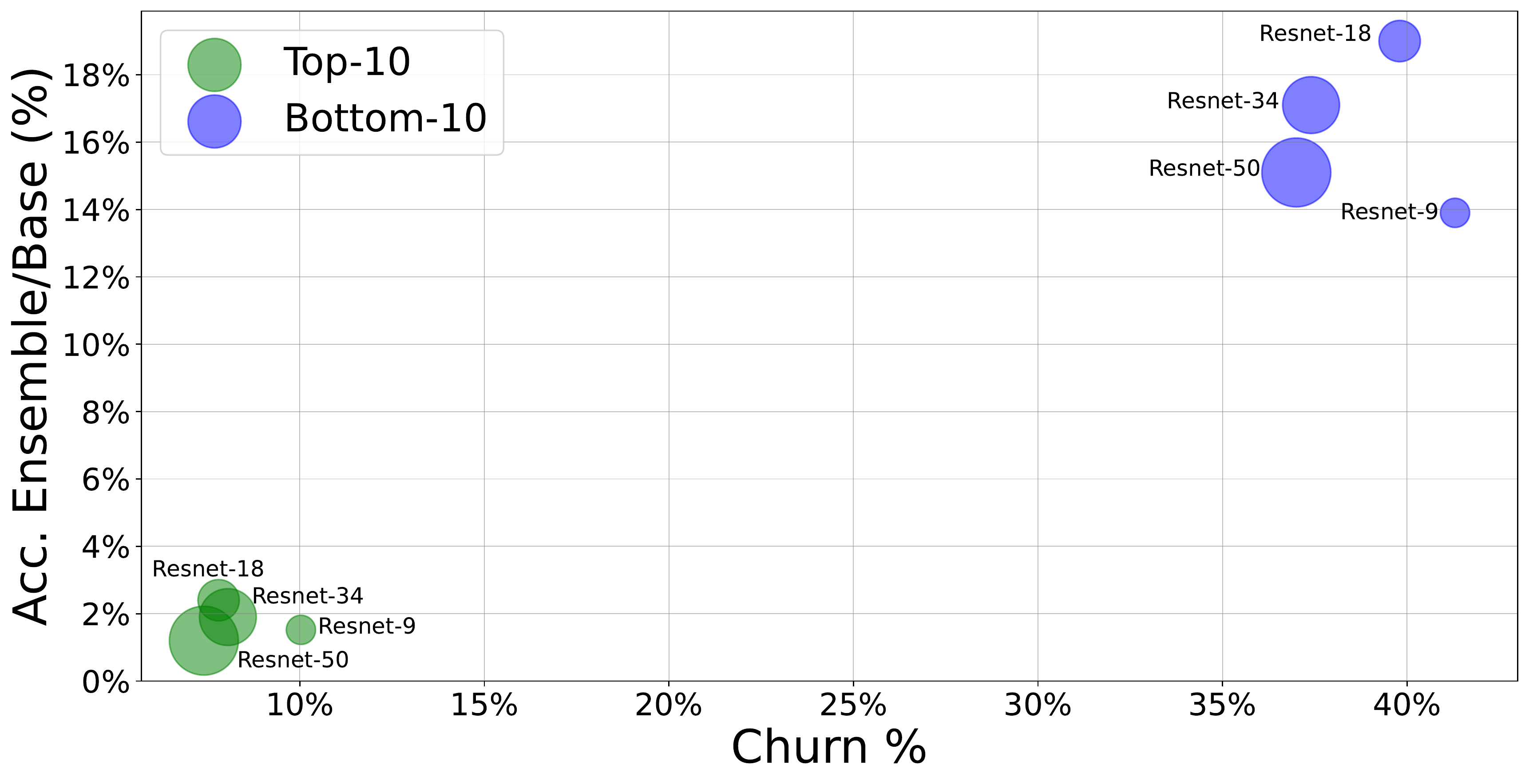}
    \end{minipage}
    \begin{minipage}{0.49\linewidth}
    \centering
    \underline{TinyImageNet}\\
    \includegraphics[width=\linewidth]{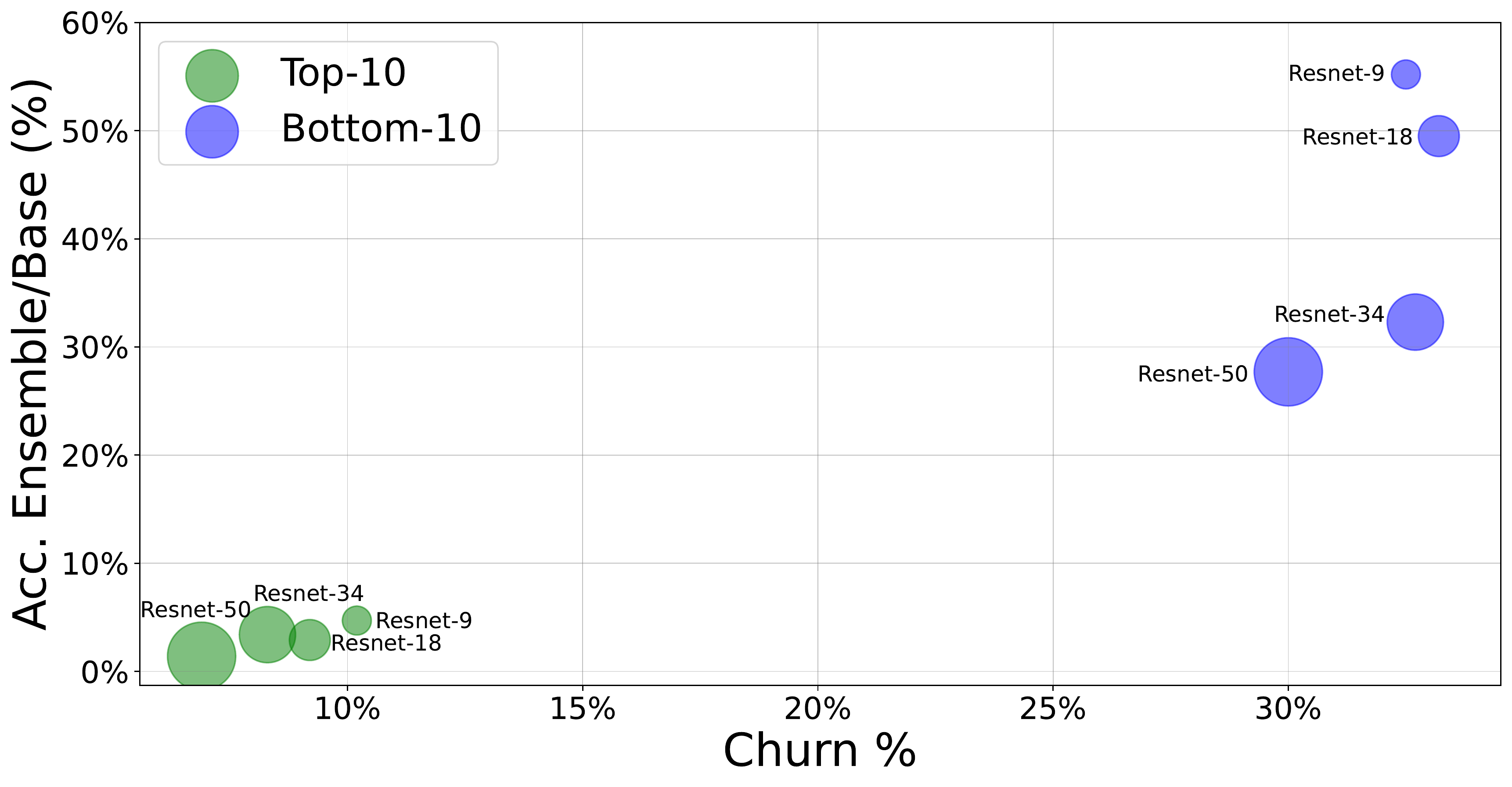}
    \end{minipage}\\
	\vspace{-0.3cm}
    \caption{Correlation between Model Disagreement and Ensemble Performance}
    \vspace{-0.5cm}
    \label{fig:churn_correlation}
\end{figure*}

\begin{figure*}[!t]
    \section{Can Different Sources of Stochasticity Improve Homogeneous Deep Ensemble Fairness?}
    \subsection{Contribution of stochasticity in ensembles}
    \centering
    \underline{Batch Sizes}\\
    \vspace{0.1cm}
    \begin{minipage}{0.01\linewidth}
        \rotatebox{90}{test accuracy \%}
    \end{minipage}
    \hspace{0.1cm}
    \begin{minipage}{0.97\linewidth}
    \begin{subfigure}{0.32\linewidth}
        \centering
        All-K classes (K=10)\\
    	\includegraphics[width=1.0\linewidth]{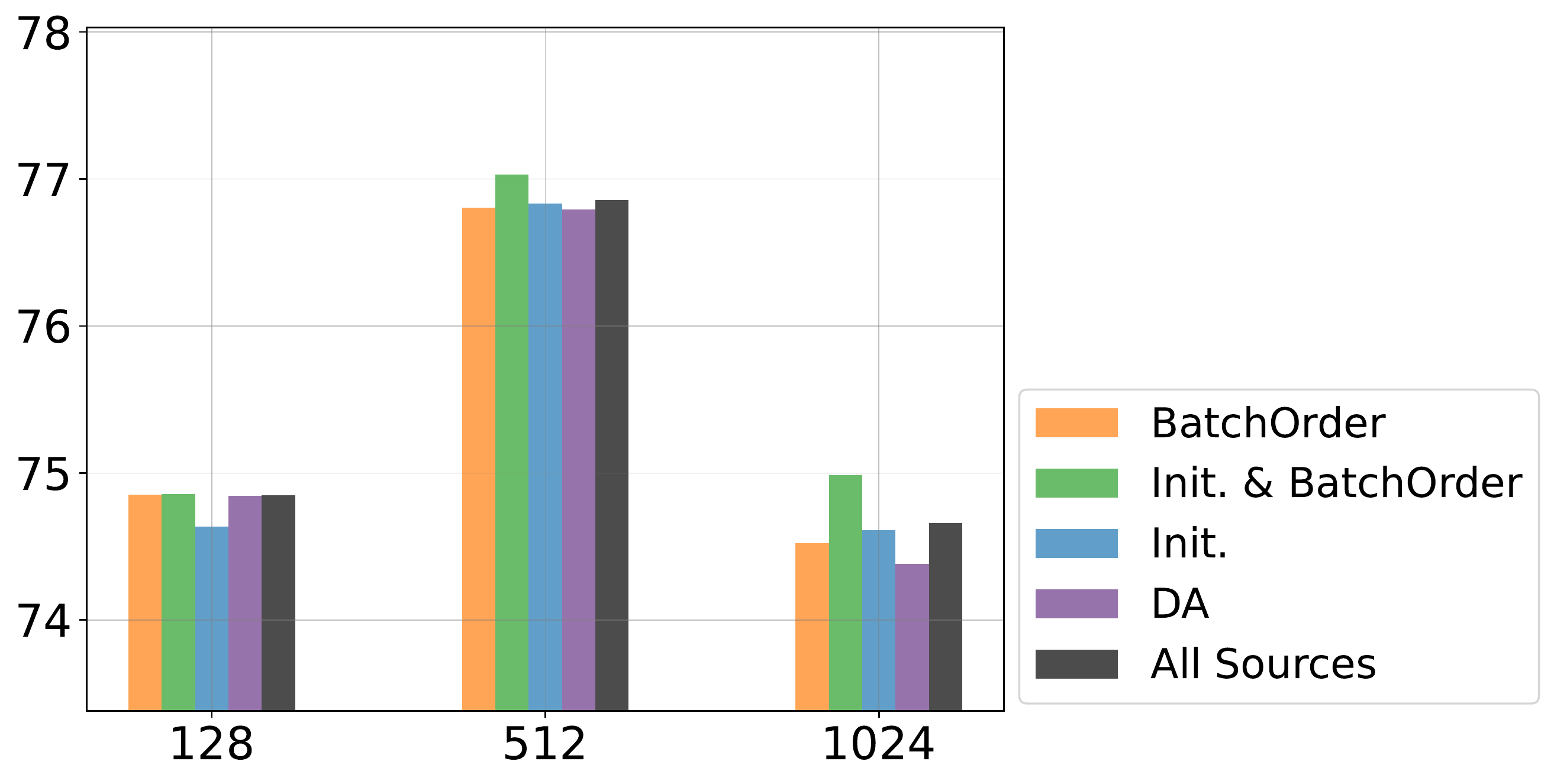}
        \\[-0.7em]
        batch size
    \end{subfigure}
    \begin{subfigure}{0.32\linewidth}
        \centering
        Top-K classes (K=10)\\
    	\includegraphics[width=1.0\linewidth]{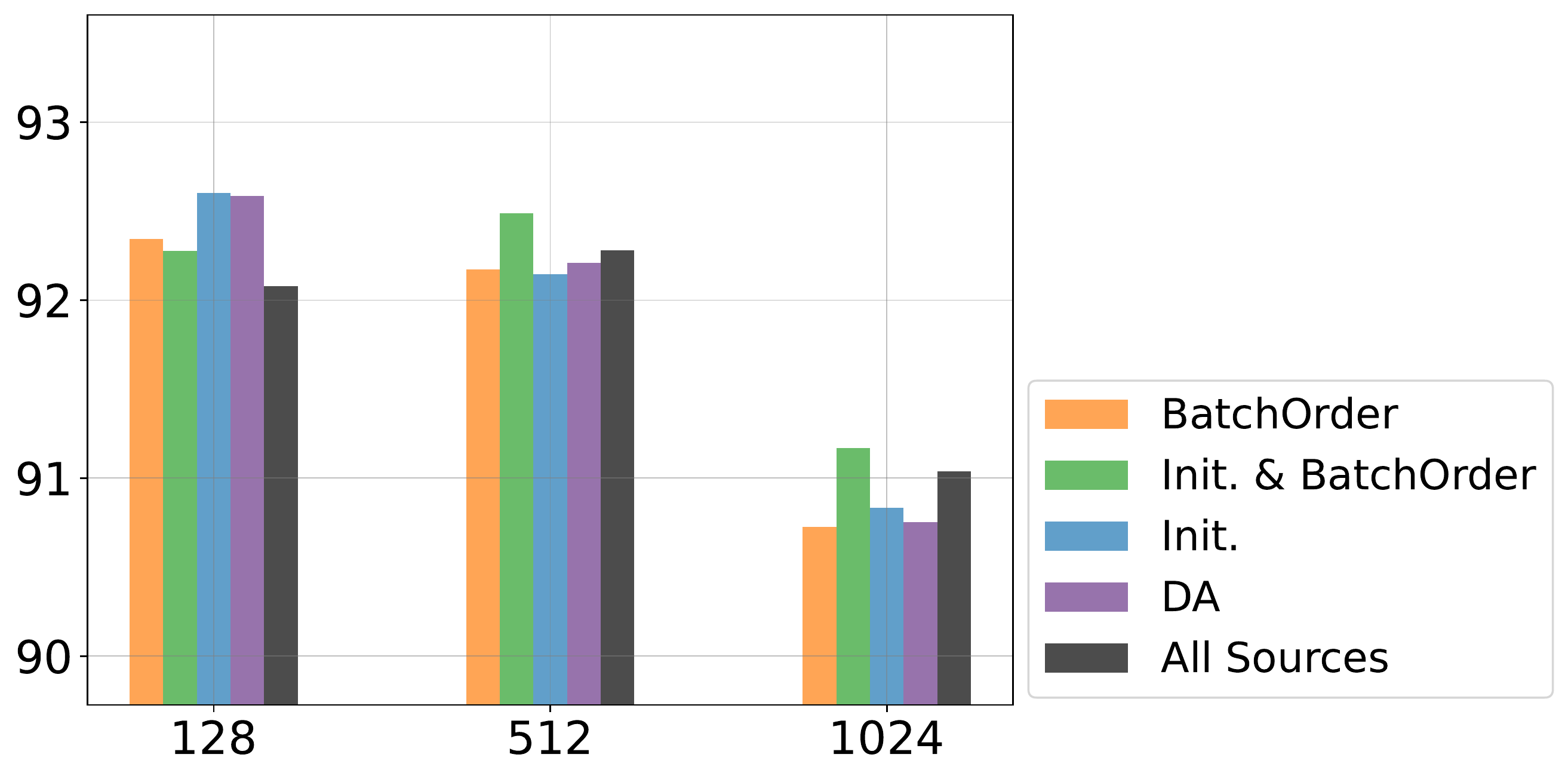}
        \\[-0.7em]
        batch size
    \end{subfigure}
    \begin{subfigure}{0.32\linewidth}
        \centering
        Bottom-K classes (K=10)\\
        \includegraphics[width=1.0\linewidth]{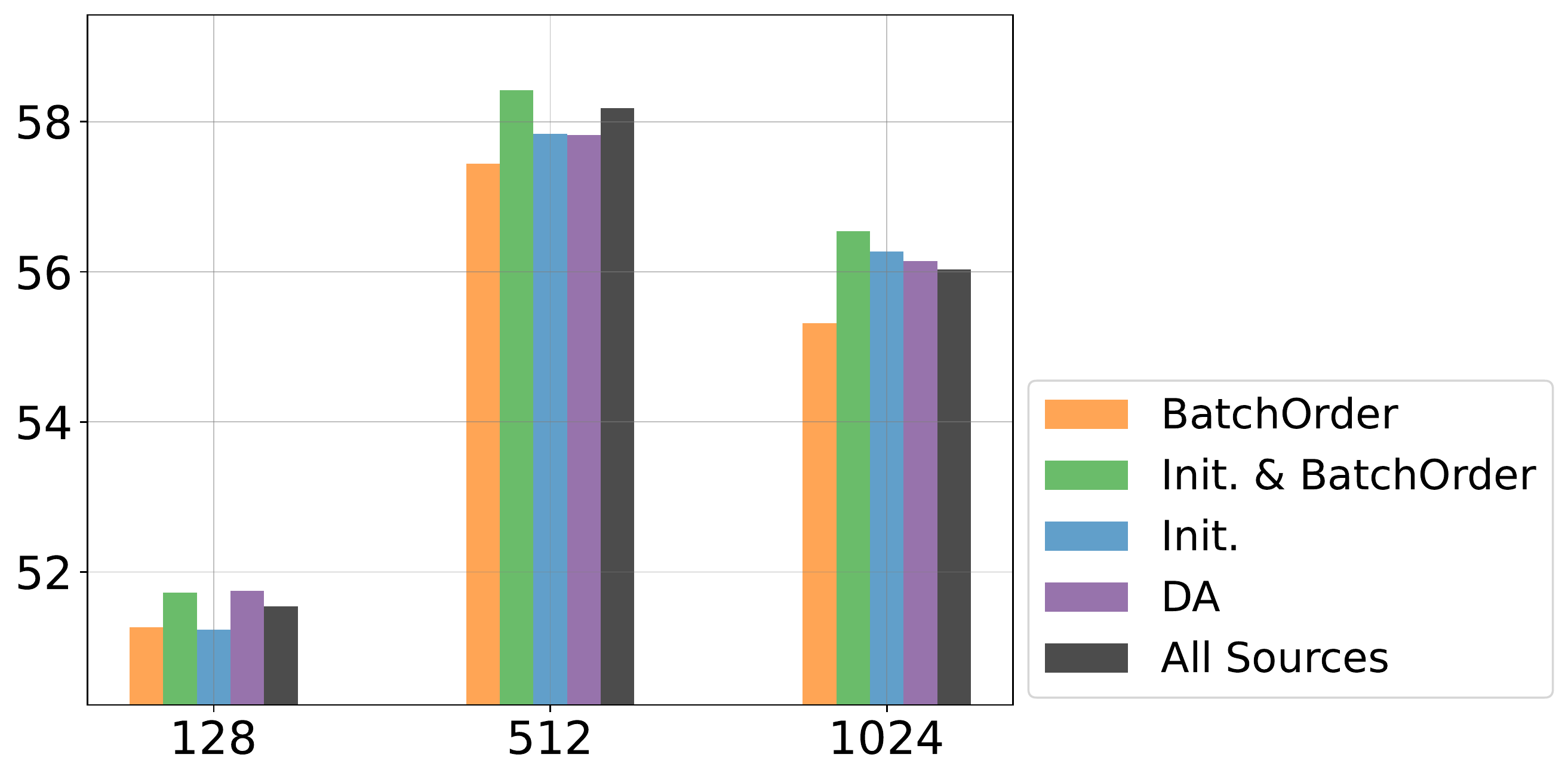}  
        \\[-0.7em]
        batch size
    \end{subfigure}
    \end{minipage}
    
    \centering
    \vspace{0.2cm}
    \underline{ResNet Architectures}\\
    \vspace{0.1cm}
    \begin{minipage}{0.01\linewidth}
        \rotatebox{90}{test accuracy \%}
    \end{minipage}
    \hspace{0.1cm}
    \begin{minipage}{0.97\linewidth}
    \begin{subfigure}{0.32\linewidth}
        \centering
        All-K classes (K=10)\\
        \includegraphics[width=1.0\linewidth]{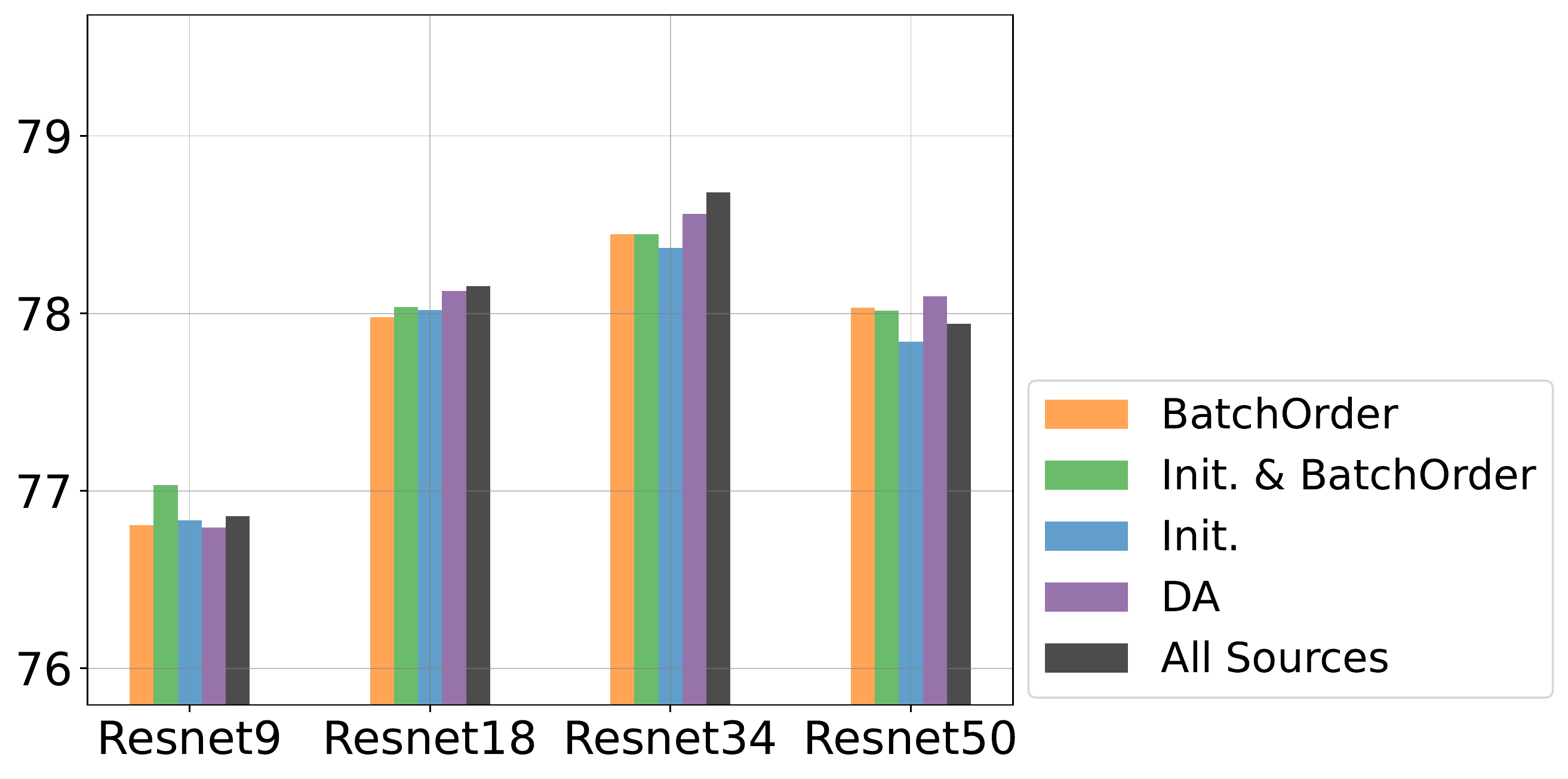}
        \\[-0.7em]
        network architecture
    \end{subfigure}
    \begin{subfigure}{0.32\linewidth}
        \centering
        Top-K classes (K=10)\\
        \includegraphics[width=1.0\linewidth]{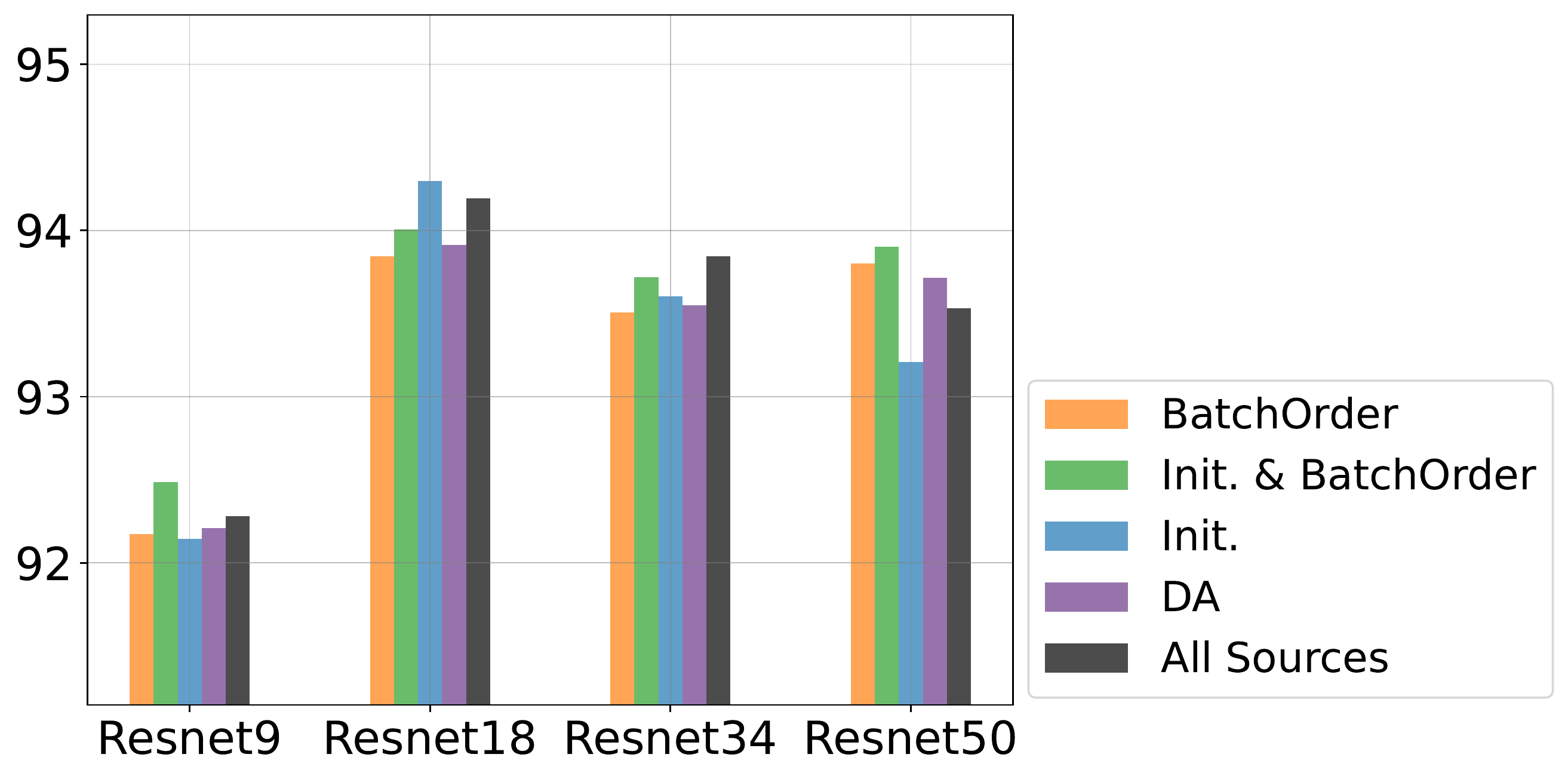}
        \\[-0.7em]
        network architecture
    \end{subfigure}
    \begin{subfigure}{0.32\linewidth}
        \centering
        Bottom-K classes (K=10)\\
        \includegraphics[width=1.0\linewidth]{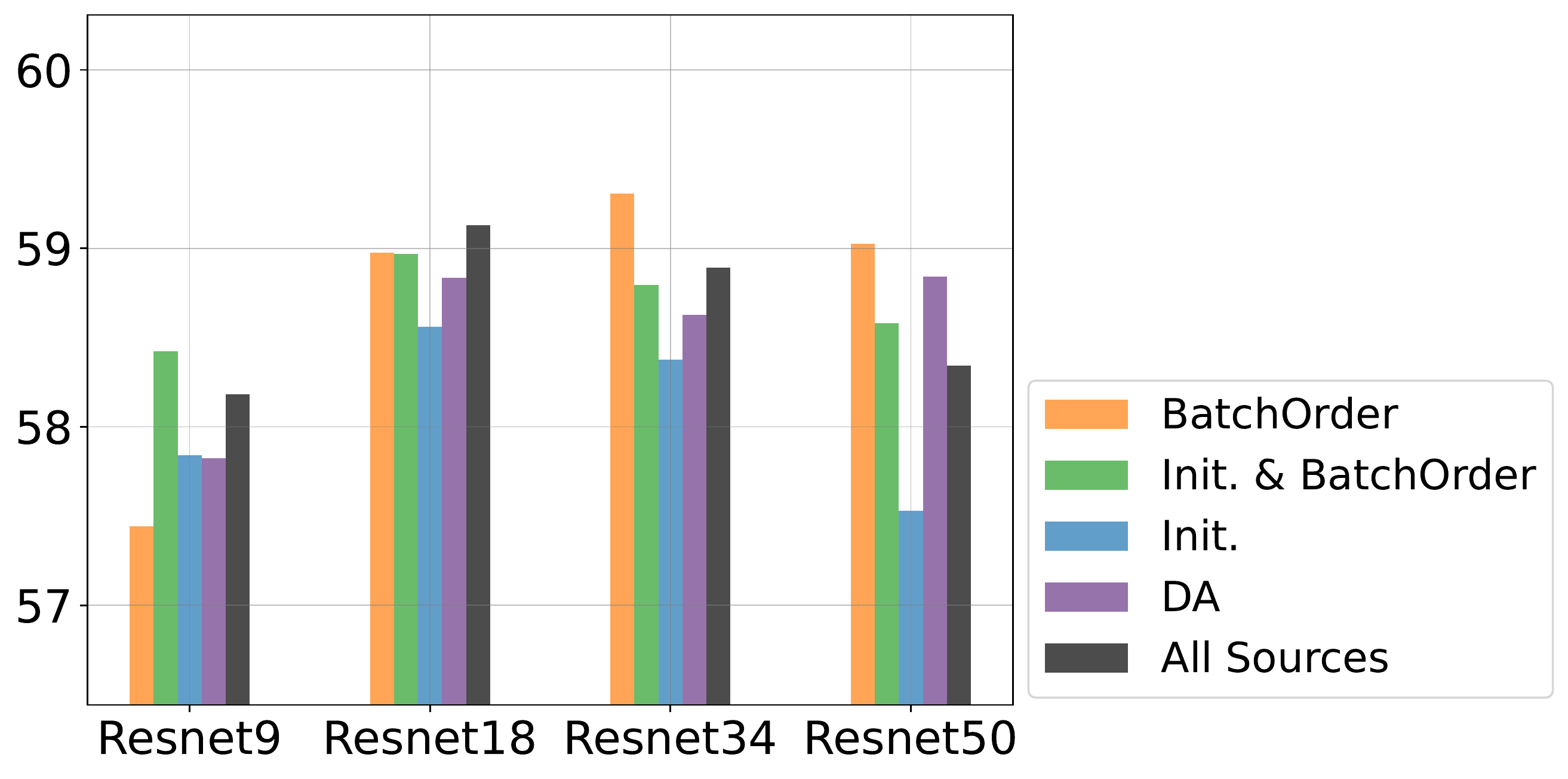}
        \\[-0.7em]
        network architecture
    \end{subfigure}
    \end{minipage}

	\caption{ Average Test Accuracy on CIFAR100 as batch and architecture size increases. Batch 512 is default.
	}
    \label{fig:CIFAR100_batchsize_arch}
\end{figure*}

\begin{figure*}[!t]
    \centering
    \underline{Learning Rate}\\
    \vspace{0.1cm}
    \begin{minipage}{0.01\linewidth}
        \rotatebox{90}{test accuracy \%}
    \end{minipage}
    \hspace{0.1cm}
    \begin{minipage}{0.97\linewidth}
    \begin{subfigure}{0.32\linewidth}
        \centering
        All-K classes (K=10)\\
    	\includegraphics[width=1.0\linewidth]{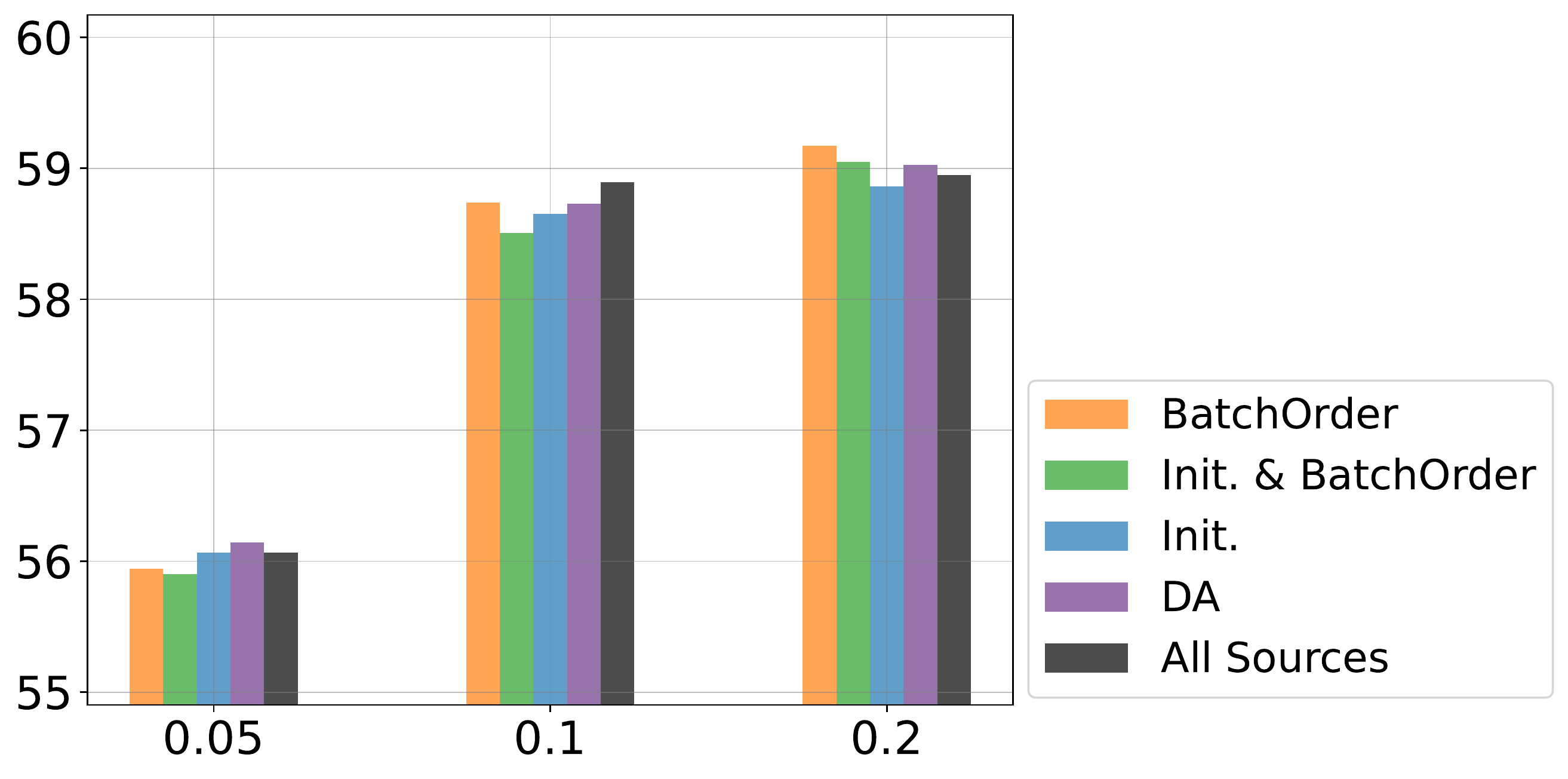}
        \\[-0.7em]
        learning rate
    \end{subfigure}
    \begin{subfigure}{0.32\linewidth}
        \centering
        Top-K classes (K=10)\\
    	\includegraphics[width=1.0\linewidth]{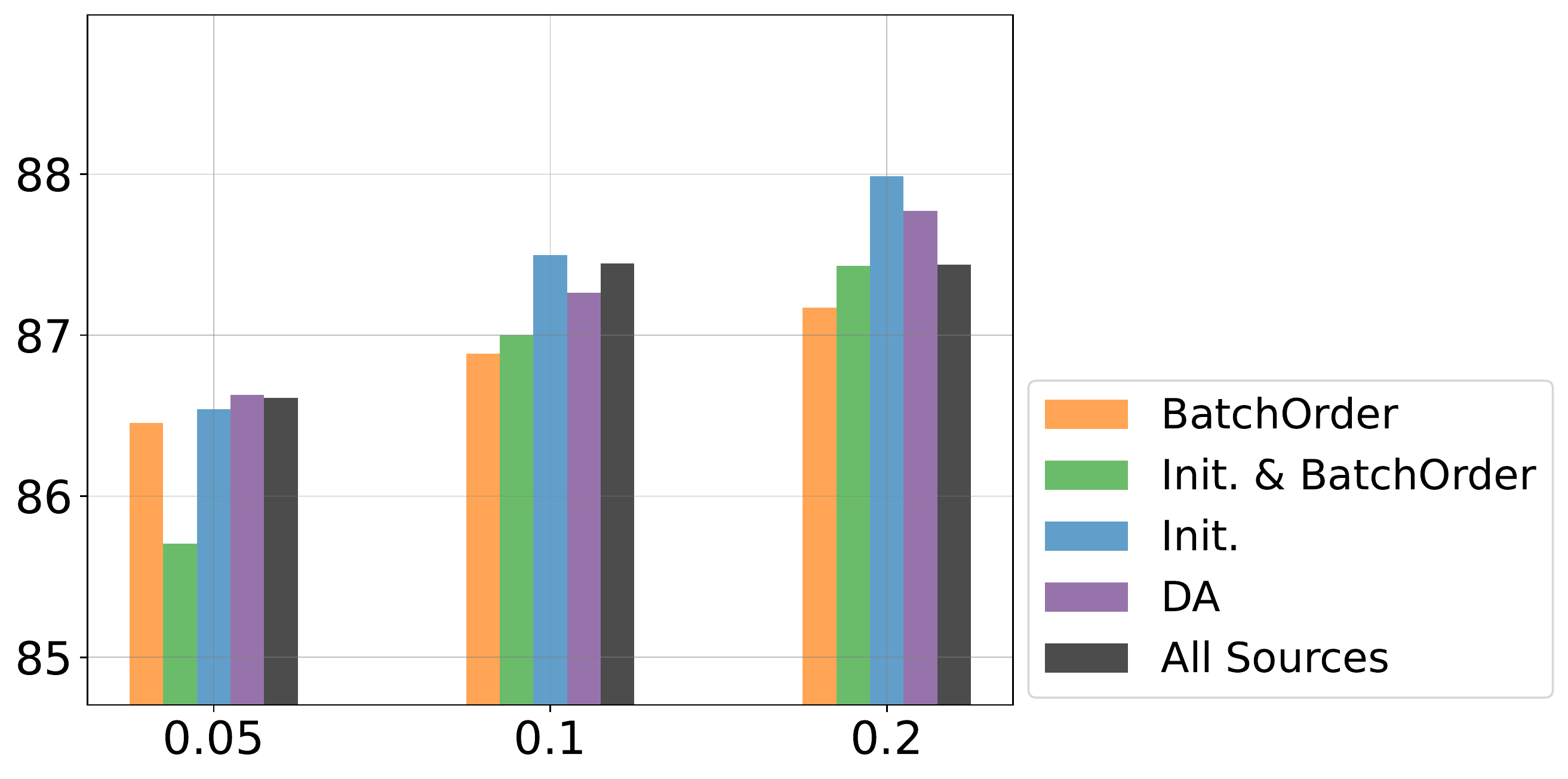}
        \\[-0.7em]
        learning rate
    \end{subfigure}
    \begin{subfigure}{0.32\linewidth}
        \centering
        Bottom-K classes (K=10)\\
        \includegraphics[width=1.0\linewidth]{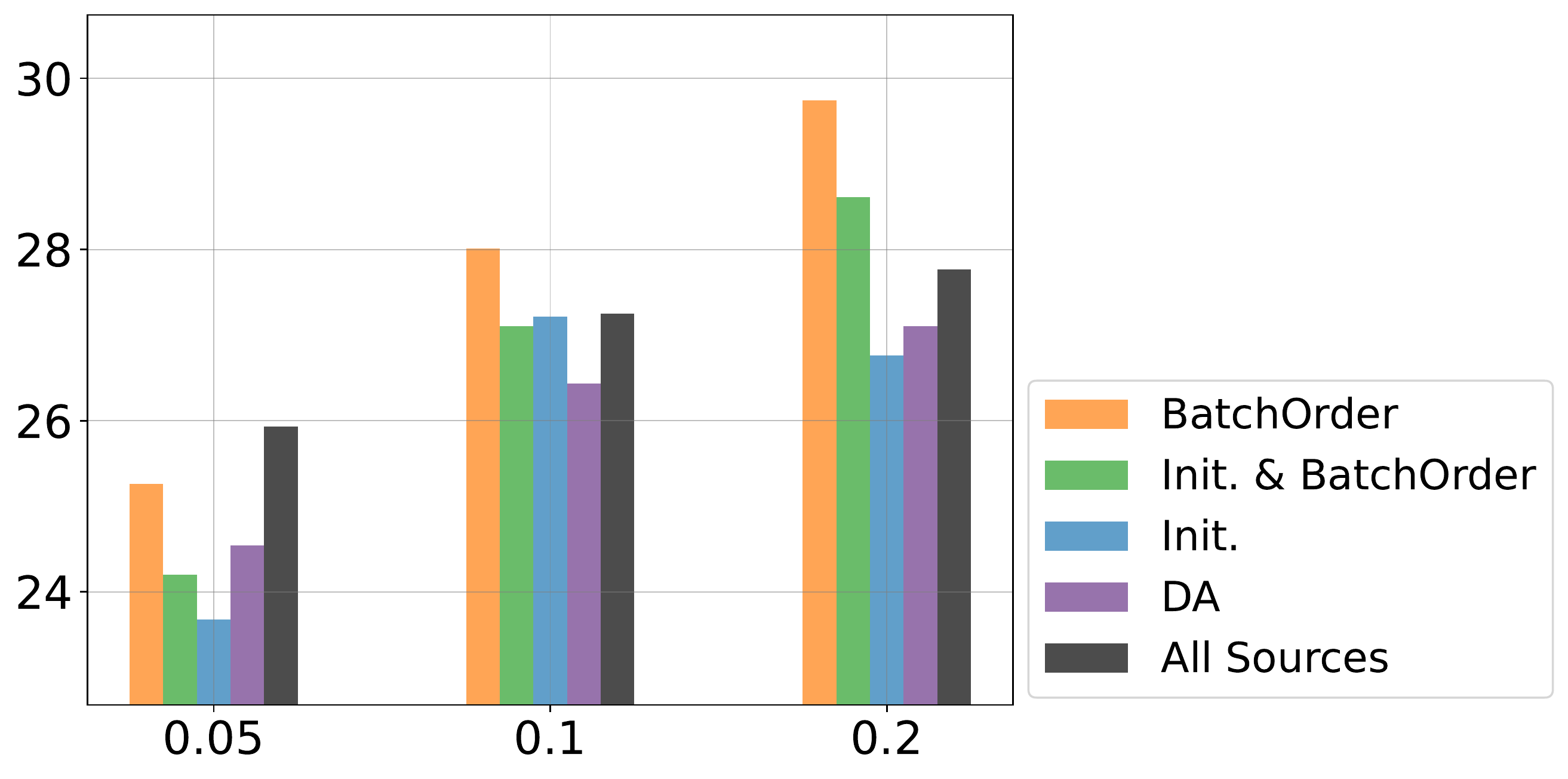}  
        \\[-0.7em]
        learning rate
    \end{subfigure}
    \end{minipage}
    
    \centering
    \vspace{0.2cm}
    \underline{Weight Decay}\\
    \vspace{0.1cm}
    \begin{minipage}{0.01\linewidth}
        \rotatebox{90}{test accuracy \%}
    \end{minipage}
    \hspace{0.1cm}
    \begin{minipage}{0.97\linewidth}
    \begin{subfigure}{0.32\linewidth}
        \centering
        All-K classes (K=10)\\
        \includegraphics[width=1.0\linewidth]{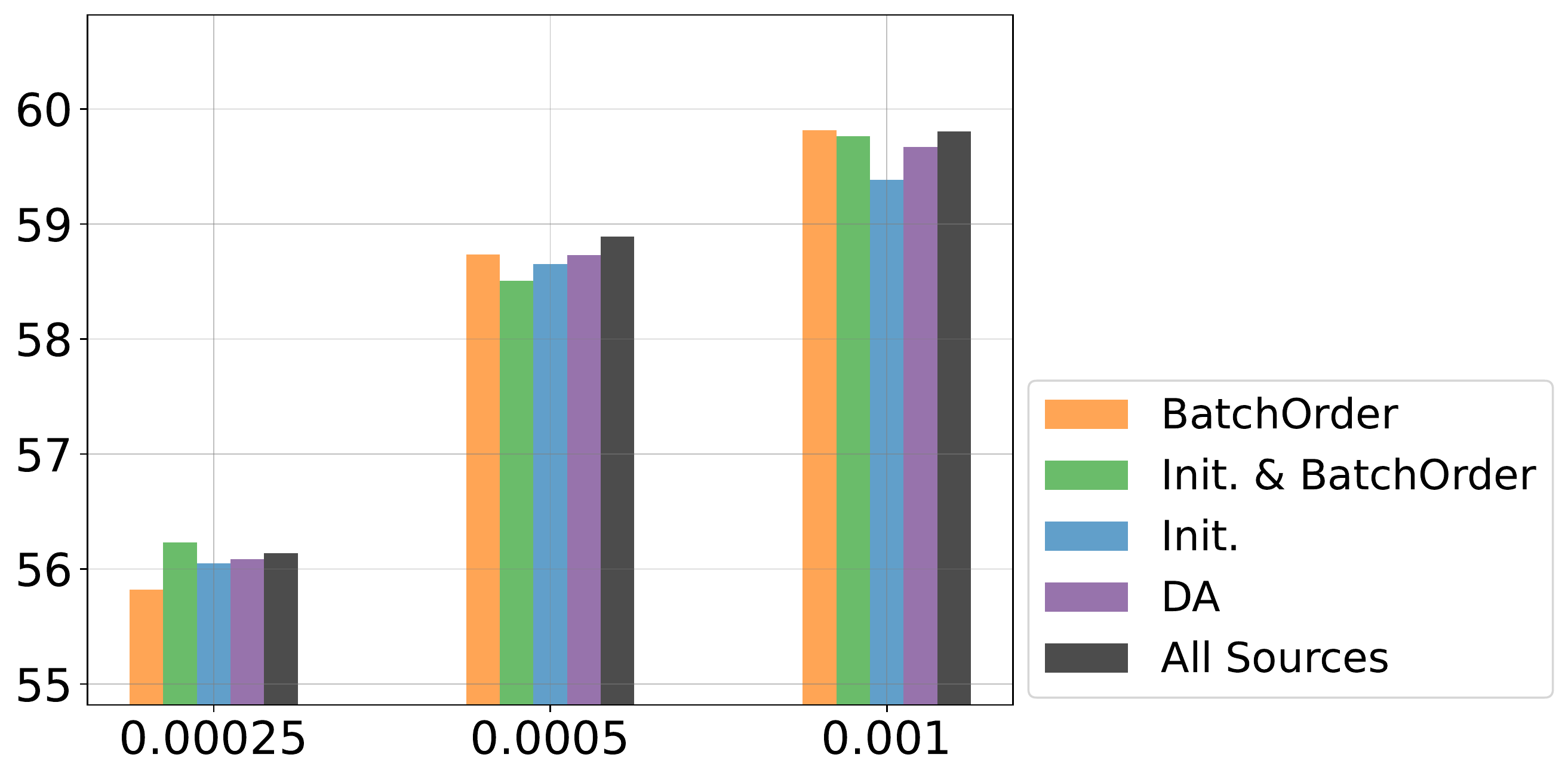}
        \\[-0.7em]
        weight decay
    \end{subfigure}
    \begin{subfigure}{0.32\linewidth}
        \centering
        Top-K classes (K=10)\\
        \includegraphics[width=1.0\linewidth]{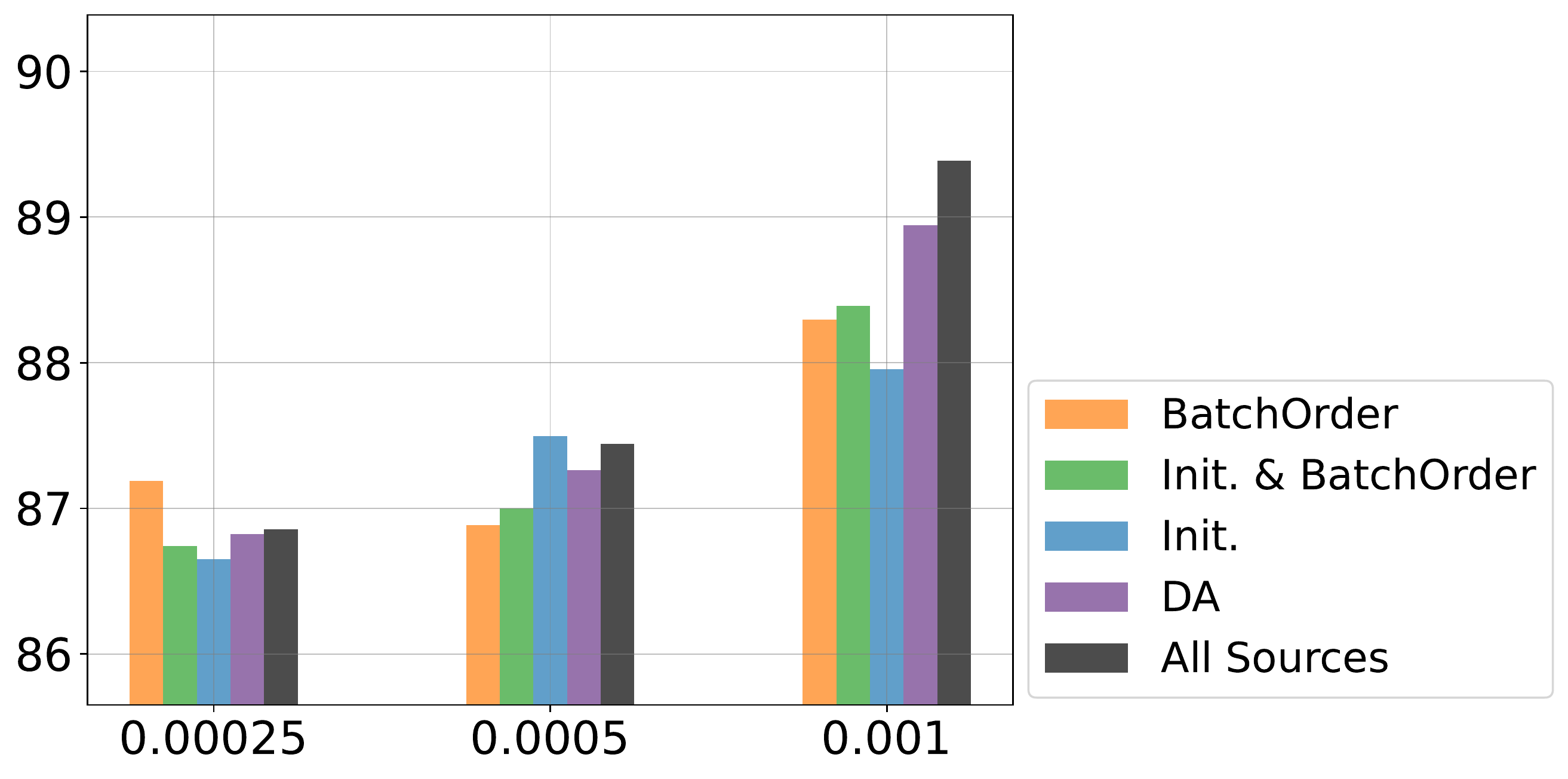}
        \\[-0.7em]
        weight decay
    \end{subfigure}
    \begin{subfigure}{0.32\linewidth}
        \centering
        Bottom-K classes (K=10)\\
        \includegraphics[width=1.0\linewidth]{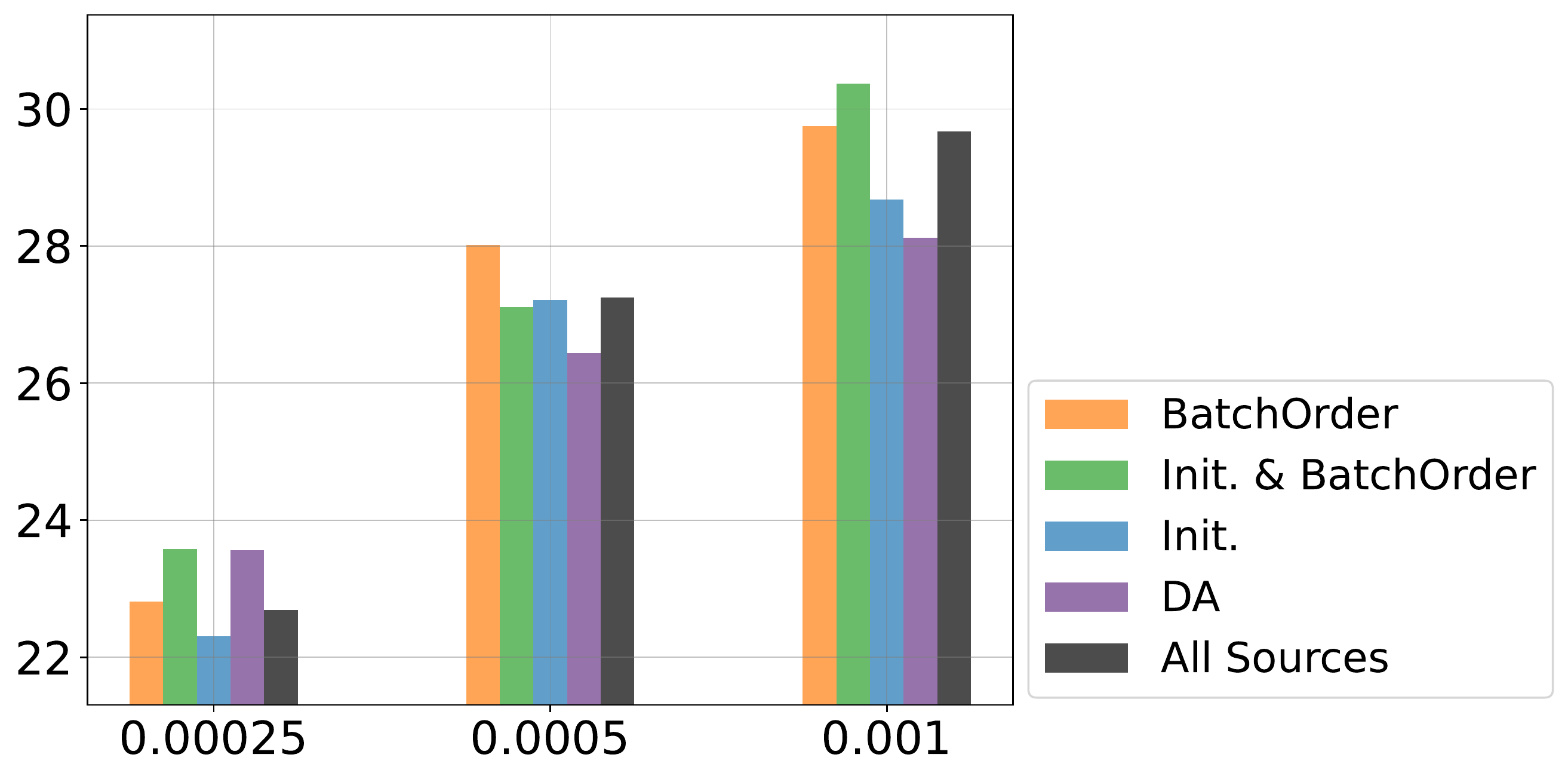}
        \\[-0.7em]
        weight decay
    \end{subfigure}
    \end{minipage}

	\caption{ Average Eval Accuracy on TinyImageNet as learning rate and weight decay increases. 0.1 is default learning rate, and 0.0005 is default weight decay.
	}
    \label{fig:TinyImagenet_lr_wd}
\end{figure*}

\begin{table}[ht!]
    \vspace{6cm}
    \section{Homogeneous ensembles in non-DNN models/non-image datasets}
    \vspace{0.2cm}

\caption{MLP ensemble performance over Adult Census Income subgroups with sensitive attributes}
\begin{tabular}{@{}llllll@{}}
                            &            & \multicolumn{4}{c}{10-model   ensemble}                                           \\ \cmidrule(l){3-6} 
                            & Base Model & BatchOrder         & Initialization     & Init \& BatchOrder & All Sources           \\ \midrule
$>$\$50k                    & \textbf{79.93}      & 79.87   $\pm$ 0.03 & 79.75   $\pm$ 0.28 & 79.57   $\pm$ 0.38 & 79.68   $\pm$ 0.21 \\
$>$\$50k Male               & 25.98      & 26.75   $\pm$ 0.19 & 26.1  $\pm$ 0.49   & \textbf{27.17   $\pm$ 0.83} & 26.96   $\pm$ 0.51 \\
$>$\$50k Female             & 27.97      & 28.63   $\pm$ 0.21 & 28.23   $\pm$ 0.41 & \textbf{28.93   $\pm$ 0.81} & 28.76   $\pm$ 0.4  \\
$>$\$50k White              & 26.48      & 27.22   $\pm$ 0.17 & 26.64   $\pm$ 0.47 & \textbf{27.61   $\pm$ 0.81} & 27.4   $\pm$ 0.48  \\
$>$\$50k Nonwhite           & 24.44      & 25.2   $\pm$ 0.45  & 24.25   $\pm$ 0.6  & \textbf{25.83   $\pm$ 0.88} & 25.62   $\pm$ 0.62 \\
$>$\$50k Black              & 24.02      & 25.29   $\pm$ 0.67 & 23.92   $\pm$ 0.74 & \textbf{25.89   $\pm$ 0.66} & 25.73   $\pm$ 0.57 \\
$>$\$50k Asian-Pac-Islander & 23.31      & 23.64   $\pm$ 0.39 & 22.82   $\pm$ 0.52 & \textbf{24.14   $\pm$ 0.96} & 23.98   $\pm$ 0.7  \\
$>$\$50k Amer-Indian-Eskimo & 26.32      & 26.32   $\pm$ 0    & 27.79   $\pm$ 3.5  & \textbf{28.85   $\pm$ 4.73} & 27.74   $\pm$ 3.14 \\
$>$\$50k Other              & 32         & 32   $\pm$ 0       & 31.44   $\pm$ 1.39 & \textbf{32.12   $\pm$ 0.68} & 32.04   $\pm$ 0.4  \\ \bottomrule
\end{tabular}
\vspace{1cm}
\label{tab:adult_census_mlp}
\end{table}

\begin{table}[ht!]
\vspace{-0.5pt}
\caption{Decision Trees ensemble performance over Adult Census Income subgroups with sensitive attributes}
\begin{tabular}{@{}lll@{}}
                            & Base Model & 10-Model Ensemble  \\ \midrule
$>$\$50k                    & 85.85      & \textbf{85.91   $\pm$ 0.02} \\
$>$\$50k Male               & 60.26      & \textbf{60.35   $\pm$ 0.04} \\
$>$\$50k Female             & 55.59      & \textbf{56.04   $\pm$ 0.09} \\
$>$\$50k White              & 59.89      & \textbf{59.99   $\pm$ 0.05} \\
$>$\$50k Nonwhite           & 56.18      & \textbf{56.73   $\pm$ 0.05} \\
$>$\$50k Black              & \textbf{52.51}      & \textbf{52.51   $\pm$ 0}    \\
$>$\$50k Asian-Pac-Islander & 61.65      & \textbf{62.41   $\pm$ 0}    \\
$>$\$50k Amer-Indian-Eskimo & \textbf{63.16}      & \textbf{63.16   $\pm$ 0}    \\
$>$\$50k Other              & 48         & \textbf{51.84   $\pm$ 0.78} \\ \bottomrule
\end{tabular}
\label{tab:adult_census_decision_trees}
\end{table}

\begin{table}[h]
    \section{Top and Bottom Classes for CIFAR100 and TinyImageNet}
    \vspace{0.2cm}
    \subsection{CIFAR100}
    \vspace{0.2cm}
    \centering
    \setlength\tabcolsep{1.4em}
    \caption{Top-10 and Bottom-10 class names for CIFAR100. The classes are from the averaged test accuracies from the 20-model ensembles.}
    \begin{tabular}{llllll}
    \hline
     ResNet9    & ResNet18   & ResNet34      & ResNet50      & VGG16      & MLP-Mixer   \\
    \hline
    \multicolumn{6}{c}{Top-10} \\
    \hline
     wardrobe   & skunk      & skunk         & orange        & road       & wardrobe    \\
     motorcycle & orange     & road          & wardrobe      & wardrobe   & motorcycle  \\
     orange     & motorcycle & orange        & motorcycle    & sunflower  & orange      \\
     skunk      & road       & sunflower     & skunk         & motorcycle & sunflower   \\
     road       & wardrobe   & motorcycle    & road          & skyscraper & road        \\
     chimpanzee & palm\_tree  & wardrobe      & sunflower     & skunk      & skyscraper  \\
     sunflower  & chimpanzee & palm\_tree     & chimpanzee    & palm\_tree  & keyboard    \\
     orchid     & sunflower  & pickup\_truck  & palm\_tree     & orange     & palm\_tree   \\
     mountain   & tractor    & aquarium\_fish & aquarium\_fish & chair      & plain       \\
     apple      & skyscraper & skyscraper    & lawn\_mower    & chimpanzee & skunk       \\
    \hline
    \multicolumn{6}{c}{Bottom-10} \\
    \hline
     man        & mouse      & shark         & girl          & possum     & mouse       \\
     shark      & bear       & possum        & lizard        & crocodile  & bowl        \\
     lizard     & shark      & crocodile     & possum        & girl       & woman       \\
     bowl       & girl       & lizard        & maple\_tree    & shark      & girl        \\
     possum     & lizard     & girl          & bear          & bear       & squirrel    \\
     shrew      & man        & man           & otter         & lizard     & possum      \\
     seal       & otter      & bowl          & bowl          & seal       & lizard      \\
     girl       & seal       & otter         & man           & boy        & boy         \\
     otter      & bowl       & seal          & boy           & otter      & otter       \\
     boy        & boy        & boy           & seal          & man        & seal        \\
    \hline
    \end{tabular}
    \label{tab:c100_k_class_names}
    \end{table}

    \begin{table}[h]
    \centering
    \subsection{TinyImageNet}
    \setlength\tabcolsep{1.4em}
    \caption{Top-10 and Bottom-10 wnid names for TinyImageNet. The names are from the averaged test accuracies from the 20-model ensembles.}
    \begin{tabular}{llllll}
    \hline
     ResNet9   & ResNet18   & ResNet34   & ResNet50   & VGG16     & ViT       \\
    \hline
    \multicolumn{6}{c}{Top-10} \\
    \hline
     n02791270 & n02791270  & n02791270  & n02791270  & n02791270 & n07875152 \\
     n02509815 & n02509815  & n02509815  & n02509815  & n03042490 & n03814639 \\
     n03976657 & n02906734  & n02906734  & n02906734  & n02509815 & n03983396 \\
     n02124075 & n03042490  & n03814639  & n03042490  & n03814639 & n03042490 \\
     n03814639 & n03814639  & n01950731  & n01950731  & n02906734 & n02823428 \\
     n03089624 & n03976657  & n03599486  & n04067472  & n01950731 & n03599486 \\
     n03983396 & n01950731  & n03042490  & n03599486  & n04398044 & n02509815 \\
     n02002724 & n04560804  & n03976657  & n03976657  & n02124075 & n02791270 \\
     n03126707 & n03599486  & n04067472  & n07579787  & n03089624 & n03126707 \\
     n03447447 & n02002724  & n03126707  & n03126707  & n04067472 & n02906734 \\
    \hline
    \multicolumn{6}{c}{Bottom-10} \\
    \hline
     n02437312 & n04532670  & n03160309  & n03544143  & n02085620 & n02927161 \\
     n04070727 & n03544143  & n01945685  & n03617480  & n04417672 & n03544143 \\
     n02268443 & n04486054  & n04417672  & n04070727  & n02268443 & n04070727 \\
     n01945685 & n02268443  & n04532670  & n03804744  & n04486054 & n01641577 \\
     n02226429 & n03160309  & n03617480  & n03160309  & n01945685 & n02094433 \\
     n02233338 & n03617480  & n01855672  & n01945685  & n02094433 & n02480495 \\
     n02480495 & n01855672  & n03804744  & n02268443  & n04070727 & n02410509 \\
     n02410509 & n02480495  & n02480495  & n02480495  & n02480495 & n04532670 \\
     n03617480 & n02123394  & n02123394  & n02123394  & n02410509 & n02950826 \\
     n02123394 & n02410509  & n02410509  & n02410509  & n02123394 & n02123394 \\
    \hline
    \end{tabular}    
    \label{tab:tinyimagenet_k_class_names}
\end{table}


\end{document}